\documentclass[10pt,journal,compsoc]{IEEEtran}

\ifCLASSOPTIONcompsoc
  \usepackage[nocompress]{cite}
\else
  \usepackage{cite}
\fi

\usepackage{epsfig}
\usepackage{graphicx}
\usepackage{amsmath}
\usepackage{amssymb}
\usepackage{booktabs}
\usepackage{algorithm}
\usepackage{algorithmic}
\usepackage{multirow}
\usepackage{subfigure}
\usepackage{amsthm}
\usepackage{url}
\usepackage{rotate}
\usepackage{bm}
\usepackage{setspace}
\usepackage{color}
\usepackage{hyperref}

\usepackage{colortbl}
\definecolor{mygray}{gray}{.9}

\def\bomega{\bm{\omega}}
\def\balpha{\bm{\alpha}}

\def\ttt{\mathtt{s}}
\def\tn{\mathtt{t}}

\ifCLASSINFOpdf
\else
\fi
\begin{document}
%
\title{Learning with Nested Scene Modeling and Cooperative Architecture Search\\ for Low-Light Vision}
%
%
%
%

\author{Risheng~Liu,~\IEEEmembership{Member,~IEEE,}
	Long~Ma,
	Tengyu~Ma,
	Xin~Fan,~\IEEEmembership{Senior Member,~IEEE,}
	and Zhongxuan~Luo
	\IEEEcompsocitemizethanks{\IEEEcompsocthanksitem R. Liu is with DUT-RU International School of Information Science \& Engineering, Dalian University of Technology, Dalian, 116024, China. (Corresponding author, e-mail: rsliu@dlut.edu.cn).
	\IEEEcompsocthanksitem L. Ma is with the School of Software Technology, Dalian University of Technology, Dalian, 116024, China. (e-mail: malone94319@gmail.com).
	\IEEEcompsocthanksitem T. Ma is with the School of Software Technology, Dalian University of Technology, Dalian, 116024, China. (e-mail:matengyu@mail.dlut.edu.cn).
	\IEEEcompsocthanksitem X. Fan is with the DUT-RU International School of Information Science \& Engineering, Dalian University of Technology, Dalian, 116024, China. (email: xin.fan@dlut.edu.cn).
	\IEEEcompsocthanksitem Z. Luo is with the School of Software Technology, Dalian University of Technology, Dalian, 116024, China. (email: zxluo@dlut.edu.cn).
}
	\thanks{Manuscript received April 19, 2005; revised August 26, 2015.}}

%
%

\markboth{Journal of \LaTeX\ Class Files,~Vol.~14, No.~8, August~2015}%
{Shell \MakeLowercase{\textit{et al.}}: Bare Advanced Demo of IEEEtran.cls for IEEE Computer Society Journals}
%



\IEEEtitleabstractindextext{%
\begin{abstract}
Images captured from low-light scenes often suffer from severe degradations, including low visibility, color cast and intensive noises, etc. These factors not only affect image qualities, but also degrade the performance of downstream Low-Light Vision (LLV) applications. A variety of deep learning methods have been proposed to enhance the visual quality of low-light images. However, these approaches mostly rely on significant architecture engineering to obtain proper low-light models and often suffer from high computational burden. Furthermore, it is still challenging to extend these enhancement techniques to handle other LLVs. To partially address above issues, we establish Retinex-inspired Unrolling with Architecture Search (RUAS), a general learning framework, which not only can address low-light enhancement task, but also has the flexibility to handle other more challenging downstream vision applications. Specifically, we first establish a nested optimization formulation, together with an unrolling strategy, to explore underlying principles of a series of LLV tasks. Furthermore, we construct a differentiable strategy to cooperatively search specific scene and task architectures for RUAS. Last but not least, we demonstrate how to apply RUAS for both low- and high-level LLV applications (e.g., enhancement, detection and segmentation). Extensive experiments verify the flexibility, effectiveness, and efficiency of RUAS.
\end{abstract}

}

\maketitle

\IEEEdisplaynontitleabstractindextext

%
\IEEEpeerreviewmaketitle


\IEEEraisesectionheading{\section{Introduction}\label{sec:introduction}}

\IEEEPARstart{L}{ow}-Light Vision (LLV) mainly refers to various visual enhancement and perception tasks in low-light scenarios and is always more challenging than classical computer vision problems. This is because that the poor visual quality of low-light observations (e.g., low visibility, color cast, and intensive noises caused by diverse degraded factors) could heavily affect information extraction and transformation in these vision tasks~\cite{liu2021retinex,wang2021hla,wu2021dannet}.


%
%
\begin{figure}[!htb]
	\centering
	\begin{tabular}{c@{\extracolsep{0.4em}}c}
		\includegraphics[width=0.47\linewidth]{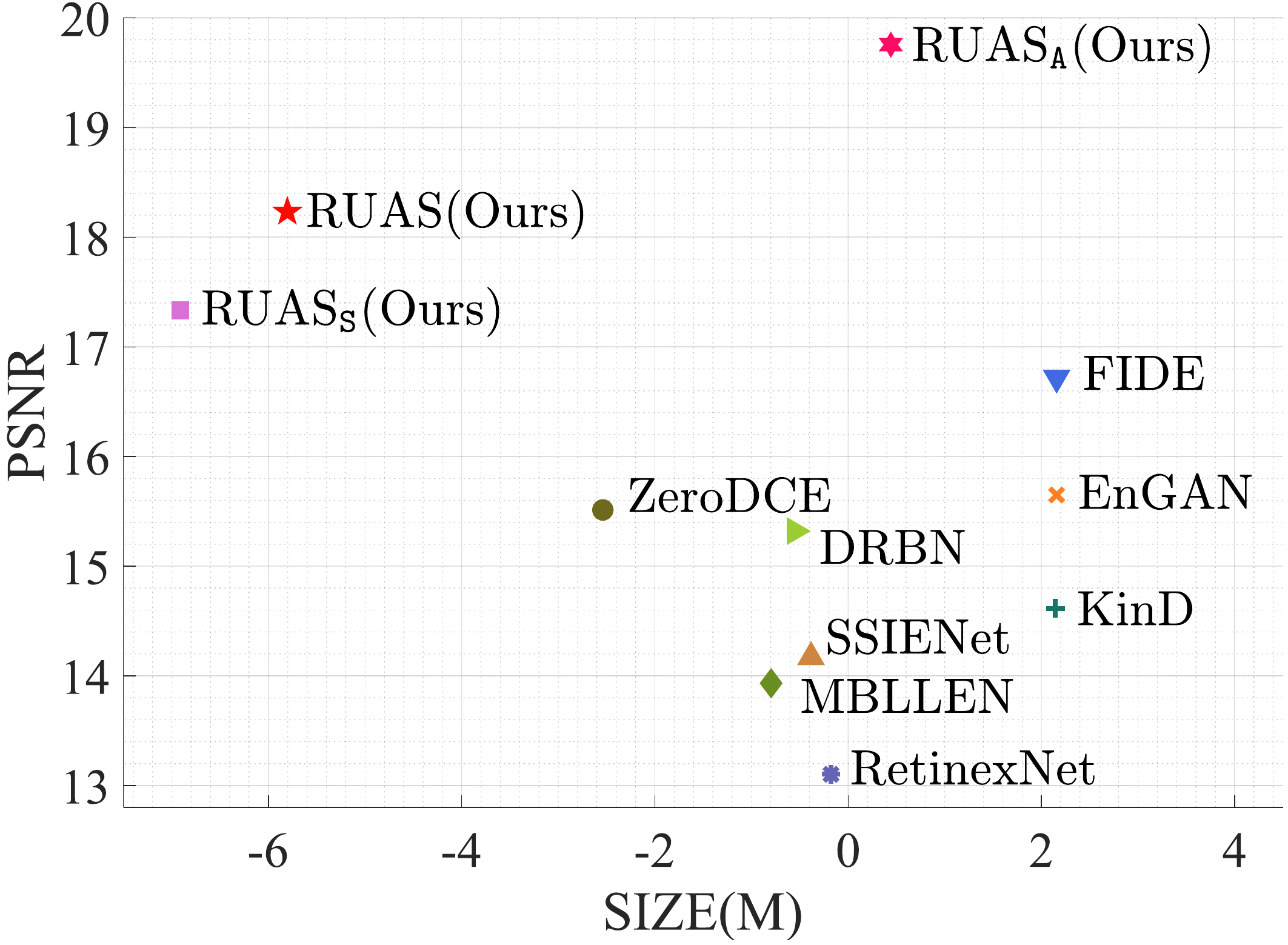}&
		\includegraphics[width=0.47\linewidth]{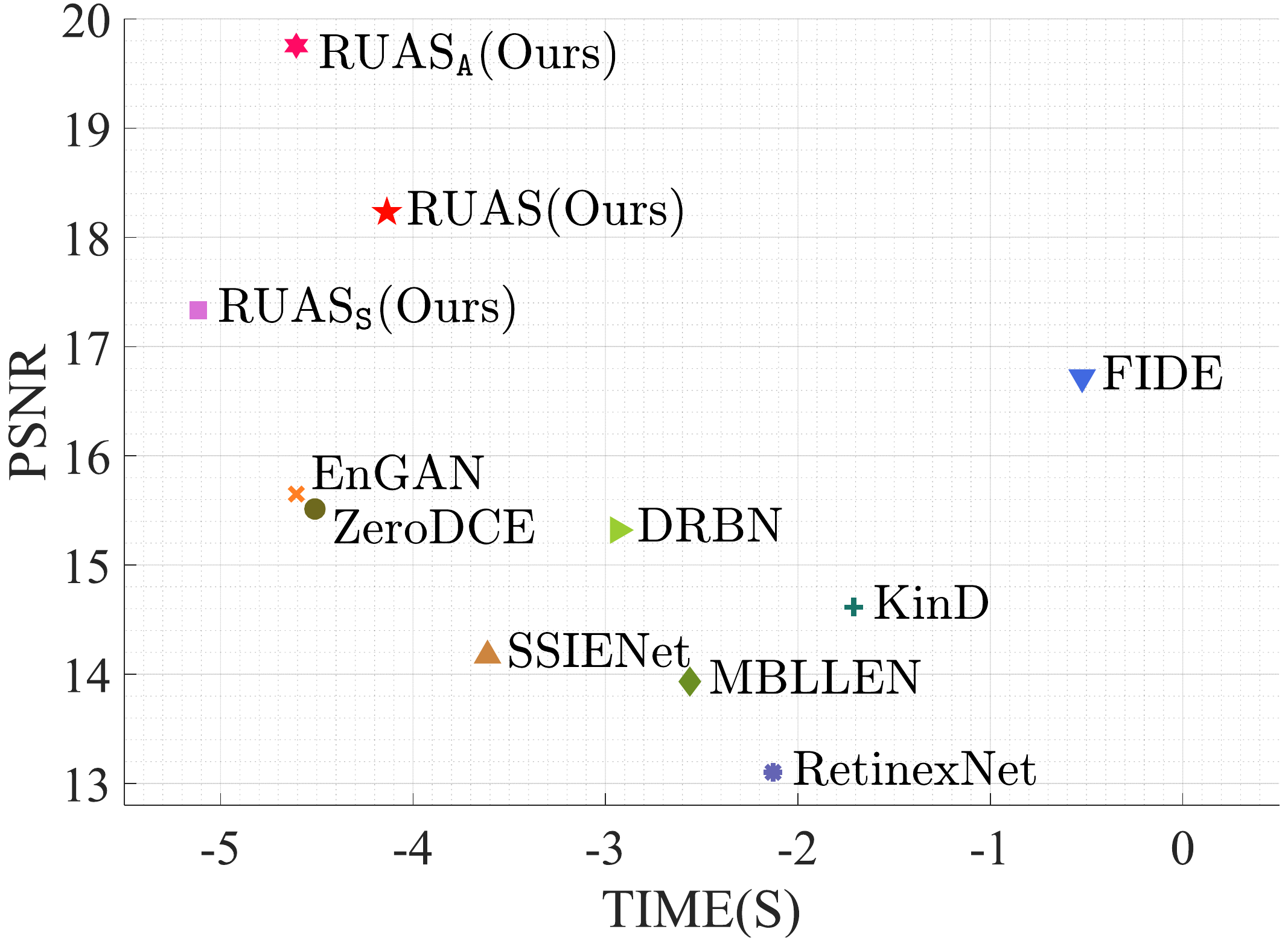}\\
		\multicolumn{2}{c}{\footnotesize (a) Image Enhancement}\\
		\includegraphics[width=0.47\linewidth]{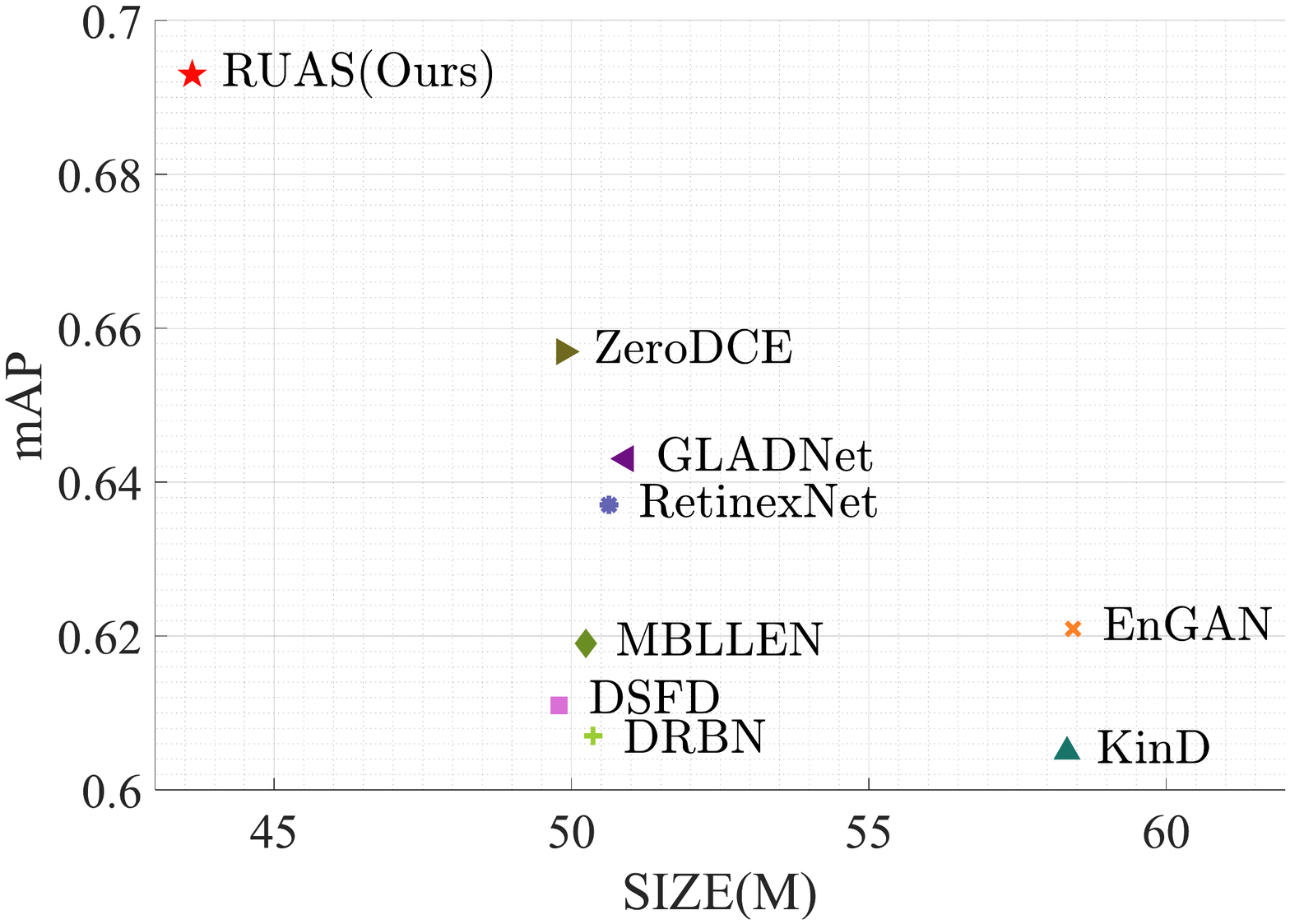}&
		\includegraphics[width=0.47\linewidth]{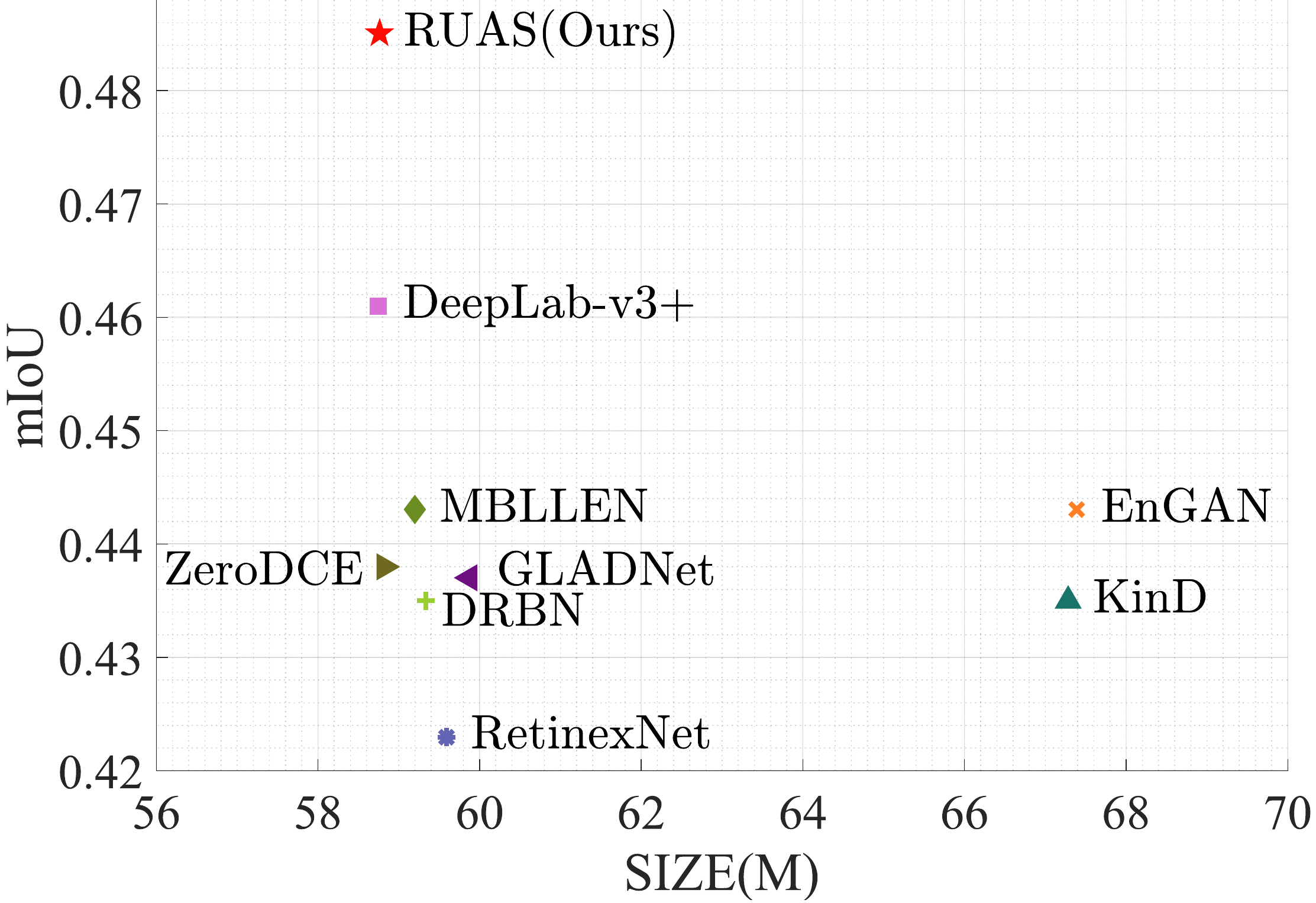}\\
		\footnotesize (b) Object Detection& \footnotesize (c) Semantic Segmentation\\
	\end{tabular}
	\caption{Illustrating averaged quantitative results of RUAS on various LLV tasks. In (a), we first compare our RUAS (with three kinds of variations) to state-of-the-art low-light enhancement approaches on LOL dataset~\cite{Chen2018Retinex}. We then demonstrate the performance of RUAS for objective detection (on DARK FACE dataset~\cite{liu2021benchmarking}) and semantic segmentation (on ACDC dataset~\cite{sakaridis2021acdc}) in (b) and (c), respectively. For two high-level tasks, we also demonstrate the overall performance of the two-stage techniques, i.e., integrate these enhancement methods with state-of-the-art detection (e.g., DSFD~\cite{li2019dsfd}) and segmentation (e.g., DeepLab-v3+~\cite{Chen2018DeepLabv3plus}) techniques. More detailed analysis can also be found in our experimental part.}
	\label{fig:FirstFigure}	
\end{figure}


In the past few years, researchers have made a considerable effort on designing deep learning models to enhance the visual quality of low-light observations. Learning with paired data~\cite{Chen2018Retinex,zhang2019kindling,liu2021underexposed,xu2020learning} is the prevalent pattern to acquire the desired model, although the physical principle (with high generality) is what they utilize in most works, which is because of the distribution limitation of supervised learning that these techniques manifest the poor generalization in the unknown real-world scenarios. Unsupervised learning paradigm for low-light image enhancement~\cite{yang2020fidelity,jiang2019enlightengan,guo2020zero} emerges as time requires, to provide a strong adaptation ability towards diverse low-light conditions. Indeed, these works realize better performance than those based on supervised learning. However, currently employed methods have mostly been developed manually, thus requiring significant architecture engineering and large-size models to explore the underlying low-light scene information. More importantly, these works with supervised/unsupervised learning are specifically designed for the purpose of improving the visual quality, hardly be extended to other types of LLVs (e.g., detection and segmentation). 
%

Very recently, benefiting from the development of intelligent techniques (e.g., autonomous driving), several perception tasks have also been considered in low-light scenarios~\cite{wang2021hla,liang2021recurrent,wu2021dannet}. 
Existing works mostly devote themselves to shrinking the gap between low-light observations and clear images, and further fine-tune perception algorithms that are designed on the regular natural environments. However, in this way, it pays too much attention to the visual quality of low-light observations, causing unfavorable outcomes of the perception algorithms because of ignoring the intrinsic interactions for scenes and tasks. In other words, how to explore the latent correspondence between low-light scenes and perception tasks may be more important. 

To partially address the above issues, we propose a general learning framework, named Retinex-inspired Unrolling with Architecture Search (RUAS)\footnote{A preliminary version of this work has been published in~\cite{liu2021retinex}.}. As shown in Fig.~\ref{fig:FirstFigure}, our method realizes the best task performance with the lowest computational resources both in low-light image enhancement and other extended high-level vision tasks (e.g., detection and segmentation). 
Concretely, taking low-light scenes into consideration, we first reformulate the low-light vision as a scene energy constrained learning formulation. Then by integrating the Retinex rule to obtain an unrolling scheme for training, we establish two kinds of training strategies from the training objective that is with an unsupervised scene-oriented loss. Finally, we provide a cooperative architecture search strategy for the scene and task module. Our contributions can be summarized as follows\footnote{Code is available at \url{https://github.com/vis-opt-group/RUAS}.}:

\begin{itemize}
	\item RUAS establishes a general and principled learning framework, which can effectively formulate the underlying low-light scene information for LLV tasks, thus not only has the ability to enhance the visual quality of low-light images, but also is flexible enough to solve other challenging perception-type vision applications.
	
	\item We establish a nested optimization model to simultaneously formulate the low-light scene feature, vision task and their intrinsic relationships. By designing a Retinex-inspired unrolling scheme and introducing an unsupervised scene-oriented loss, we obtain a prior and data aggregated training strategy for RUAS.
	
	
	\item For architecture search, we develop a differentiable strategy, which is able to discover cooperative scene and task architectures from a compact search space, thus has the ability to not only obtain memory and computation efficient models, but also offer flexibility for different kinds of LLVs.

	\item We experimentally demonstrate that RUAS can be effectively applied to both low- and high-level LLV applications (e.g., enhancement, detection and segmentation) with favorable performance against the state-of-the-art approaches. 
\end{itemize}

%
%
%
%
%

\begin{figure*}[t]
	\centering
		\begin{tabular}{c}
			\includegraphics[width=0.98\linewidth]{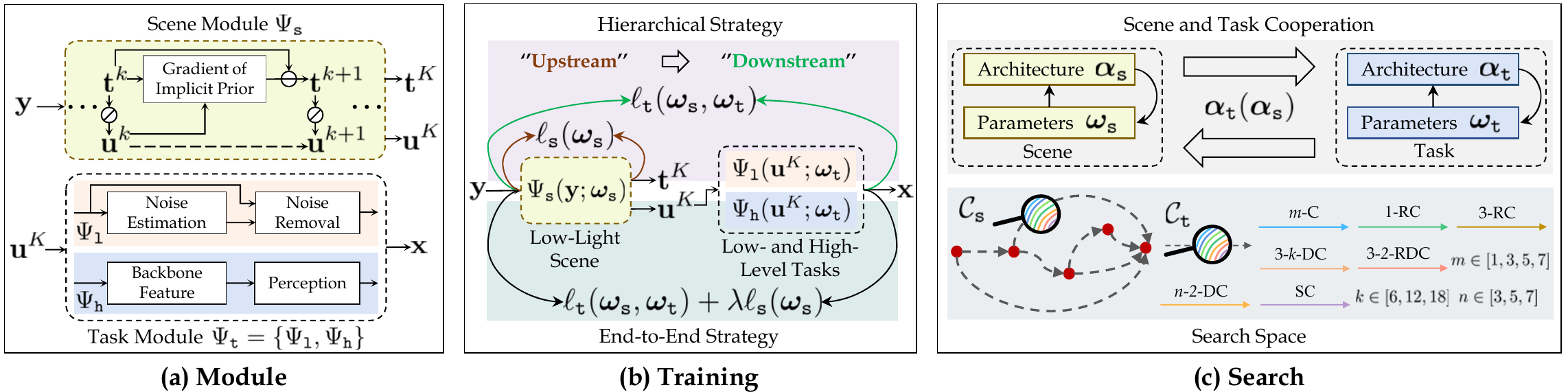}\\
		\end{tabular}	
	\caption{Illustrating the modules, training and search strategies of RUAS. (a) first demonstrates our scene and task modules. Here we introduce $\Psi_\mathtt{l}$ and $\Psi_\mathtt{h}$ to denote modules for low- and high-level tasks, respectively. In (b), we further illustrate two kinds of training strategies (i.e., end-to-end and hierarchical) for RUAS. Finally, we show the search space and our cooperative architecture search strategy in (c).}
	\label{fig:flowchart}
\end{figure*}

\section{Related Work}
In this section, we comprehensively review two related topics including low-level LLVs and high-level LLVs.  

\subsection{Low-Level LLVs: Enhancing the Visual Quality of Low-Light Scenes}
Low-level LLVs concentrate on the improvement of visual quality for the given low-light observations. 
With the development of deep learning, low-light image enhancement has achieved a significant performance boost. 

In view of lacking paired dataset captured in real-world, training the network model on the synthetic/retouched image pairs was a commonly-used way in the early stage. 
LLNet~\cite{lore2017llnet} utilized a variant of the stacked sparse denoising autoencoder to brighten the low-light images and denoising simultaneously, and trained the model on the synthetic dataset acquired by performing the Gamma adjustment.  Shen~\emph{et al.}~\cite{shen2017msr} treated the multi-scale Retinex as the cascaded Gaussian convolution with residual connection to establish a multi-scale convolutional neural network MSRNet and trained it on the paired dataset that was retouched by using Photoshop.  GLADNet~\cite{wang2018gladnet} proposed a global illumination-aware and detail-preserving network and also adopted the same synthetic data generation manner with MSRNet. Recently, LightenNet~\cite{li2018lightennet} learned the mapping between the observation and desired illumination and constructed a synthetic dataset by performing Retinex theory for training. Ren~\emph{et al.}~\cite{ren2019low} designed a deep hybrid network to simultaneously learn the global content and salient structure for clear images, they also darkened images by using gamma correction for training. The work in~\cite{lv2019attention} proposed to construct a large-scale dataset that contained multiple low-light levels by using a series of low-light simulation strategies. This work further built a multi-branch decomposition-and-fusion enhancement network. Wang~\emph{et al.}~\cite{wang2019underexposed} designed an image-to-illumination network architecture based on the bilateral learning framework as presented in~\cite{gharbi2017deep} and built a paired dataset by expert-retouching for training.  
However, since the difference in data distributions, these models with supervised learning mechanisms that were trained on simulated datasets mostly cannot consistently perform excellently in real-world scenarios. 

To better simulate real-world scenarios, there emerged two patterns including acquiring paired datasets in real-world environments for training~\cite{chen2018learning,Chen2018Retinex,hai2021r2rnet} and designing the unsupervised learning mechanisms~\cite{zhang2020self,guo2020zero,ma2021learning}. As for the first pattern, SID~\cite{chen2018learning} was a representative raw dataset that had been widely applied to industrial and academic communities. The work in~\cite{xu2020learning} developed a frequency-based decomposition-and-enhancement network based on the attention to context encoding module, it trained the network on a new paired dataset acquired by recreating the SID dataset. 
Chen~\emph{et al.} constructed a famous paired low-light dataset (named LOL) that was captured by adjusting the exposure time and proposed the RetinexNet for estimating the reflectance and illumination, further performed the denoising process. Similarly, the paper in~\cite{hai2021r2rnet} further constructed a large-scale paired dataset on real-world scenarios and proposed a new architecture for brightening and denoising simultaneously. There exist a lot of deep learning techniques by training their models on the well-known LOL dataset. Such as, the work in~\cite{zhang2019kindling} designed an end-to-end network that was similar to RetinexNet but connected the feature-level illumination and reflectance in the decomposition step. Although these works indeed ameliorate performance for those methods trained on the synthetic dataset. The counterfeit low-light condition by adjusting the exposure level still cannot cater to the actual demand. 

The unsupervised learning paradigm aimed to learn the desired outputs by introducing different knowledge, instead of supervised signals. EnGAN~\cite{jiang2019enlightengan} designed an attention module on U-Net~\cite{ronneberger2015u} and can be trained with only low/normal-light images (unnecessarily paired). Zhang~\emph{et al.}~\cite{zhang2020self} established a self-supervised CNN to simultaneously output the illumination and reflectance. The work in~\cite{guo2020zero} proposed a zero-reference curve estimation CNN to address the LLIE task. A recursive band network was proposed in~\cite{yang2020fidelity} and trained by a semi-supervised strategy. Ma~\emph{et al.}~\cite{ma2021learning} developed a context-sensitive decomposition network with supervised and unsupervised manners for realizing high-efficient low-light image enhancement. 
Nonetheless, discovering state-of-the-art neural network architectures requires substantial effort. Very recently, Liu~\emph{et al.}~\cite{liu2021retinex} introduced the neural architecture search in the unrolling scheme to acquire the high-efficient network model, but it needs to manually switch the task mode for different scenarios which reduces the flexibility. 

\subsection{High-Level LLVs: Low-Light Scene Perception}
High-level LLVs aims at understanding low-light scenes by extracting valuable semantic-level information, rather than pixel-level treatment focused by low-level LLVs. Generally, high-level LLVs possesses powerful practicability because it can pander to the cognitive process of humans, to support more intelligent applications. 

Recently, few efforts had been made by researchers for target-specific low-light object detection, e.g., face detection. 
A significant event was UG$^2$+ Challenge\footnote{\url{http://www.ug2challenge.org/}} competition which had been held for consecutive four years. This competition offered the DARK FACE dataset~\cite{yang2020advancing,liu2021benchmarking} which was a large-scale low-light face detection dataset with annotation labels. A mainstream, top-performing strategy in this competition was to combine enhancement and detection. 
Very recently, Wang~\emph{et al.}~\cite{wang2021hla} established a joint high-low adaptation framework through a bidirectional low-level and multi-task high-level adaptation scheme. But since the intention of unsupervised learning, the performance was unsatisfying.  
Further, Liang~\emph{et al.}~\cite{liang2021recurrent} developed an end-to-end ``detection with enhancement'' network. Although it attempted to enlarge the relationship between enhancement and detection by the joint training strategy, unfortunately, the accuracy improvement was still slight.

Nighttime semantic segmentation is the other representative task in LLVs. The work in~\cite{dai2018dark} provided a progressive mechanism via the bridge of twilight time to adapt to the nighttime. Sakaridis \emph{et al.}~\cite{sakaridis2019guided, sakaridis2020map} developed a curriculum adaptation framework to gradually adapt semantic segmentation models from daytime to nighttime. The paper in~\cite{wu2021dannet} handled nighttime semantic segmentation by designing a multi-target domain adaptation network via adversarial learning. 
Sakaridis \emph{et al.} also created an Adverse Conditions Dataset with Correspondences (ACDC)~\cite{sakaridis2021acdc} for semantic segmentation on adverse visual conditions. Actually, existing techniques mostly concentrate on bridging the source and the target domain. 

In a word, it is still difficult for these recently developed high-level-LLVs-aimed works to significantly improve the performance because of giving up modeling the particularity of low-light scenarios, leading to ignoring the latent correspondence between scene and task and deep exploitation in architecture design for them.


\section{Learning with Nested Scene Modeling}\label{sec:Model}
%

In classical learning methods, one may straightforwardly build a network architecture $\mathbf{x}=\Psi(\mathbf{y};\bm{\omega})$ (with input $\mathbf{y}$, output $\mathbf{x}$ and parameters $\bm{\omega}$) to model the vision task and then train its parameters by solving the optimization model $\min_{\bm{\omega}}\mathcal{L}(\Psi(\mathbf{y};\bm{\omega}))$ (with a loss function $\mathcal{L}$ on collected image set)\footnote{Please notice that here we actually simplify the notation by omitting expectation on the data set.}. However, we argue that these approaches may have difficulty in explicitly formulating complex low-light scene information for establishing network architectures and designing training strategies. So in this section, we propose a new nested learning framework, together with a Retinex-inspired unrolling scheme and an  adaptive training strategy, to  enforce rich priors of low-light scene as explicit constraints for vision tasks in low-light scenarios.  

\subsection{A Scene Energy Constrained Learning Formulation} 

We firstly demonstrate how to formulate LLV tasks from the nested optimization perspective. Specifically, by introducing a latent variable $\mathbf{u}$ to represent our desired features (extracted from the low-light observation $\mathbf{y}$), we actually consider vision tasks in low-light scenarios as the following nested  learning problem:
\begin{eqnarray}
	&&\quad\min\limits_{\bm{\omega}_{\mathtt{t}}} \ell_{\mathtt{t}}\big(\Psi_{\mathtt{t}}(\mathbf{u};\bm{\omega}_{\mathtt{t}})\big),\label{eq:task-objectivce}\\
	&&s.t. \ \mathbf{u}\in\arg\min\limits_{\mathbf{u}}F(\mathbf{u}):=f(\mathbf{u};\mathbf{y}) + g(\mathbf{u}),\label{eq:scene-constraint}
\end{eqnarray}
where $\Psi_{\mathtt{t}}(\mathbf{u};\bm{\omega}_{\mathtt{t}})$ denotes a pre-constructed network module for our considered vision task  (with input $\mathbf{u}$ and parameters $\bm{\omega}_{\mathtt{t}}$), the outer-level objective $\ell_{\mathtt{t}}$ denotes the corresponding training loss and the inner-level objective $F=f+g$ is a low-light scene recovery energy with fidelity $f$ and prior $g$. In fact, the inner-level subprobem in Eq.~\eqref{eq:scene-constraint} just aims to recover the latent feature $\mathbf{u}$ and is embedded in the training process of our considered vision task (formulated in Eq.~\eqref{eq:task-objectivce}). Hereafter, we call the outer- and inner-level subproblems as Task Module (TM) and Scene Module (SM), respectively.

It should be pointed out that due to the nested optimization structure, it is actually challenging to directly solve Eqs.~\eqref{eq:task-objectivce}-\eqref{eq:scene-constraint}. Thus in the following subsections, we firstly introduce an unrolling scheme to construct explicit learnable SM and then design an unsupervised regularization to guide our joint SM-constrained TM training process.


\subsection{Retinex-inspired Unrolling for Training}\label{sec:RUS}

Classic Retinex theory~\cite{fu2015probabilistic,fu2016weighted} assumes that the low-light observation $\mathbf{y}$ can be decomposed into two components, which represent reflectance and illumination, respectively. In general, reflectance makes content in darkness visible and can be derived from the estimation of illumination and the Retinex principle. Inspired by them, we would like to define the latent low-light scene feature $\mathbf{u}$ which reconstructs the underexposed observation $\mathbf{y}$ by removing the illumination component $\mathbf{t}$. 
That is, we enforce $\mathbf{y} = \mathbf{u}\otimes\mathbf{t}$, where $\otimes$ represents the element-wise multiplication. Indeed, our feature component $\mathbf{u}$ should be understood as the generation of reflectance which can be used for a variety of LLV tasks.

Now we formulate our inner-level energy model from the above Retinex-inspired perspective. In particular, we specify  its fidelity term $f$ as the $\ell_2$ reconstruction error $\|\mathbf{u}\otimes\mathbf{t}-\mathbf{y}\|^2$ and consider $g$ as an implicit but learnable prior on $\mathbf{t}$ (i.e., only enforce prior on the illumination-like feature component). Then by adopting the cascaded unrolling scheme proposed in~\cite{liu2019convergence} on the inner-level subproblem in Eq.~\eqref{eq:scene-constraint}, we can numerically transform this nested scene reconstruction model as the following iterative dynamic system, i.e., 
\begin{equation}
	\Psi_{\mathtt{s}}^{k+1}:\left\{
	\begin{aligned}
	\mathbf{t}^{k+1}&={\mathbf{t}}^{k}-\nabla g(\mathbf{t}^{k}), \ (\mbox{with} \ {\mathbf{t}}^0=\max_{z\in\Omega(x)}\mathbf{y}(z)), \\
	\mathbf{u}^{k+1}&=\mathbf{y}\oslash\mathbf{t}^{k+1}, \  k=0,1,\cdots,K-1, \\
	\end{aligned}
	\right.
\end{equation}
where $\oslash$ is the element-wise
division operation, and $\Omega(x)$ denotes  a local region centered at pixel $x$. Here we actually first initialize the illumination-like feature component $\mathbf{t}^0$ using the local information of the observation $\mathbf{y}$, and then propagate $(\mathbf{u}, \mathbf{t})$ based on the inner-level composite objective $F=f+g$. Finally, by parameterizing the gradient of the implicit prior (i.e., $\nabla g$) with a CNN architecture\footnote{The details will be introduced in Sec~\ref{sec:CAS}.} $\mathcal{C}_{\mathtt{s}}$ at each stage, we construct the energy-derived learnable system (i.e., explicit scene constraints) as follows: 
\begin{equation}
	\Psi_{\mathtt{s}}(\mathbf{y};\bm{\omega_{\mathtt{s}}}):=\Psi_{\mathtt{s}}^{K}\circ\cdot\cdot\cdot\circ\Psi_{\mathtt{s}}^{1} \ (\mbox{with parameters} \  \bm{\omega_{\mathtt{s}}}),
\end{equation}
which can successfully embed low-light scene information into the task-specific training process formulated in Eq.~\eqref{eq:task-objectivce}. The computational flow for $\Psi_{\mathtt{s}}$ can be seen in the top row of Fig.~\ref{fig:flowchart} (a). 

Since it has been recognized that enhancing structural information in illumination-like components can improve the visual quality of the low-light scene feature~\cite{guo2017lime,zhang2021beyond},
we consider the following unsupervised scene-oriented objective $\ell_{\mathtt{s}}(\bm{\omega}_{\mathtt{s}})=\|\mathbf{t}^K(\bm{\omega}_{\mathtt{s}})-\mathbf{y}\|^2 + \eta\|\mathbf{t}^K(\bm{\omega}_{\mathtt{s}})\|_{\mathtt{RTV}},$
in which we utilize the so-called relative total variation norm $\|\cdot\|_{\mathtt{RTV}}$ (with a trade-off parameter $\eta$)~\cite{xu2012structure} as our structure-preserving prior on $\mathbf{t}^K$ to constrain the training of $\bm{\omega}_{\mathtt{s}}$. Then we can formulate the general training objective as follows: 
\begin{equation}
	\begin{array}{c}
	\min\limits_{\bm{\omega}:=(\bm{\omega}_{\mathtt{s}},\bm{\omega}_{\mathtt{t}})}  \ell_{\mathtt{t}}(\Psi_{\mathtt{t}}(\mathbf{u}^K(\bm{\omega}_{\mathtt{s}});\bm{\omega}_{\mathtt{t}})) + \lambda\ell_{\mathtt{s}}(\bm{\omega}_{\mathtt{s}}), \\
s.t. \ (\mathbf{u}^K,\mathbf{t}^K)=\Psi_{\mathtt{s}}(\mathbf{y};\bm{\omega}_{\mathtt{s}}),
	\end{array}
\label{eq:task-training}
\end{equation}
where $\lambda\geq0$ represents a balancing parameter.

In this work, we actually consider two kinds of strategies to solve Eq.~\eqref{eq:task-training}, i.e., ``end-to-end" and ``hierarchical". For end-to-end strategy, we just perform training on $\Psi:=\Psi_{\mathtt{t}}\circ\Psi_{\mathtt{s}}$ using $\ell_{\mathtt{t}}+\lambda\ell_{\mathtt{s}}$ (i.e., consider $\Psi$ as a single model when solving Eq.~\eqref{eq:task-training}). In contrast, our hierarchical scheme considers 
$\Psi_{\mathtt{s}}$ (general scene reconstruction problem) as the ``upstream task" and first train it on the given unlabeled dataset by using $\ell_{\mathtt{s}}$. The goal of this step is to learn a model from the dataset itself in an unsupervised manner to provide effective enough scene features. Then we formulate $\Psi$ (specific vision problem) as a ``downstream task" and train $\Psi$ by minimizing $\ell_{\mathtt{t}}$ (with pre-trained $\Psi_{\mathtt{s}}$) on the same dataset used in the previous step but contains the task-specific labels. In this stage, we fine-tune the overall model parameters in $\Psi$ with the labeled dataset. Actually, the training process of the newly-designed hierarchical strategy is a little bit similar to the training idea in self-supervised learning~\cite{jing2020self}. 
As shown in Fig.~\ref{fig:flowchart} (b), we plot the diagram of these two kinds of training strategies for better understanding.

\section{Cooperative Architecture Search}\label{sec:CAS}
In the previous section, we have introduced the learning process for LLVs with low-light scene module. In essence, it bridges the newly-designed scene module and vision task by a two-stage training strategy. Further, the architectures for these two parts are important for deciding the parameter space used in the training process. Indeed, we can design their architecture manually. However, the latent correspondence between scene and task is complex and hard to explicitly present. That is to say, we should not only perform the intrinsic relationship in the parameters learning but also the architecture design. 
Therefore, we introduce a cooperative search strategy to automatically design architectures for our SM and TM in this section. Specifically, we follow similar idea as that in differentiable NAS approaches (e.g.,~\cite{liu2018darts,xu2019pc,liang2019darts+}) to consider computation cells as the building block of our architectures and represent it as directed acyclic graph. We introduce the vectorized form $\bm{\alpha}=\{\bm{\alpha}_{\mathtt{s}},\bm{\alpha}_{\mathtt{t}}\}$ to encode architectures for $\Psi$ and $\Phi$, respectively. In the following, we first design the search space and then propose our cooperative search strategy for $\Psi=(\Psi_{\mathtt{s}},\Psi_{\mathtt{t}})$ in detail.

\begin{table}[t]
	\renewcommand\arraystretch{1.2} 
	\setlength{\tabcolsep}{0.2mm}
	\caption{The overall search space. The rows with the gray shadow represent the search space for SM and low-level TM. The search space for high-level TM contains all the operations except for residual-type convolution and skip connection. }
	\centering
	\begin{tabular}{|c|c|c|c|}
		\hline
		\footnotesize Candidate Operations&\footnotesize Abbreviation& \footnotesize Kernel Size&\footnotesize Dilation Rate\\
		\hline
		 \multirow{4}{*}{\footnotesize Convolution} &\cellcolor{mygray}\footnotesize 1-C&\cellcolor{mygray}\footnotesize 1$\times$1 &\cellcolor{mygray}\footnotesize - \\
		 ~ &\cellcolor{mygray}\footnotesize 3-C&\cellcolor{mygray}\footnotesize 3$\times$3 &\cellcolor{mygray}\footnotesize - \\
		 ~ &\footnotesize 5-C& \footnotesize 5$\times$5 &\footnotesize - \\
		 ~ &\footnotesize 7-C& \footnotesize 7$\times$7 &\footnotesize - \\
		\hline
		 \multirow{2}{*}{\footnotesize Residual Convolution}&\cellcolor{mygray}\footnotesize 1-RC&\cellcolor{mygray}\footnotesize 1$\times$1 &\cellcolor{mygray}\footnotesize - \\
		~&\cellcolor{mygray}\footnotesize 3-RC& \cellcolor{mygray}\footnotesize 3$\times$3 &\cellcolor{mygray}\footnotesize - \\
		\hline
		\multirow{6}{*}{\footnotesize Dilation Convolution}  &\cellcolor{mygray}\footnotesize 3-2-DC&\cellcolor{mygray}\footnotesize 3$\times$3 &\cellcolor{mygray}\footnotesize 2 \\
		~&\footnotesize 3-6-DC& \footnotesize 3$\times$3 &\footnotesize 6 \\
		~&\footnotesize 3-12-DC& \footnotesize 3$\times$3 &\footnotesize 12 \\
		~&\footnotesize 3-18-DC& \footnotesize 3$\times$3 &\footnotesize 18 \\
		 ~&\footnotesize 5-2-DC& \footnotesize 5$\times$5 &\footnotesize 2 \\
		 ~&\footnotesize 7-2-DC& \footnotesize 7$\times$7 &\footnotesize 2 \\
		 \hline
		\footnotesize Residual Dilation Convolution &\cellcolor{mygray} \footnotesize 3-2-RDC&\cellcolor{mygray}\footnotesize 3$\times$3 &\cellcolor{mygray}\footnotesize 2 \\
		\hline
		\footnotesize Skip Convolution &\cellcolor{mygray}\footnotesize SC&\cellcolor{mygray}\footnotesize - &\cellcolor{mygray}\footnotesize - \\
		\hline
	\end{tabular}
	\label{tab:searchspace}
\end{table}

\subsection{Search Space}\label{sec:searchspace}
Here we construct the search spaces for SM and TM. The overall search space can be found in Table~\ref{tab:searchspace} and the bottom row in Fig.~\ref{fig:flowchart} (c). Next, we will introduce them carefully. 

As for scene module, 
we employ the feature distillation techniques~\cite{liu2020residual} to define a distillation cell as $\mathcal{C}_\mathtt{s}$, which is a directed acyclic graph with five nodes and each node connects to the next and the last nodes. In fact, each node in the cell is a latent representation and each direct edge is associated with some operation. The connection to the last node exactly realizes the feature information distillation, to largely reduce parameters. The candidate operations include 
1$\times$1 and 3$\times$3 Convolution (1-C and 3-C), 1$\times$1 and 3$\times$3 Residual Convolution (1-RC and 3-RC), 3$\times$3 Dilation Convolution with dilation rate of 2 (3-2-DC), 3$\times$3 Residual Dilation Convolution with dilation rate of 2 (3-2-RDC), and Skip Connection (SC).

In our constructed framework, the task module is closely associated with the specific low-light vision task. To describe the search space for $\balpha_\mathtt{t}$ clearly, we considered two types of tasks including low-level and high-level LLVs. For the former, task module has the close research target with scene module (i.e., improving the visual quality), so we define the search space of $\balpha_\mathtt{t}$ by using the space of  $\balpha_\mathtt{s}$, to improve the information exploitation. 
As for the high-level LLVs, we define a direct task-oriented cell $\mathcal{C}_\mathtt{t}$\footnote{The concrete form can be found in Sec.~\ref{sec:App}} acquired from the inherent architecture pattern, instead of a newly-designed topological structure. The task-oriented cell in high-level LLVs contains multiple candidate operations including
$m\times m$ ($m\in[1,3,5,7]$) Convolution ($m$-C). $n\times n$ ($n\in[3,5,7]$) Dilation Convolution with dilation rate of $2$ ($n$-2-DC), and $3\times 3$ Dilation Convolution with dilation rate of $k$ (3-$k$-DC, $k\in[6,12,18]$).

\begin{algorithm}[t]
	\caption{Cooperative Architecture Search Strategy}\label{alg:search}
	\begin{algorithmic}[1]
		\REQUIRE 
		The search space $\mathcal{A}$, the training and validation datasets $\mathcal{D}_{\mathtt{tr}}$ and $\mathcal{D}_{\mathtt{val}}$ and necessary parameters.
		\ENSURE The searched architecture. 
		\STATE Initialize $\balpha=\{\balpha_{\ttt},\balpha_{\tn}\}$ and $\bomega=\{\bomega_{\ttt},\bomega_{\tn}\}$.
		\WHILE{not converged}
		\STATE // Update $\balpha_{\ttt}$ and $\bomega_{\ttt}$ for SM.
		\WHILE{not converged}
		\STATE  $\balpha_{\ttt}^{\dagger}\leftarrow\balpha_{\ttt}-\nabla_{\balpha_{\ttt}}\mathcal{L}_{\mathtt{val}}^{\ttt}(\balpha_{\ttt};\bomega_{\ttt}-
		\nabla_{\bomega_{\ttt}}\mathcal{L}_{\mathtt{tr}}^{\ttt})-\beta\nabla_{\balpha_{\ttt}}\mathcal{L}_{\mathtt{val}}^{\tn}(\balpha_{\tn}(\balpha_{\ttt});\bomega_{\tn})$.
		\STATE $\bomega_{\ttt}^{\dagger}\leftarrow\bomega_{\ttt}-\nabla_{\bomega_{\ttt}}\mathcal{L}_{\mathtt{tr}}^{\ttt}(\bomega_{\ttt};\balpha_{\ttt}^{\dagger})$.
		\ENDWHILE
		\STATE // Update $\balpha_{\tn}$ and $\bomega_{\tn}$ for TM.
		\WHILE{not converged}
		\STATE  $\balpha_{\tn}^{\dagger}\leftarrow\balpha_{\tn}-\nabla_{\balpha_{\tn}}\mathcal{L}_{\mathtt{val}}^{\tn}(\balpha_{\tn}(\balpha_{\ttt}^{\dagger});\bomega_{\tn}-
		\nabla_{\bomega_{\tn}}\mathcal{L}_{\mathtt{tr}}^{\tn})$.
		\STATE $\bomega_{\tn}^{\dagger}\leftarrow\bomega_{\tn}-\nabla_{\bomega_{\tn}}\mathcal{L}_{\mathtt{tr}}^{\tn}(\bomega_{\tn};\balpha_{\tn}^{\dagger})$.
		\ENDWHILE
		\ENDWHILE
		\RETURN  Architecture derived based on $\balpha_{\ttt}^*$ and $\balpha_{\tn}^*$.
	\end{algorithmic}
\end{algorithm}

\begin{figure*}[t]
	\centering
	\begin{tabular}{c@{\extracolsep{0.37em}}c@{\extracolsep{0.15em}}c@{\extracolsep{0.37em}}c@{\extracolsep{0.15em}}c@{\extracolsep{0.37em}}c@{\extracolsep{0.15em}}c}
		\includegraphics[width=0.136\linewidth]{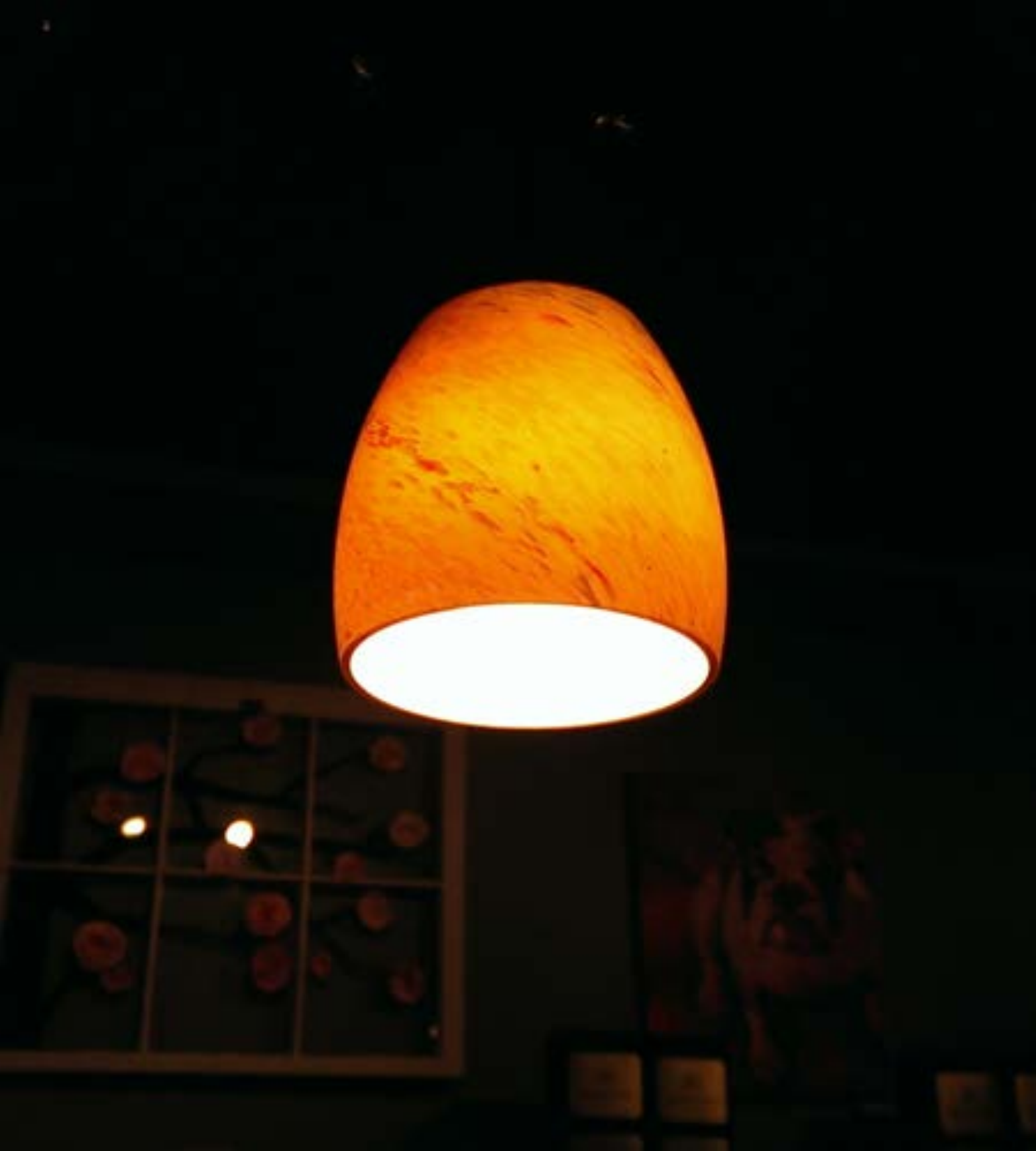}&
		\includegraphics[width=0.136\linewidth]{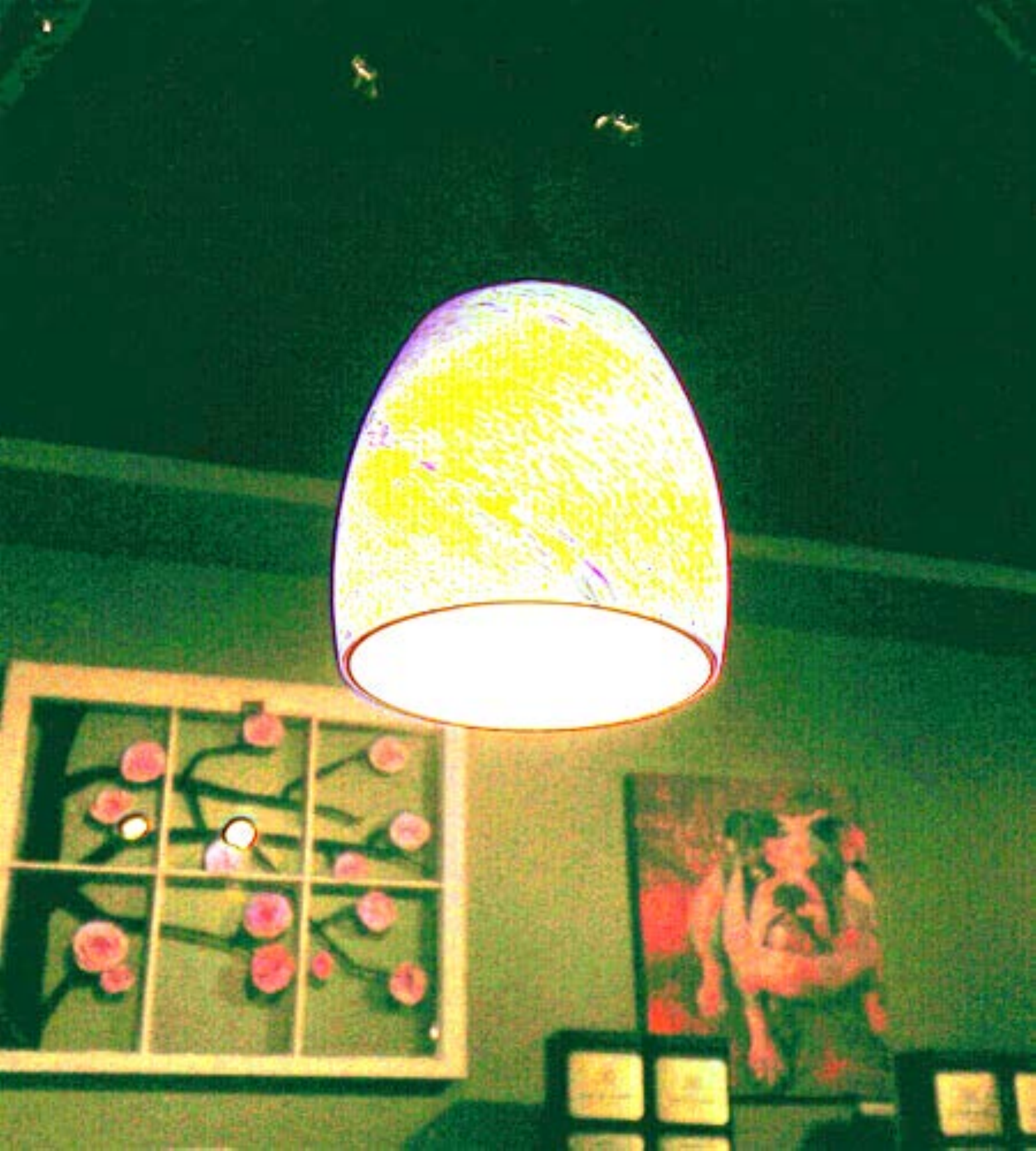}&
		\includegraphics[width=0.136\linewidth]{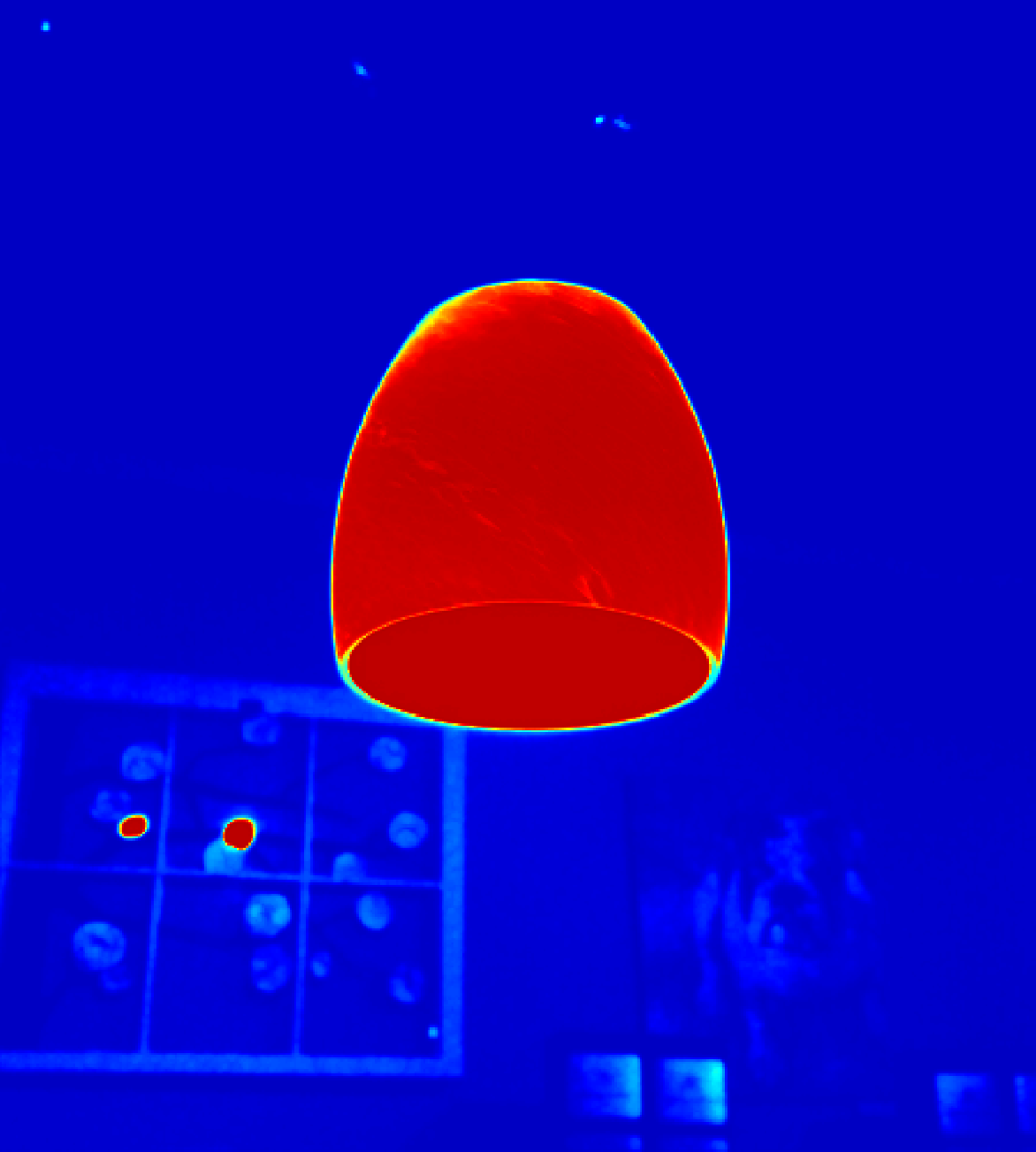}&
		\includegraphics[width=0.136\linewidth]{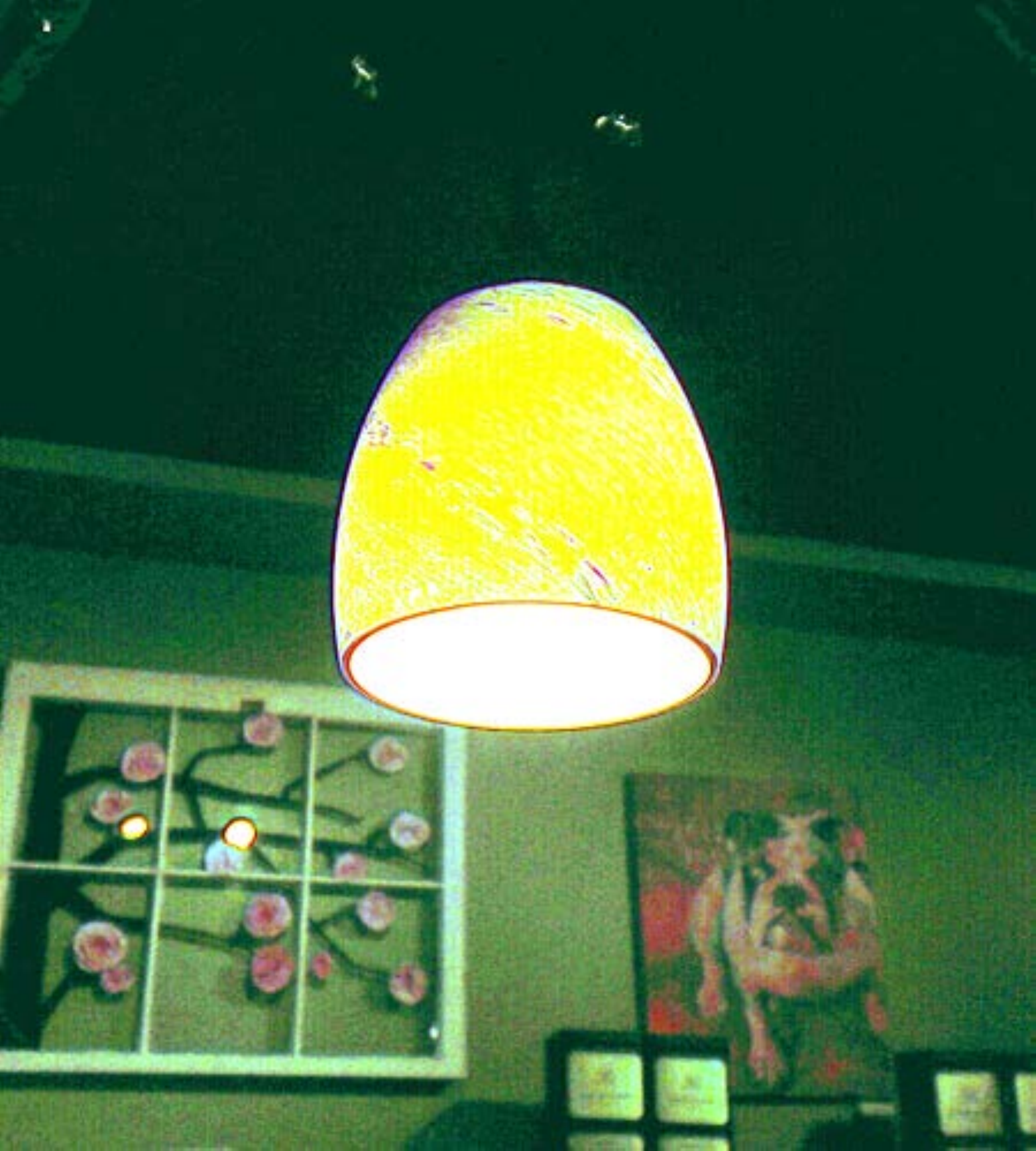}&
		\includegraphics[width=0.136\linewidth]{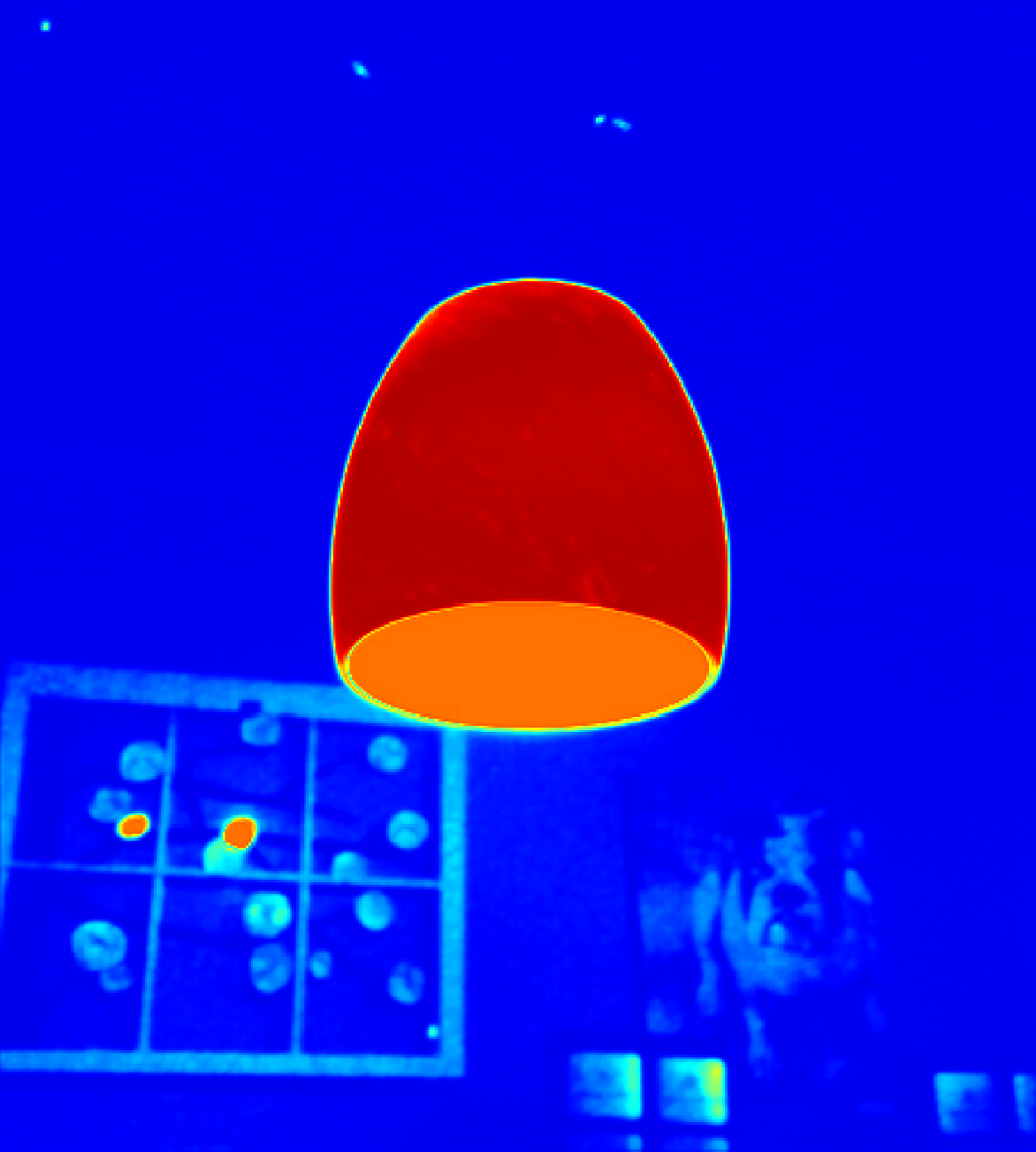}&
		\includegraphics[width=0.136\linewidth]{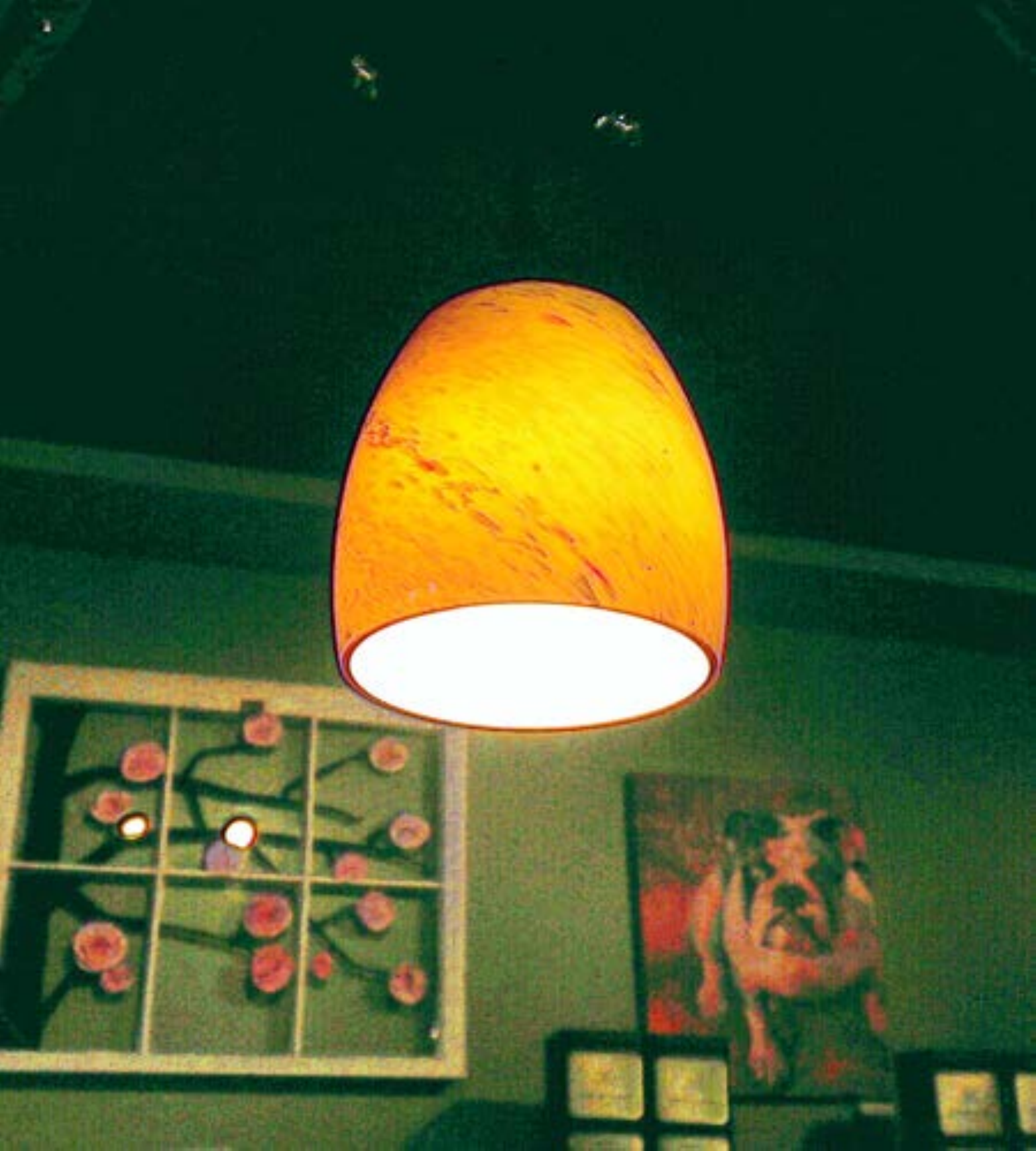}&
		\includegraphics[width=0.136\linewidth]{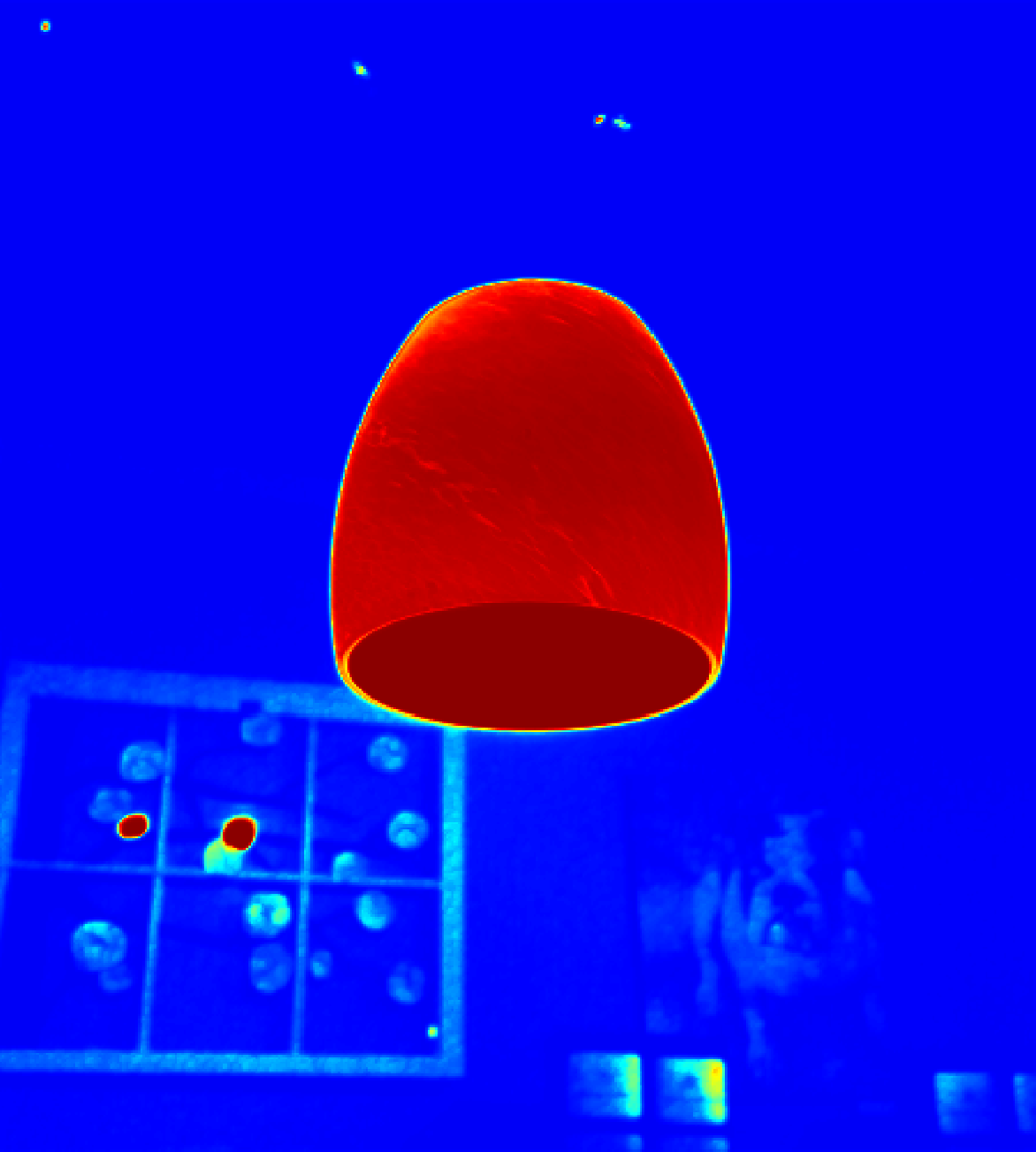}\\
		\footnotesize (a) Input &	\multicolumn{2}{c}{\footnotesize (b) Fix warm-start as $\hat{\mathbf{u}}_\mathtt{r}^{0}$}&\multicolumn{2}{c}{\footnotesize (c) Update $\hat{\mathbf{u}}_\mathtt{r}^{k}$ w/o rectification} &	\multicolumn{2}{c}{\footnotesize (d) Update $\hat{\mathbf{u}}_\mathtt{r}^{k}$ w/ rectification}\\
	\end{tabular}
	\caption{Ablation study of the effect of different warm-start strategies on low-level LLV. Subfigures (b)-(d) plot the enhanced results and corresponding estimated illumination maps for different settings. }
	\label{fig: teffect}
\end{figure*}

\subsection{Search Strategy}
Different from classical differential methods that learn $\bomega$ and $\balpha$ in an end-to-end fashion by directly applying gradient-based NAS techniques, which completely ignore the intrinsic relationship between $\balpha_{\ttt}$ and $\balpha_{\tn}$ that is our core focus. Concretely, the intrinsic relationship is that, $\balpha_{\ttt}$ decides the parameter space for presenting the SM, to further decide the solution space of the generated result of the SM. Likewise, there also exist this connection between $\balpha_{\tn}$ and the output of TM. As mentioned above, the outputs of SM and TM exist the latent correspondence, while $\balpha_{\ttt}$ and $\balpha_{\tn}$ indirectly decide these outputs. Therefore, as demonstrated in the top row in Fig.~\ref{fig:flowchart} (c), they should be defined in a more interrelated manner, rather than naive or simple strategy. 

To this end, we propose to search architectures for  ${\bm{\alpha}_{\ttt}}$ and ${\bm{\alpha}_{\tn}}$ cooperatively. We aim to solve the following model by formulating the search process of TM and SM as a cooperative game
\begin{equation}
	\min\limits_{{\bm{\alpha}_{\tn}}\in\mathcal{A}}\left\{\min\limits_{{\bm{\alpha}_{\ttt}}\in\mathcal{A}}
	\mathcal{L}_{\mathtt{val}}(\bm{\alpha}_{\ttt},\bm{\alpha}_{\tn};\bomega_{\ttt}^*,\bomega_{\tn}^*)\right\},\label{eq:loss_val}
\end{equation}
where we denote $\mathcal{L}_{\mathtt{val}}$ as a cooperative loss on the validation dataset, i.e., 
\begin{equation}
	\mathcal{L}_{\mathtt{val}}:=\mathcal{L}^{\mathtt{s}}_{\mathtt{val}}(\bm{\alpha}_{\mathtt{s}};\bm{\omega}_{\mathtt{t}}^{*})+\beta\mathcal{L}^{\mathtt{t}}_{{\mathtt{val}}}(\bm{\alpha}_{\mathtt{t}}(\bm{\alpha}_{\mathtt{s}});\bm{\omega}_{\mathtt{s}}^{*}),
\end{equation} 
where $\beta\geq 0$ is a trade-off parameter, $\mathcal{L}^{\ttt}_{\mathtt{val}}$ and $\mathcal{L}^{\tn}_{{\mathtt{val}}}$ denote the losses on task and scene module, respectively. Note that here we just reuse the structure-preserving loss in Eq.~\eqref{eq:task-training} as $\mathcal{L}^{\ttt}_{{\mathtt{val}}}$. Since TM is defined based on the output of SM, we should also consider $\balpha_{\ttt}$ as parameters of $\balpha_{\tn}$ in $\mathcal{L}^{\tn}_{{\mathtt{val}}}$. 
In fact, by analogy with the generative adversarial learning task~\cite{goodfellow2014generative}, it should be understood that the optimization problem in Eq.~\eqref{eq:loss_val} actually considers a cooperative (``min-min''), rather than an adversarial (``min-max'') objective.


As for $\bomega_{\ttt}^*$ (and $\bomega_{\tn}^*$), we assume that they are only associated with the architecture  $\balpha_{\ttt}$ (and $\balpha_{\tn}$). That is, they can be obtained by minimizing the following models 
\begin{equation}
	\left\{\begin{array}{l}
		\bomega_{\ttt}^*=\arg\min\limits_{\bomega_{\ttt}}\mathcal{L}_{\mathtt{tr}}^{\ttt}(\bomega_{\ttt};\balpha_{\ttt}),\\
		\bomega_{\tn}^*=\arg\min\limits_{\bomega_{\tn}}\mathcal{L}_{\mathtt{tr}}^{\tn}(\bomega_{\tn};\balpha_{\tn}),
	\end{array}\right.\label{eq:loss_tr}
\end{equation}
where $\mathcal{L}_{\mathtt{tr}}^{\ttt}$ and $\mathcal{L}_{\mathtt{tr}}^{\tn}$ are the training losses for task and scene modules, respectively. 

Alg.~\ref{alg:search} summarizes the overall search process\footnote{Here we skip some regular numerical parameters (e.g., initialization, learning rate and stopping criterion) to simplify our notations.}. In particular, we update $\balpha_{\ttt}$ (with the current $\balpha_{\tn}$) for SM (Steps 4-7) and update $\balpha_{\tn}$ (based on the updated $\balpha_{\ttt}$) for TM (Steps 9-12). As for each part, we just adopt the widely used one-step finite difference technique~\cite{forsythe1960finite} to approximately calculate gradients for the upper-level variables (Steps 5 and 10).

\begin{figure}[t]
	\centering
	\begin{tabular}{c@{\extracolsep{0.3em}}c@{\extracolsep{0.3em}}c@{\extracolsep{0.3em}}c} 
		\includegraphics[width=0.233\linewidth]{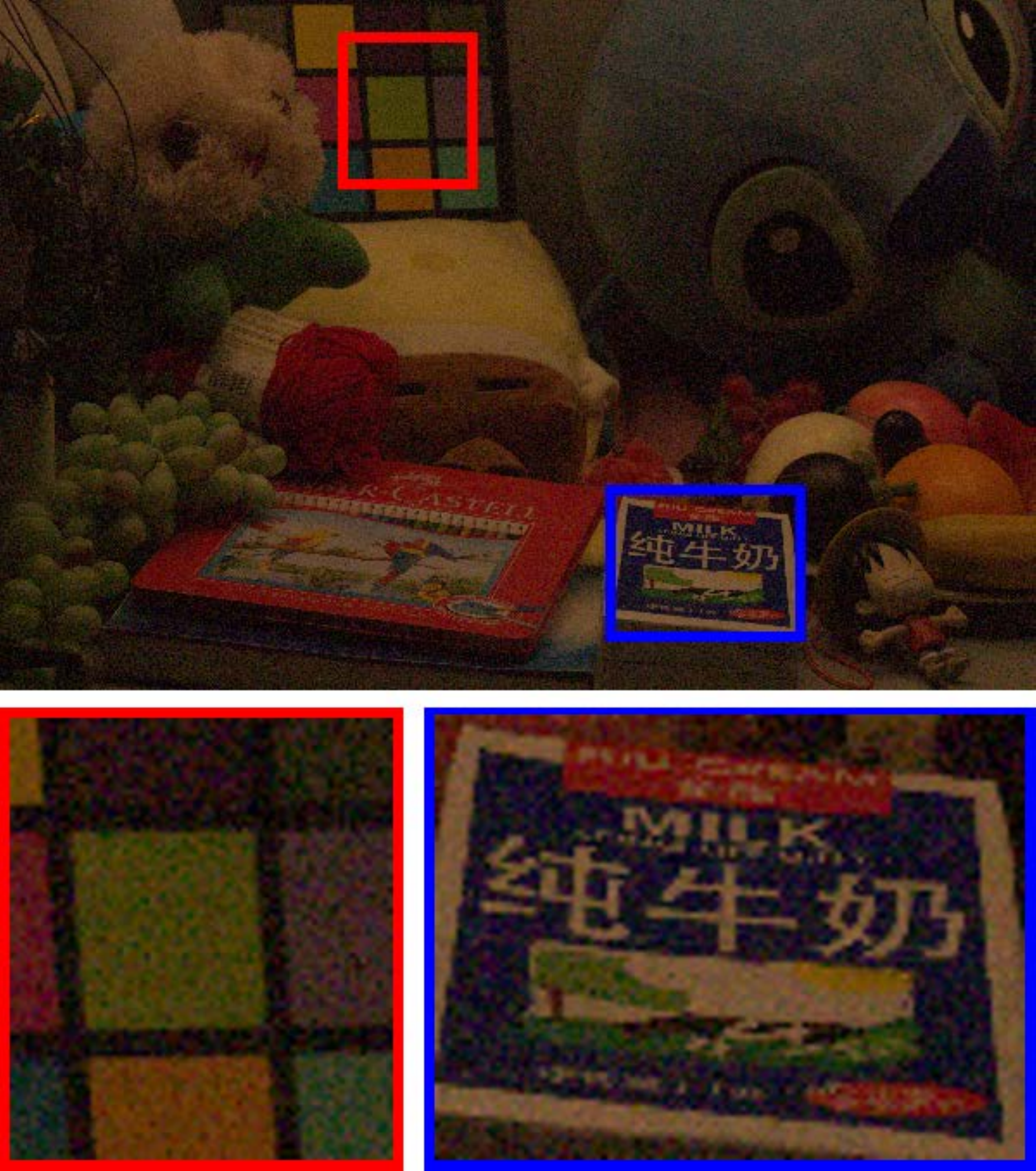}&	
		\includegraphics[width=0.233\linewidth]{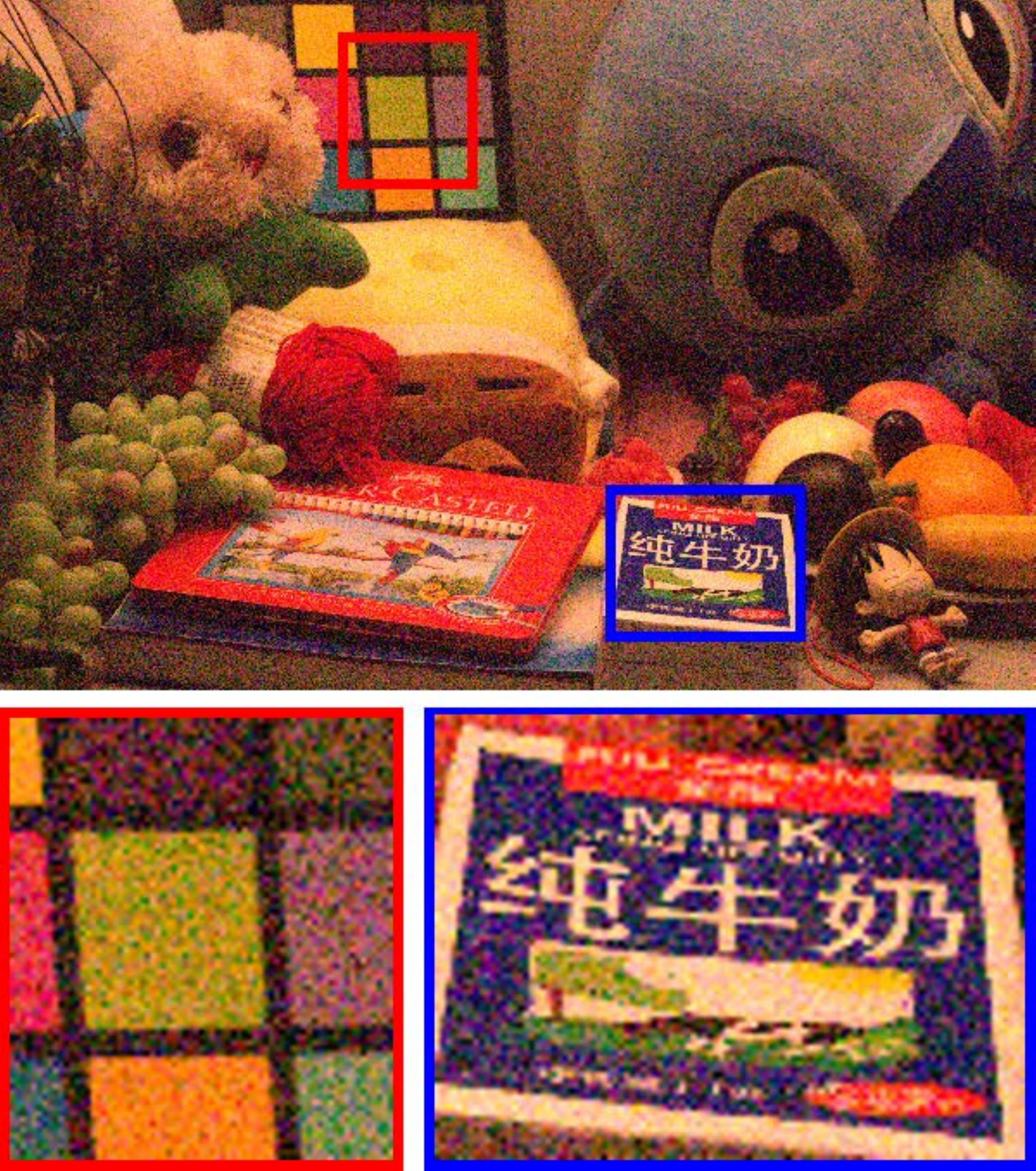}&
		\includegraphics[width=0.233\linewidth]{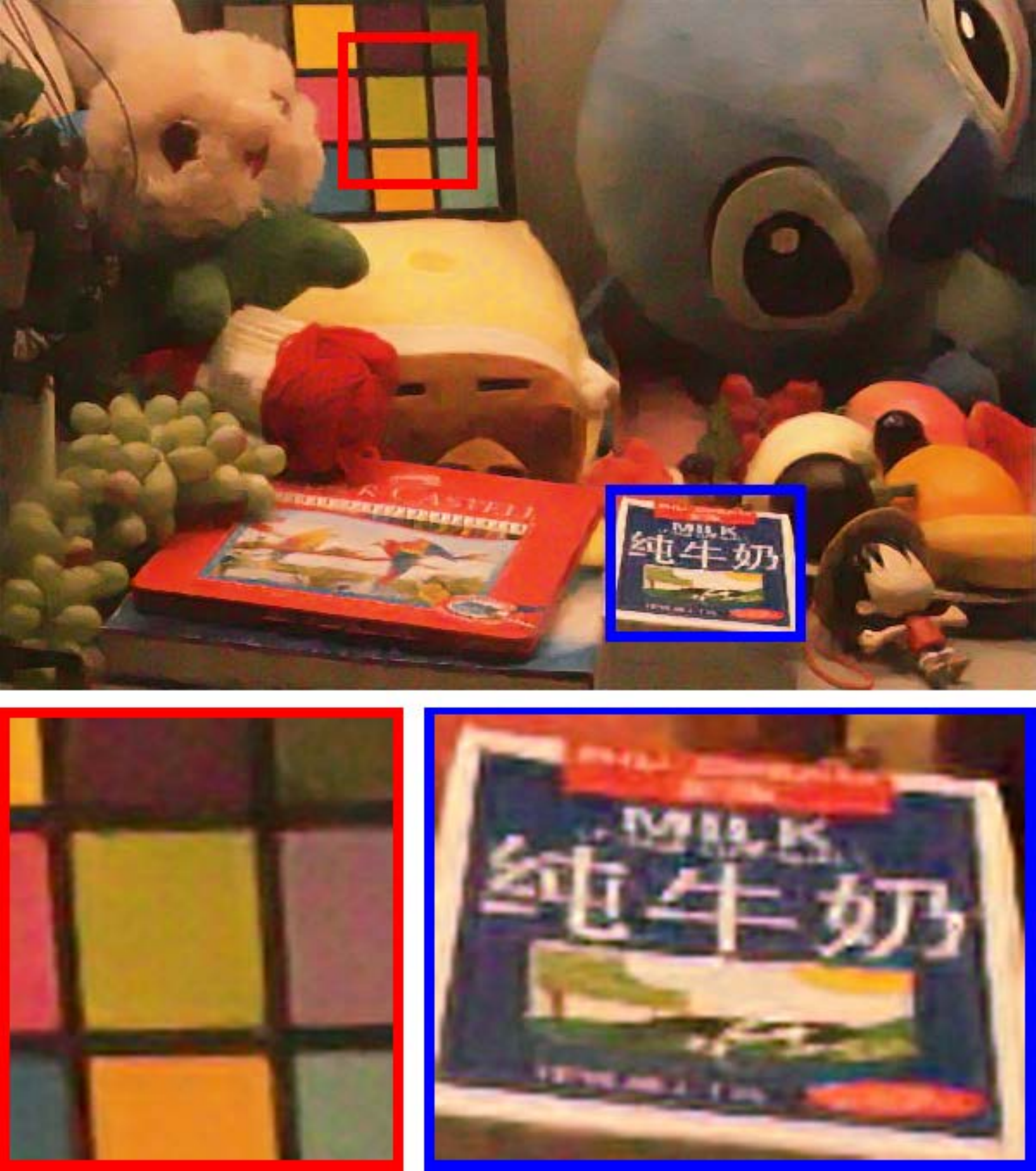}&
		\includegraphics[width=0.233\linewidth]{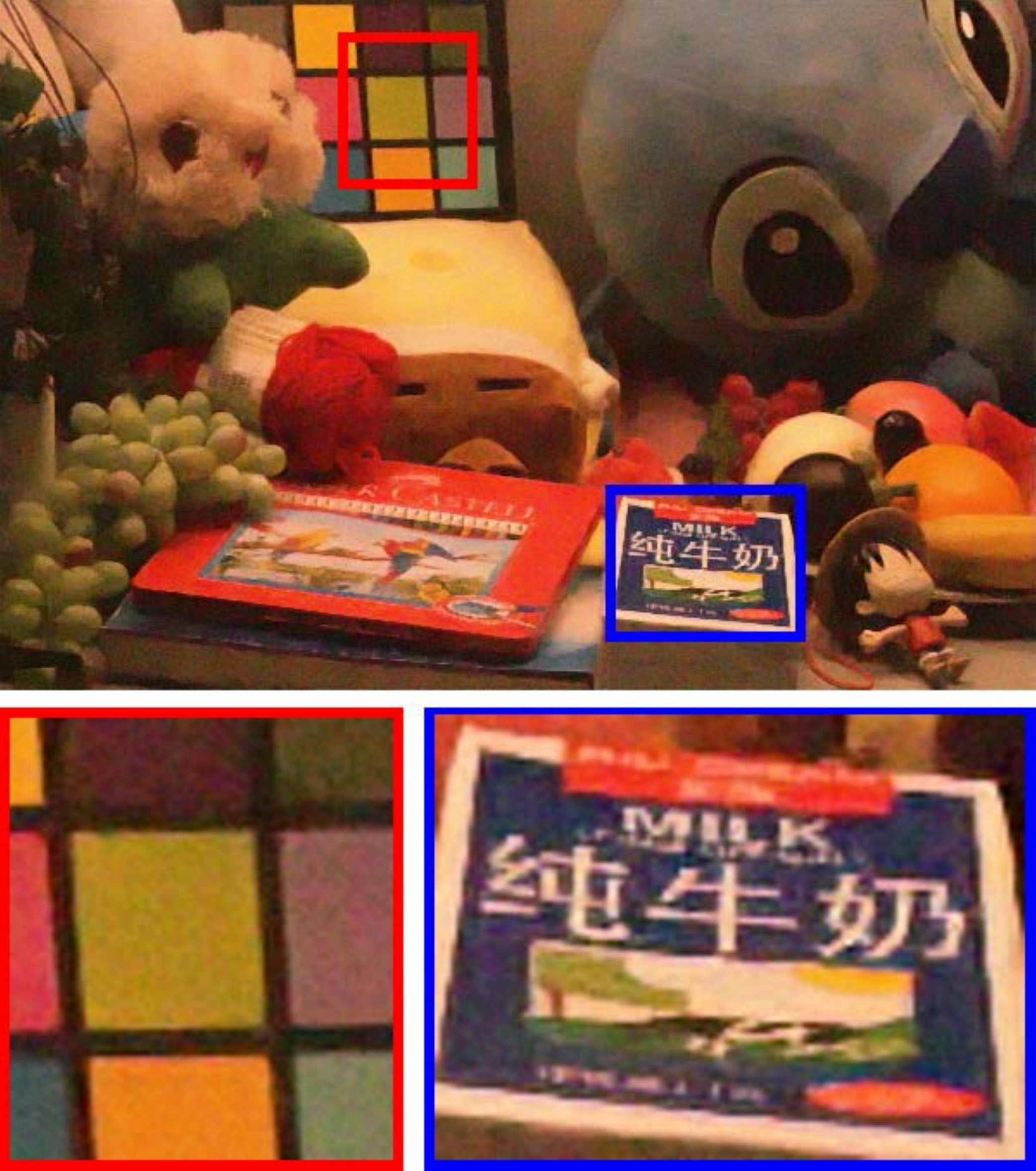}\\
		\footnotesize PSNR/SSIM&\footnotesize 16.348/0.510&\footnotesize 21.156/0.860&\footnotesize 22.281/0.850\\
		\includegraphics[width=0.233\linewidth]{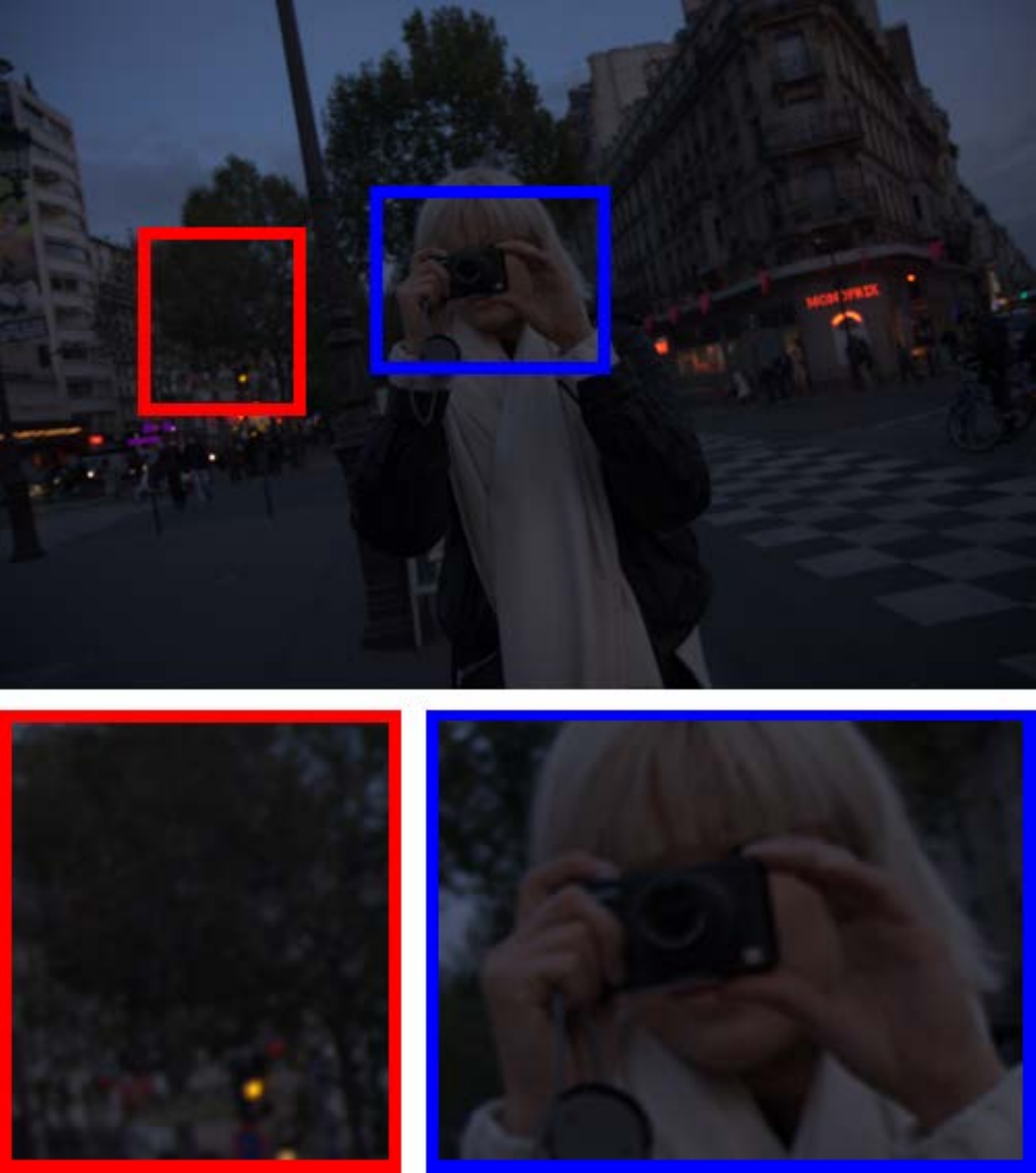}&	
		\includegraphics[width=0.233\linewidth]{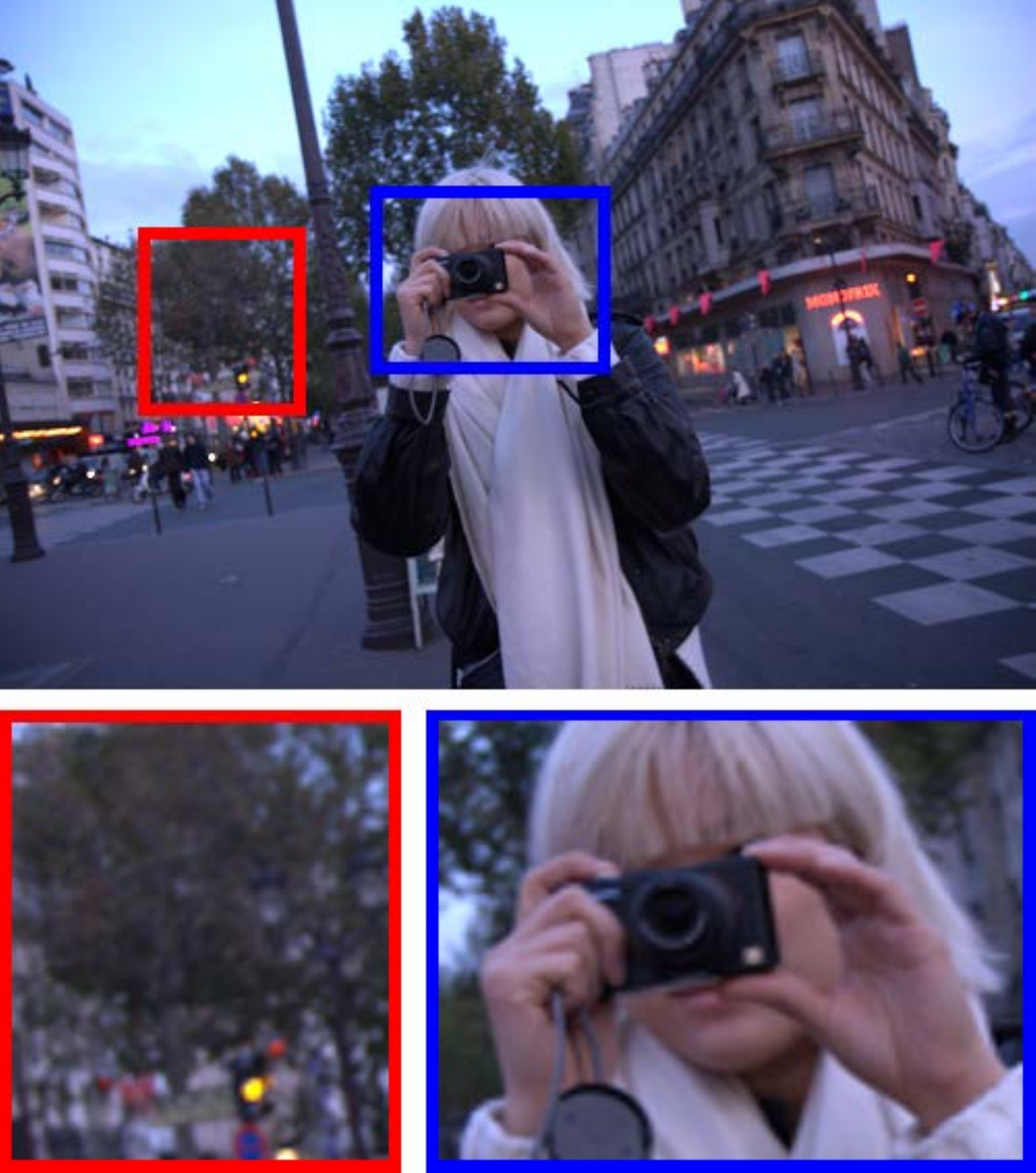}&
		\includegraphics[width=0.233\linewidth]{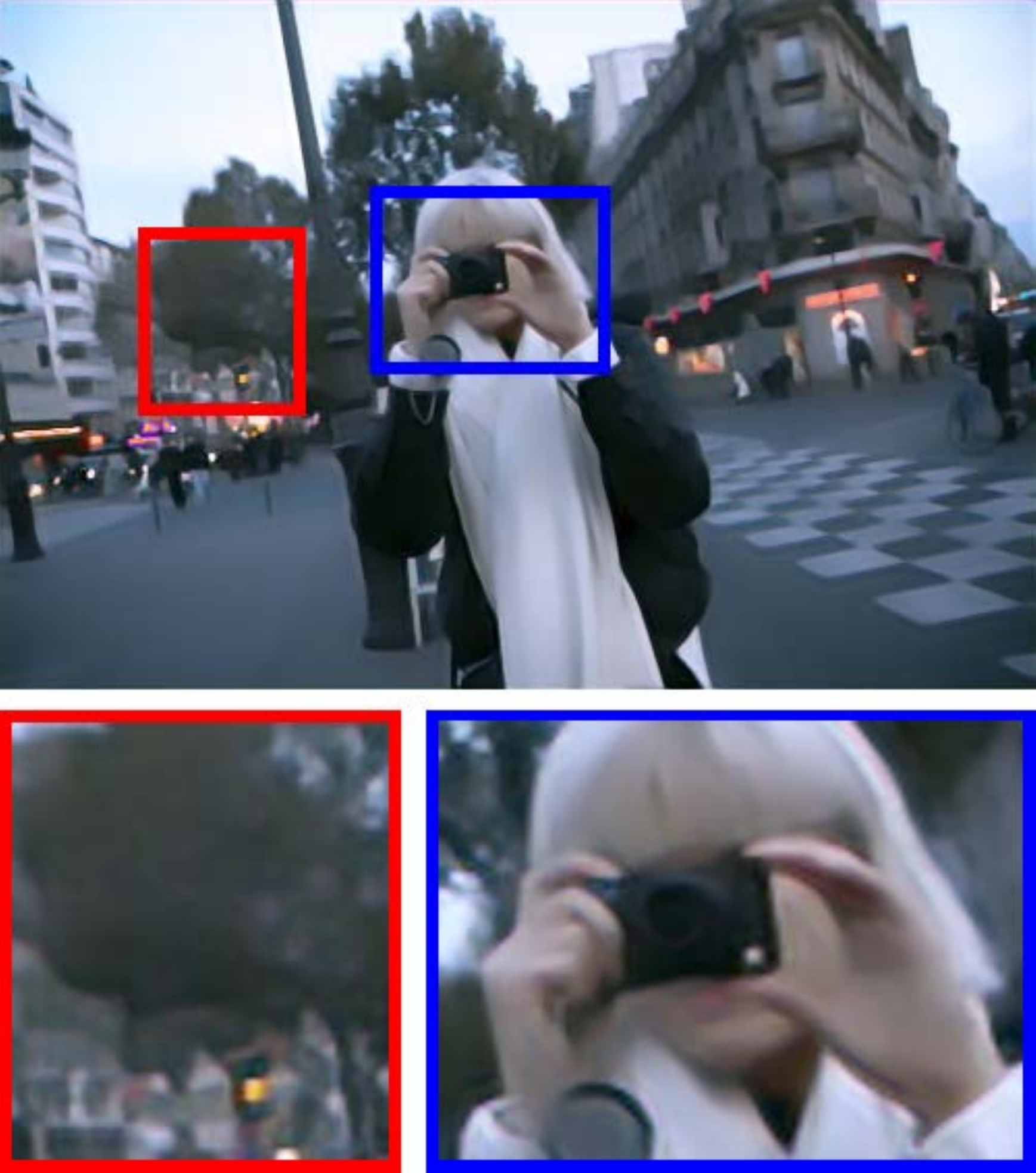}&
		\includegraphics[width=0.233\linewidth]{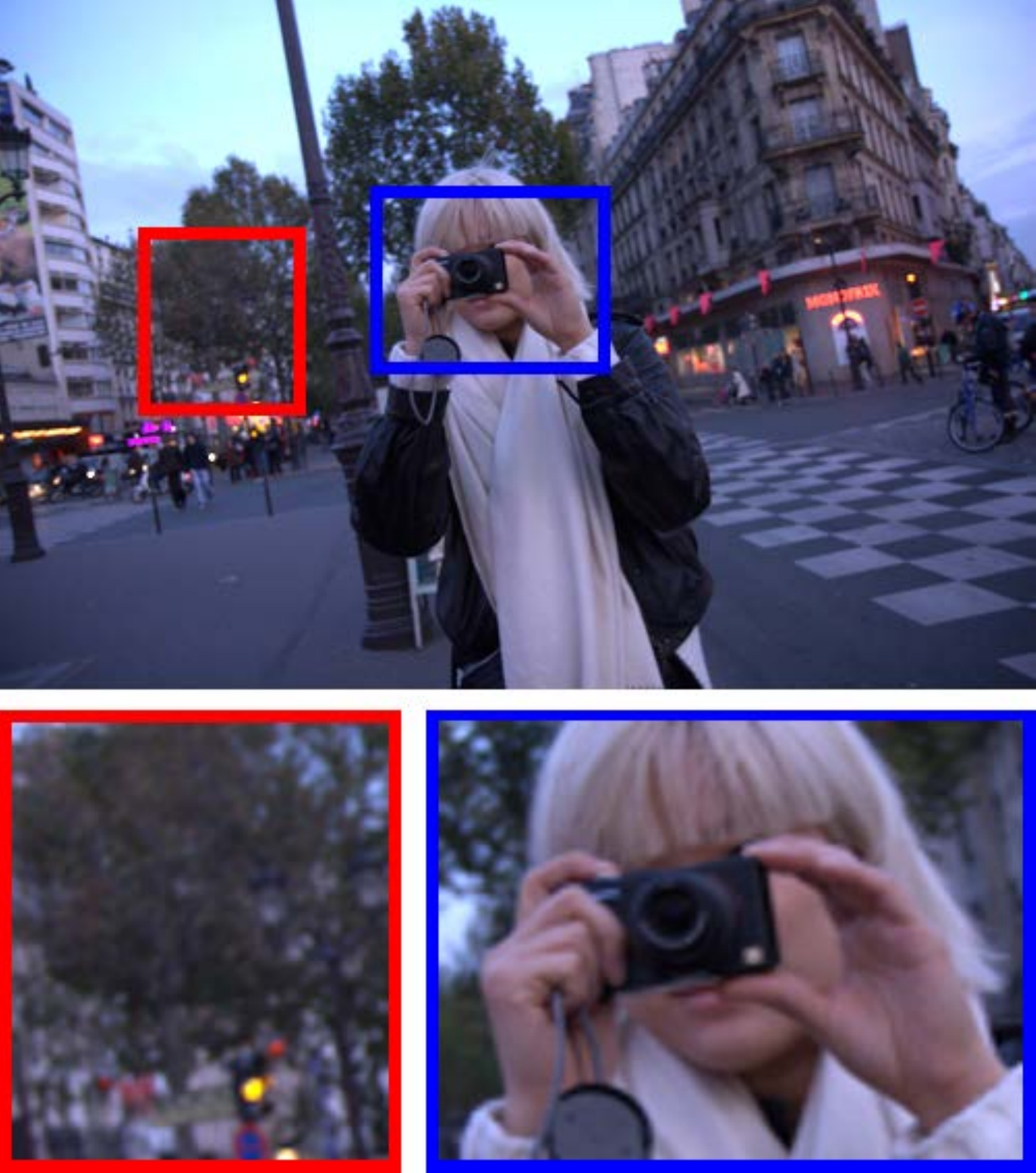}\\
		\footnotesize PSNR/SSIM&\footnotesize  25.144/0.929 &\footnotesize 19.707/0.775 &\footnotesize 25.144/0.929\\
		\footnotesize Input&\footnotesize RUAS$_\mathtt{S}$&\footnotesize RUAS&\footnotesize RUAS$_\mathtt{A}$\\
	\end{tabular}
	\caption{Visual results among different versions of RUAS.  }
	\label{fig:Versions}
\end{figure}

\section{Applications for Various LLVs}\label{sec:App}
In this section, we detail how to apply our constructed algorithm to enhance visual quality in low-light scenarios. We also make an extension for low-light scene perception. The overall architecture for task module can be found in the bottom row of Fig.~\ref{fig:flowchart} (a). 

\subsection{Enhancing Visual Quality in Low-light Scenario}\label{sec:lowlevel}

Enhancing visual quality of low-light images is one of the most important LLV task. Now we demonstrate how to address this problem within our RUAS framework.
Specifically, we design two components as our TM in this task, i.e., noise estimation and removal modules.  

\textbf{Noise estimation module.} The goal of introducing this part is to endow the adaptation ability towards different scenarios, preventing degrading the visual quality after performing the noise removal module, e.g., the case without noises. This module can be formulated as $\bm{\theta}=\Psi_{\mathtt{e}}(\mathbf{u}^{K};\bm{\omega}_{\mathtt{e}})$, where $\bm{\theta}$ denotes the estimated noise map, and $\Psi_{\mathbf{e}}$ represents the noise estimation network (with learnable parameters $\bm{\omega_{\mathtt{e}}}$). Here we adopt a simple network architecture for $\Psi_{\mathbf{e}}$,  it is composed of five layers of convolution. 

\textbf{Noise removal module.} After acquiring the noise map $\bm{\theta}$, we define the noise removal as $\mathbf{x}=\Psi_{\mathtt{r}}(\mathbf{u}^{K},\bm{\theta};\bm{\omega}_{\mathtt{r}})$, where $\Psi_{\mathbf{r}}$ represents the noise removal network (with learnable parameters $\bm{\omega_{\mathtt{r}}}$). Actually, this part is closely associated with the noise estimation. That is to say, if $\mathbf{u}$ exist noises that are determined by the noise estimation module, we can perform this part. Otherwise, noise removal module is prohibited.  
Note that the architecture for this module is constructed by the aforementioned search process, and the search space for $\Psi_{\mathbf{r}}$ is completely identical to SM. 

As for connecting these two task-related modules, we first perform $\mathbf{u}^K=\Psi_{\mathtt{s}}(\mathbf{y};\bm{\omega}_{\mathtt{s}})$ to generate the result $\mathbf{u}^{K}$ with brightness enhancement. Then we train $\bm{\omega}_{\mathtt{e}}$ based on $\mathbf{u}^K$ to obtain $\bm{\theta}^*=\Psi_{\mathtt{e}}(\mathbf{u}^K;\bm{\omega}_{\mathtt{e}}^*)$. Finally, we train $\Psi(\mathbf{y},\bm{\theta}^*;\bm{\omega}_{\mathtt{r}},\bm{\omega}_{\mathtt{s}}):=\Psi_{\mathtt{r}}\circ\Psi_{\mathtt{s}}$ with the estimated noise level $\bm{\theta}^*$. If $\|\bm{\theta}^*\|_1\leq\epsilon$, we may directly ignore the noise removal module $\Psi_{\mathtt{r}}$.
Moreover, we define $\hat{\mathbf{t}}^{k}=\max_{z\in\Omega(x)}{\mathbf{t}^{k}}(z)-\gamma\mathbf{r}_k$, where $\mathbf{r}^k=\mathbf{u}^k-\mathbf{y}$ (with penalty parameter $0<\gamma\leq1$) represents the residual rectification. It is used to reformulate the update process of $\mathbf{t}$, described as $\mathbf{t}^{k+1}=\hat{\mathbf{t}}^{k}-\nabla g(\hat{\mathbf{t}}^{k}), \ (\mbox{with} \ {\mathbf{t}}^0=\max_{z\in\Omega(x)}\mathbf{y}(z)),$
The principle behind the term is that the illumination is at least the maximal value of a certain location and can be used to handle non-uniform illuminations. As for the residual rectification, we actually introduce this term to adaptively suppress some overexposed pixels for $\hat{\mathbf{t}}^{k}$ during the propagation\footnote{The analytical experiment can be found in Fig.~\ref{fig: teffect}.}. 
Note that we define three versions for low-light image enhancement to fully reveal the effectiveness of our method, including RUAS$_\mathtt{S}$ (only scene module), RUAS (without noise estimation module), and RUAS$_\mathtt{A}$ (the full version).

\begin{table}[t]
	\renewcommand\arraystretch{1.2} 
	\setlength{\tabcolsep}{3.2mm}
	\caption{Quantitative results among different numbers of $K$ on the MIT-Adobe 5K dataset. }
	\centering
	\begin{tabular}{|c|c|c|c|c|c|}
		\hline
		\footnotesize Model&\footnotesize $K=1$&\footnotesize $K=2$ &\footnotesize $K=3$&\footnotesize $K=4$&\footnotesize $K=5$\\
		\hline
		\footnotesize PSNR&\footnotesize  13.013&\footnotesize 20.926&\footnotesize 21.018&\footnotesize  20.857&\footnotesize  20.846\\ 
		\hline
		\footnotesize SSIM&\footnotesize  0.772&\footnotesize 0.823&\footnotesize 0.855&\footnotesize  0.826&\footnotesize  0.828\\ 
		\hline
		\footnotesize LPIPS&\footnotesize  0.205&\footnotesize 0.172&\footnotesize 0.139&\footnotesize  0.172&\footnotesize  0.162\\ 
		\hline
	\end{tabular}
	\label{tab:Stage}
\end{table}

\begin{table}[t]
	\renewcommand\arraystretch{1.2} 
	\setlength{\tabcolsep}{3.3mm}
	\caption{Quantitative results between naively determined architectures (with supernet and single type of convolution) and our searched RUAS on MIT-Adobe 5K dataset. }
	\centering
	\begin{tabular}{|c|c|c|c|c|c|}
		\hline
		\footnotesize Model&\footnotesize PSNR&\footnotesize SSIM &\footnotesize SIZE&\footnotesize FLOPs&\footnotesize TIME\\
		\hline
		\footnotesize Supernet&\footnotesize  18.940&\footnotesize 0.819&\footnotesize 0.038&\footnotesize  9.113&\footnotesize  0.136\\ 
		\hline
		\footnotesize 1-C&\footnotesize  18.829&\footnotesize 0.816&\footnotesize 0.002&\footnotesize  0.495&\footnotesize  0.009\\ 
		\hline
		\footnotesize 3-C&\footnotesize  19.107&\footnotesize 0.823&\footnotesize 0.010&\footnotesize  2.309&\footnotesize 0.010 \\ 
		\hline
		\footnotesize 1-RC&\footnotesize  19.151&\footnotesize 0.823&\footnotesize 0.002&\footnotesize  0.495&\footnotesize  0.012\\ 
		\hline
		\footnotesize 3-RC&\footnotesize  19.307&\footnotesize 0.810&\footnotesize 0.010&\footnotesize  2.309&\footnotesize  0.010\\ 
		\hline
		\footnotesize 3-2-DC&\footnotesize  19.154&\footnotesize 0.833&\footnotesize 0.010&\footnotesize  2.309&\footnotesize  0.010\\ 
		\hline
		\footnotesize 3-2-RDC&\footnotesize  19.391&\footnotesize 0.825&\footnotesize 0.010&\footnotesize  2.309&\footnotesize 0.015 \\ 
		\hline
		\footnotesize RUAS&\footnotesize 21.018&\footnotesize 0.855&\footnotesize 0.003&\footnotesize  0.832&\footnotesize  0.009\\ 
		\hline
	\end{tabular}
	\label{tab:SceneModelingSearch}
\end{table}

\begin{table}[t]
	\renewcommand\arraystretch{1.2} 
	\setlength{\tabcolsep}{0.4mm}
	\caption{Quantitative results of different search strategy. }
	\centering
	\begin{tabular}{|c|c|c|c|}
		\hline
		\footnotesize Strategy&\footnotesize Global search&\footnotesize Independent search &\footnotesize Cooperative search\\
		\hline
		\footnotesize PSNR/SSIM&\footnotesize  14.496/0.543&\footnotesize  15.841/0.584&\footnotesize  18.226/0.717\\ 
		\hline
		\footnotesize SIZE/FLOPs&\footnotesize  1.001G/0.004M&\footnotesize  1.220G/0.005M&\footnotesize 0.832G/0.003M\\ 
		\hline
	\end{tabular}
	\label{tab:SearchStrategy}
\end{table}

\begin{figure}[!htb]
	\centering
	\begin{tabular}{c}
		\includegraphics[width=0.96\linewidth]{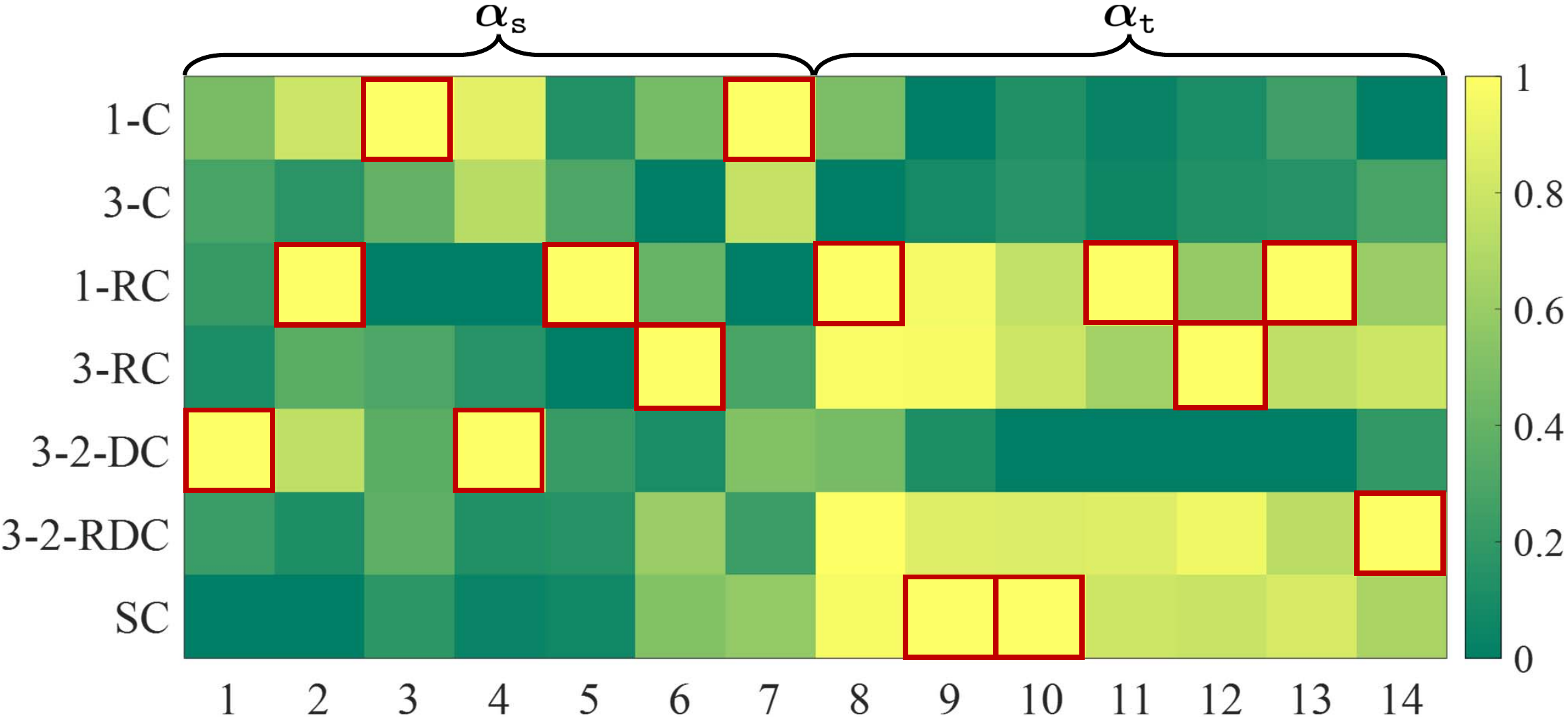}\\
		{\footnotesize (a) Global search}\\
		\includegraphics[width=0.96\linewidth]{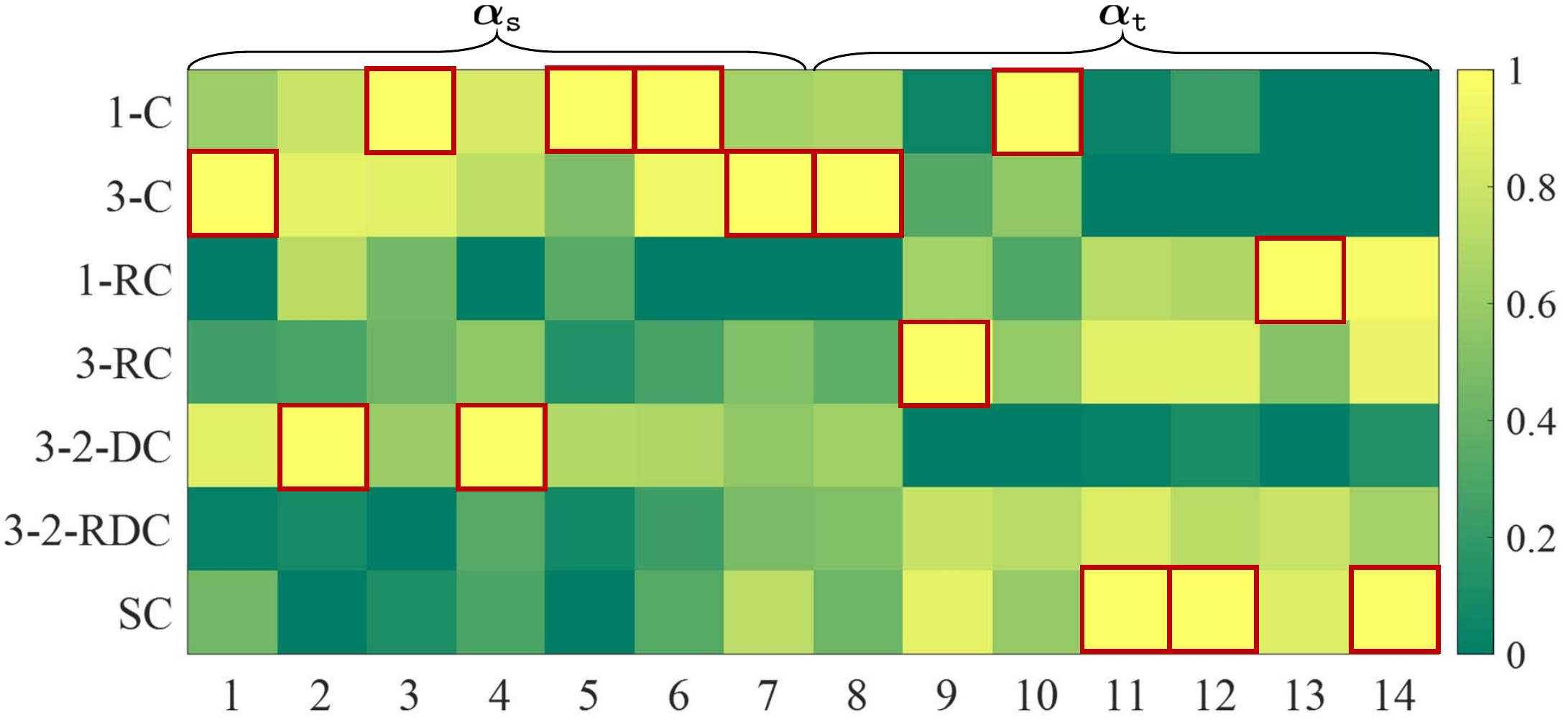}\\
		{\footnotesize (b) Independent search}\\
		\includegraphics[width=0.96\linewidth]{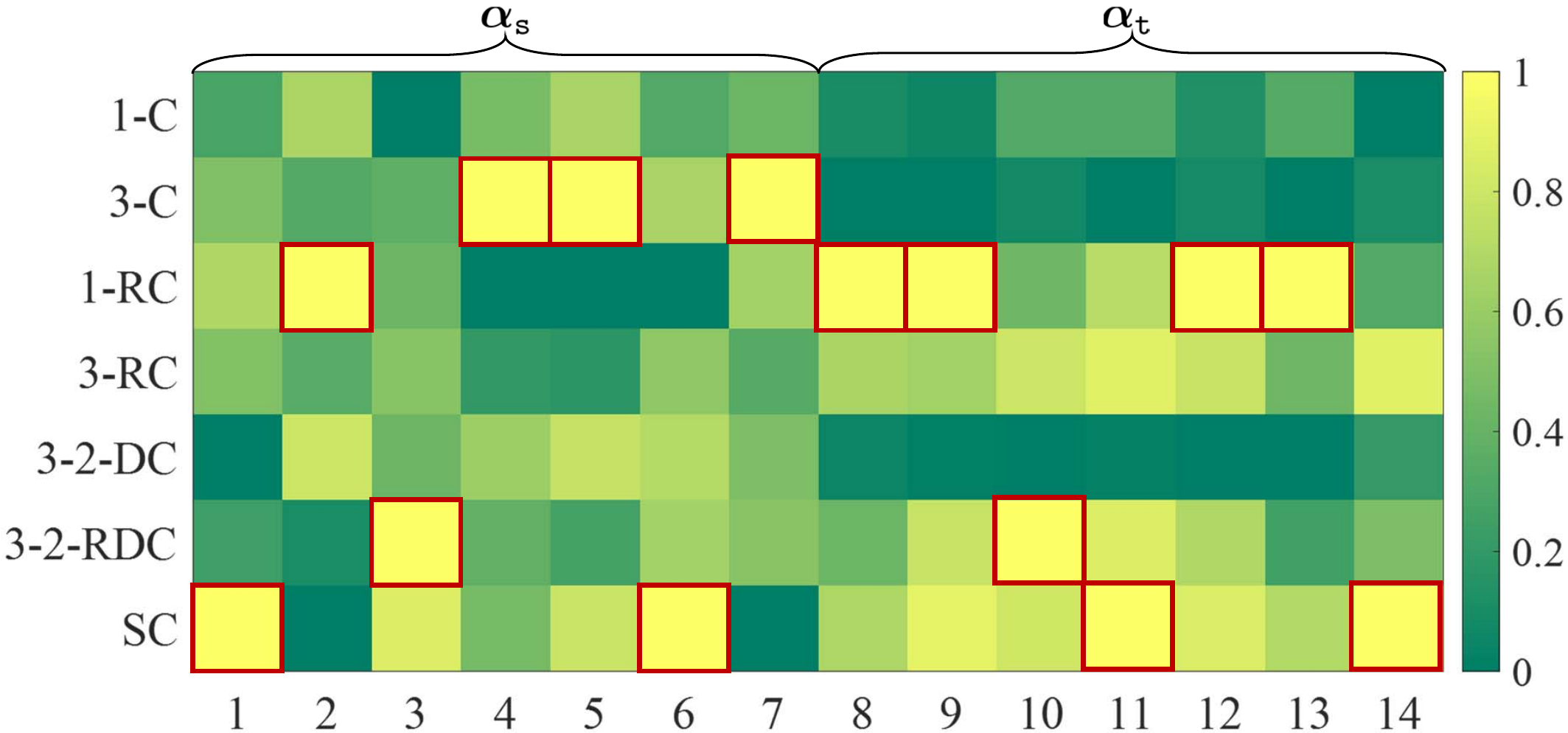}\\
		{\footnotesize (c) Cooperative search}\\\
	\end{tabular}	
	\caption{The searched cells of different search strategies in low-light image enhancement. Since we share cell for all the stages, thus only one cell is plotted. The top and bottom rows are the searched architectures of $\balpha_{\ttt}$ and $\balpha_{\tn}$, respectively.} 
	\label{fig: SearchStrategy}
\end{figure}

%
%
%
%

\subsection{Extension for Low-Light Scene Perception}\label{sec:highlevel}

Now we demonstrate how to extend RUAS to address high-level LLV tasks, such as object detection and semantic segmentation.

As described in Sec.~\ref{sec:searchspace}, the task module in high-level LLVs defines the task-oriented cell in the search phase, rather than a self-defined topology used in low-level LLVs. It is because that the architecture for them in regular natural environments presents a general paradigm. In other words, the specific modular component in the general paradigm is indispensable. As a result of that, we construct the search space for $\balpha_\mathtt{t}$ in the feature aggregation phase, not the whole architecture. More specifically, we consider searching the neck and head except for the backbone in the low-light object detection, the feature conversion except for the encoder-decoder in the low-light semantic segmentation.  
As for the training and validation loss in high-level LLVs, we directly utilize the commonly-adopted task-specific loss in general high-level tasks.

\begin{table*}[t]
	\caption{Quantitative results (PSNR, SSIM, and LPIPS) of state-of-the-art methods and ours on the MIT-Adobe 5K and LOL datasets. The best result is in {\textcolor{red}{\textbf{red}}} whereas the second best one is in {\textcolor{blue}{\textbf{blue}}}.}
	\renewcommand\arraystretch{1.2} 
	\setlength{\tabcolsep}{0.6mm}
	\centering
	\begin{tabular}{|c|c|cccccccccc||ccc|}
		\hline
		\footnotesize Datasets&\footnotesize Metrics&\footnotesize MBLLEN&\footnotesize GLADNet&\footnotesize RetinexNet&\footnotesize EnGAN&\footnotesize SSIENet&\footnotesize KinD&\footnotesize DeepUPE&\footnotesize ZeroDCE&\footnotesize FIDE&\footnotesize DRBN&\footnotesize RUAS$_\mathtt{S}$&\footnotesize RUAS&\footnotesize RUAS$_\mathtt{A}$\\
		\hline
		\multirow{3}{*}{\footnotesize MIT}&\footnotesize PSNR& \footnotesize 15.587&\footnotesize 16.728&	\footnotesize 12.685&\footnotesize 15.014&\footnotesize 10.324&\footnotesize 17.169&\footnotesize 18.779&\footnotesize 16.463&\footnotesize 17.170&\footnotesize 15.954&\footnotesize \textcolor{red}{\textbf{21.018}}&\footnotesize \textcolor{blue}{\textbf{20.830}}& \footnotesize\textcolor{red}{\textbf{21.018}}\\
		\cline{2-15}
		~&\footnotesize SSIM&\footnotesize 0.713&\footnotesize 0.764&\footnotesize 0.644&\footnotesize 0.768&\footnotesize 0.620&\footnotesize 0.696&\footnotesize 0.822&\footnotesize 0.764&\footnotesize 0.696&\footnotesize 0.704&\footnotesize\textcolor{red}{\textbf{0.855}}&\footnotesize \textcolor{blue}{\textbf{0.854}}&\footnotesize \textcolor{red}{\textbf{0.855}}\\
		\cline{2-15}
		~&\footnotesize LPIPS&\footnotesize 0.307&\footnotesize  0.693&\footnotesize 0.338&\footnotesize 0.209&\footnotesize 0.323 &\footnotesize 0.229&\footnotesize 0.183&\footnotesize 0.193&\footnotesize 0.324&\footnotesize 0.292& \footnotesize\textcolor{red}{\textbf{0.139}}&\footnotesize \textcolor{blue}{\textbf{0.141}}&\footnotesize \textcolor{red}{\textbf{0.139}}\\
		\hline
		\multirow{6}{*}{\footnotesize LOL}&\footnotesize PSNR&\footnotesize 13.931&\footnotesize 16.188&\footnotesize 13.096&\footnotesize 15.644&\footnotesize 14.176&\footnotesize 14.616&\footnotesize 13.041&\footnotesize 15.512&\footnotesize 16.718&\footnotesize 15.324&\footnotesize 16.038&\footnotesize \textcolor{blue}{\textbf{18.226}}&\footnotesize \textcolor{red}{\textbf{19.751}}\\
		\cline{2-15} 
		~&\footnotesize SSIM&\footnotesize 0.489&\footnotesize 0.605&\footnotesize 0.429&\footnotesize 0.578&\footnotesize 0.534&\footnotesize 0.636&\footnotesize 0.483&\footnotesize 0.553&\footnotesize 0.673&\footnotesize0.699&\footnotesize 0.513&\footnotesize \textcolor{blue}{\textbf{0.717}}&\footnotesize \textcolor{red}{\textbf{0.735}}\\
		\cline{2-15}
		~&\footnotesize LPIPS&\footnotesize 0.697&\footnotesize  \textcolor{red}{\textbf{0.205}}&\footnotesize 0.864&\footnotesize 0.647&\footnotesize 0.675&\footnotesize 0.463&\footnotesize 0.677&\footnotesize 0.718&\footnotesize 0.510&\footnotesize  0.362&\footnotesize 0.661&\footnotesize \textcolor{blue}{\textbf{0.354}}&\footnotesize 0.393\\
		\cline{2-15}
		~&\footnotesize SIZE(M)&\footnotesize 0.450&\footnotesize 1.128&\footnotesize 0.838&\footnotesize 8.636&\footnotesize 0.682&\footnotesize 8.540&\footnotesize 1.0482&\footnotesize 0.079&\footnotesize 8.621&\footnotesize 0.577&\footnotesize \textcolor{red}{\textbf{0.001}}&\footnotesize \textcolor{blue}{\textbf{0.003}}&\footnotesize 0.037\\
		\cline{2-15}
		~&\footnotesize FLOPs(G)&\footnotesize 19.956&\footnotesize 252.141&\footnotesize 136.015&\footnotesize 61.010&\footnotesize 34.607&\footnotesize 29.130&\footnotesize 44.226&\footnotesize 5.207&\footnotesize 57.240&\footnotesize 37.790&\footnotesize \textcolor{red}{\textbf{0.281}}&\footnotesize \textcolor{blue}{\textbf{0.832}}&\footnotesize 8.833\\
		\cline{2-15}
		~&\footnotesize TIME(S)&\footnotesize 0.077&\footnotesize 0.025&\footnotesize 0.119&\footnotesize \textcolor{blue}{\textbf{0.010}}&\footnotesize 0.027&\footnotesize 0.181&\footnotesize 0.019&\footnotesize 0.011&\footnotesize 0.594&\footnotesize 0.053&\footnotesize \textcolor{red}{\textbf{0.006}}&\footnotesize 0.016&\footnotesize \textcolor{blue}{\textbf{0.010}}\\
		\hline
	\end{tabular}
	\label{tab: LLIEquan}
\end{table*}

\begin{figure*}[t]
	\centering
	\begin{tabular}{c@{\extracolsep{0.25em}}c@{\extracolsep{0.25em}}c@{\extracolsep{0.25em}}c@{\extracolsep{0.25em}}c@{\extracolsep{0.25em}}c@{\extracolsep{0.25em}}c@{\extracolsep{0.25em}}c}
		\includegraphics[width=0.116\linewidth]{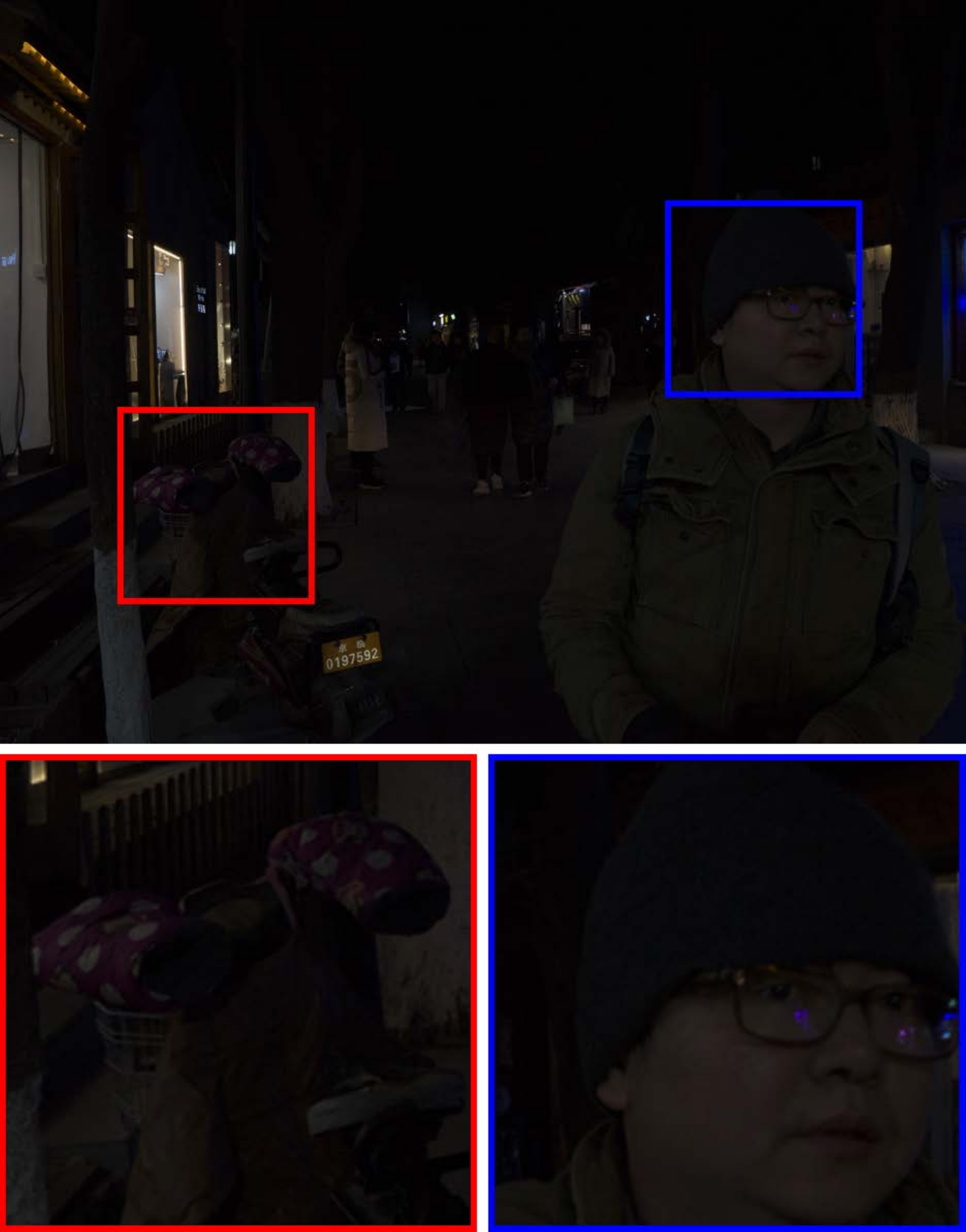}&
		\includegraphics[width=0.116\linewidth]{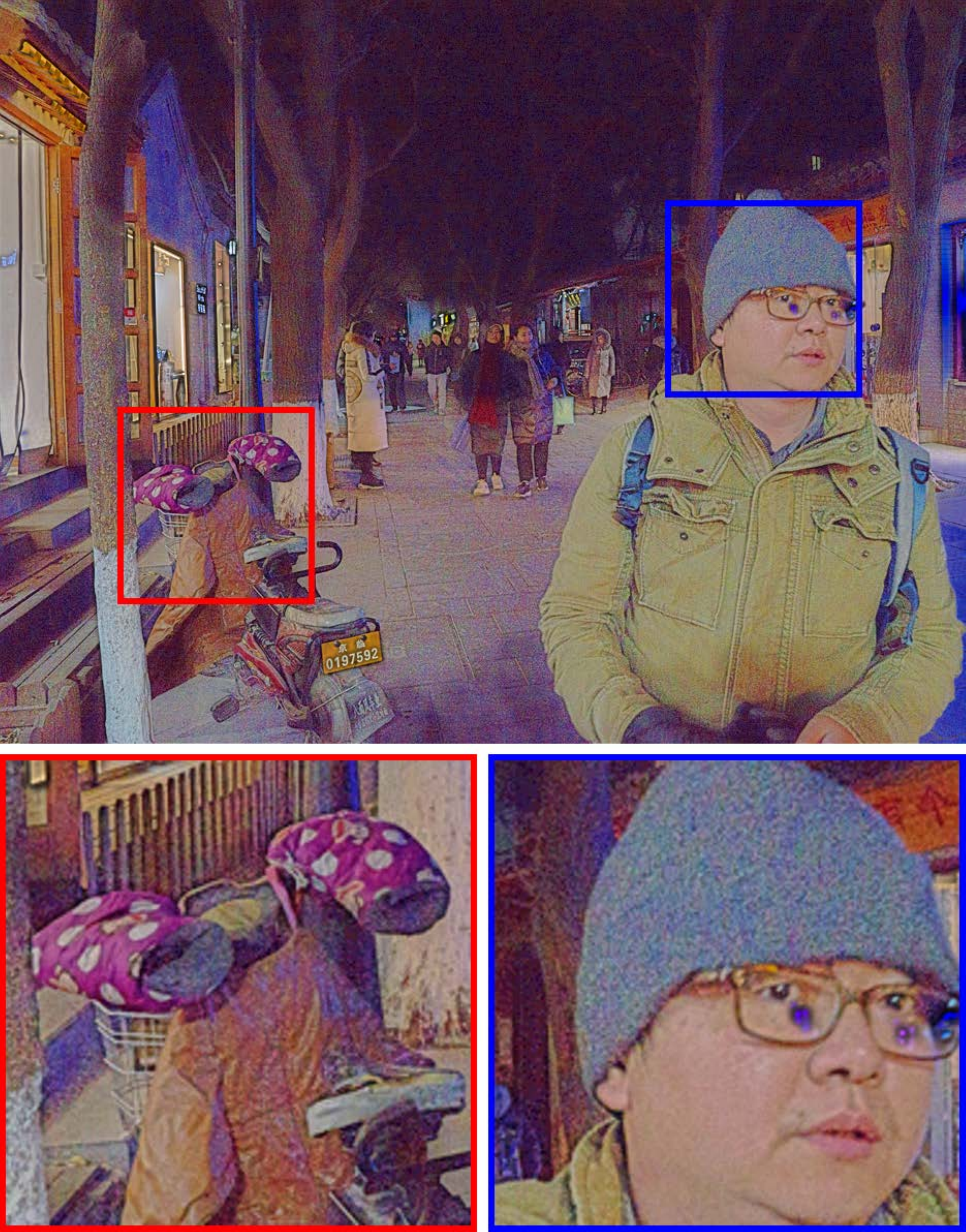}&
		\includegraphics[width=0.116\linewidth]{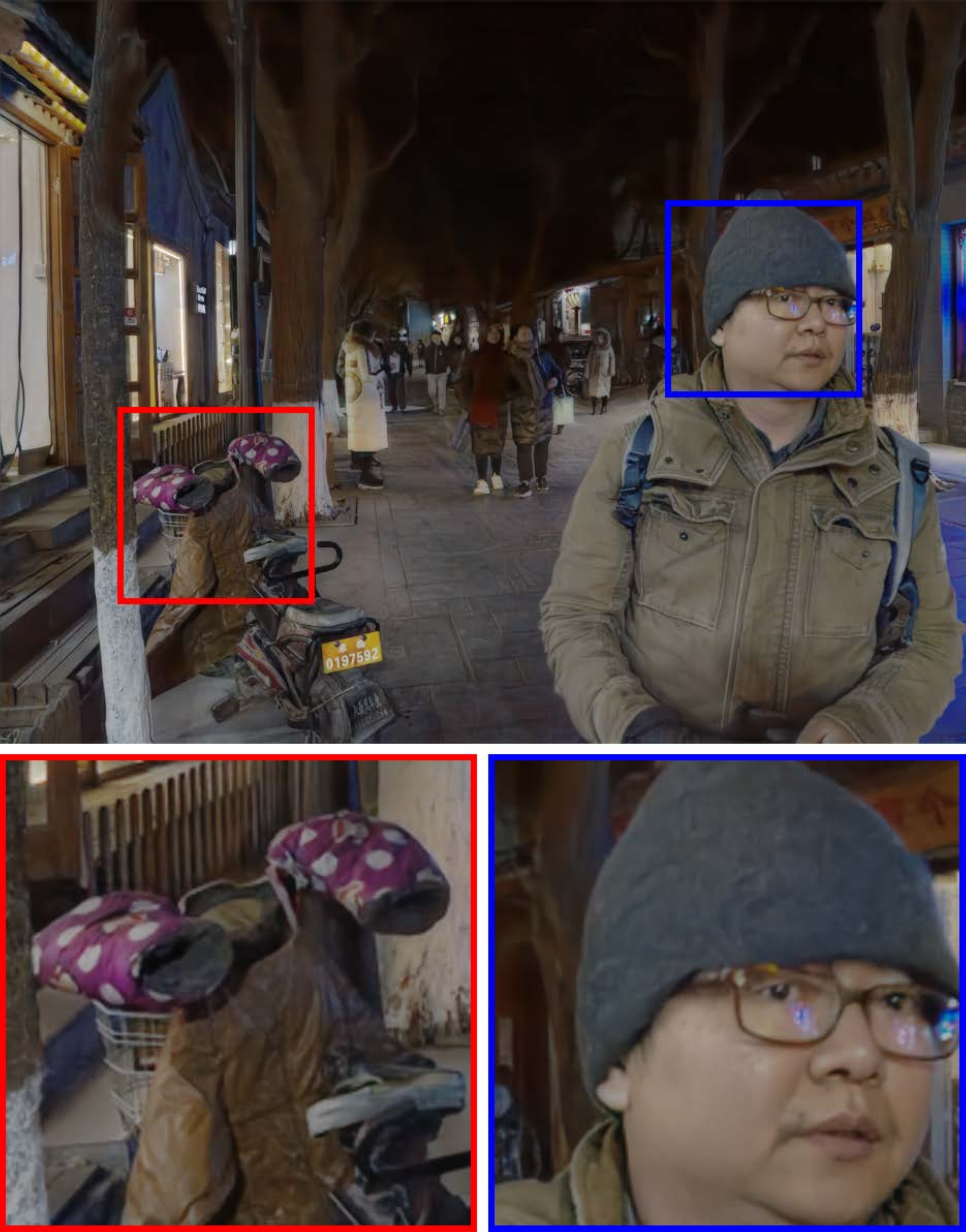}&
		\includegraphics[width=0.116\linewidth]{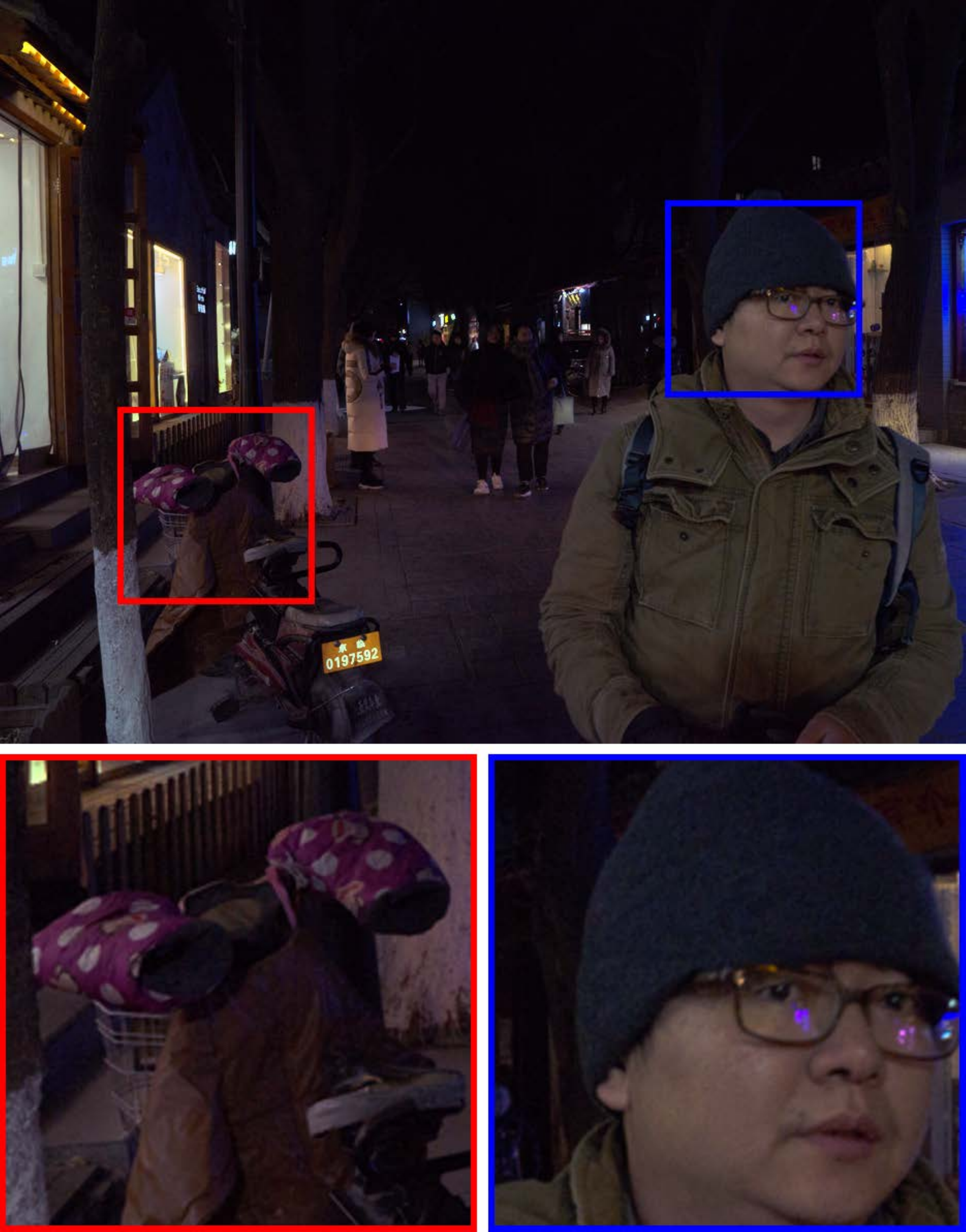}&
		\includegraphics[width=0.116\linewidth]{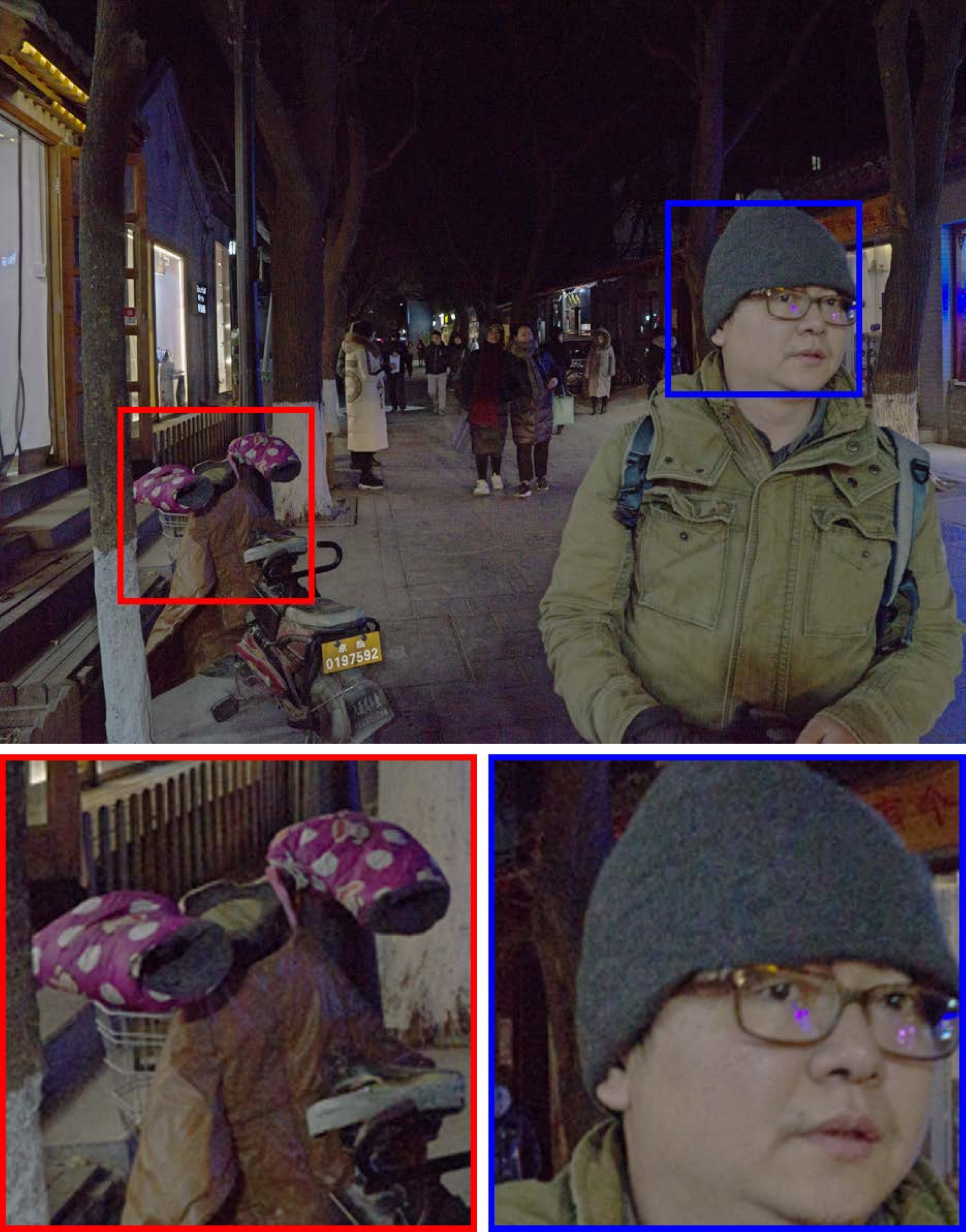}&
		\includegraphics[width=0.116\linewidth]{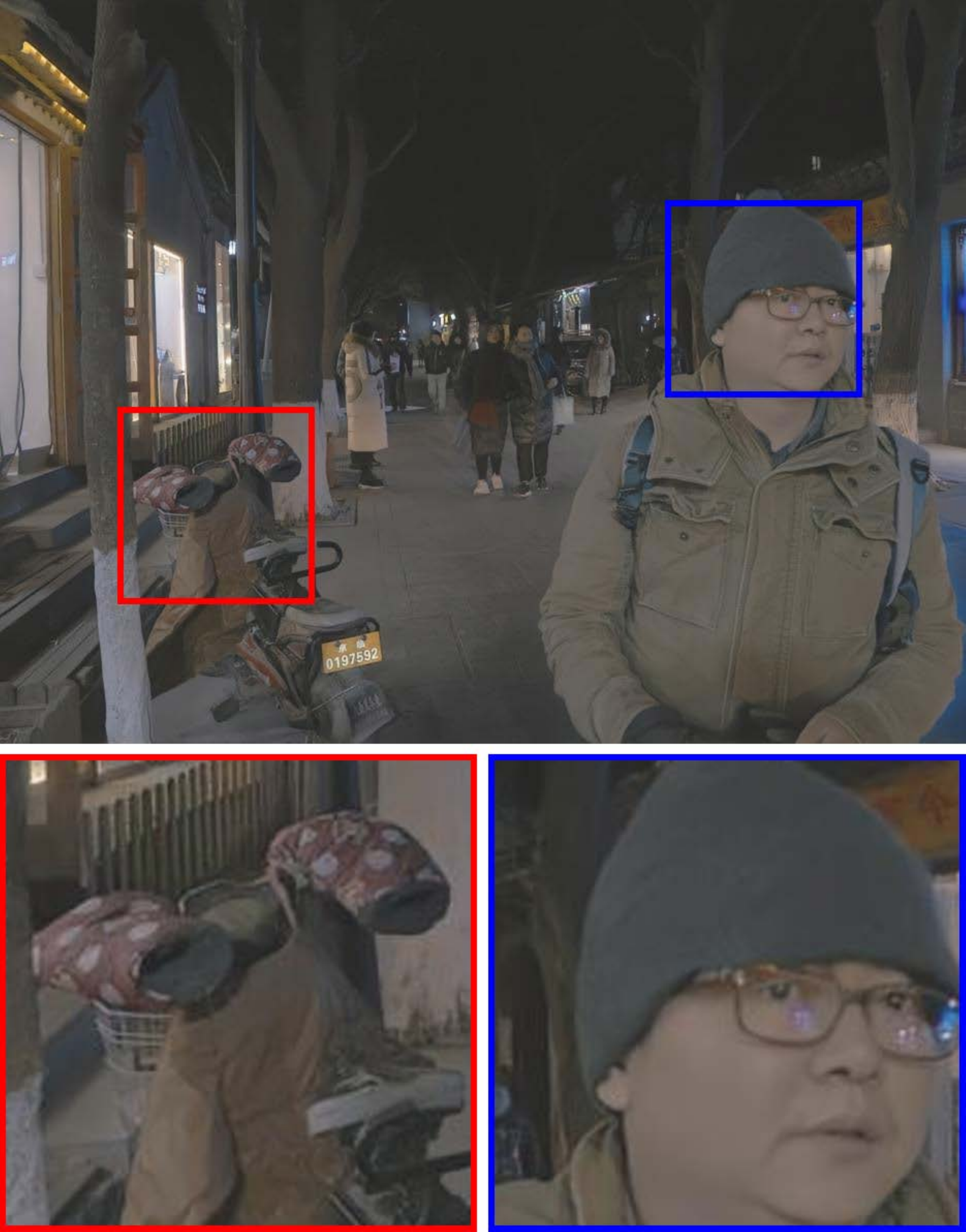}&
		\includegraphics[width=0.116\linewidth]{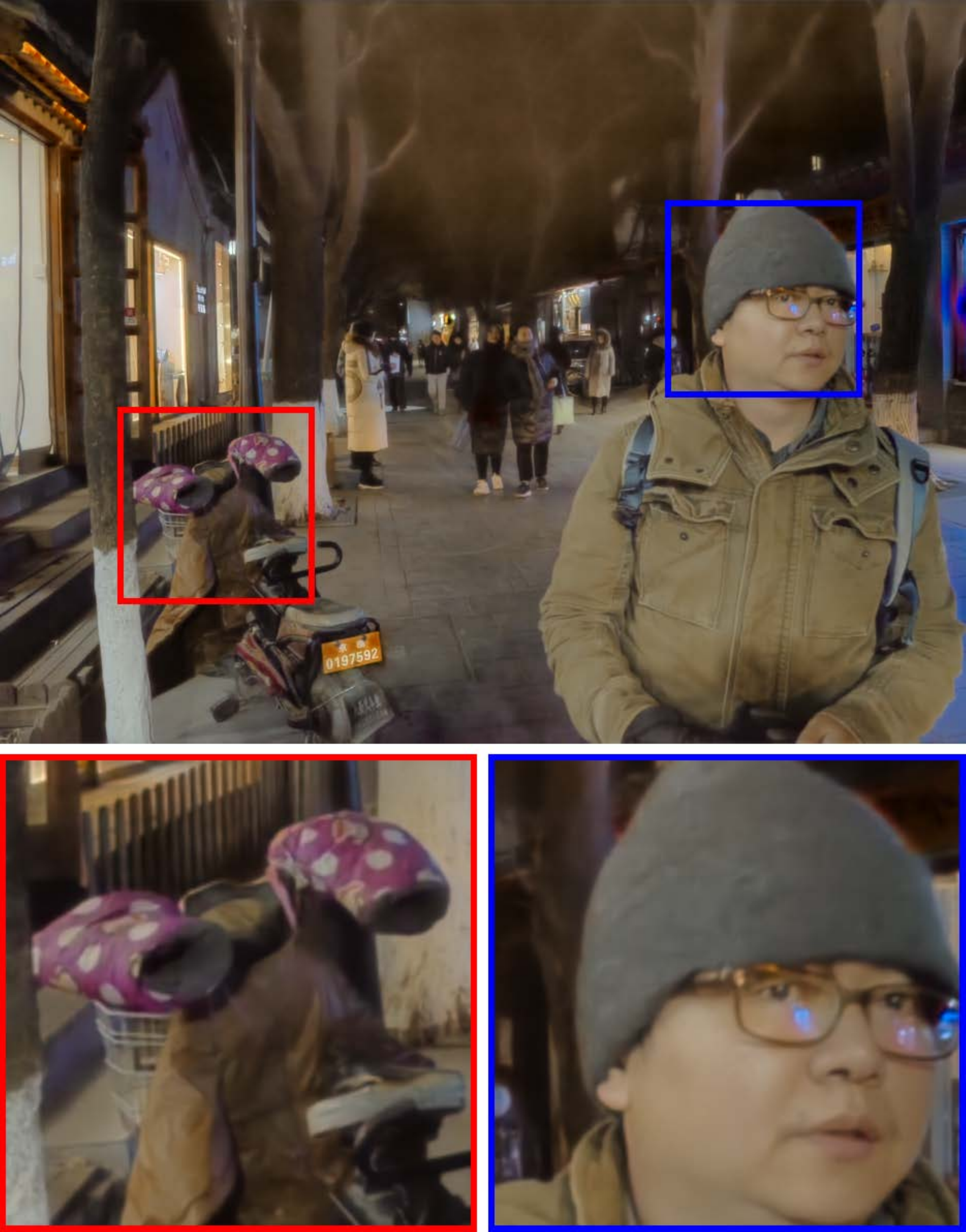}&
		\includegraphics[width=0.116\linewidth]{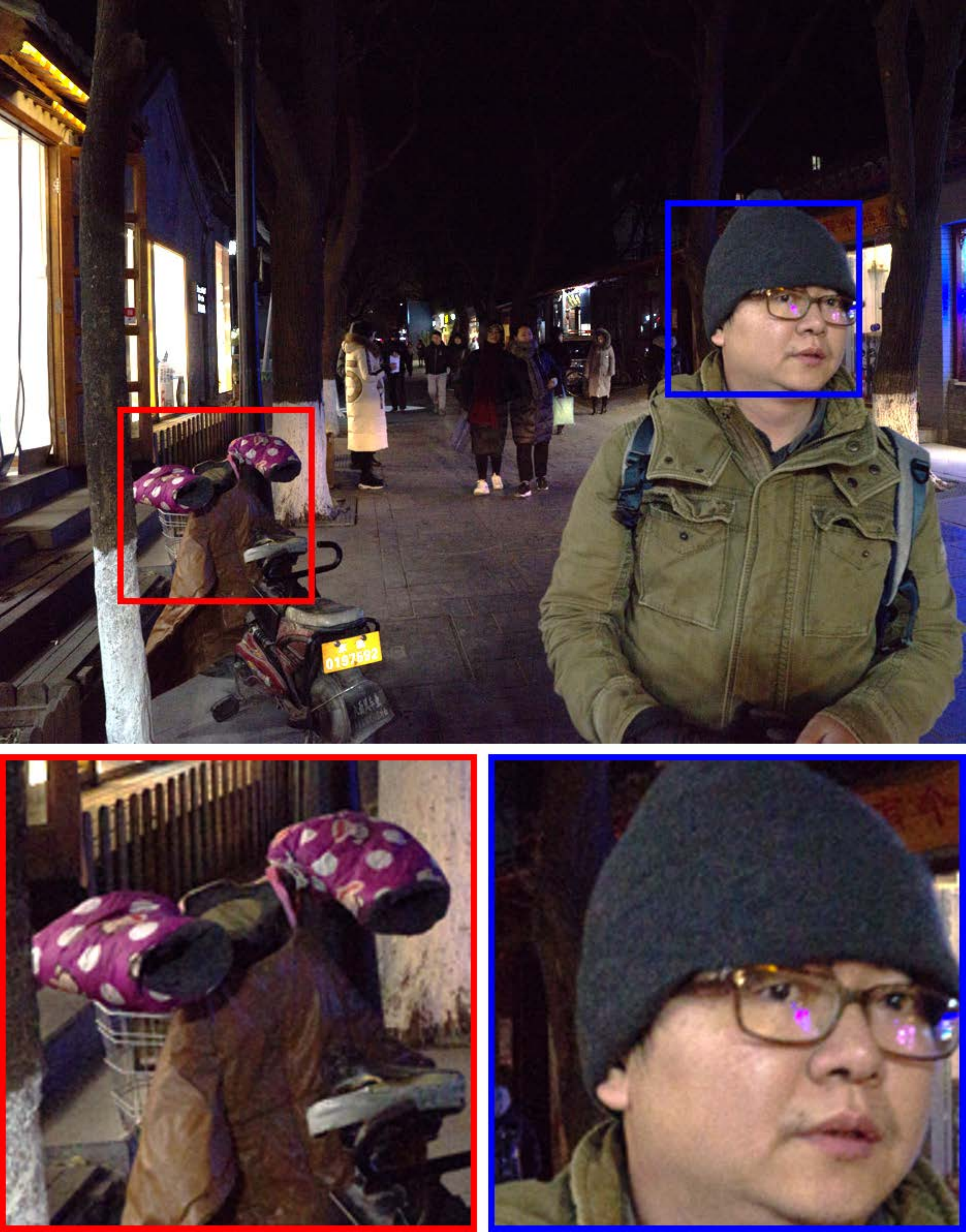}\\
		\includegraphics[width=0.116\linewidth]{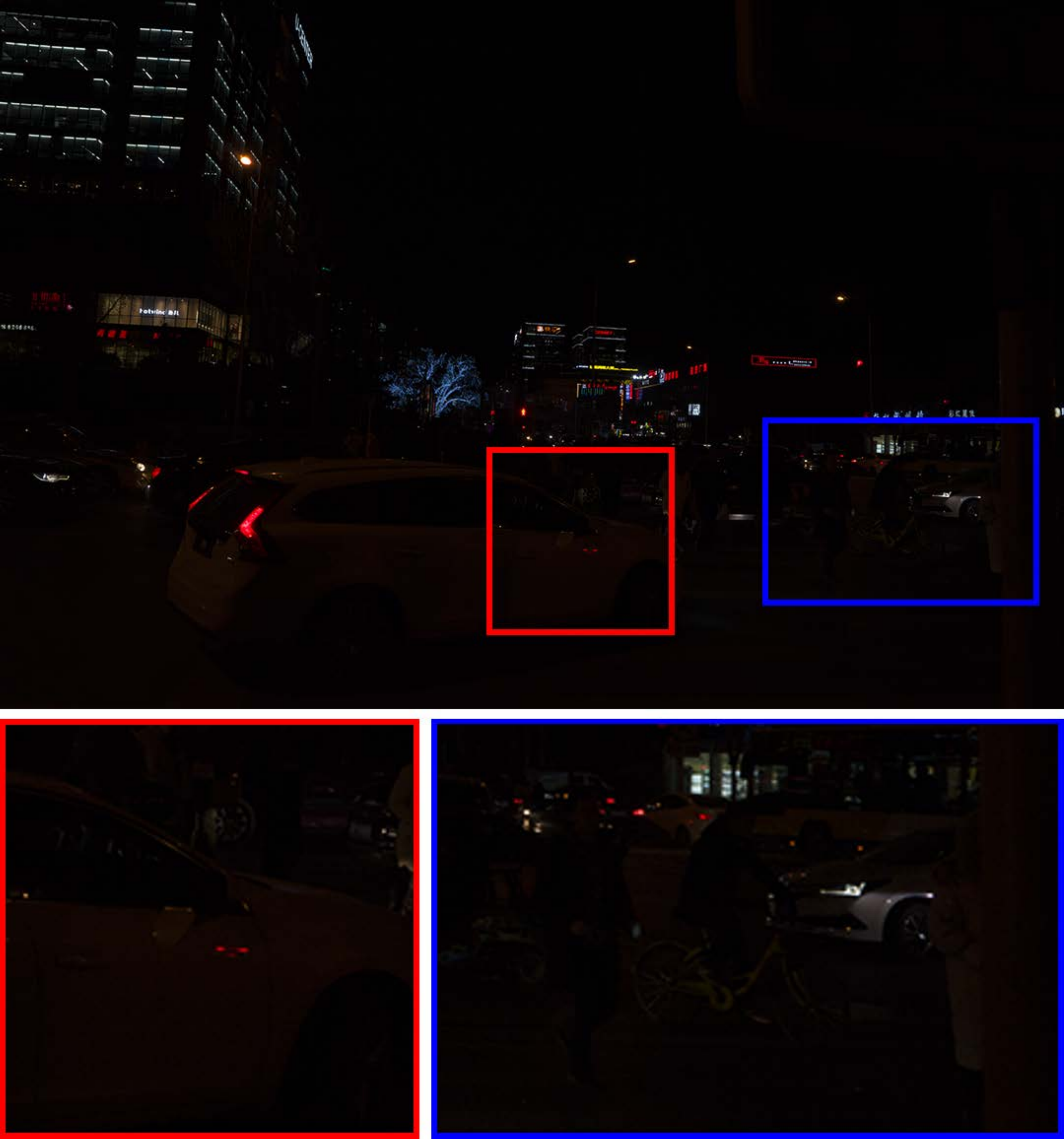}&
		\includegraphics[width=0.116\linewidth]{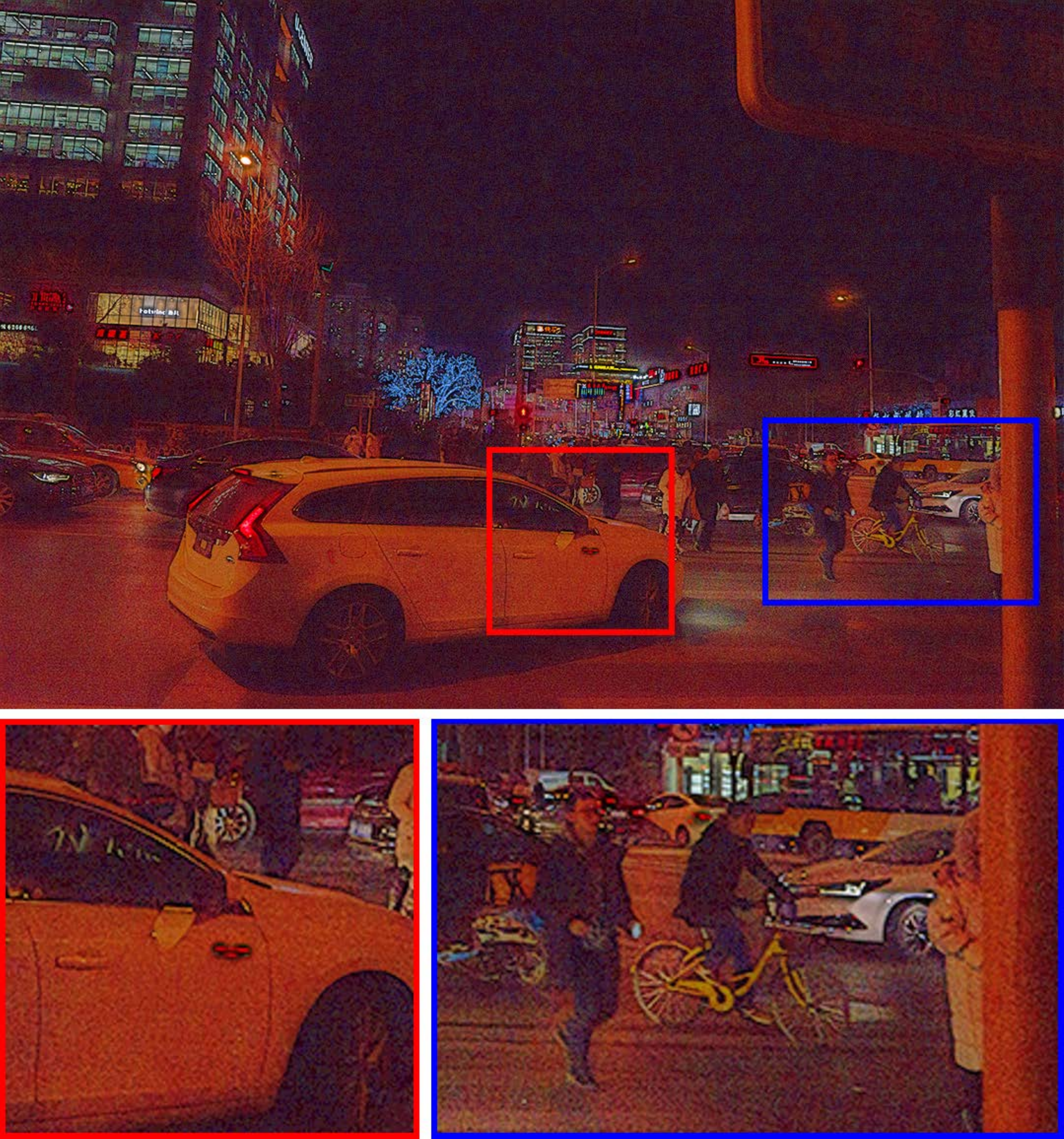}&
		\includegraphics[width=0.116\linewidth]{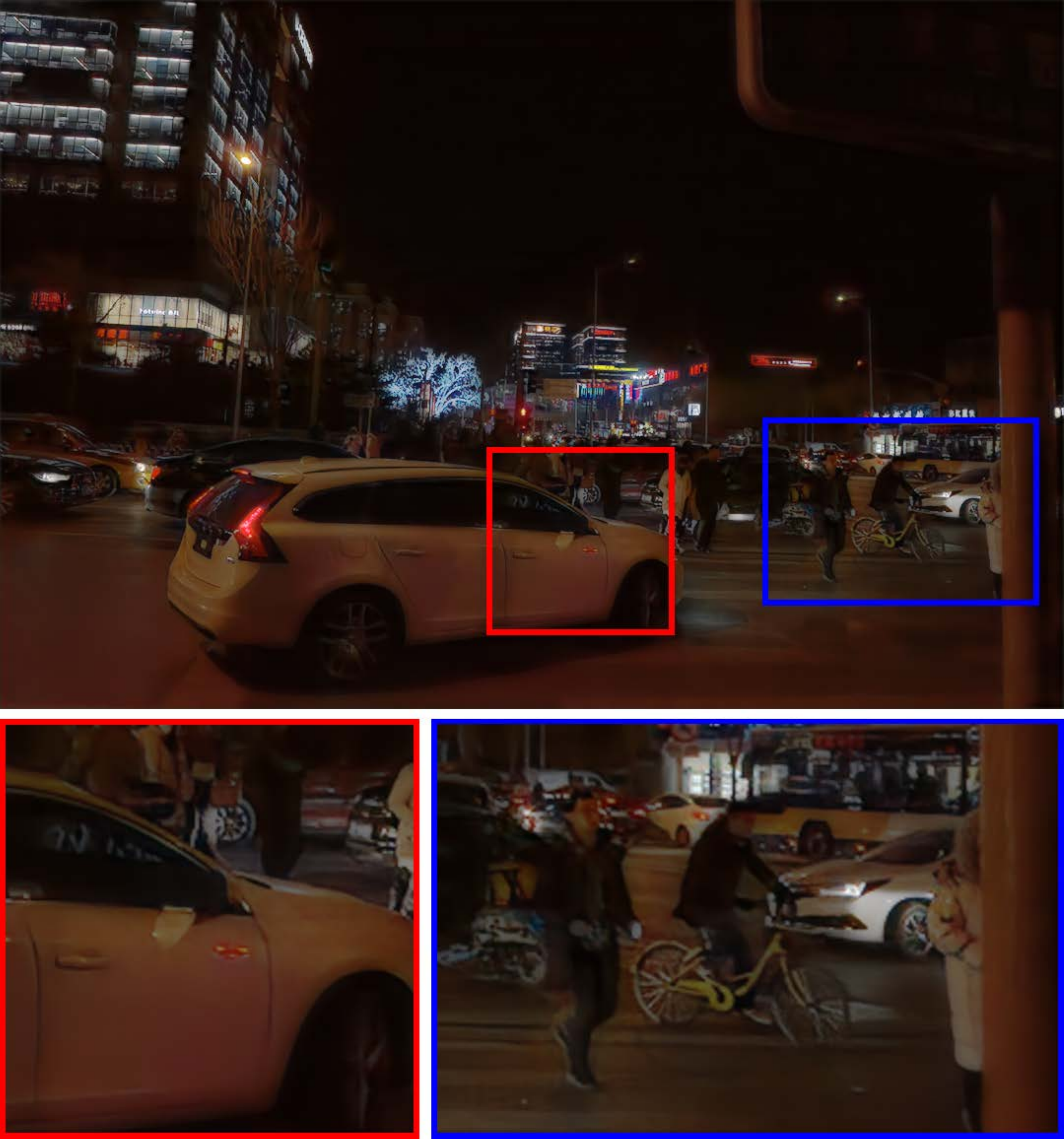}&
		\includegraphics[width=0.116\linewidth]{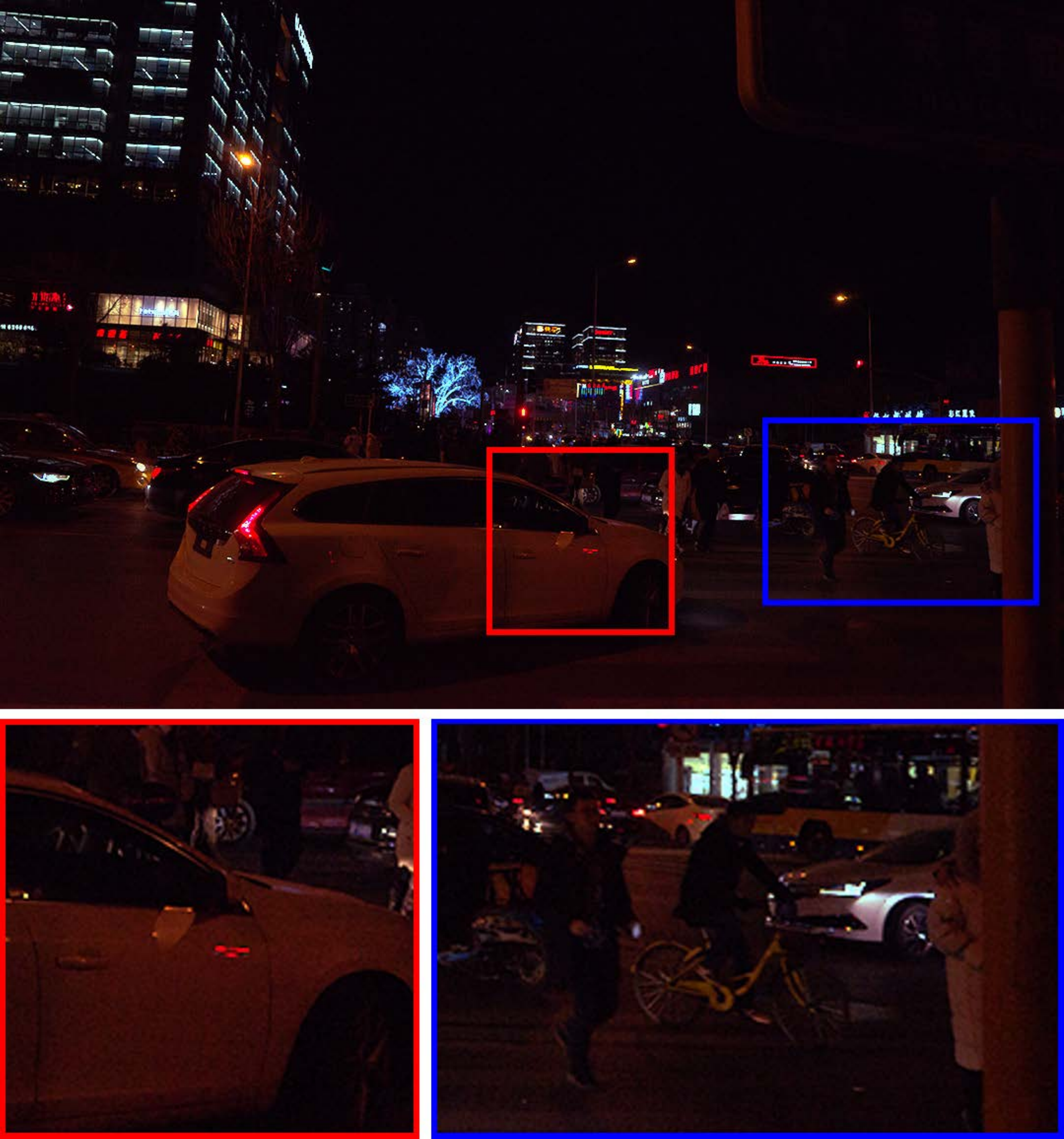}&
		\includegraphics[width=0.116\linewidth]{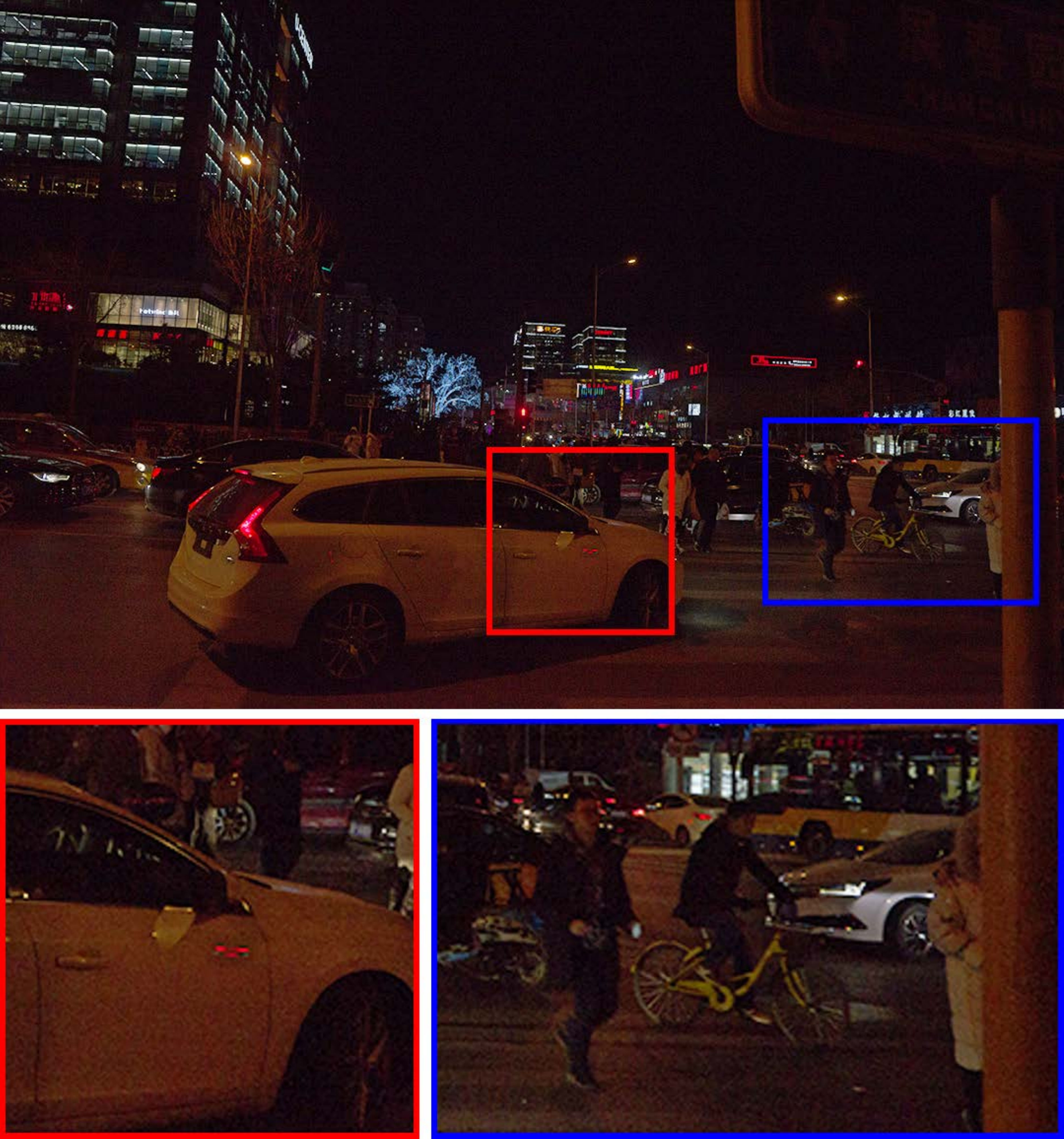}&
		\includegraphics[width=0.116\linewidth]{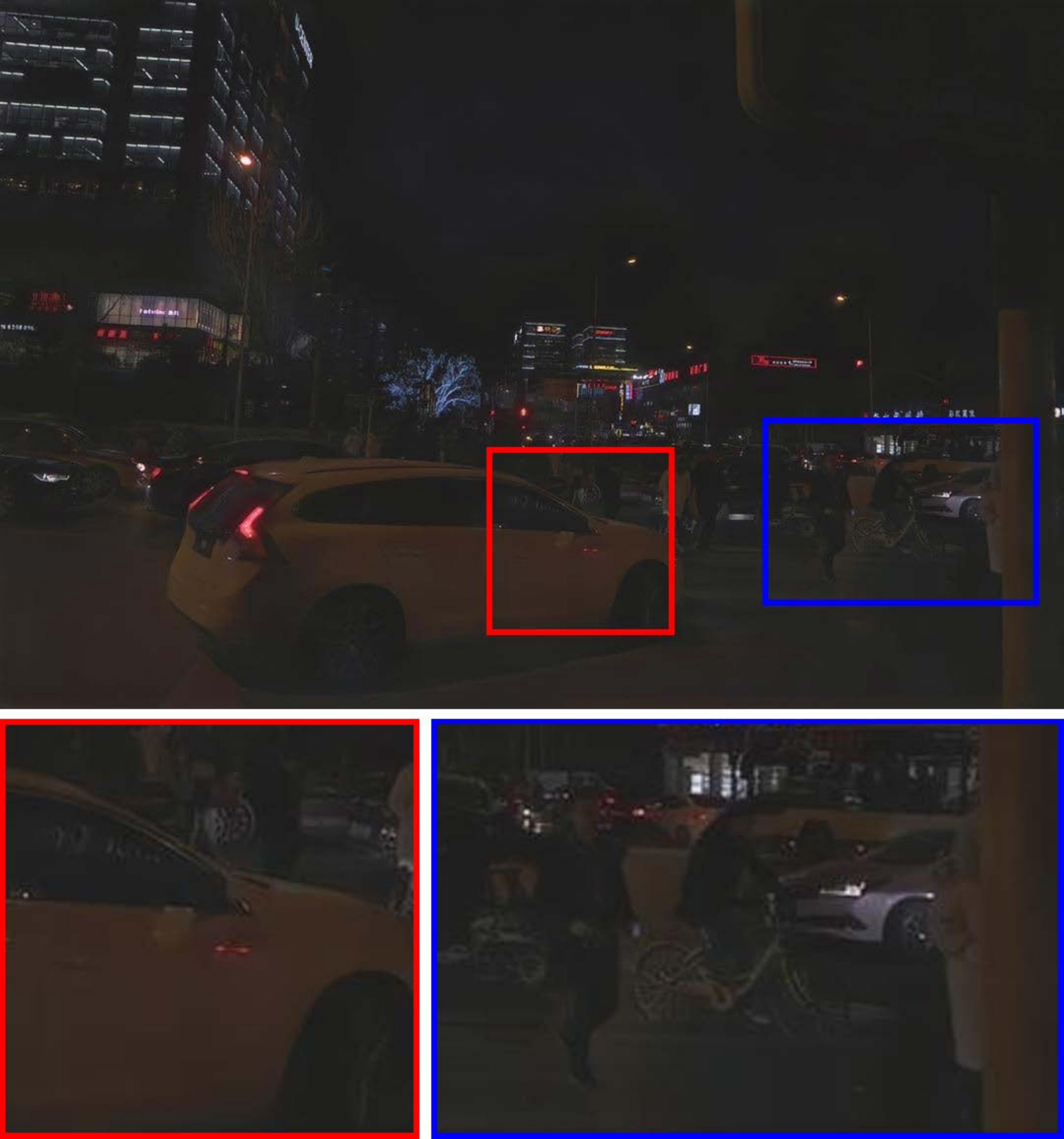}&
		\includegraphics[width=0.116\linewidth]{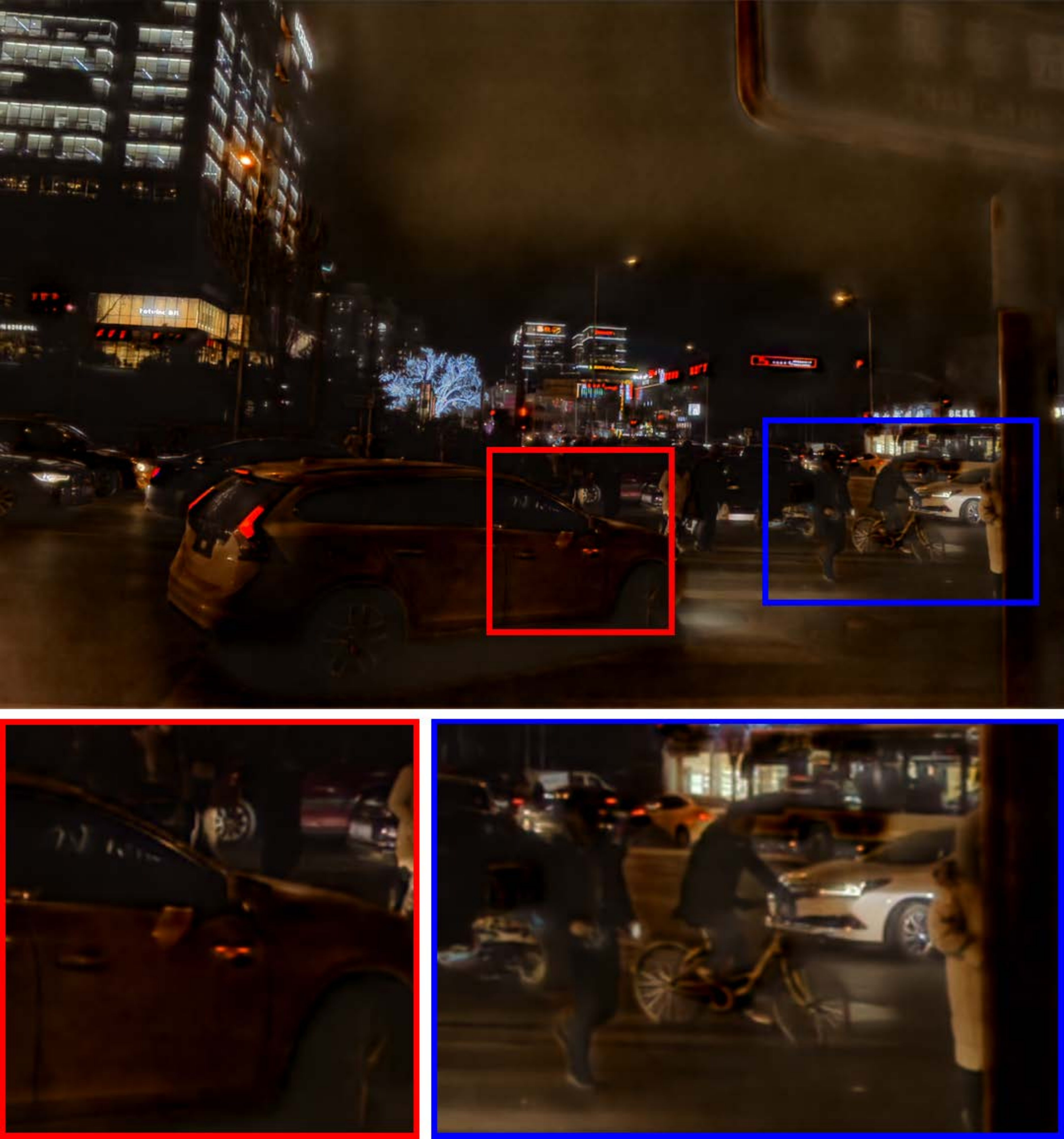}&
		\includegraphics[width=0.116\linewidth]{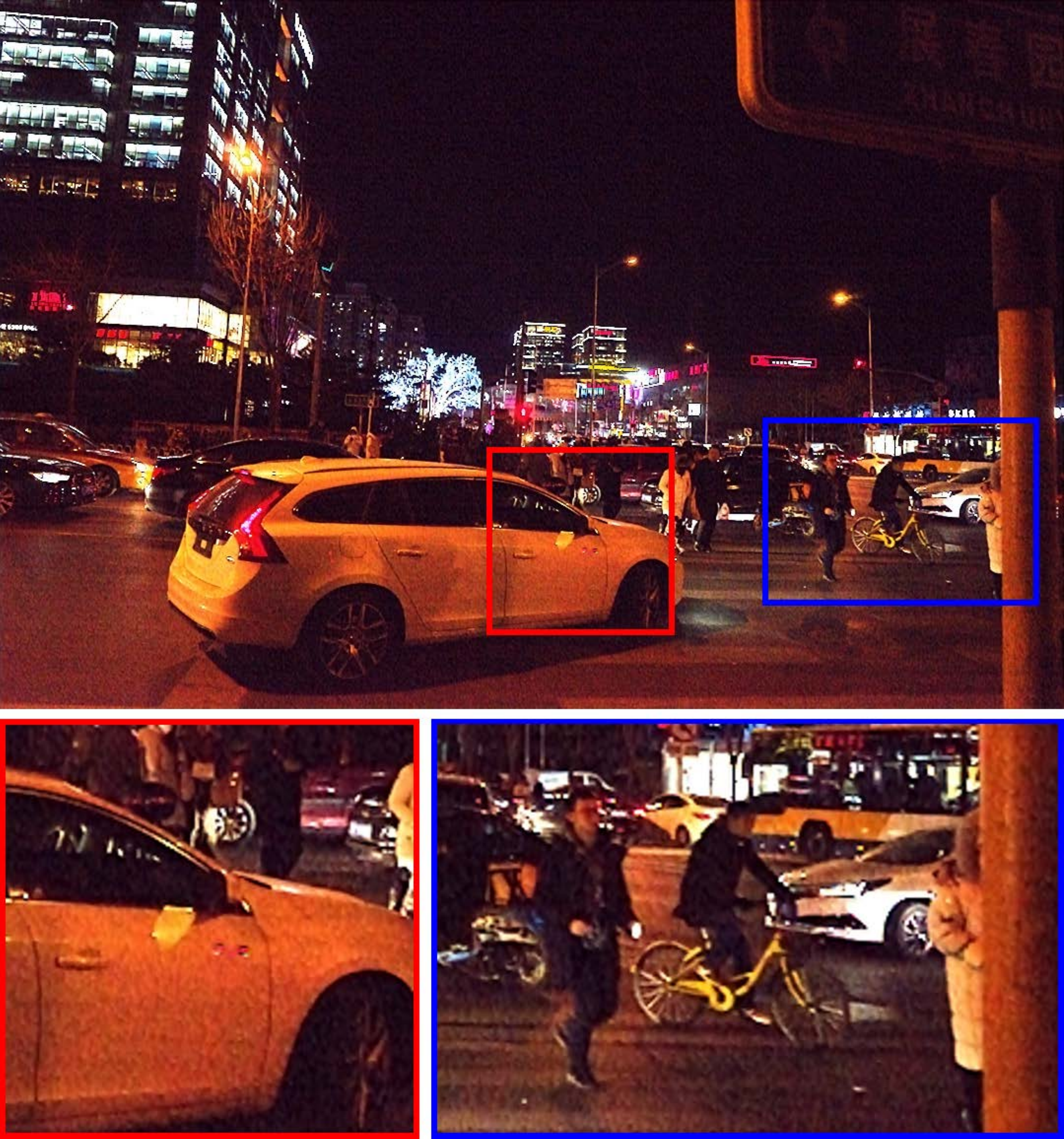}\\
		\includegraphics[width=0.116\linewidth]{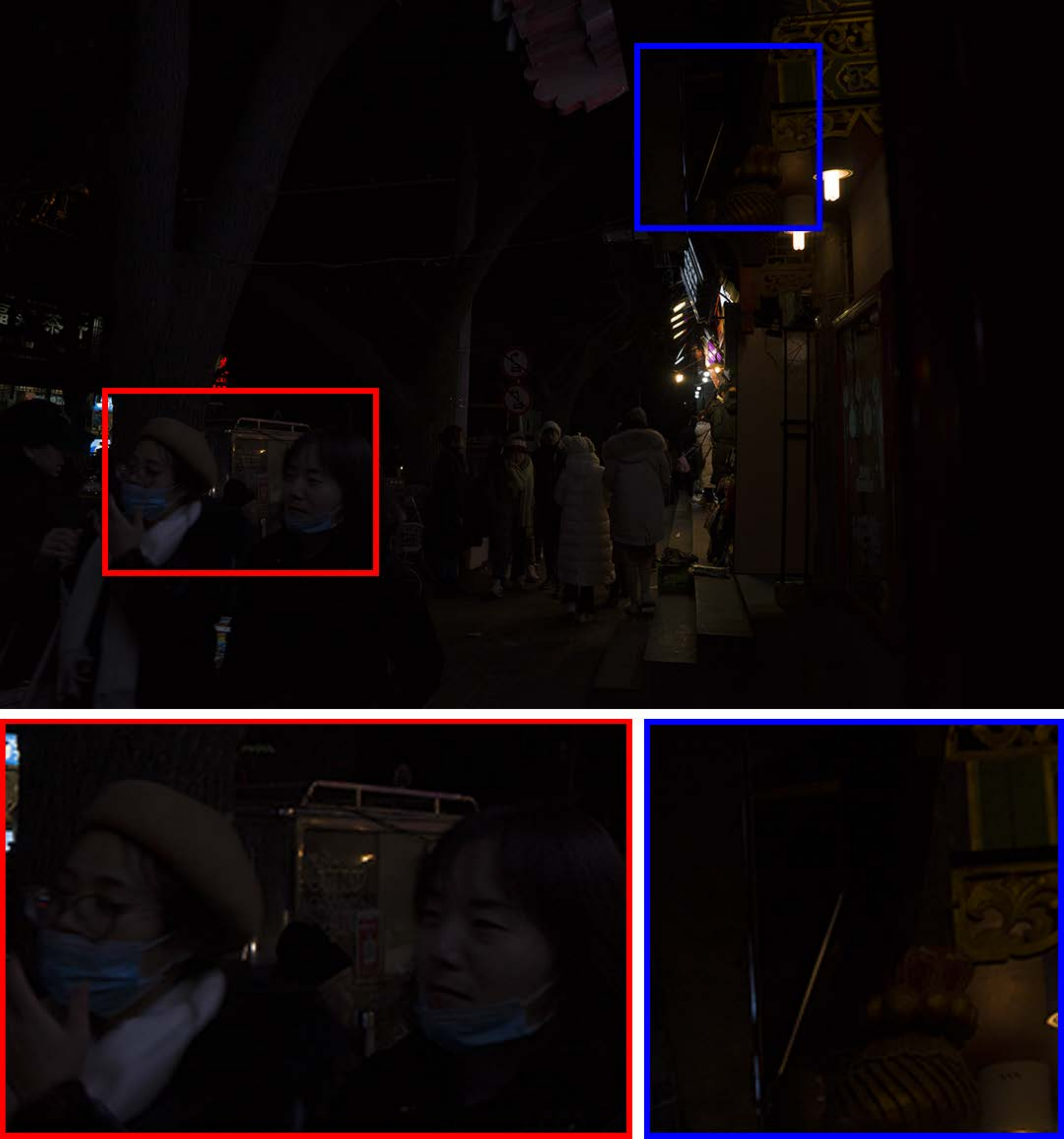}&
		\includegraphics[width=0.116\linewidth]{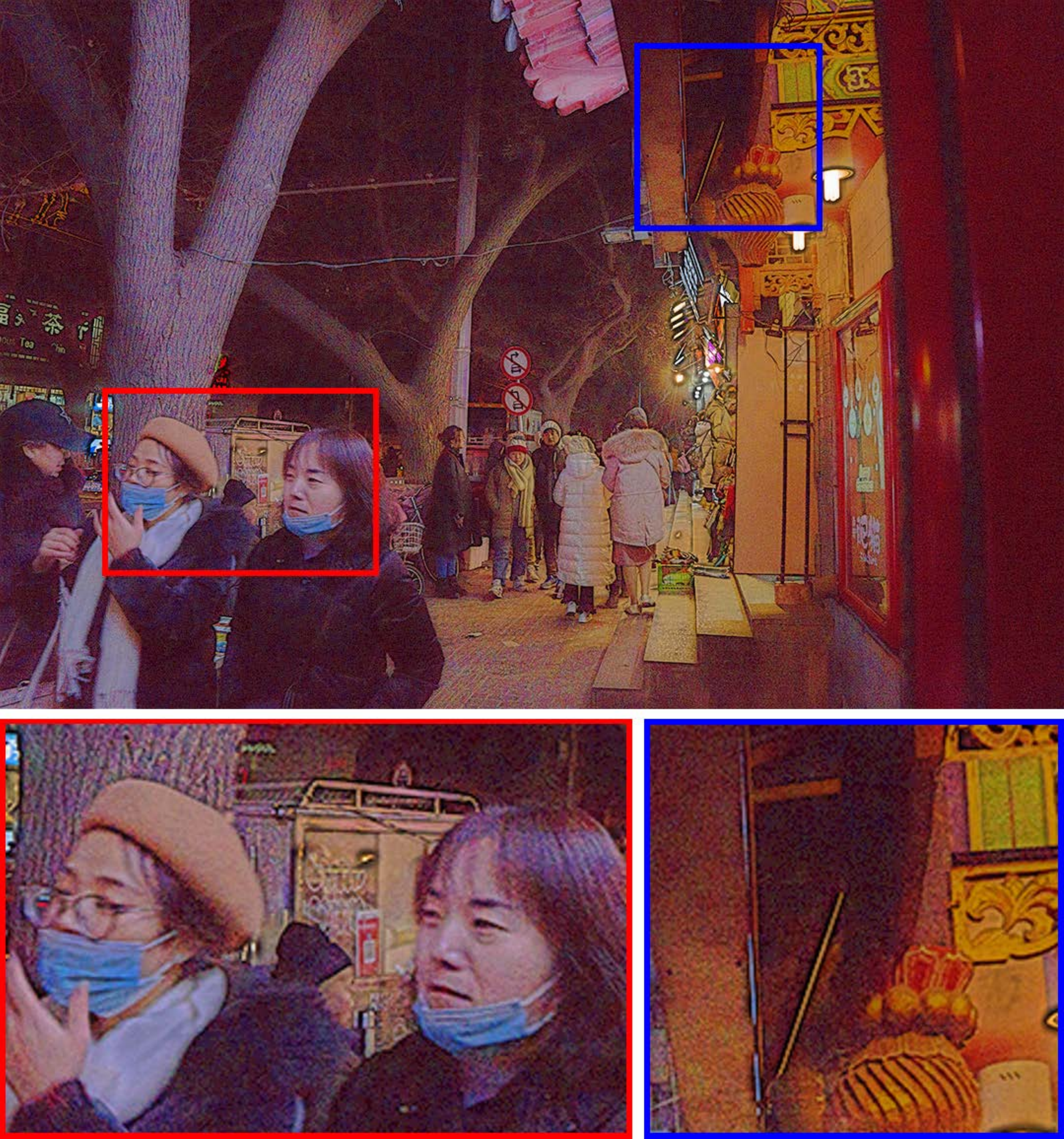}&
		\includegraphics[width=0.116\linewidth]{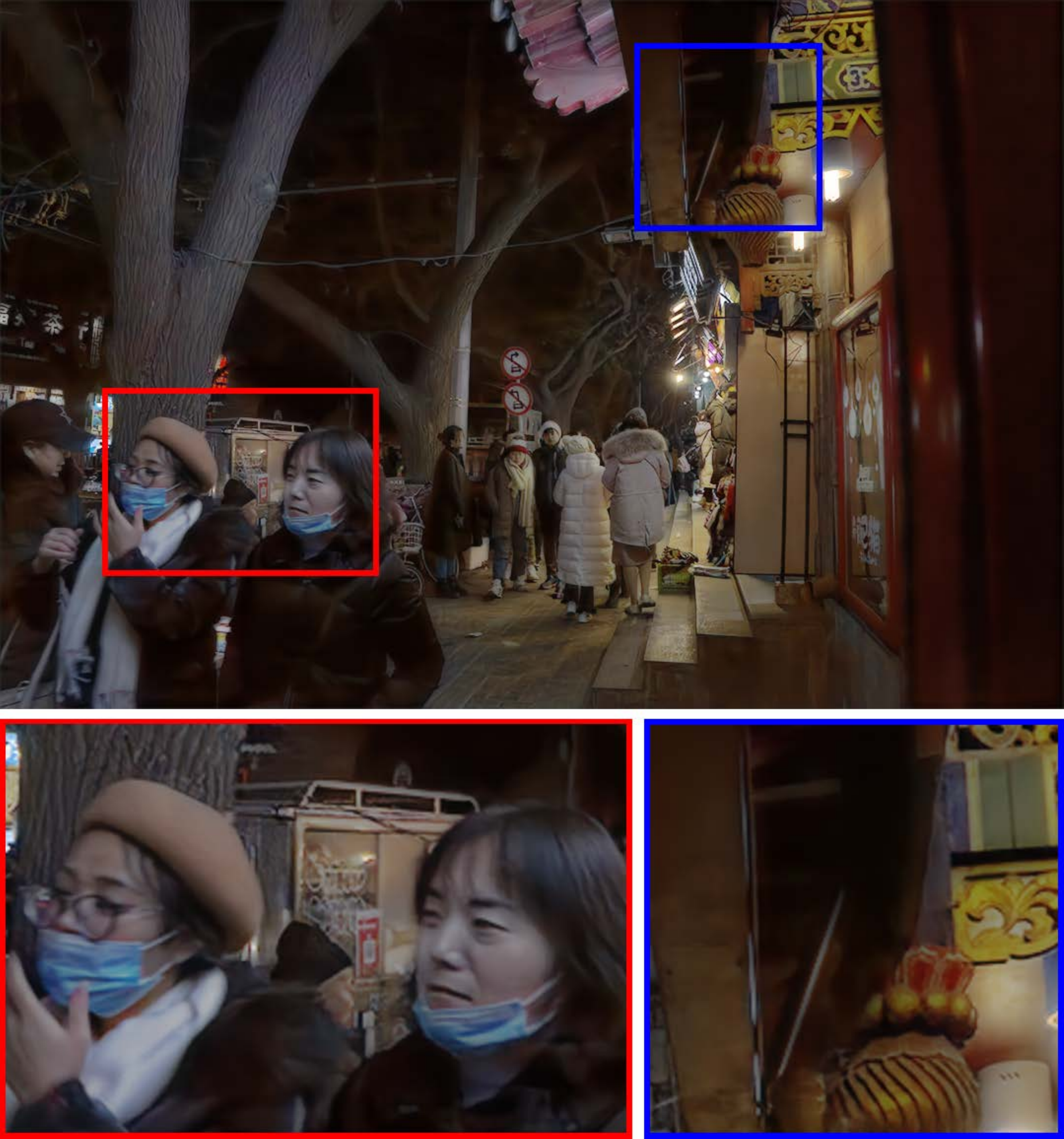}&
		\includegraphics[width=0.116\linewidth]{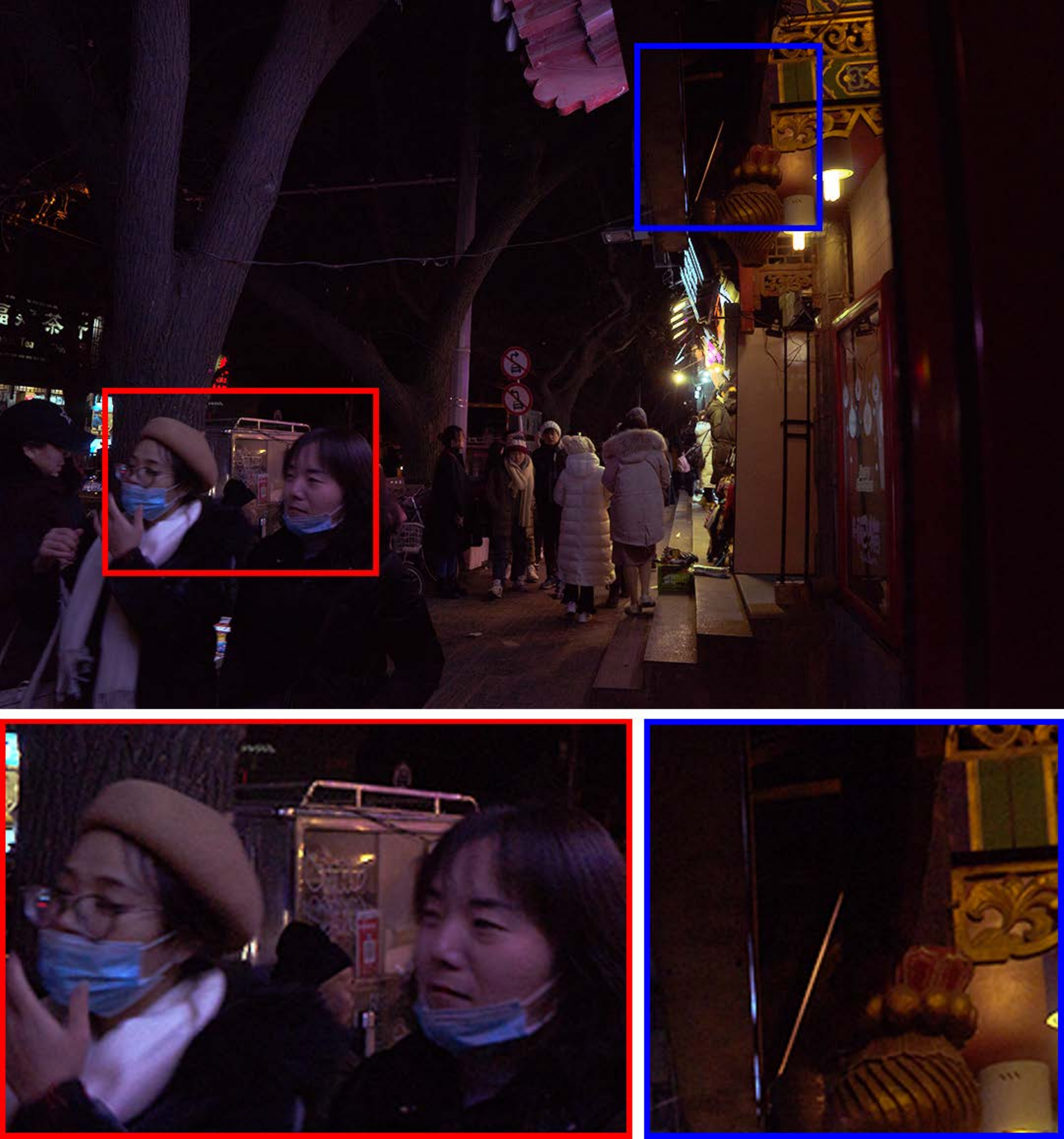}&
		\includegraphics[width=0.116\linewidth]{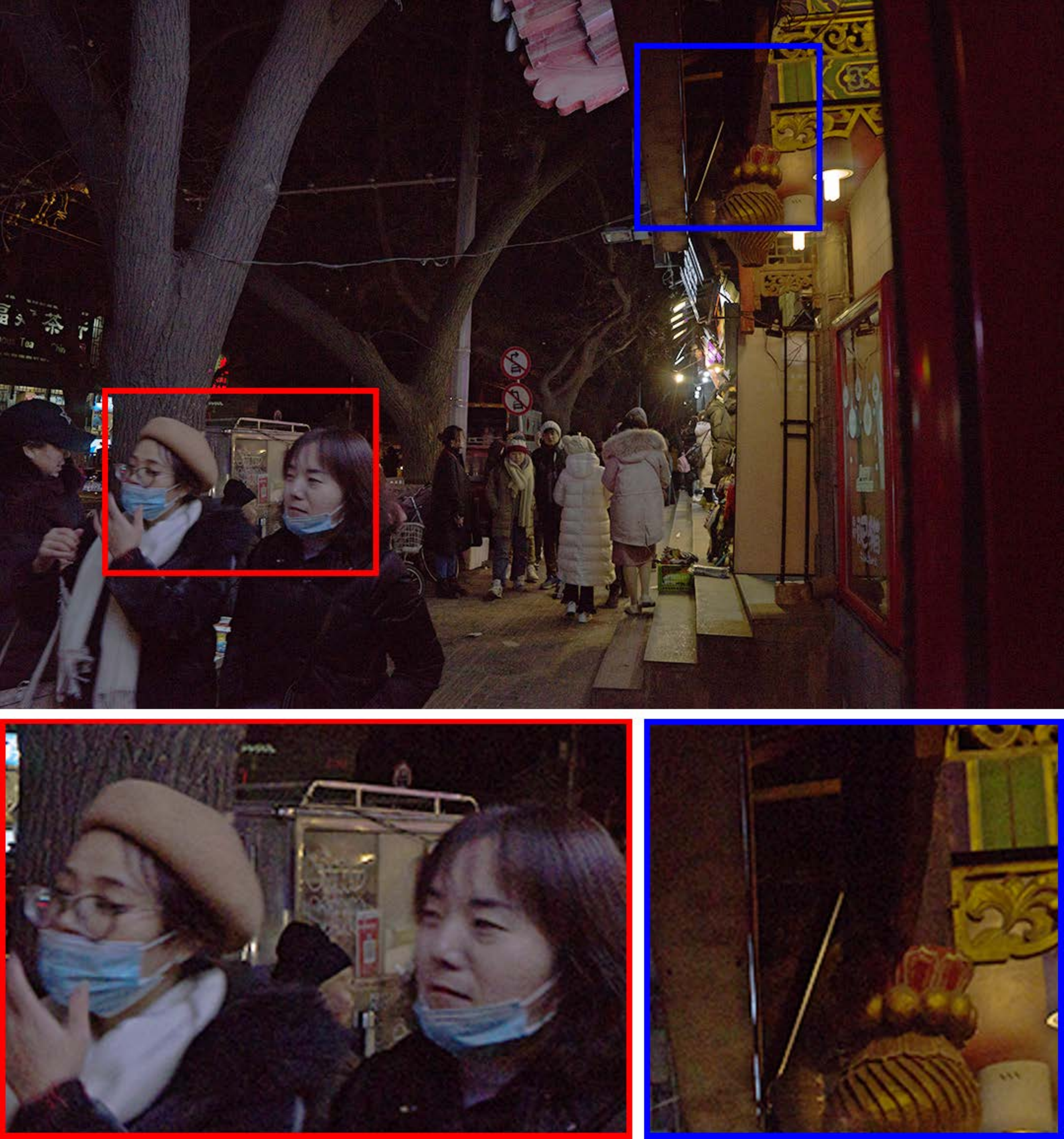}&
		\includegraphics[width=0.116\linewidth]{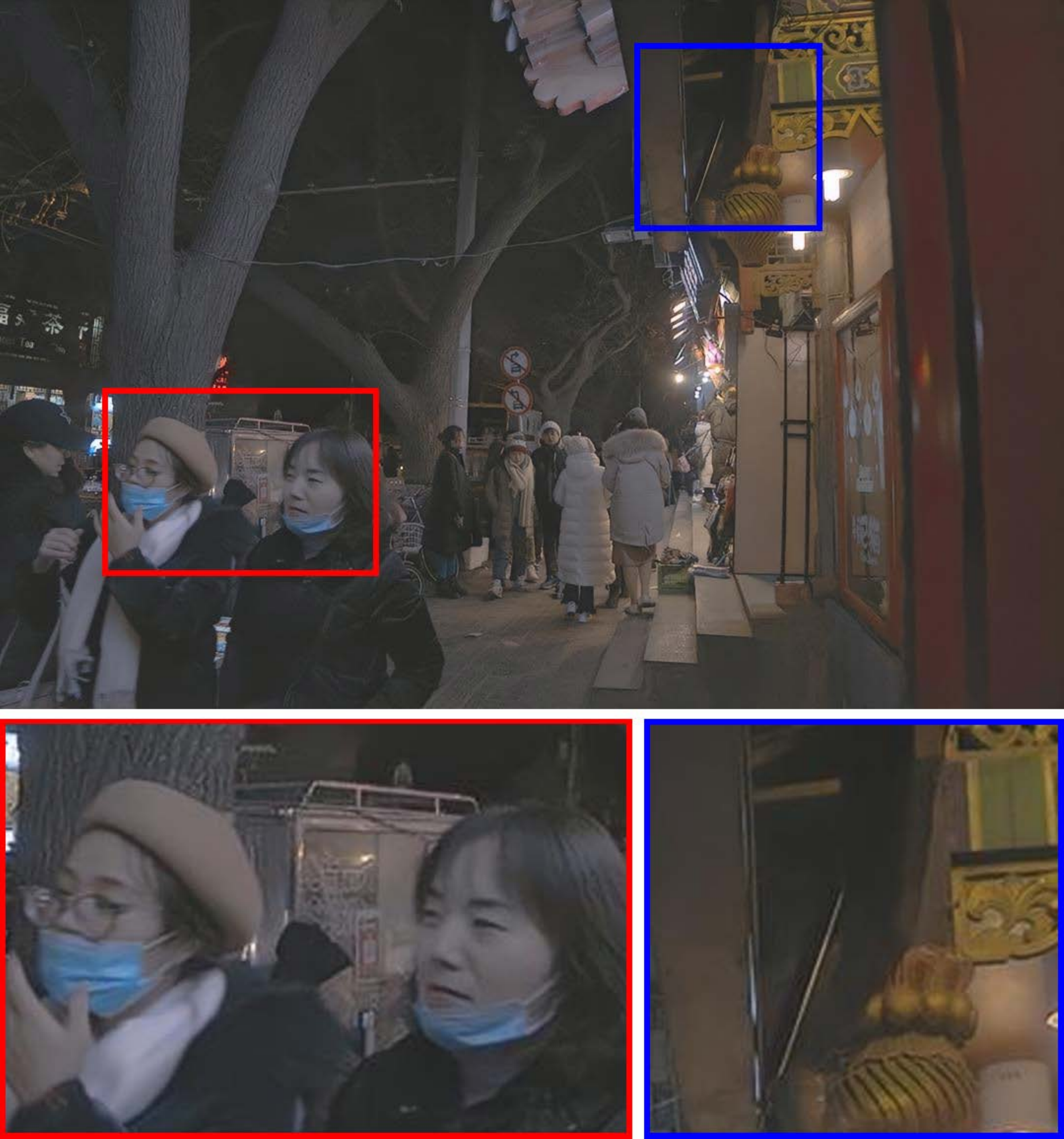}&
		\includegraphics[width=0.116\linewidth]{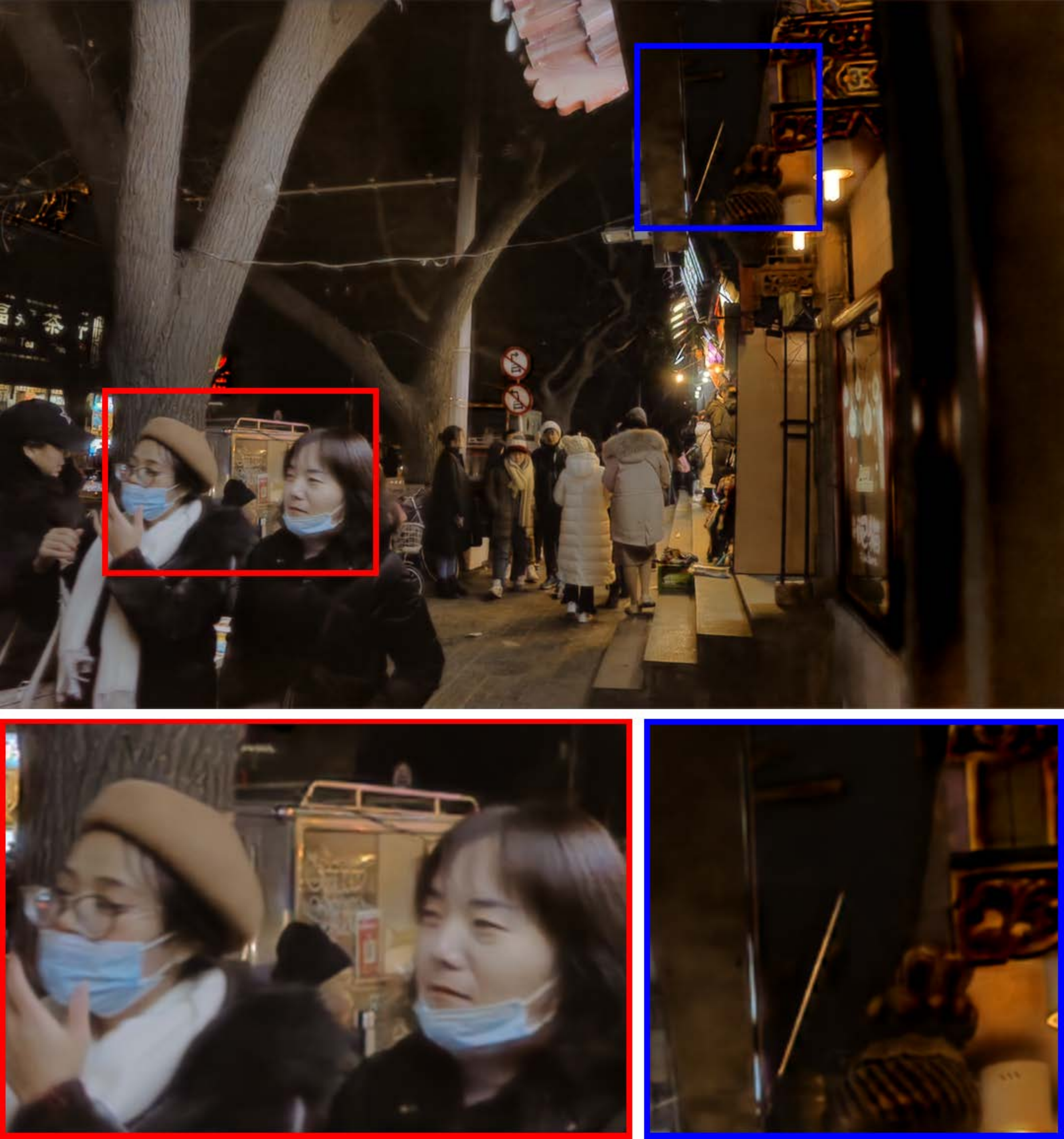}&
		\includegraphics[width=0.116\linewidth]{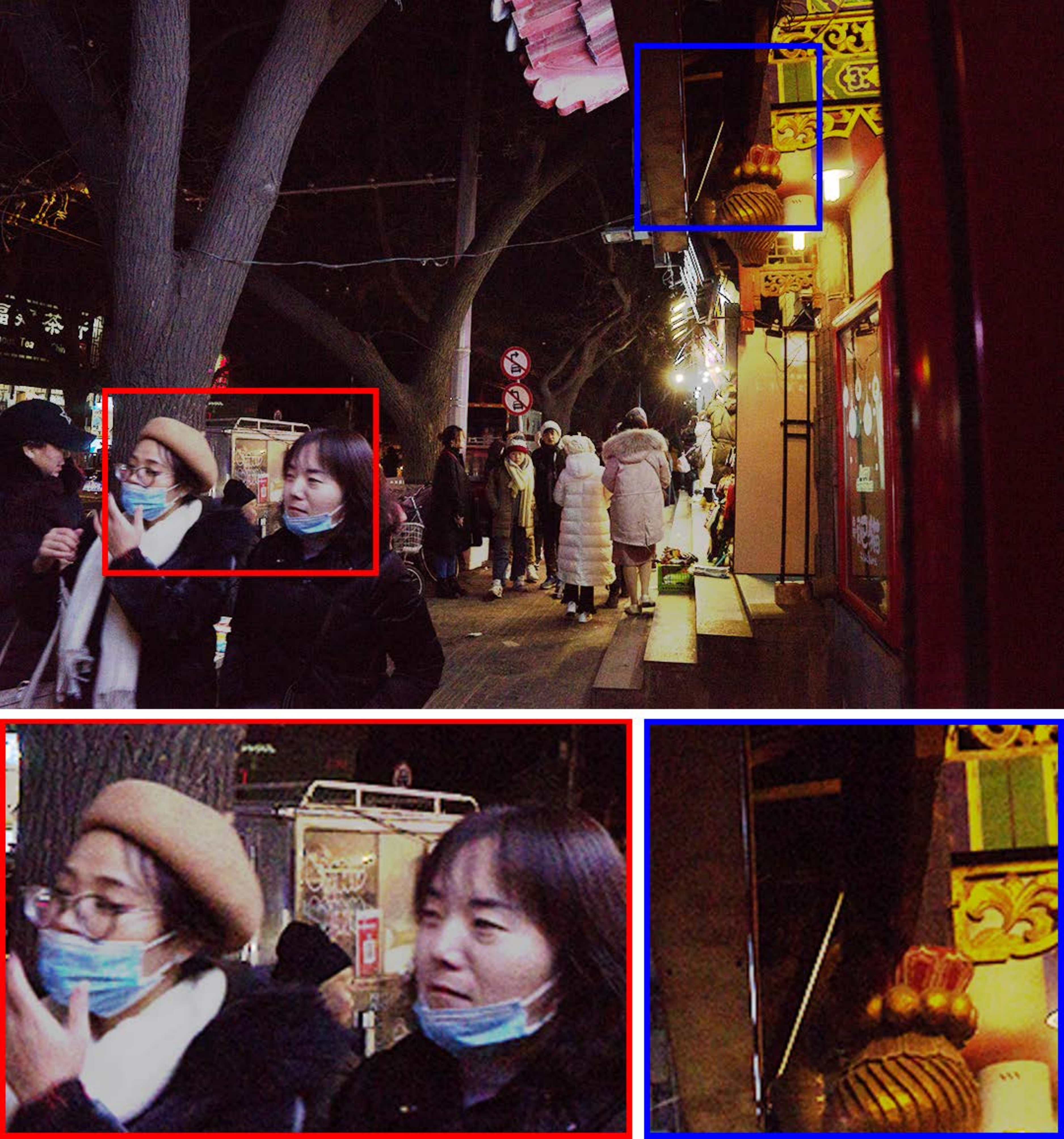}\\
		\footnotesize{Input}&\footnotesize RetinexNet&\footnotesize{KinD}&\footnotesize DeepUPE&\footnotesize{ZeroDCE}&\footnotesize FIDE&\footnotesize DRBN&\footnotesize Ours\\
	\end{tabular}
	\caption{Visual results of state-of-the-art methods and our RUAS on the DARK FACE dataset. Red and blue boxes indicate the obvious differences. }
	\label{fig:DarkFace}
\end{figure*}

\section{Experiments Results}\label{sec: experiments}
In this section, we performed experiments on the task of low-level LLVs including implementation details, algorithmic analyses, and comparison with other state-of-the-art methods. 
\subsection{Implementation Details}
We evaluated the performance of low-light image enhancement by adopting two widely-used datasets including MIT-Adobe 5K~\cite{fivek} (contained 5000 images with five expert-retouched reference images, in which the reference images retouched by expert C are usually used for calculating the numerical scores) and LOL datasets~\cite{Chen2018Retinex} (included 789 pairs of low-light and normal-light reference images).

We sampled 500 underexposure images from MIT-Adobe 5K for searching and training, and sampled 100 image pairs for testing. 
For LOL Dataset, 100 image pairs were randomly sampled for evaluating and the remaining 689 low-light images are used for searching and training. 
We adopted the well-known PSNR, SSIM and LPIPS~\cite{zhang2018perceptual} as our evaluated metrics.
We also evaluated the visual performance in the DarkFace~\cite{yang2020advancing} and ExtremelyDarkFace (used as the sub-challenge in the CVPR 2020 UG2+Challenge\footnote{\url{http://cvpr2020.ug2challenge.org/dataset20_t1.html}}) datasets. 

In the architecture search phase, we considered the same search space (with 3 fundamental cell structures) for SM and TM, but define their cells with different channel widths (i.e., 3 for SM and 6 for TM). As for the numerical parameters, we set the maximum epoch as 20, the batch size as 1, and chose the initial learning rate as $3\times10^{-4}$. The momentum parameter was randomly sampled from (0.5, 0.999) and the weight decay was set as $10^{-3}$. As for the training phase (with searched architecture), we used the same training losses as that in the search phase.

To fully prove the effectiveness of low-light image enhancement, we compared RUAS with eleven recently-proposed state-of-the-art low-light image enhancement approaches including MBLLEN~\cite{lv2019attention}, GLADNet~\cite{wang2018gladnet}, RetinexNet~\cite{Chen2018Retinex}, DeepUPE~\cite{wang2019underexposed}, SSIENet~\cite{zhang2020self}, EnGAN~\cite{jiang2019enlightengan},  ZeroDCE~\cite{guo2020zero}, FIDE~\cite{xu2020learning}, DRBN~\cite{yang2020fidelity}, and KinD~\cite{zhang2019kindling}. Additionally, we conducted all the experiments on a PC with a single TITAN X GPU and Intel Core i7-7700 3.60GHz CPU.

\begin{figure*}[t]
	\centering
	\begin{tabular}{c@{\extracolsep{0.3em}}c@{\extracolsep{0.3em}}c@{\extracolsep{0.3em}}c@{\extracolsep{0.3em}}c}
		\includegraphics[width=0.19\linewidth]{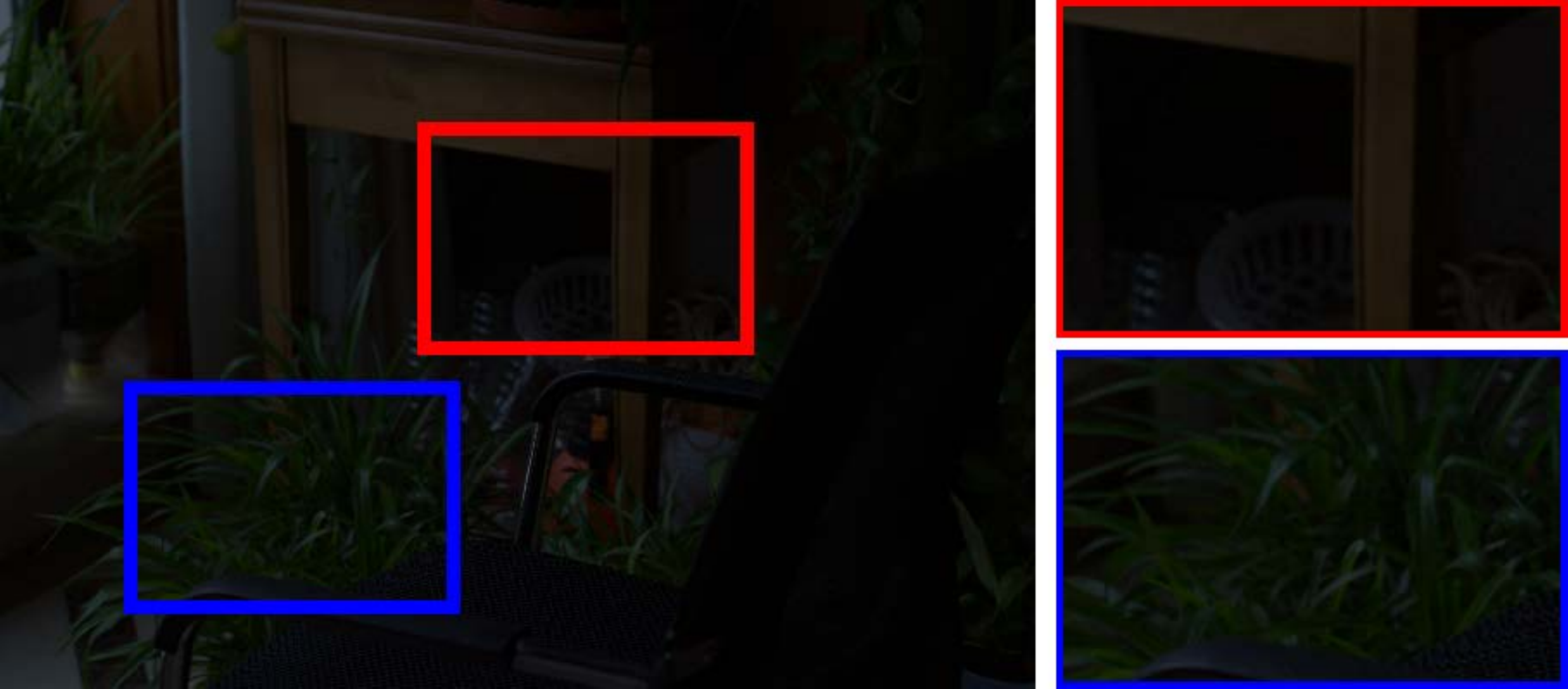}&
		\includegraphics[width=0.19\linewidth]{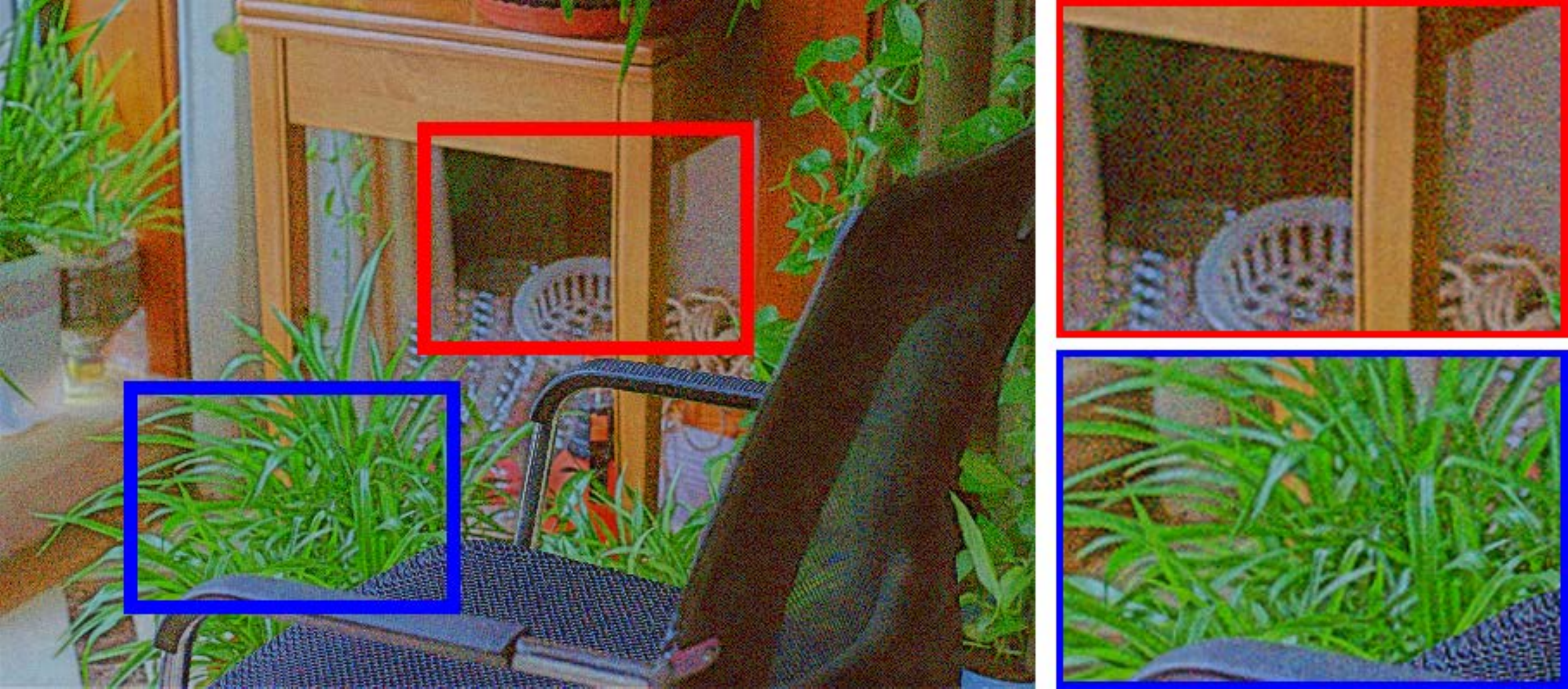}&
		\includegraphics[width=0.19\linewidth]{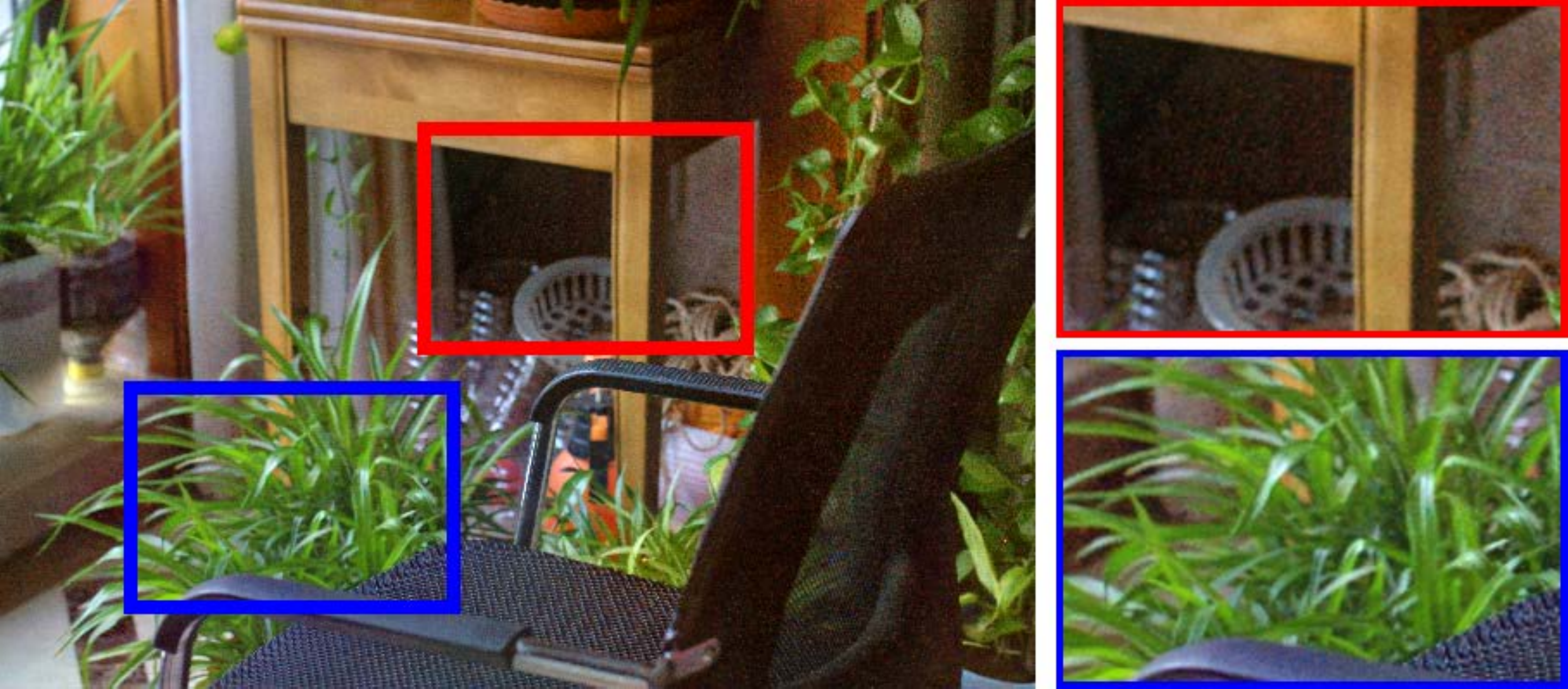}&
		\includegraphics[width=0.19\linewidth]{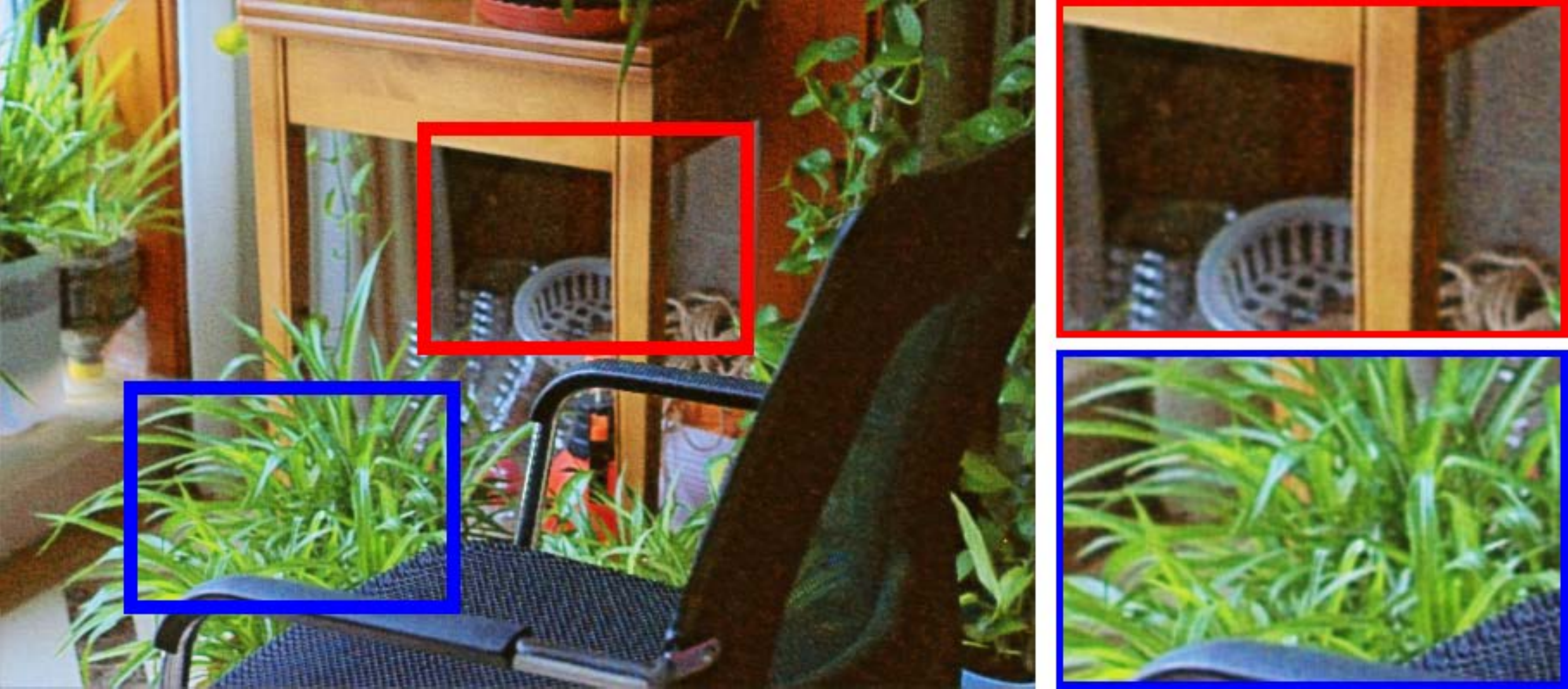}&
		\includegraphics[width=0.19\linewidth]{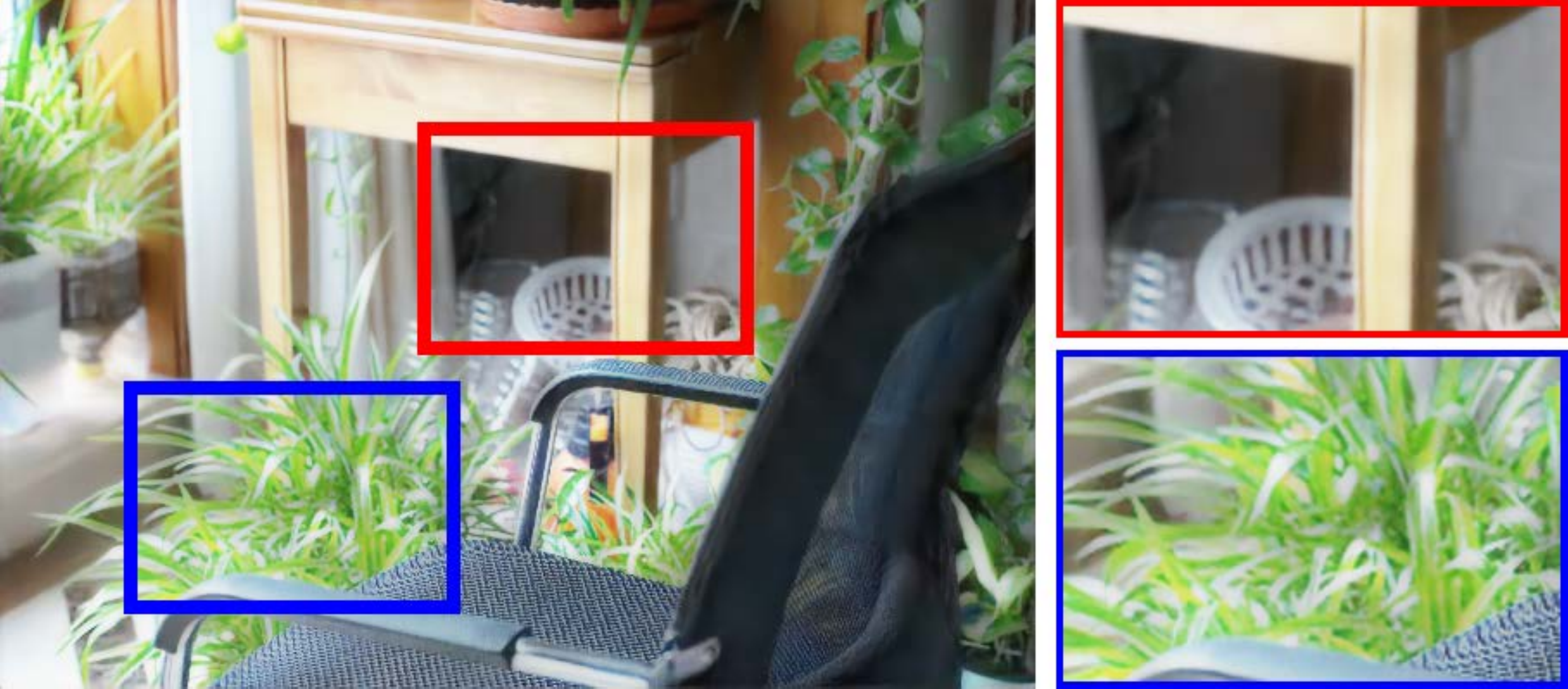}\\
		\footnotesize Input&\footnotesize{RetinexNet}&\footnotesize EnGAN&\footnotesize{SSIENet}&\footnotesize KinD\\
		\includegraphics[width=0.19\linewidth]{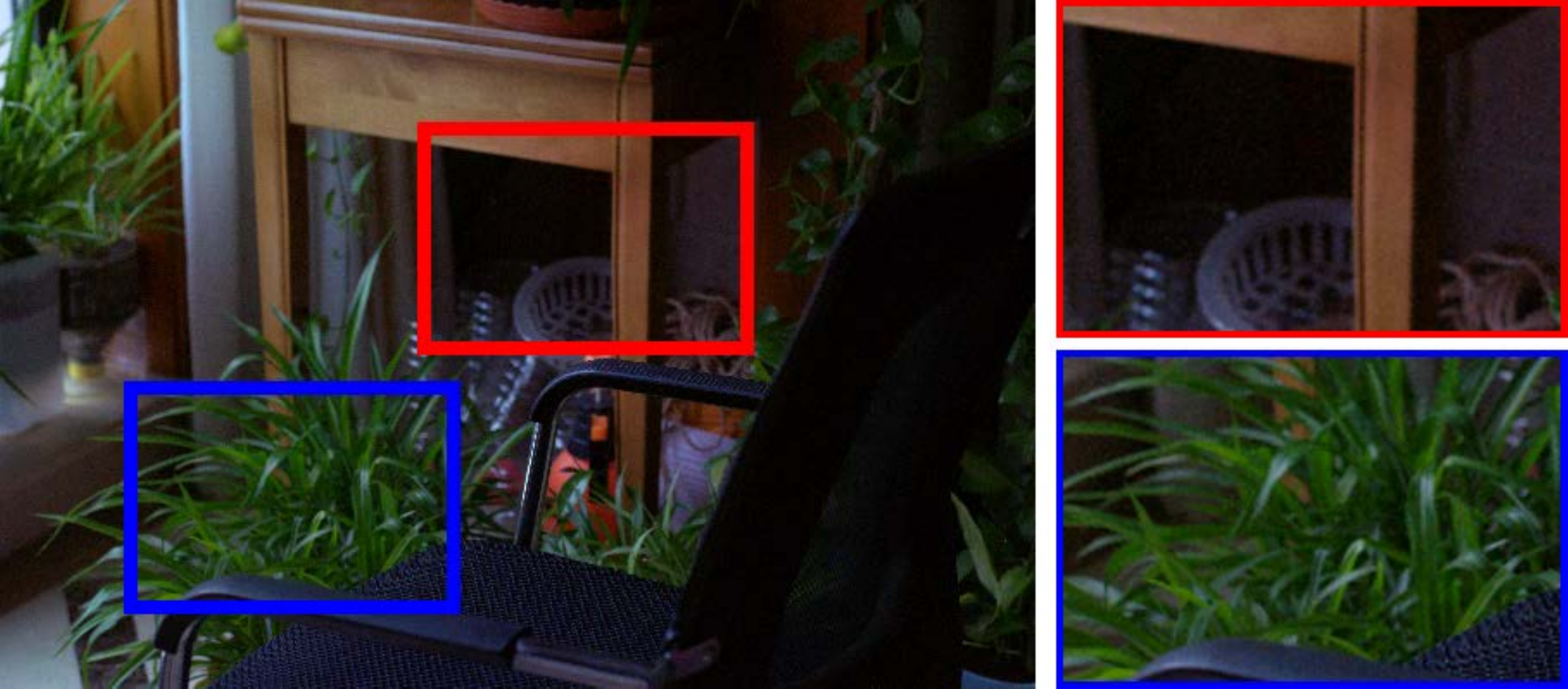}&
		\includegraphics[width=0.19\linewidth]{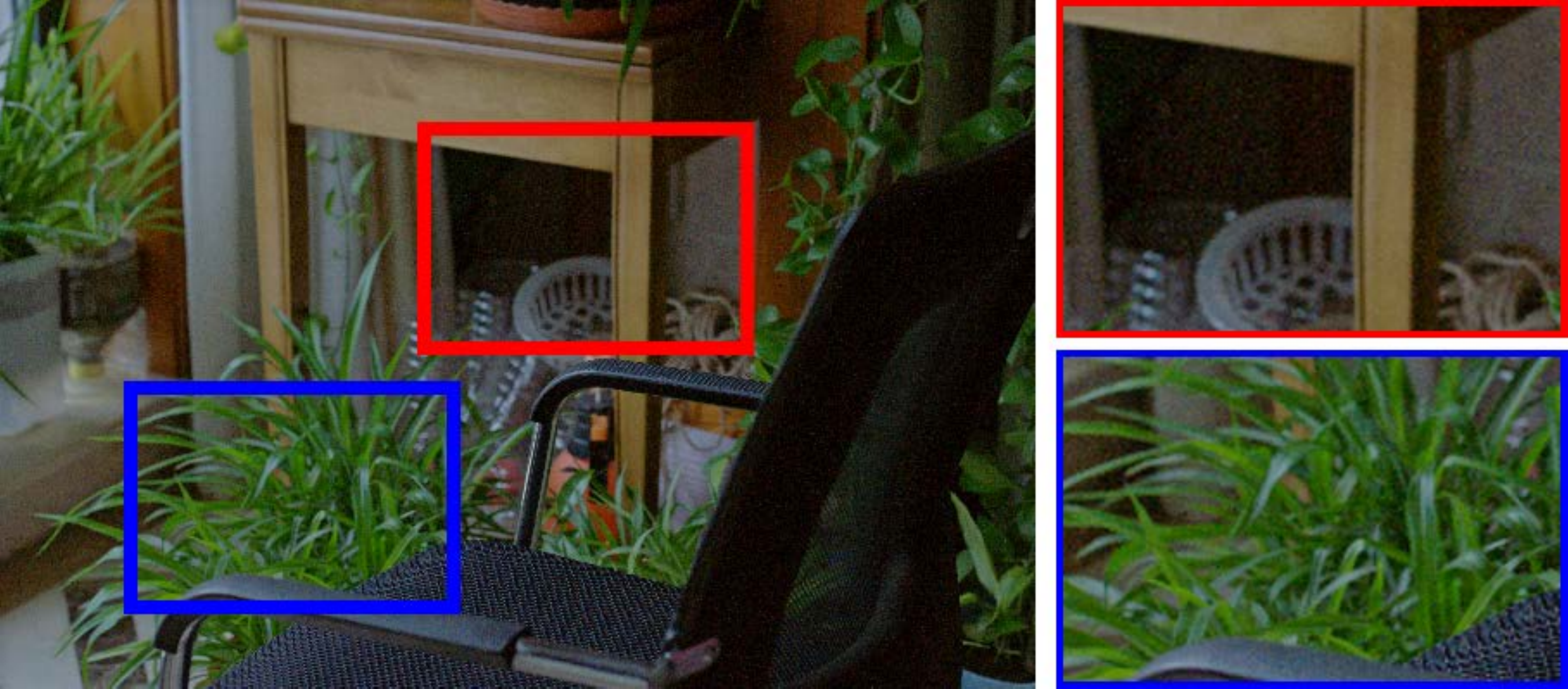}&
		\includegraphics[width=0.19\linewidth]{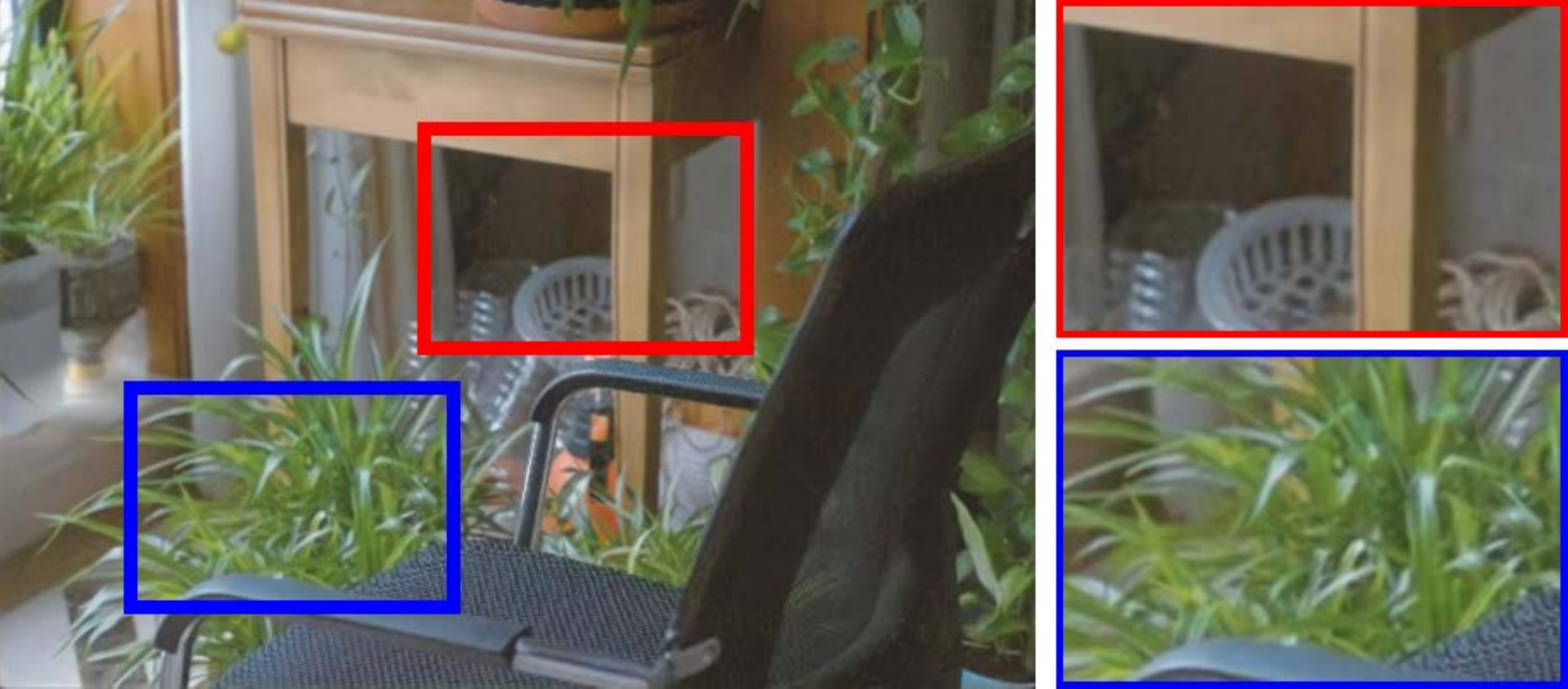}&
		\includegraphics[width=0.19\linewidth]{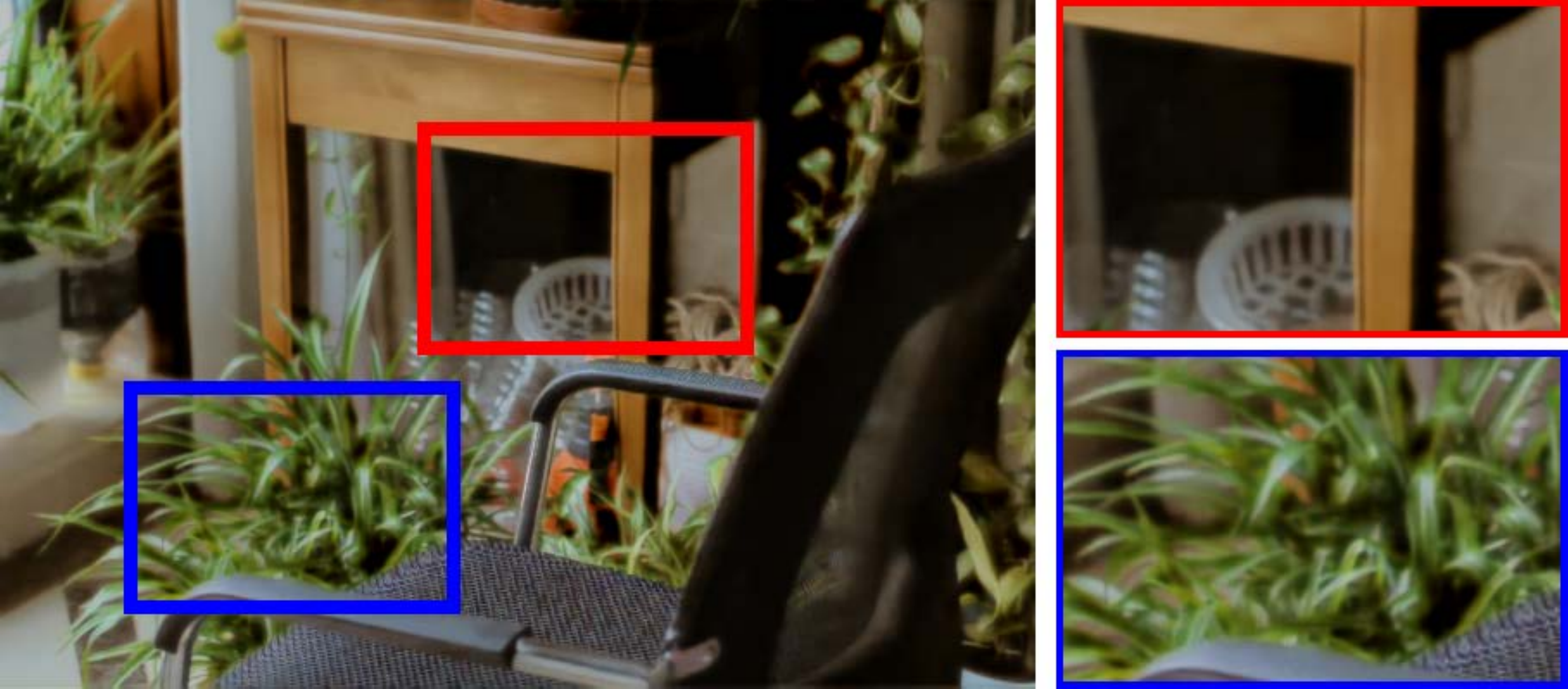}&
		\includegraphics[width=0.19\linewidth]{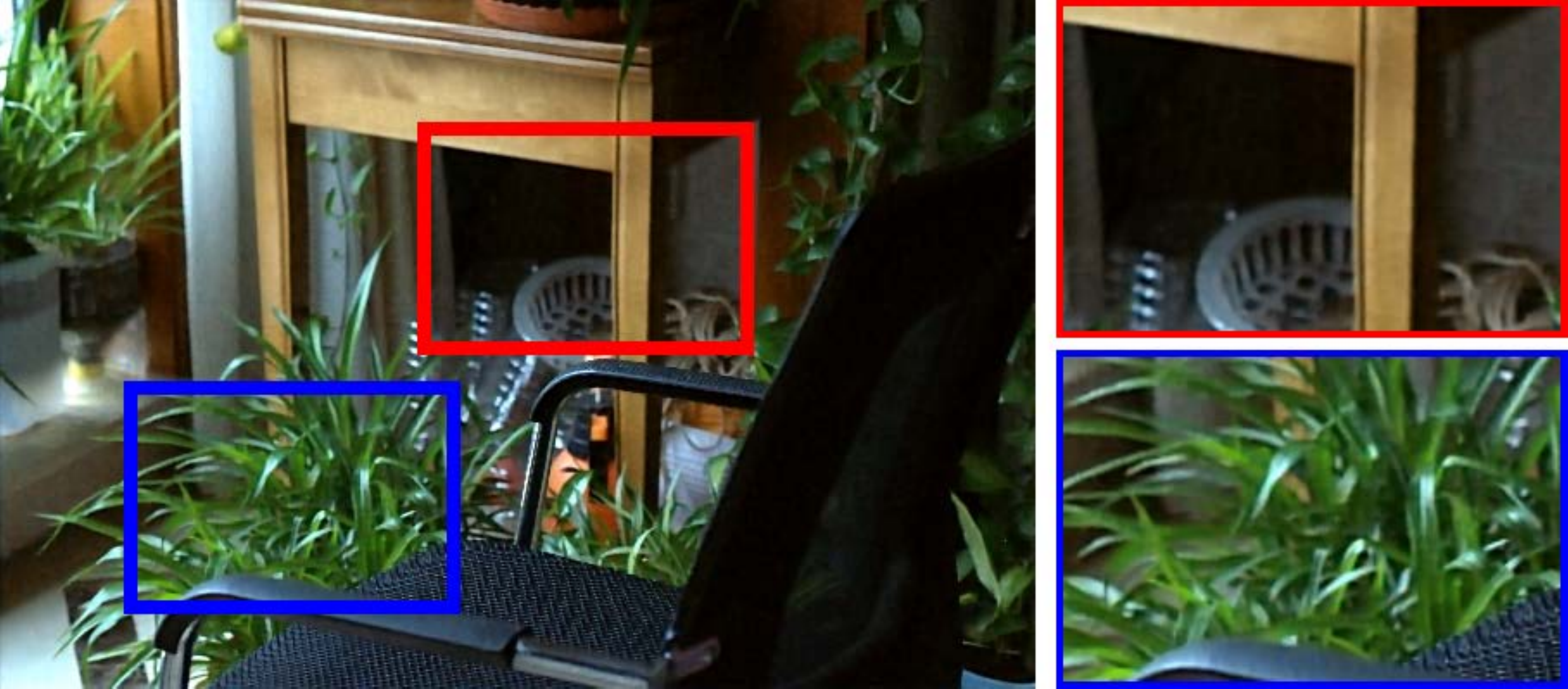}\\
		\footnotesize DeepUPE&\footnotesize{ZeroDCE}&\footnotesize FIDE&\footnotesize DRBN&\footnotesize {Ours}\\	
		\includegraphics[width=0.19\linewidth]{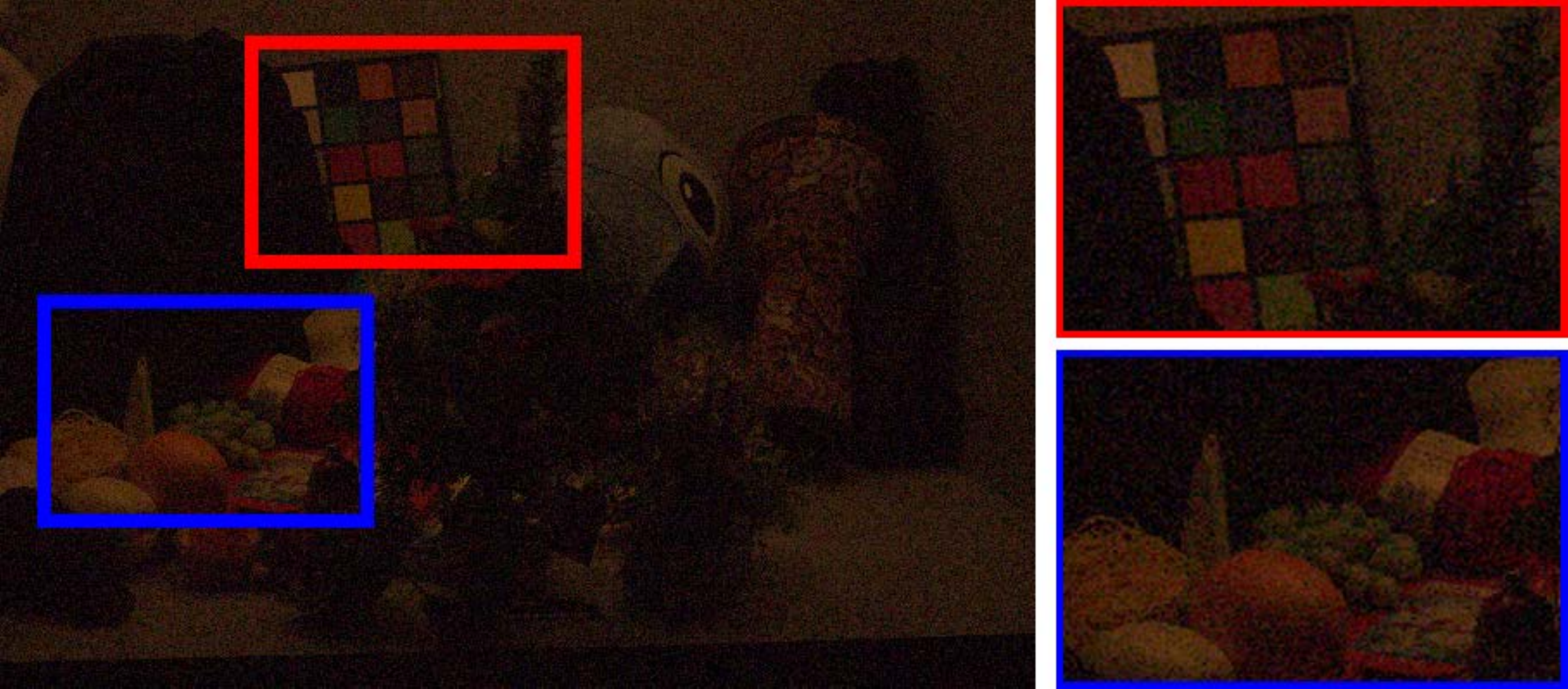}&
		\includegraphics[width=0.19\linewidth]{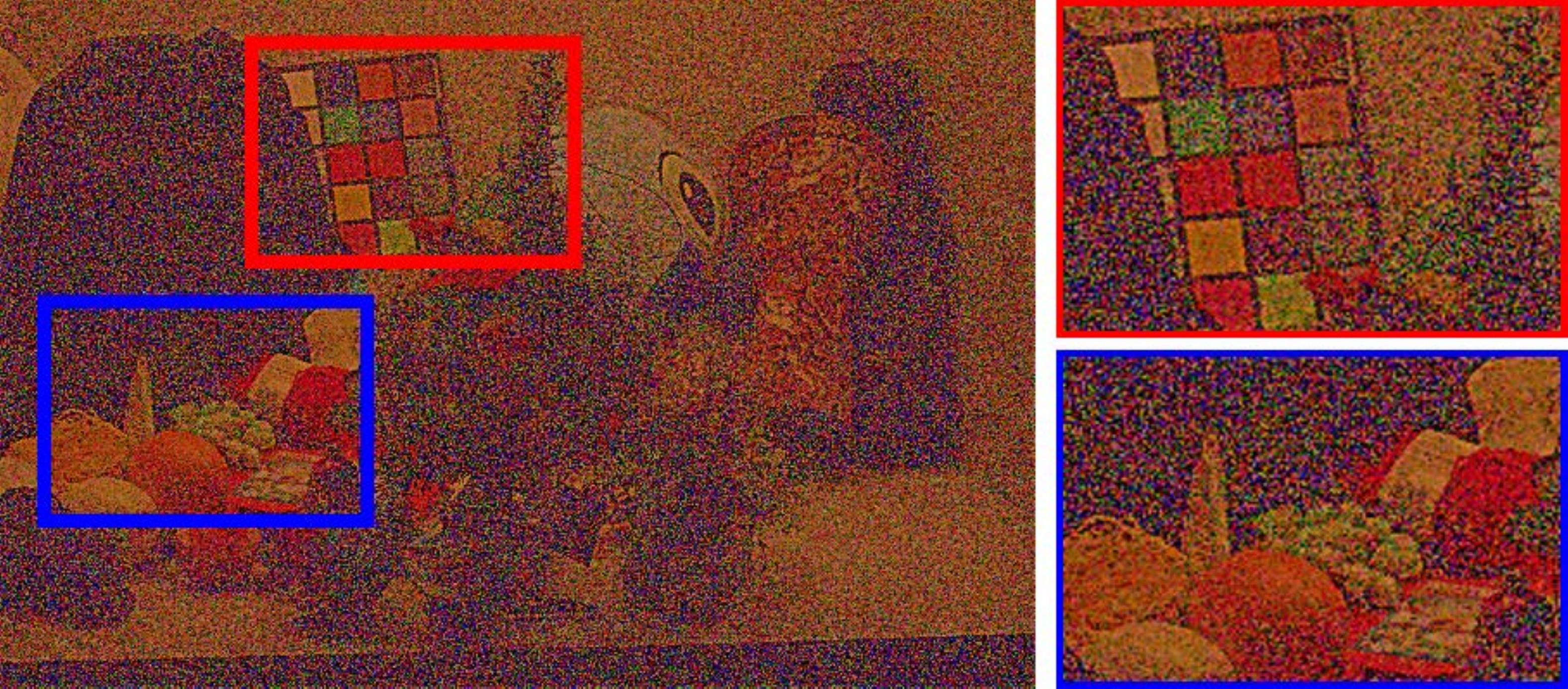}&
		\includegraphics[width=0.19\linewidth]{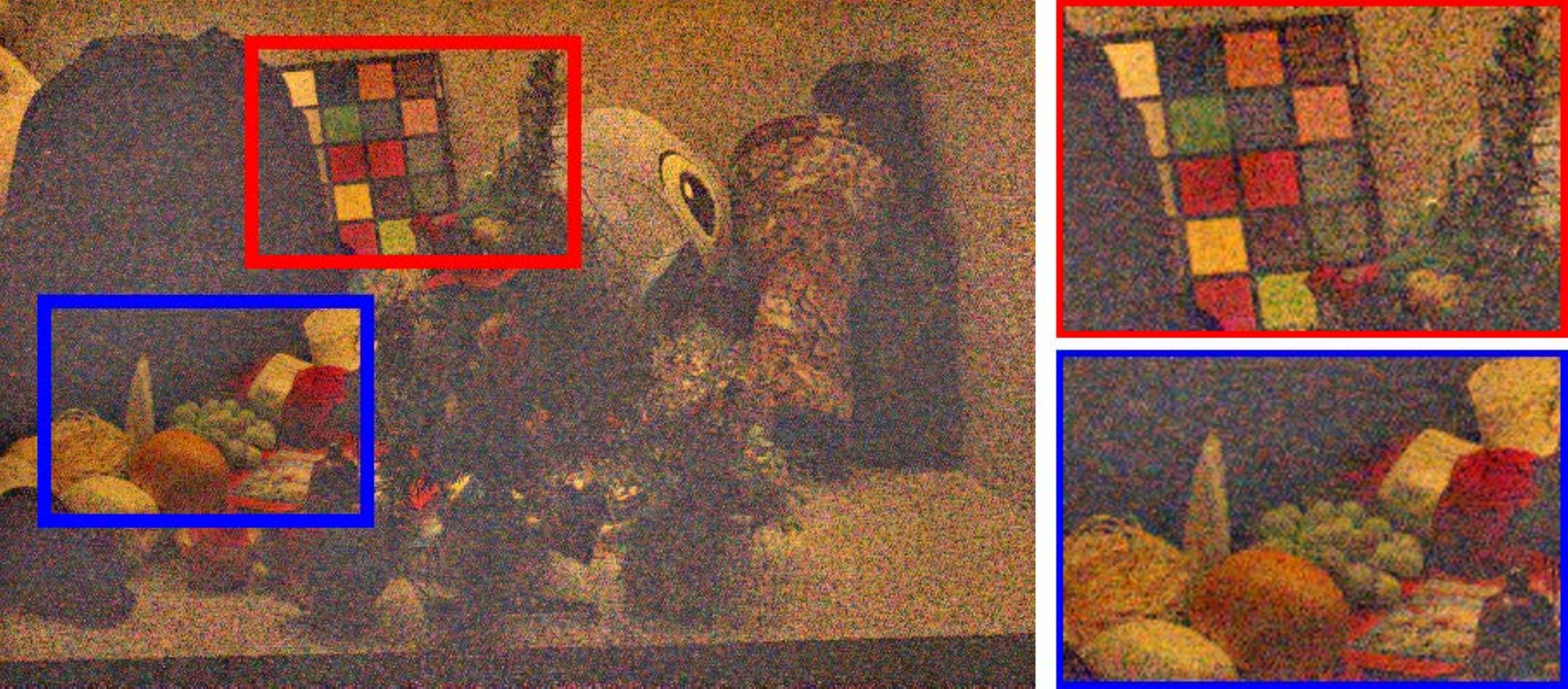}&
		\includegraphics[width=0.19\linewidth]{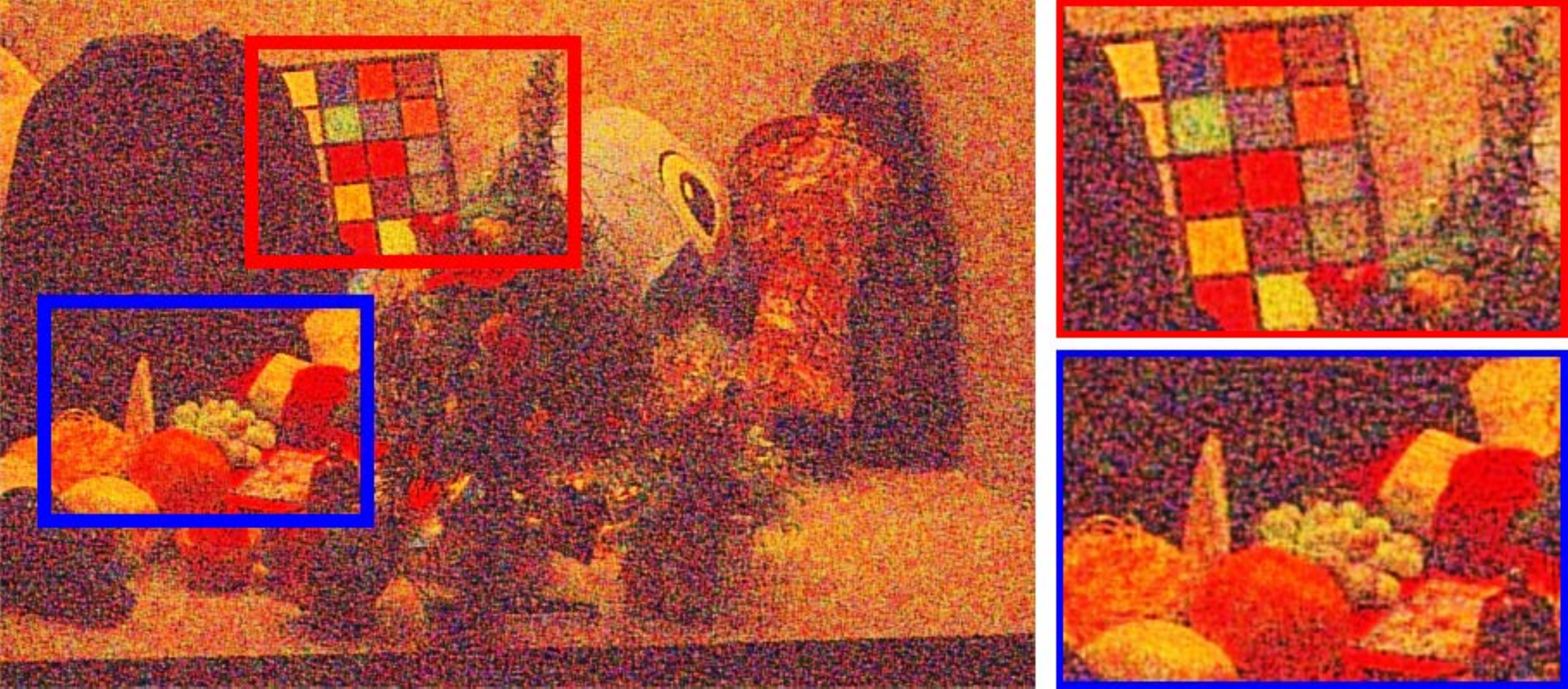}&
		\includegraphics[width=0.19\linewidth]{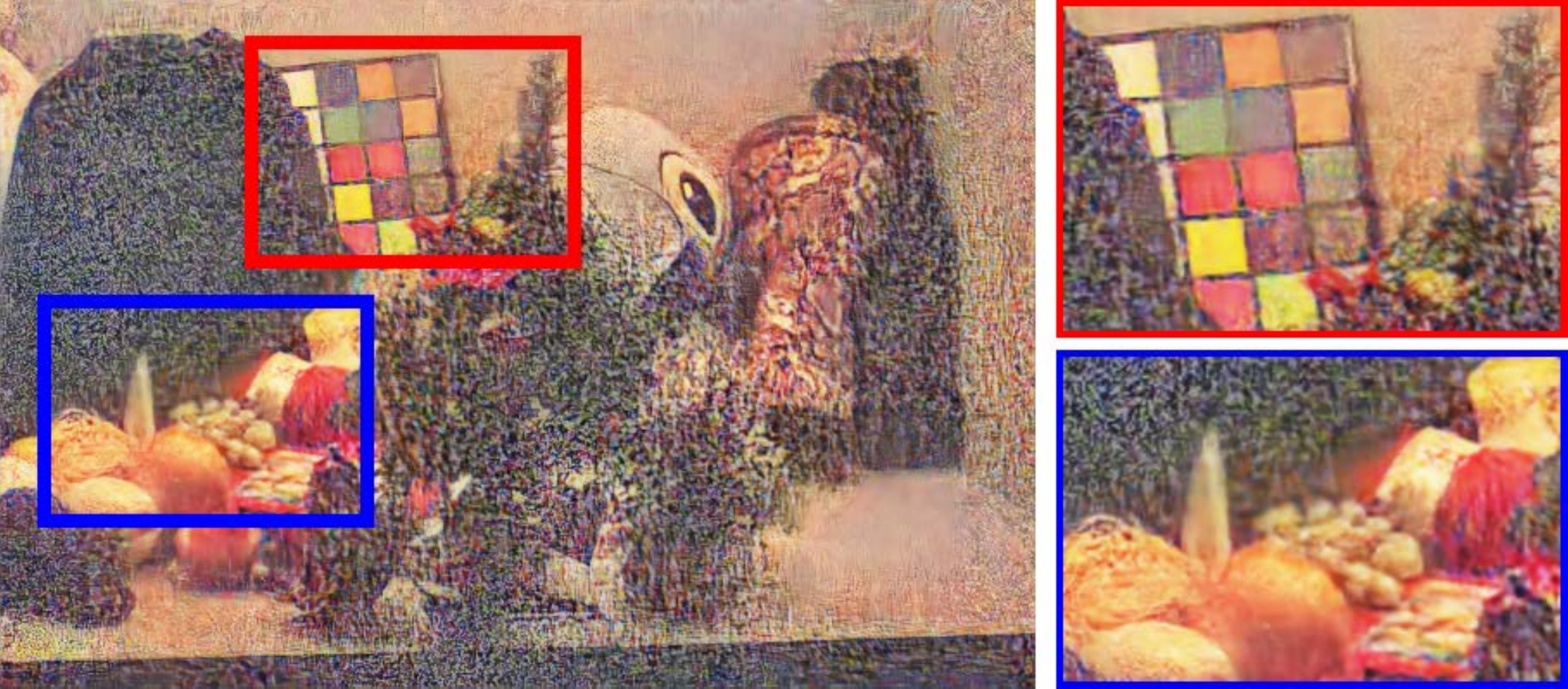}\\
		\footnotesize Input&\footnotesize{RetinexNet}&\footnotesize EnGAN&\footnotesize{SSIENet}&\footnotesize KinD\\
		\includegraphics[width=0.19\linewidth]{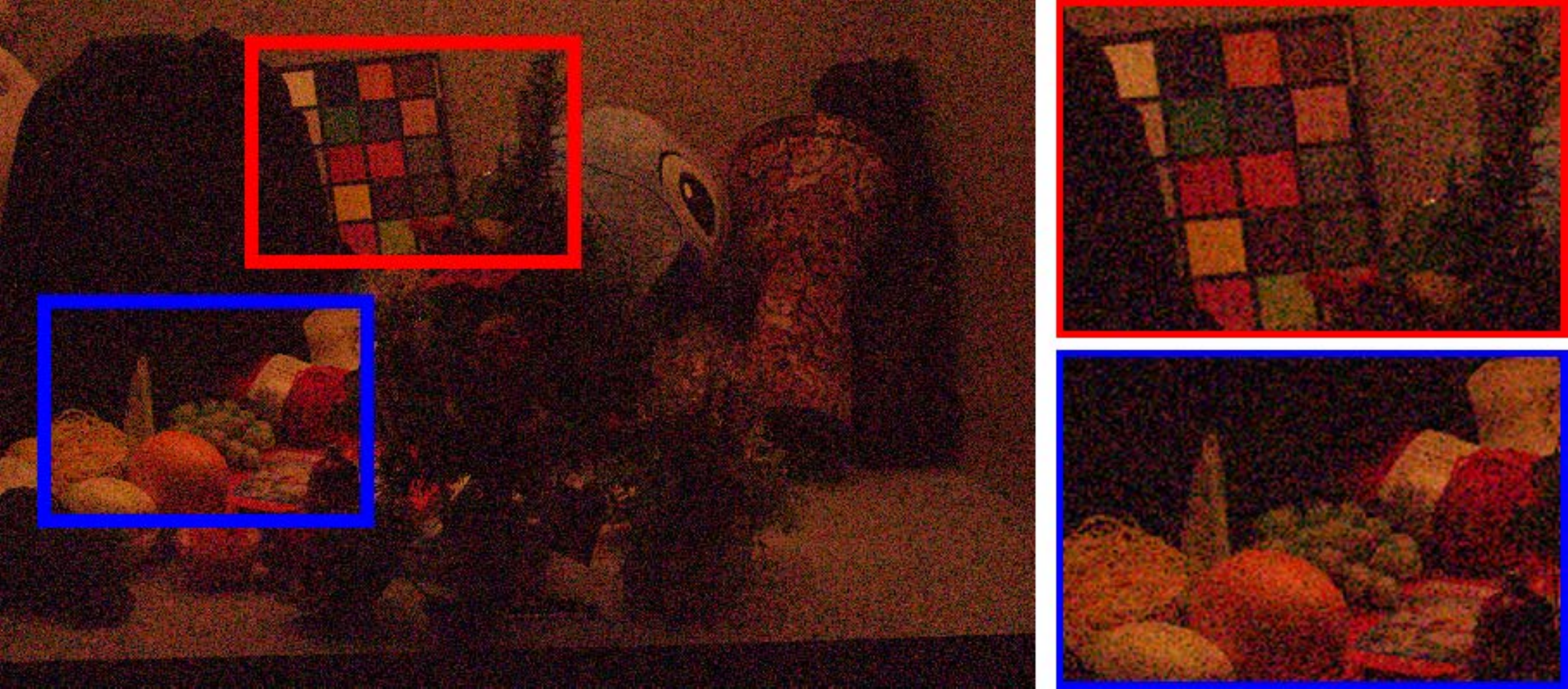}&
		\includegraphics[width=0.19\linewidth]{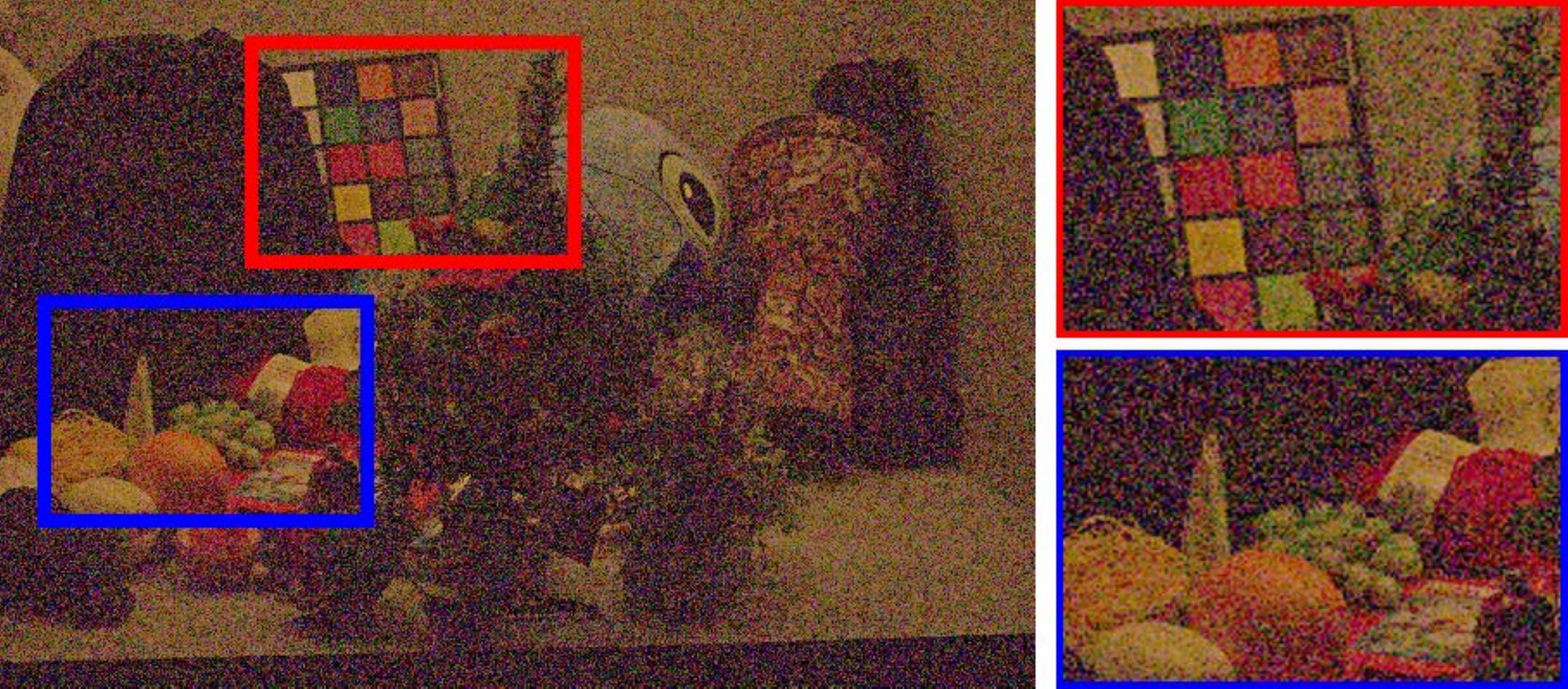}&
		\includegraphics[width=0.19\linewidth]{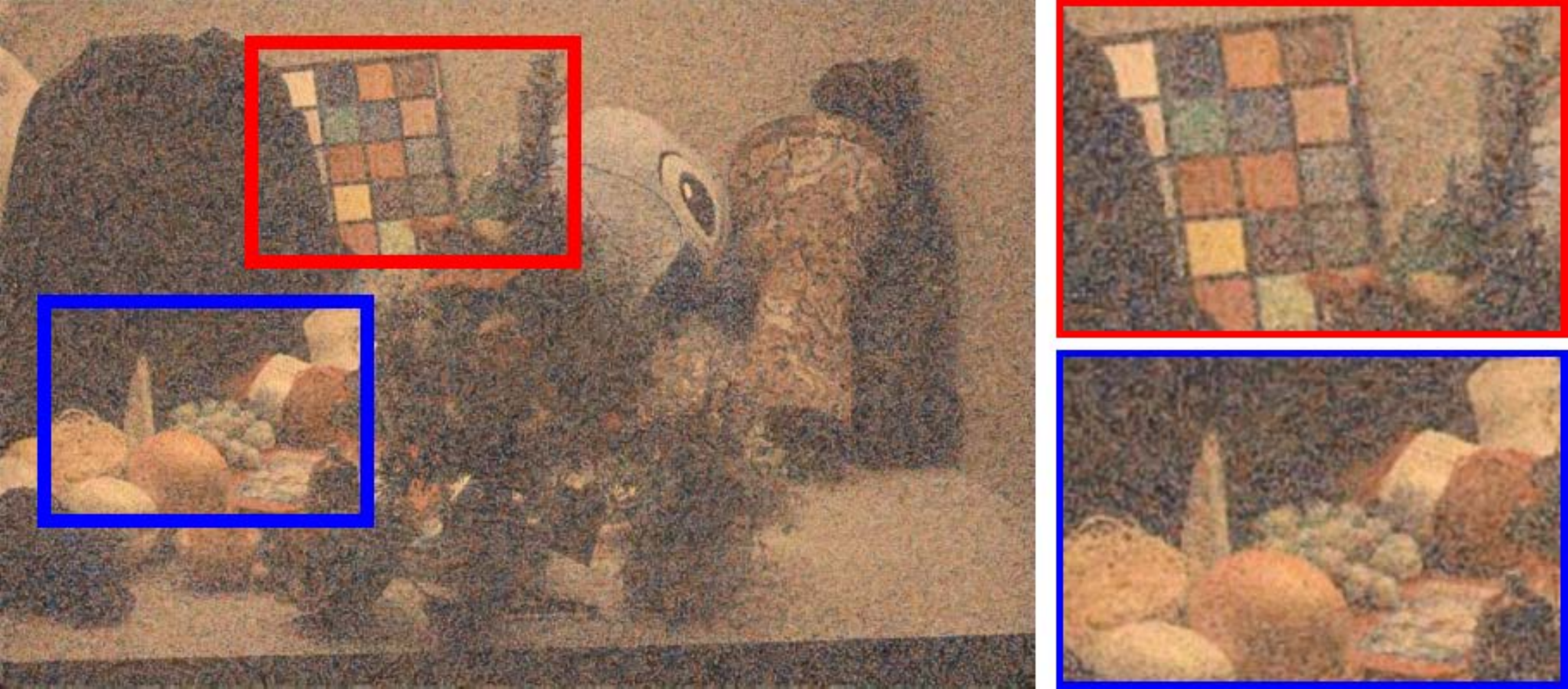}&
		\includegraphics[width=0.19\linewidth]{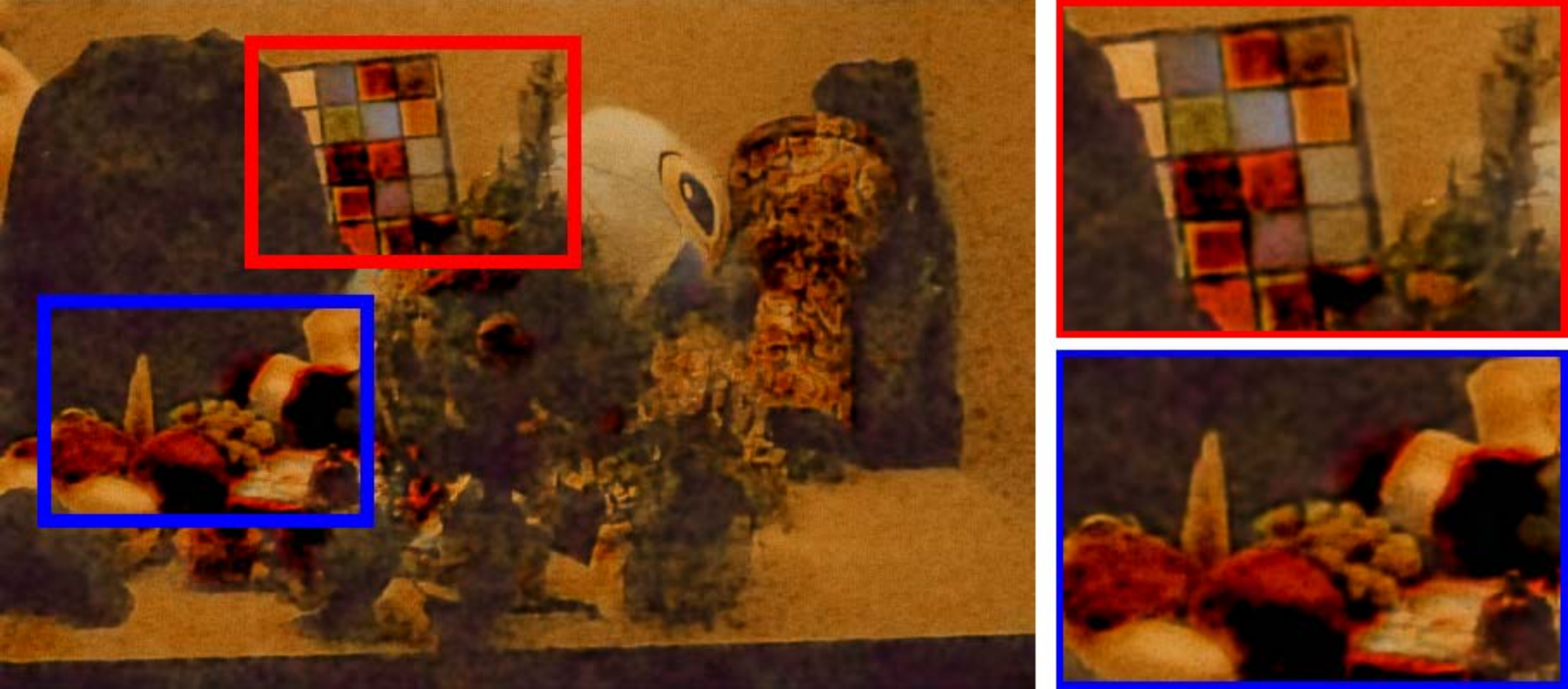}&
		\includegraphics[width=0.19\linewidth]{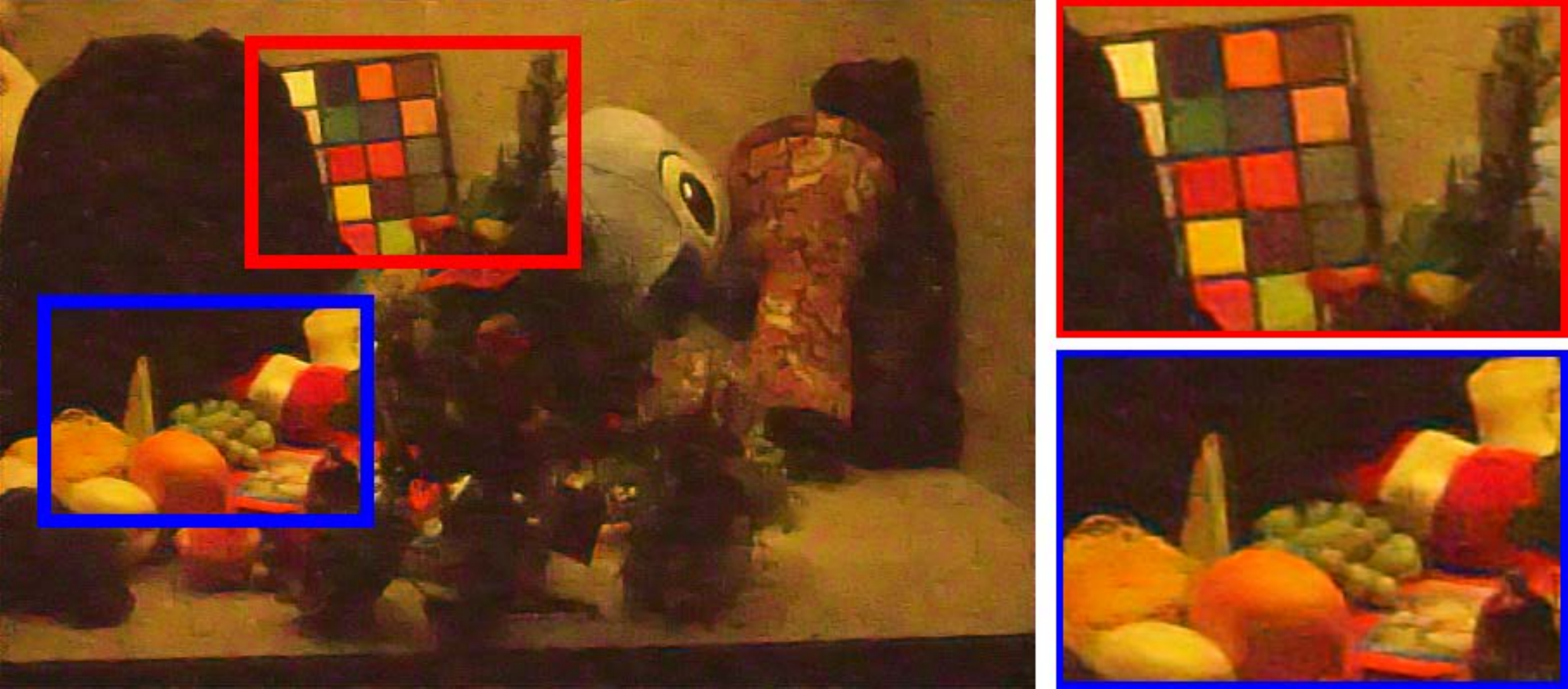}\\
		\footnotesize DeepUPE&\footnotesize{ZeroDCE}&\footnotesize FIDE&\footnotesize DRBN&\footnotesize {Ours}\\		
	\end{tabular}	
	\caption{Visual results of state-of-the-art methods and our RUAS on the LOL dataset. Red and blue boxes indicate the obvious differences. }
	\label{fig:LOL}
\end{figure*}

\begin{figure*}[t]
	\centering
	\begin{tabular}{c@{\extracolsep{0.3em}}c@{\extracolsep{0.3em}}c@{\extracolsep{0.3em}}c@{\extracolsep{0.3em}}c@{\extracolsep{0.3em}}c@{\extracolsep{0.3em}}c@{\extracolsep{0.3em}}c} 
		\includegraphics[width=0.115\linewidth]{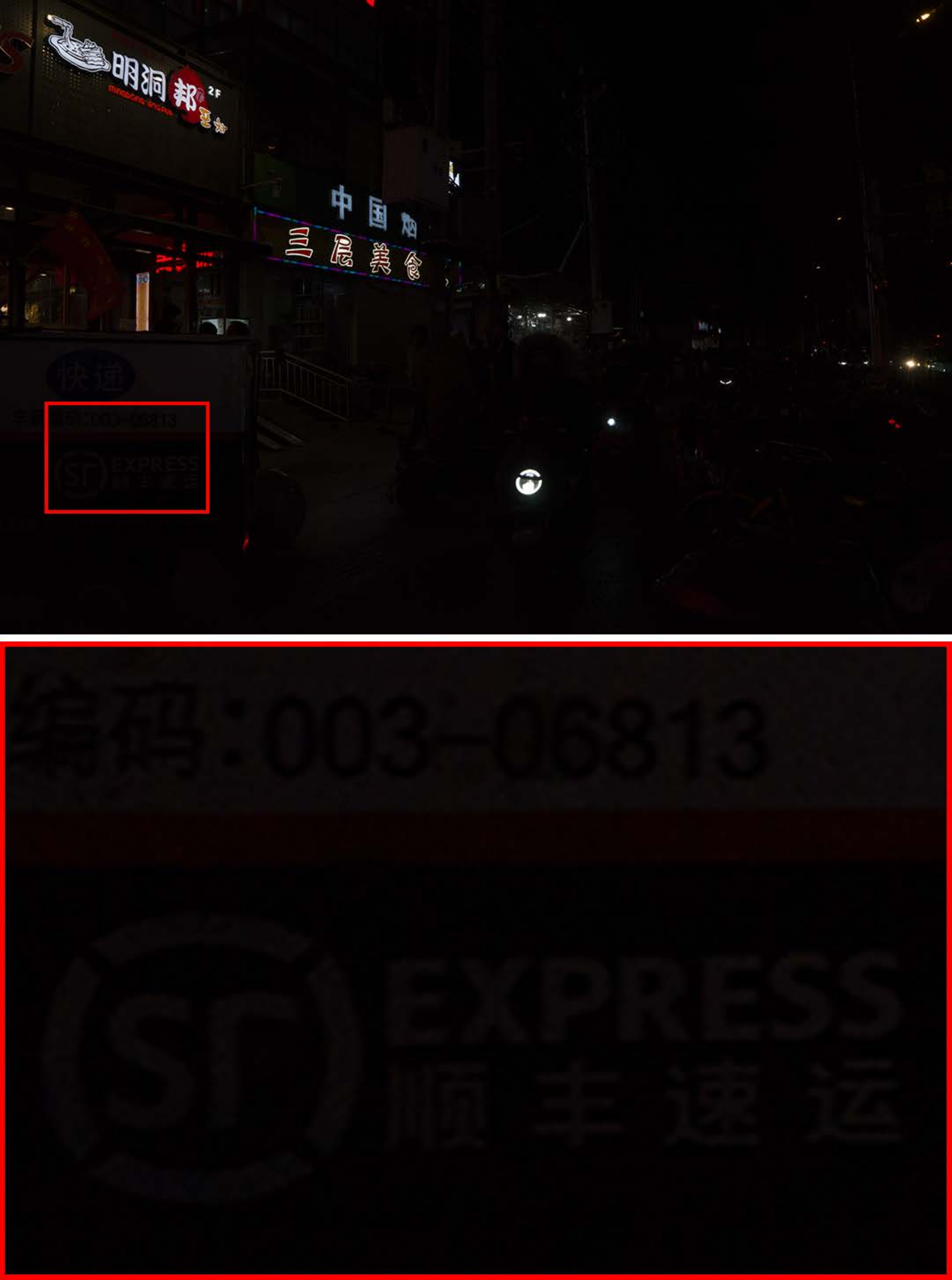}&
		\includegraphics[width=0.115\linewidth]{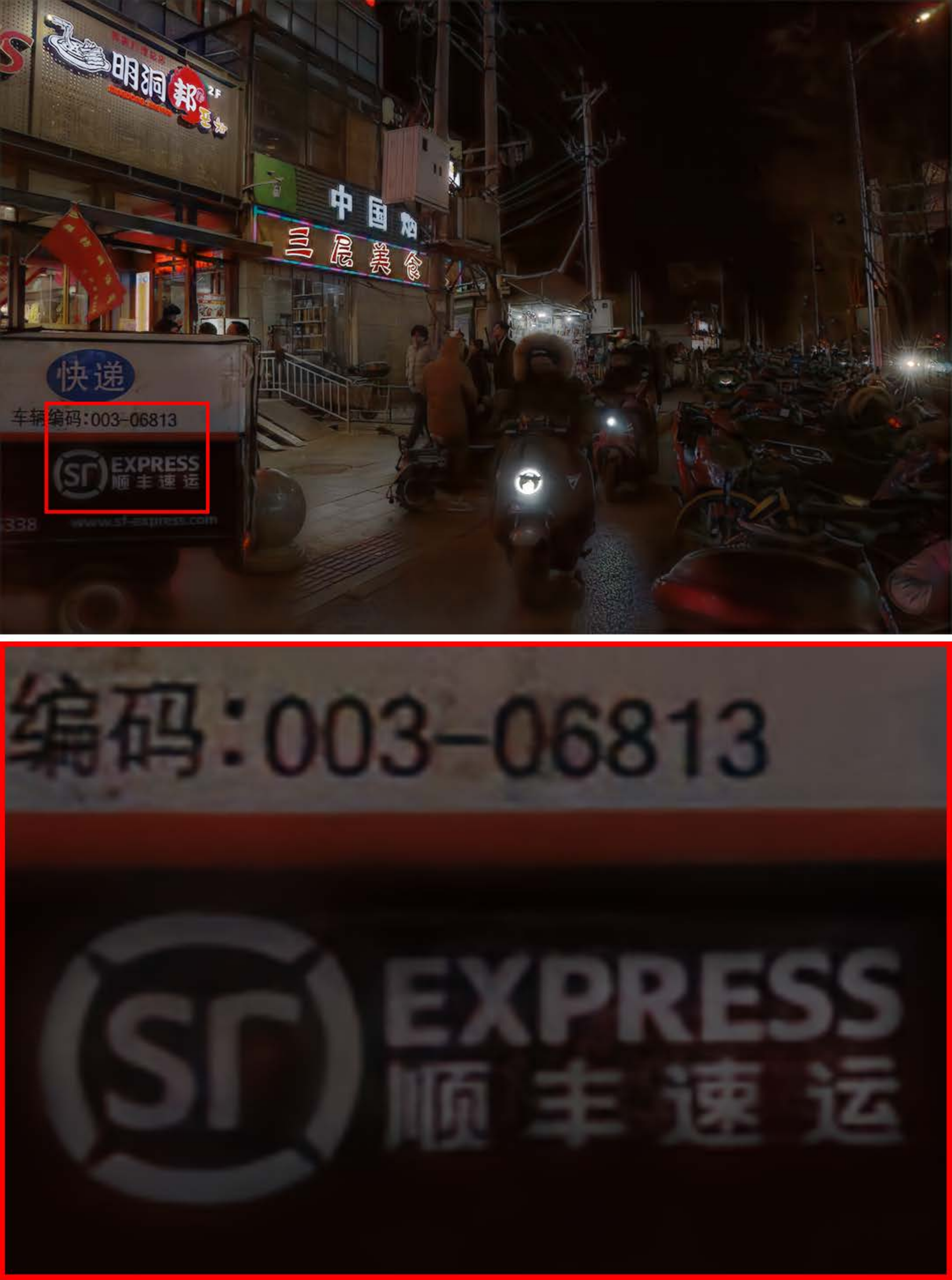}&
		\includegraphics[width=0.115\linewidth]{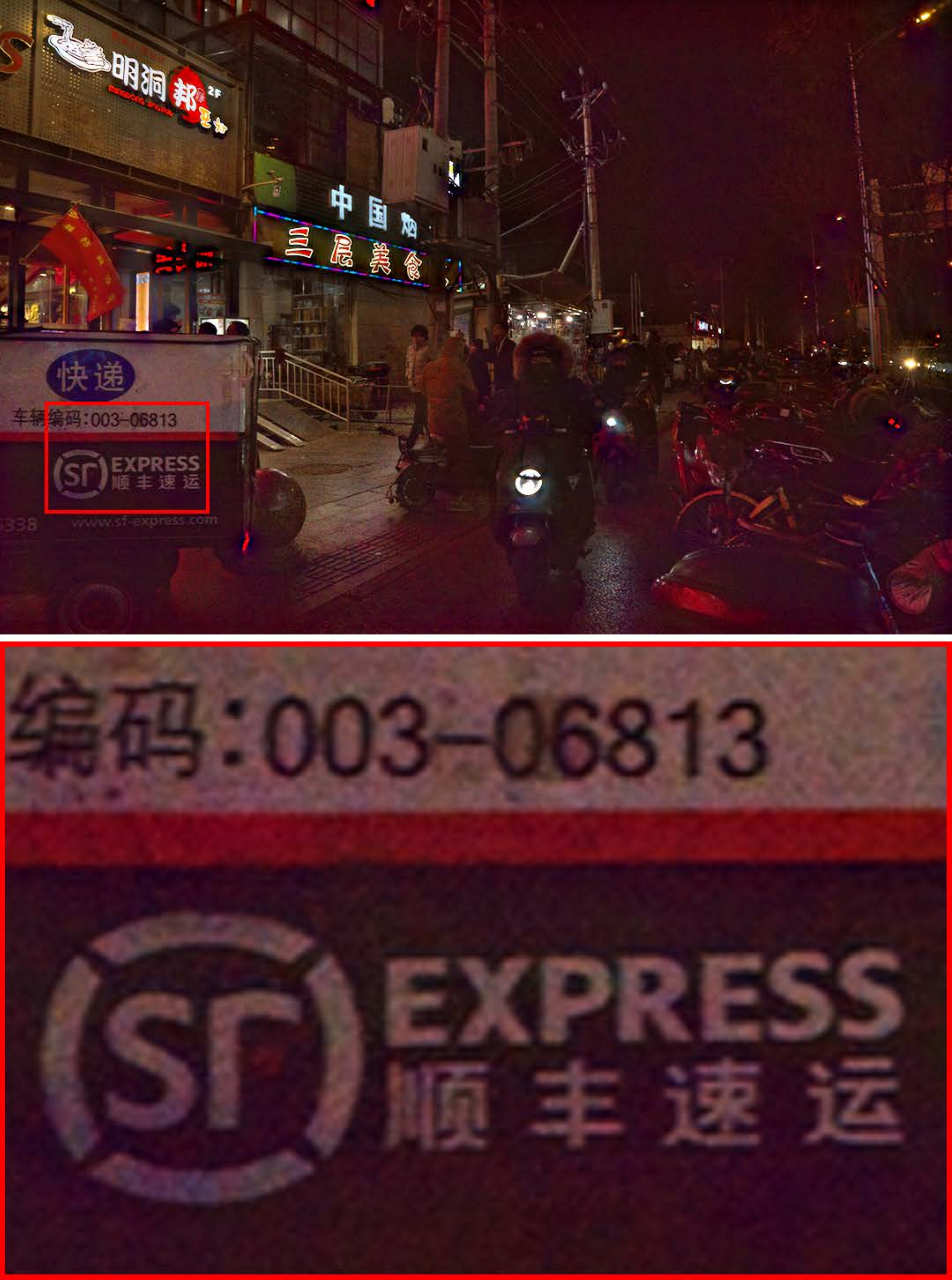}&
		\includegraphics[width=0.115\linewidth]{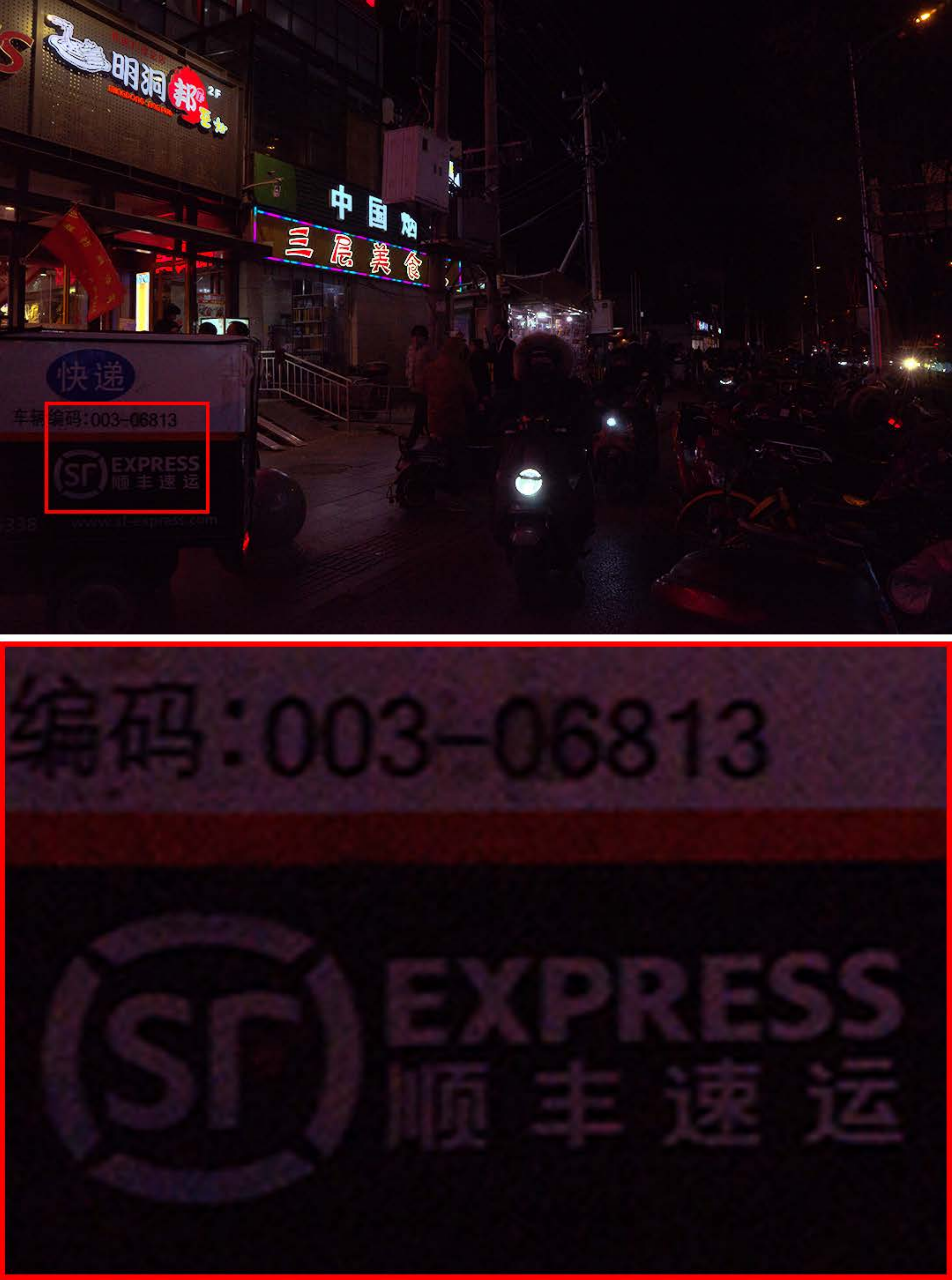}&
		\includegraphics[width=0.115\linewidth]{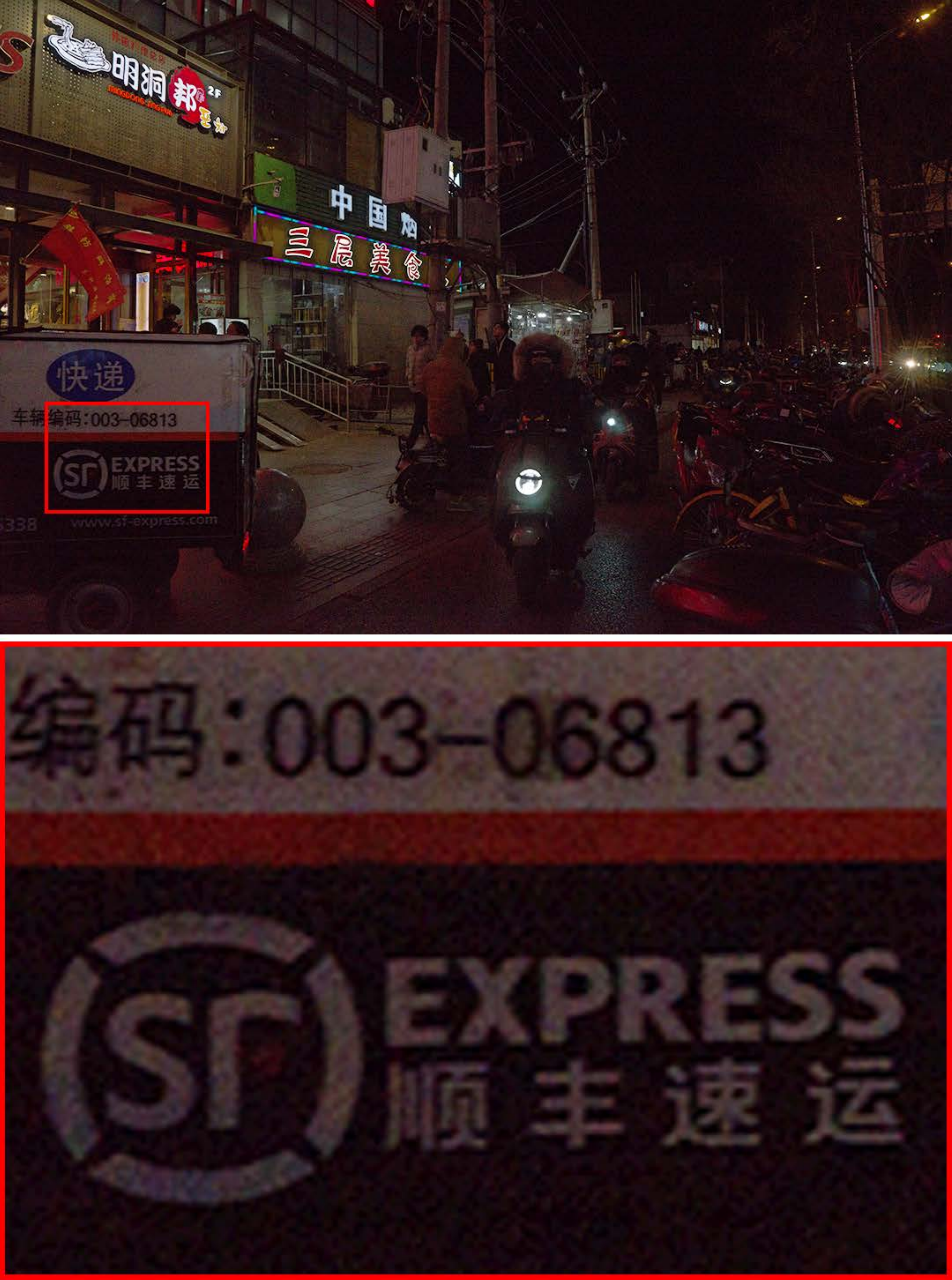}&
		\includegraphics[width=0.115\linewidth]{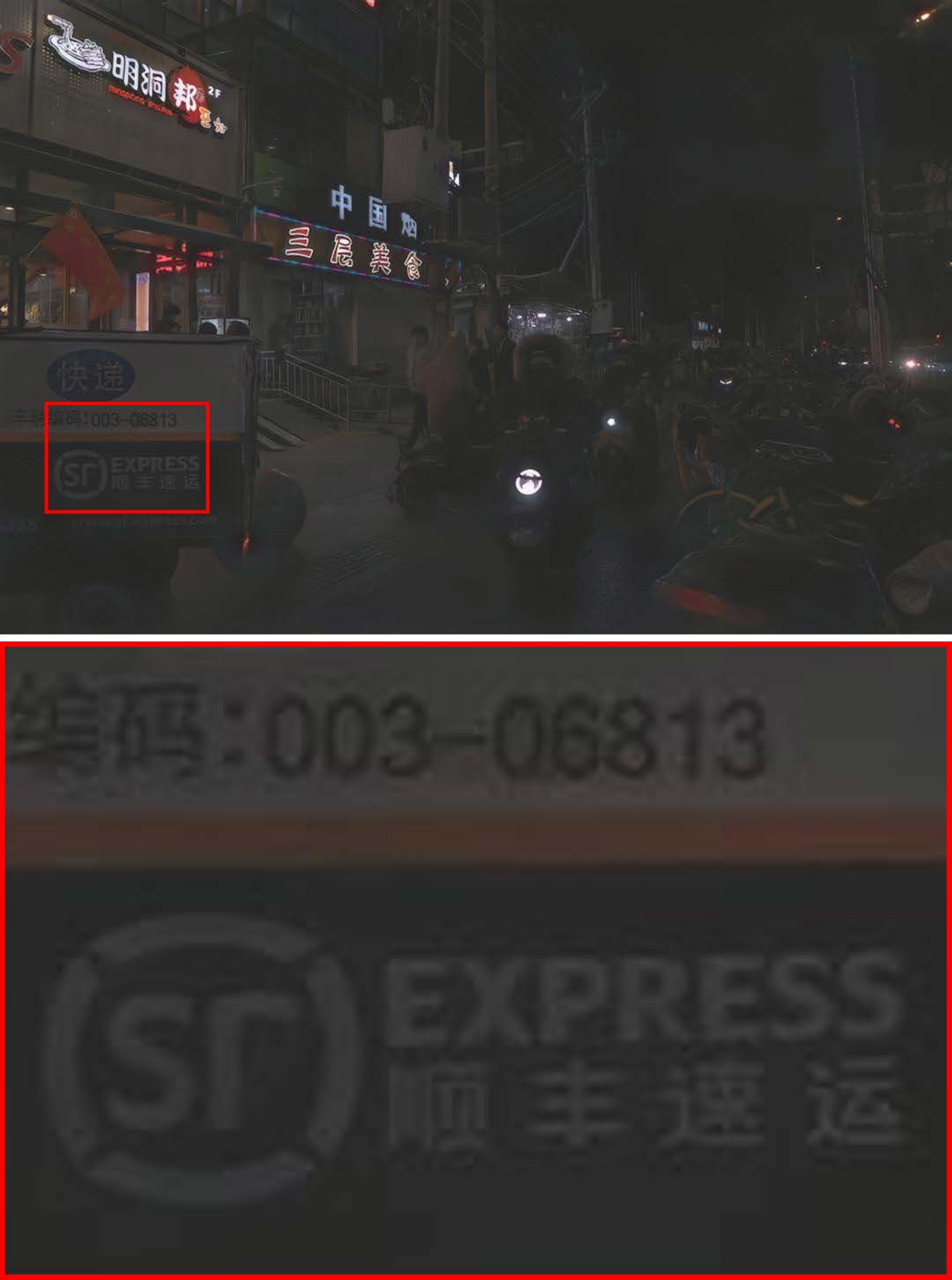}&
		\includegraphics[width=0.115\linewidth]{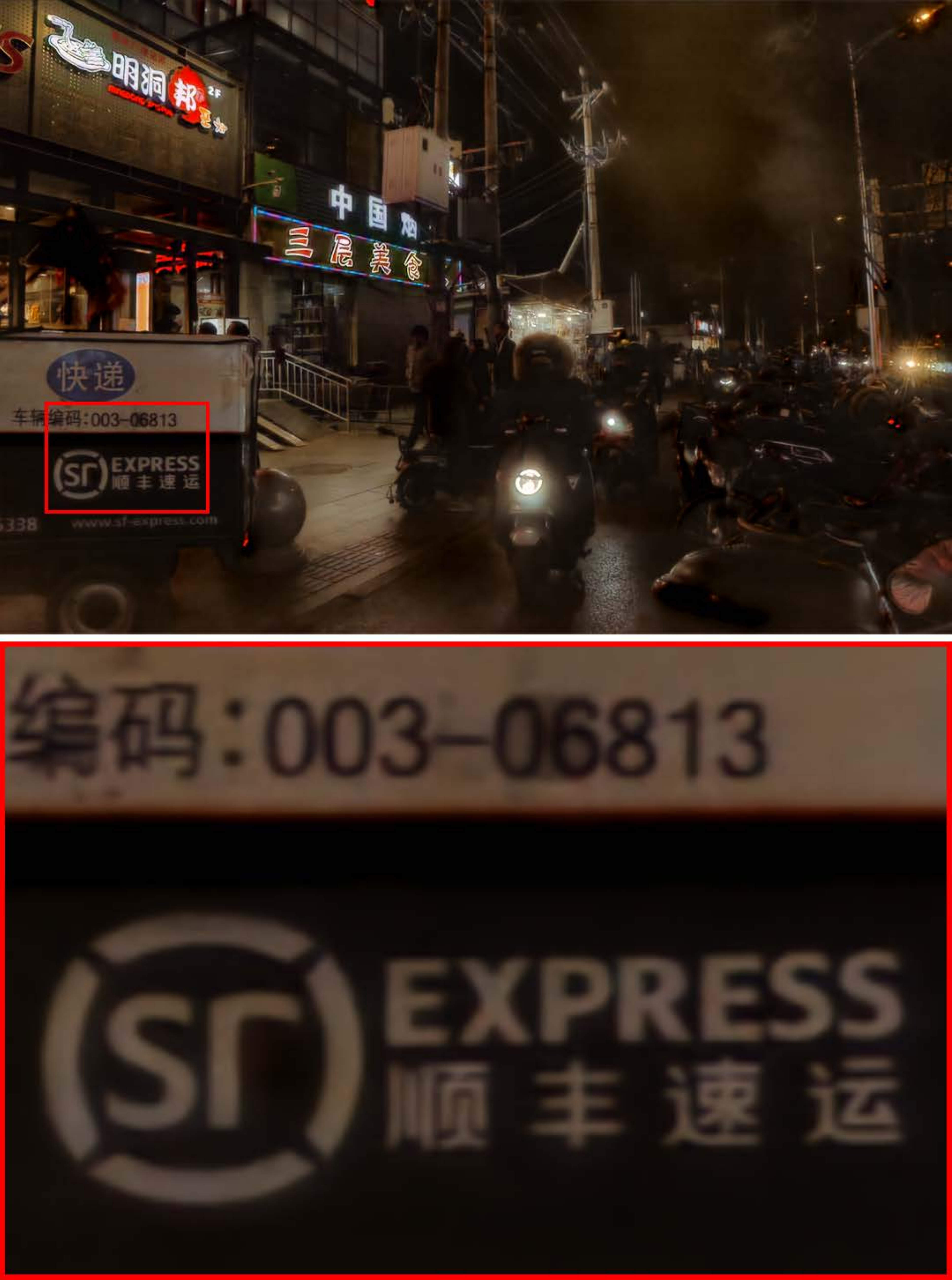}&
		\includegraphics[width=0.115\linewidth]{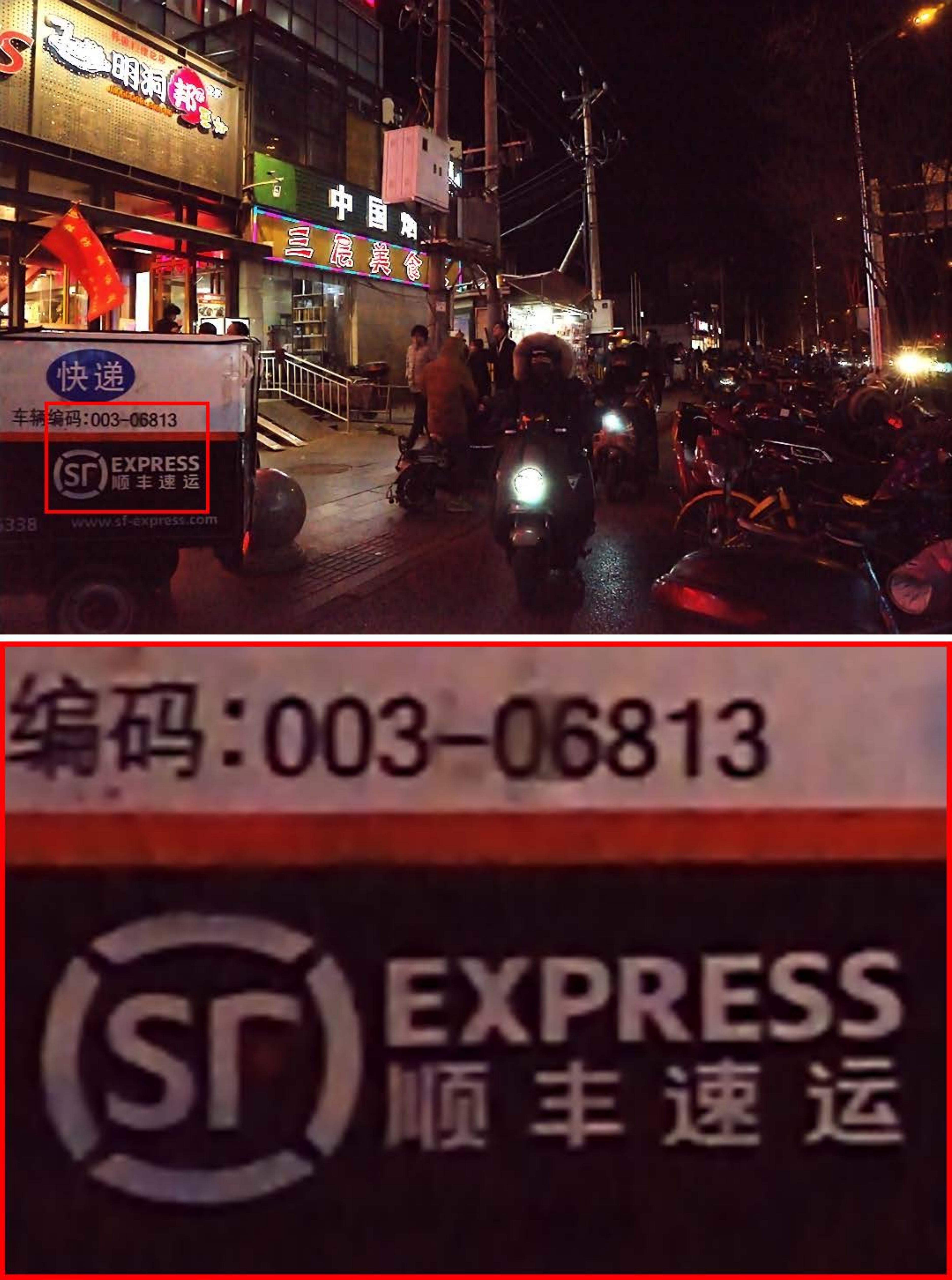}\\
		\includegraphics[width=0.115\linewidth]{Figures/Comparisons/DarkFace0/2/Input}&
		\includegraphics[width=0.115\linewidth]{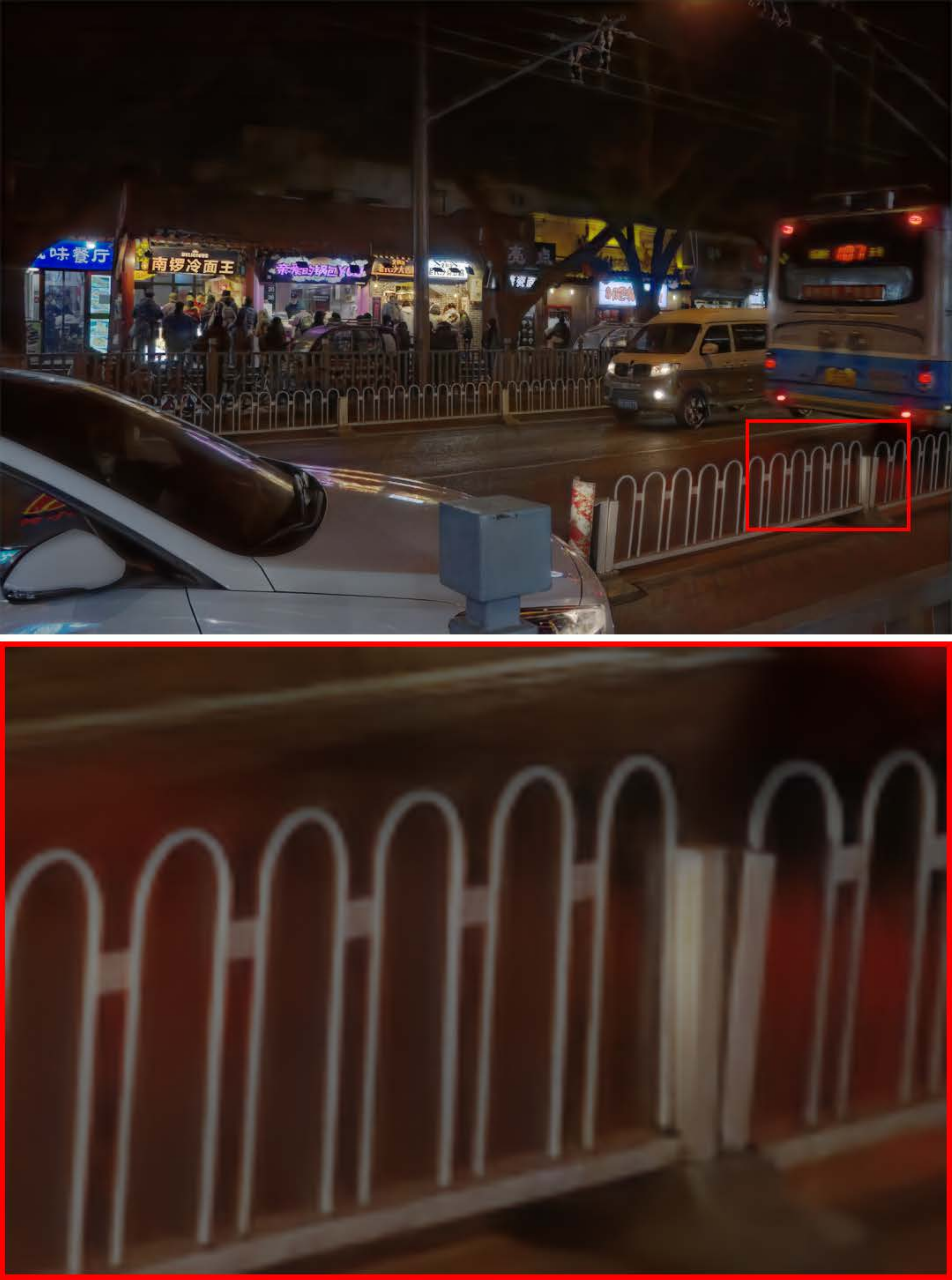}&
		\includegraphics[width=0.115\linewidth]{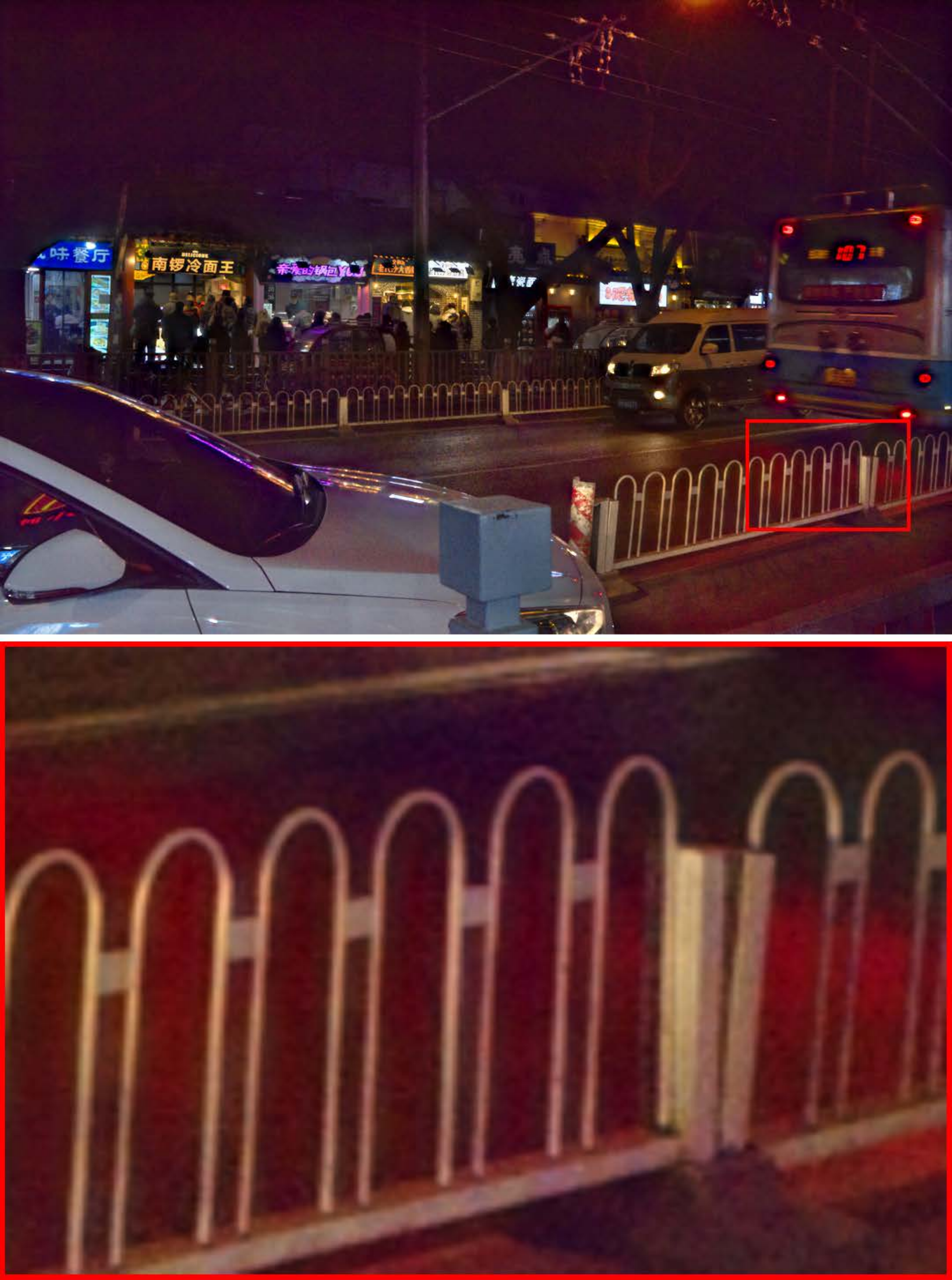}&
		\includegraphics[width=0.115\linewidth]{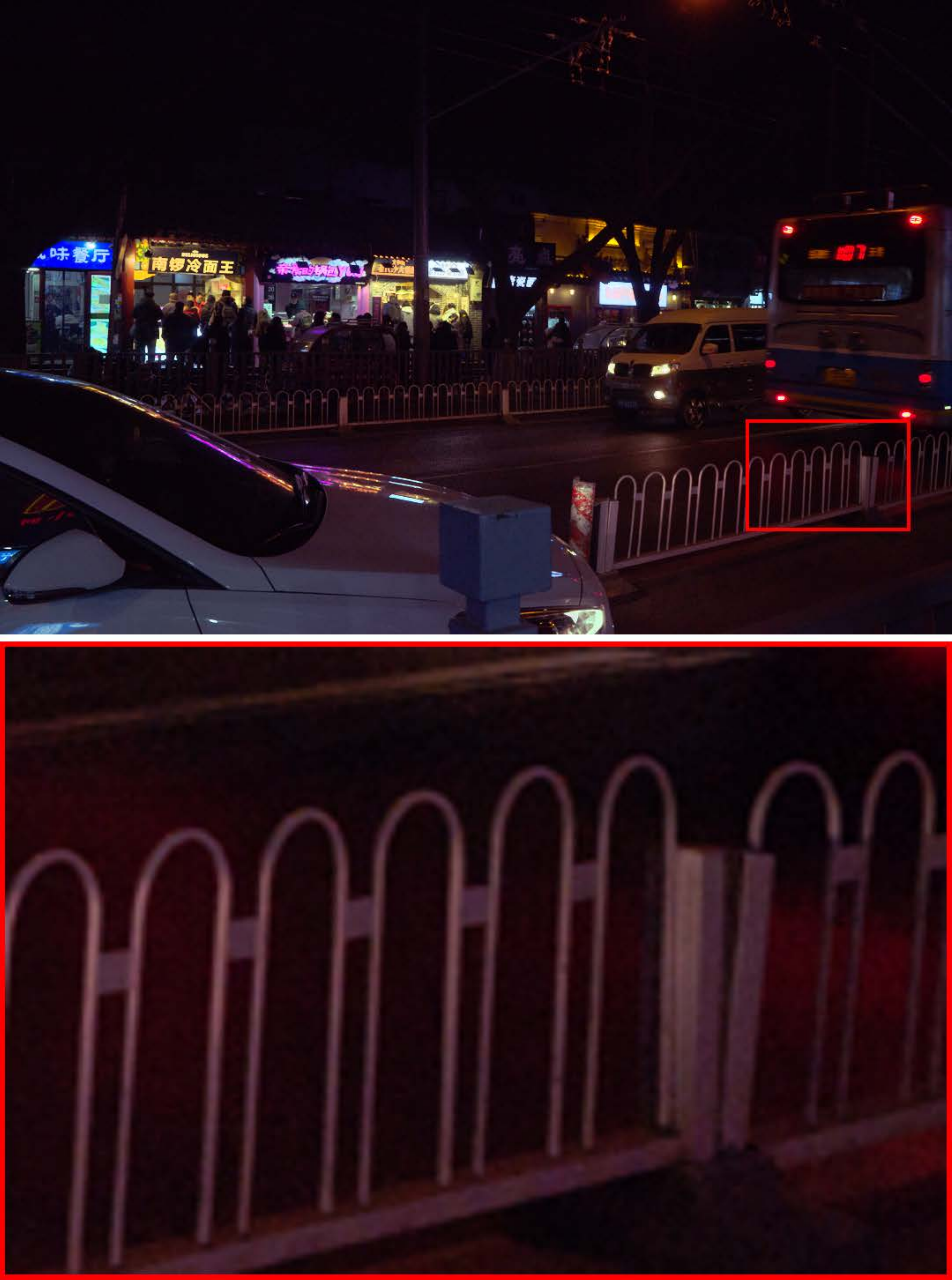}&
		\includegraphics[width=0.115\linewidth]{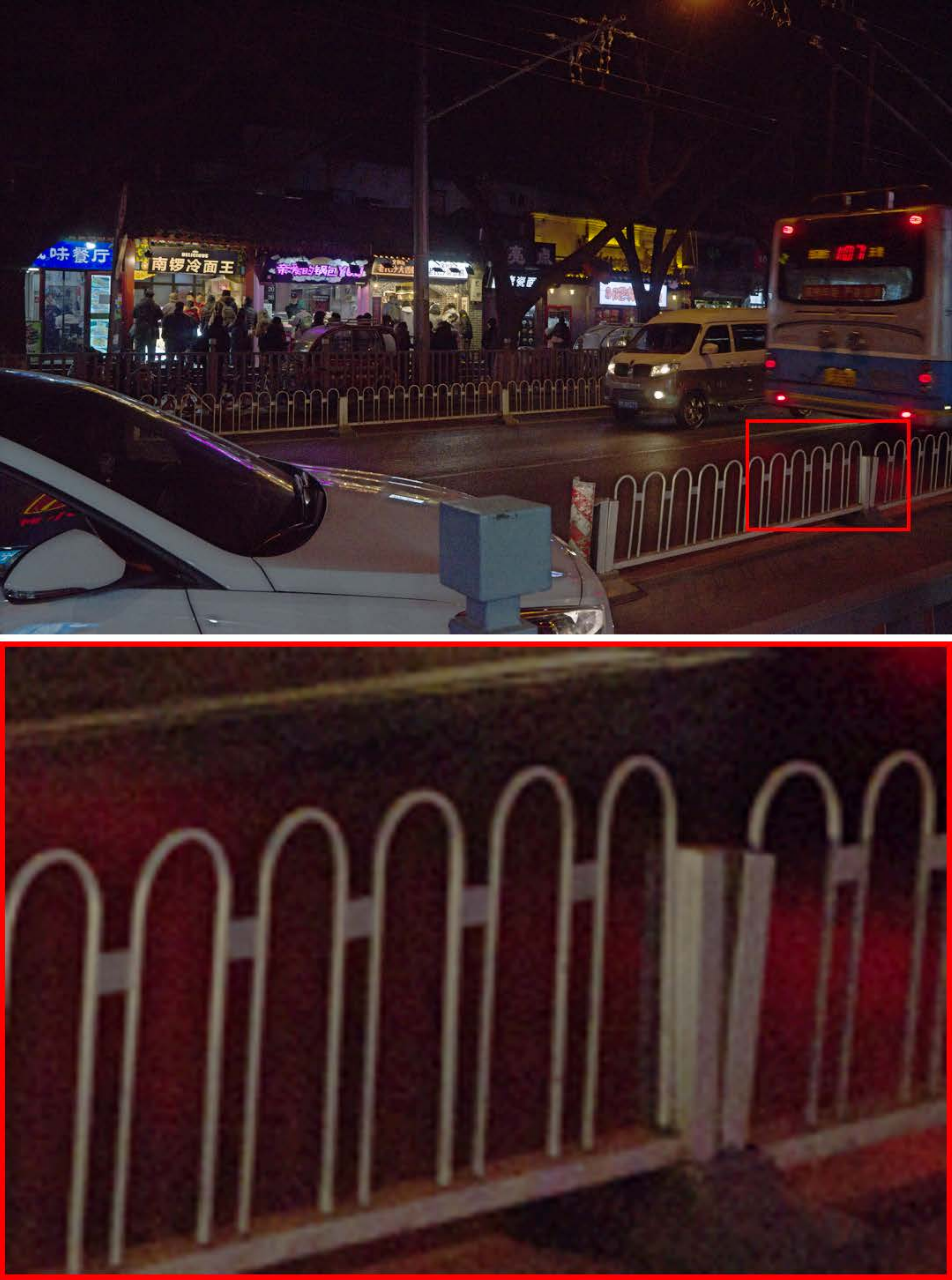}&
		\includegraphics[width=0.115\linewidth]{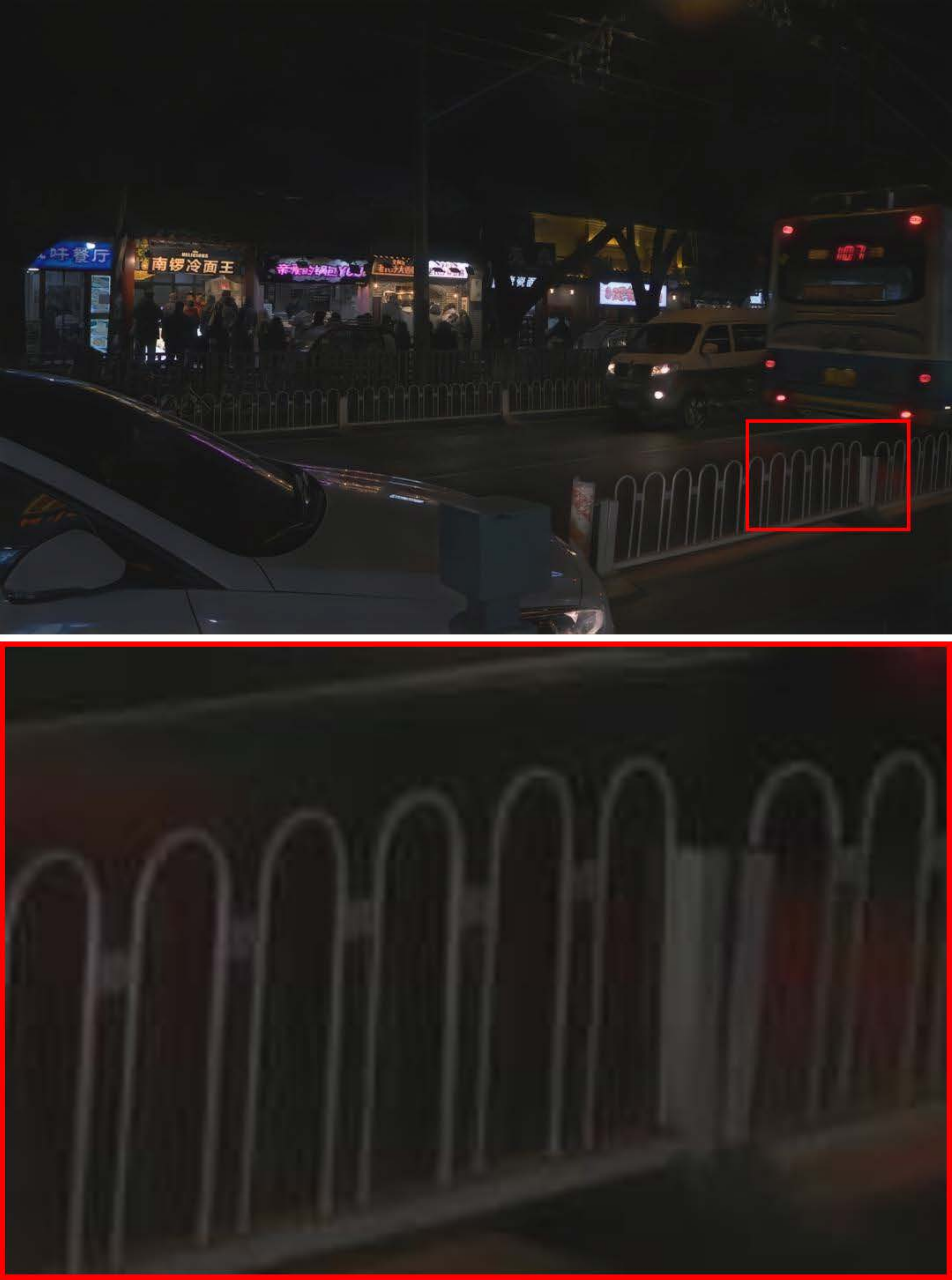}&
		\includegraphics[width=0.115\linewidth]{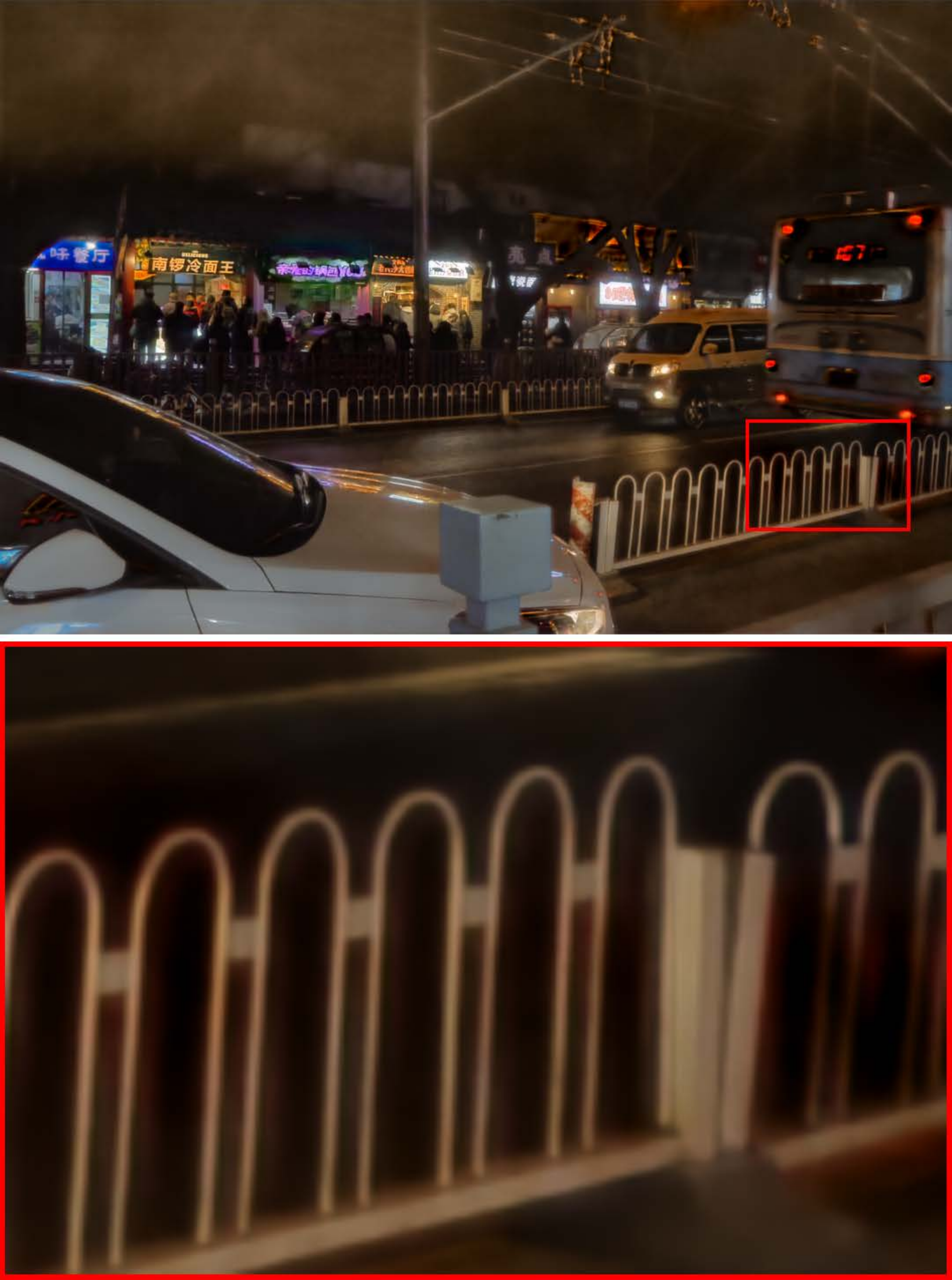}&
		\includegraphics[width=0.115\linewidth]{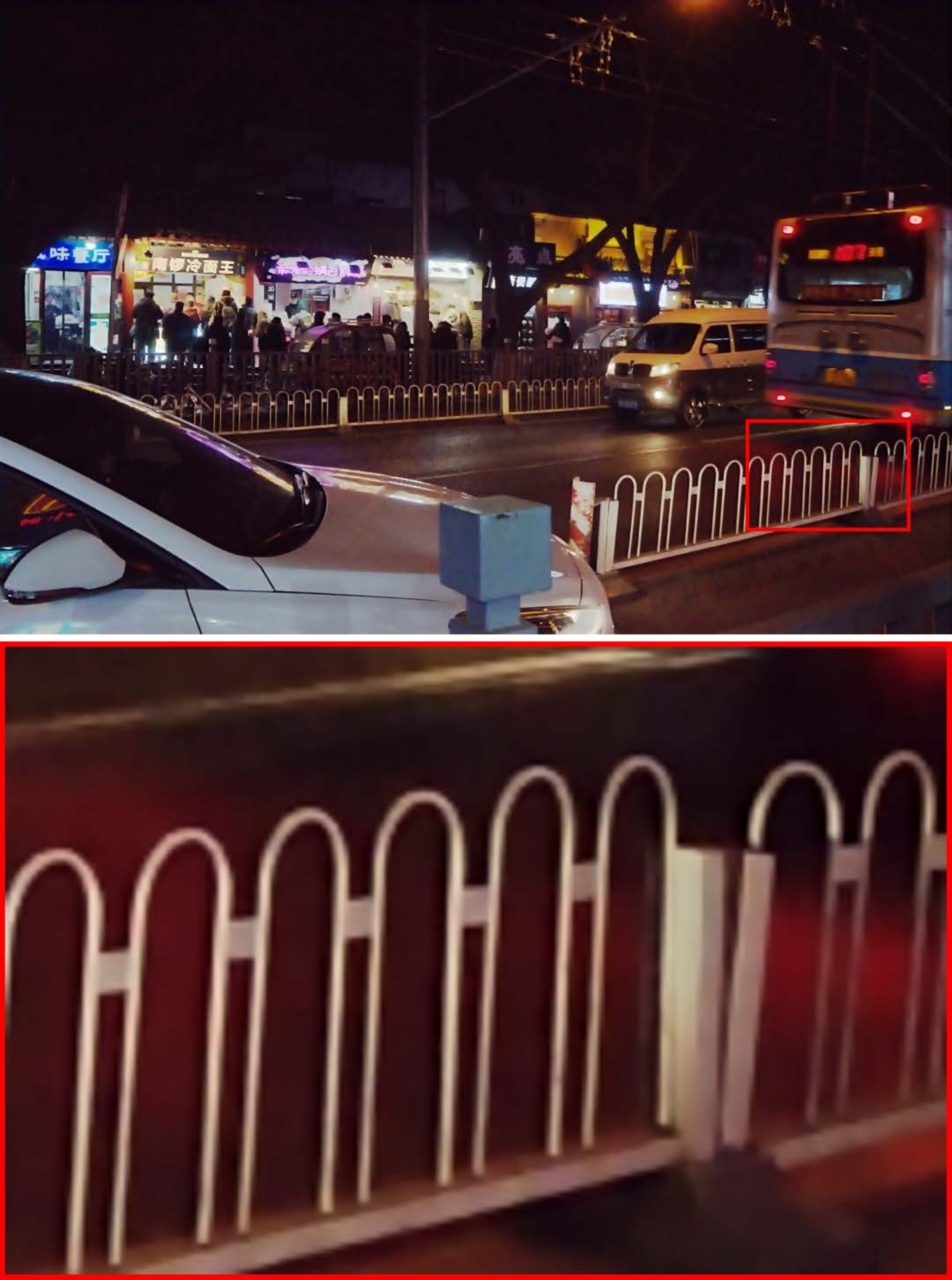}\\
		\footnotesize Input&\footnotesize KinD&\footnotesize EnGAN&\footnotesize DeepUPE&\footnotesize ZeroDCE&\footnotesize FIDE&\footnotesize DRBN&\footnotesize Ours\\
	\end{tabular}
	\caption{Visual results of state-of-the-art methods and our RUAS on the DARK FACE dataset. Red box indicates the obvious differences.}
	\label{fig:DarkFace1}
\end{figure*}

\begin{figure*}[t]
	\centering
	\begin{minipage}{0.30\textwidth}
		\vspace{0.15cm}
		\subfigure{
			\begin{minipage}{1\textwidth}
				\includegraphics[width=1\textwidth]{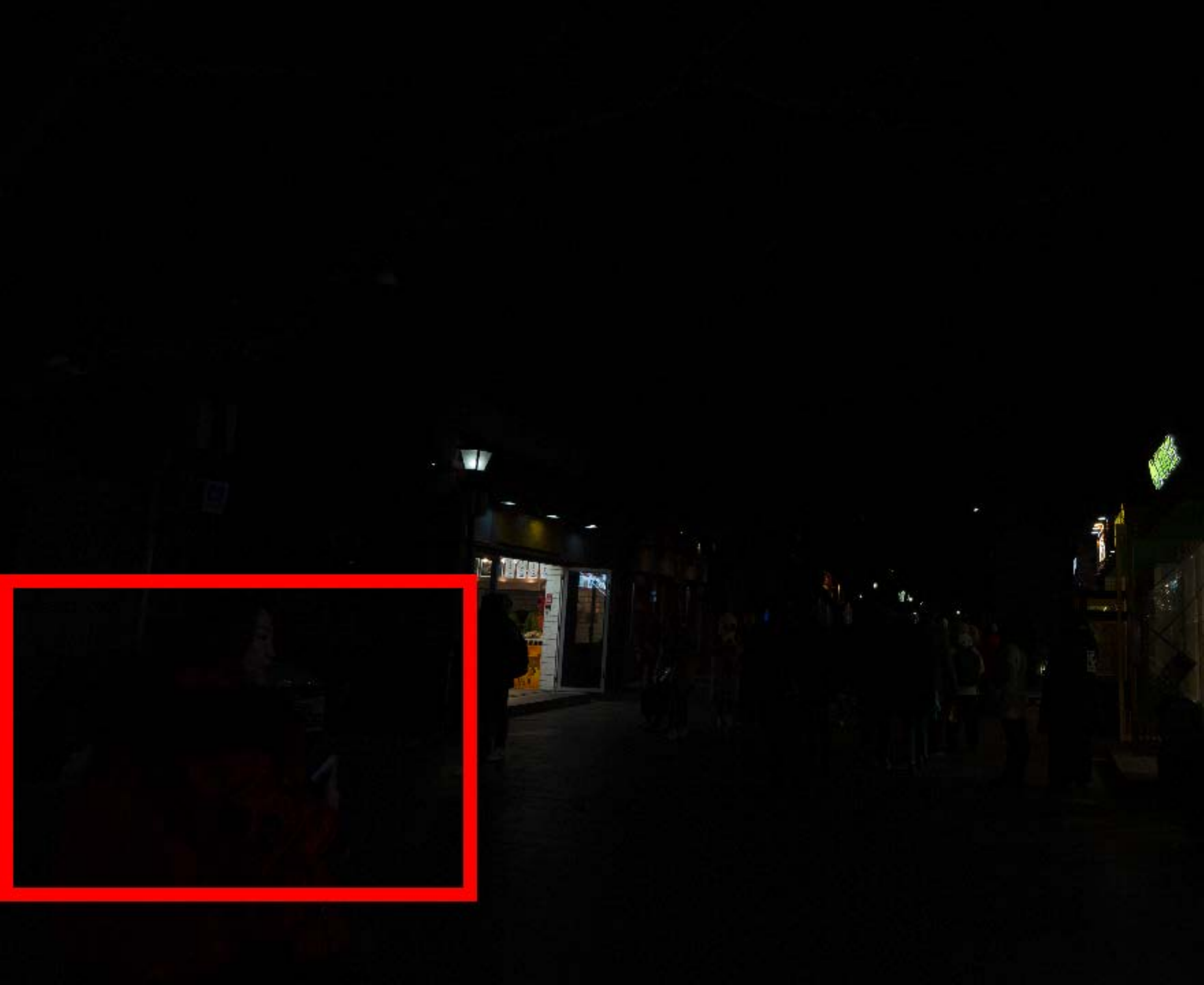}
				\centering  \footnotesize Input\\
			\end{minipage}
		}
	\end{minipage}
	\begin{minipage}{0.16\textwidth}
		\subfigure{
			\begin{minipage}{1\textwidth}
				\vspace{0.1cm}
				\includegraphics[width=1\textwidth]{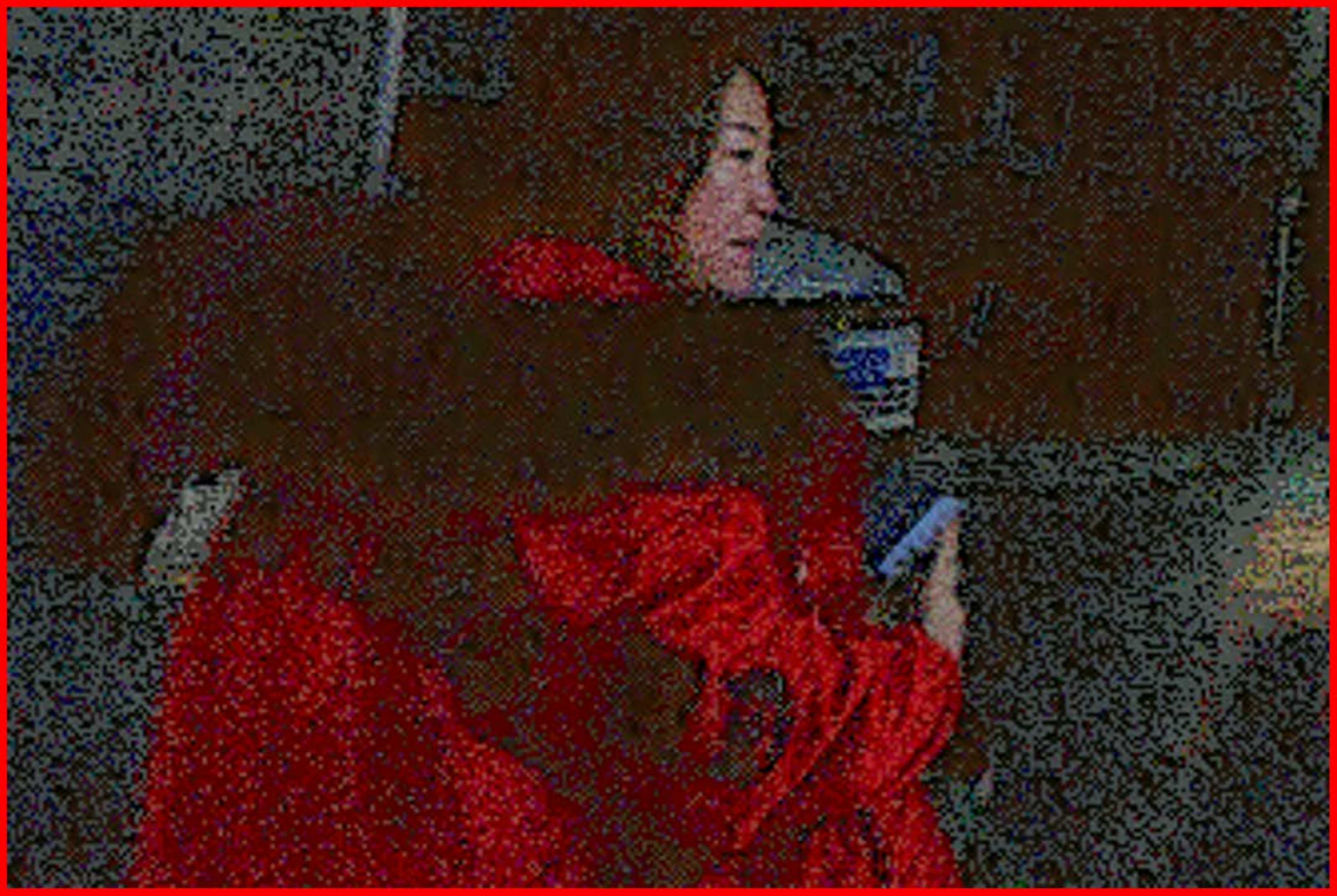}
				\centering \footnotesize RetinexNet \vspace{-0.7em}\\
			\end{minipage}
		}
		\subfigure{
			\begin{minipage}{1\textwidth}
				\vspace{0.1cm}
				\includegraphics[width=1\textwidth]{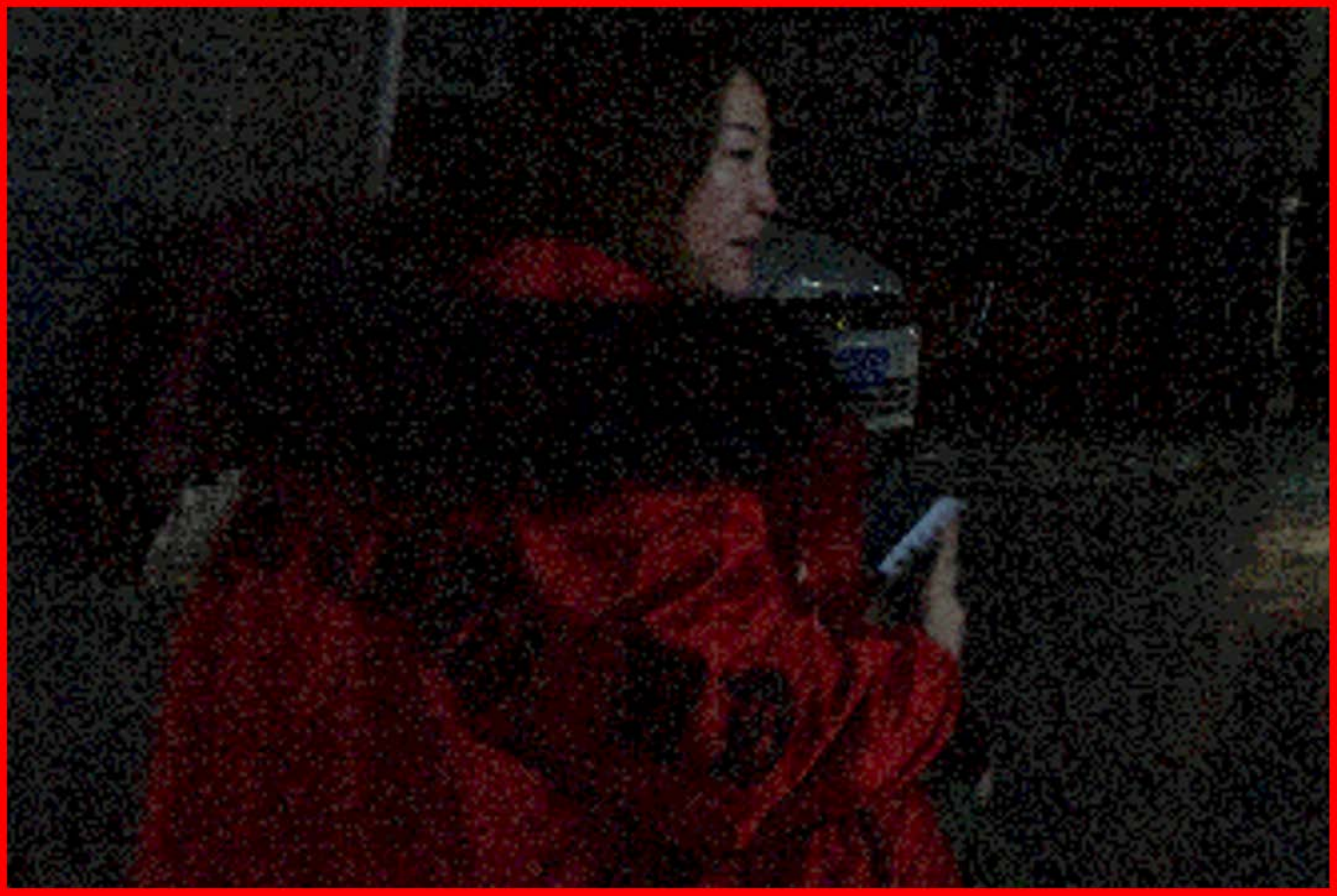}
				\centering \footnotesize ZeroDCE\\
			\end{minipage}
		}
	\end{minipage}
	\begin{minipage}{0.16\textwidth}
		\subfigure{
			\begin{minipage}{1\textwidth}
				\vspace{0.1cm}
				\includegraphics[width=1\textwidth]{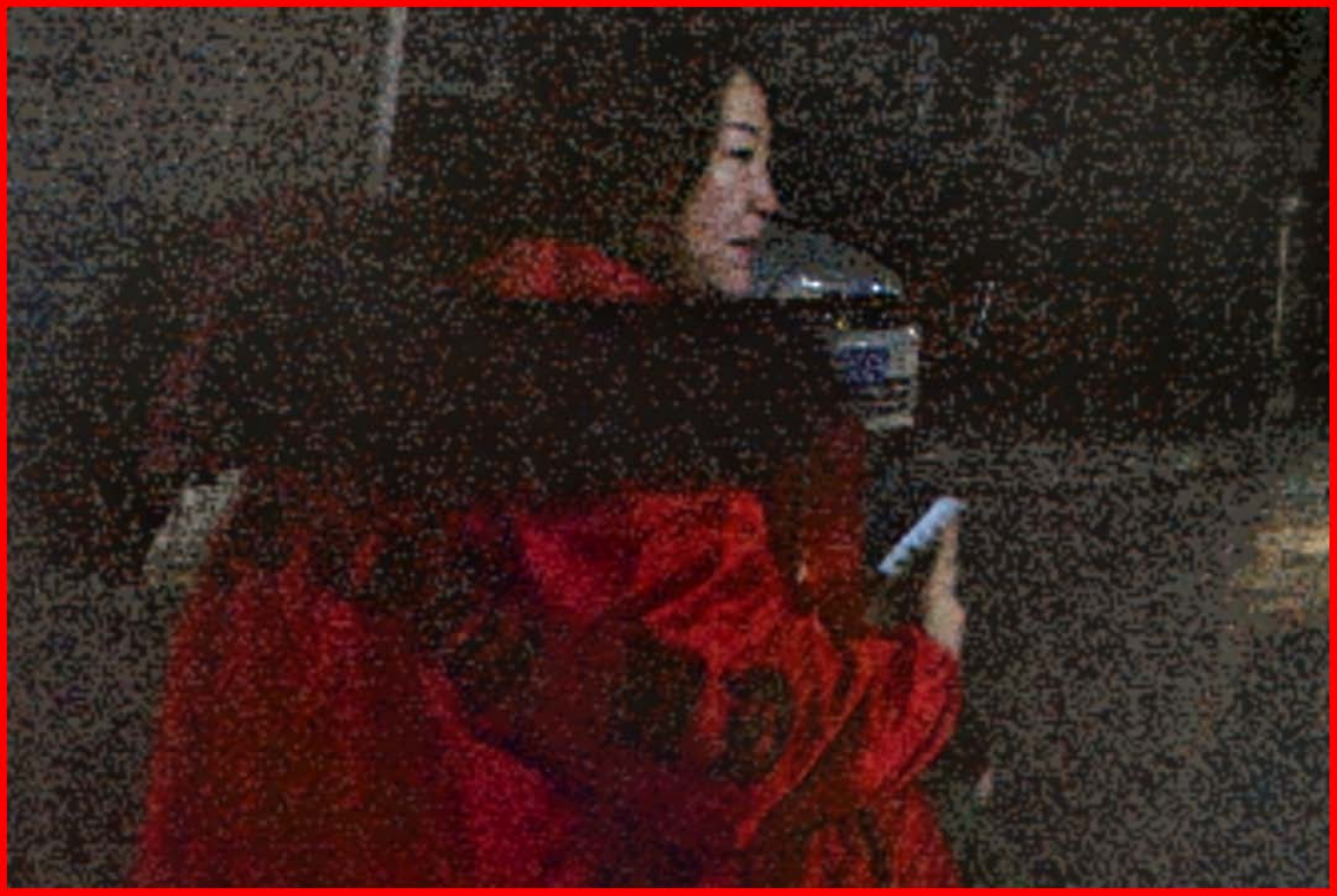}
				\centering \footnotesize  EnGAN \vspace{-0.7em}\\
			\end{minipage}
		}
		\subfigure{
			\begin{minipage}{1\textwidth}
				\vspace{0.1cm}
				\includegraphics[width=1\textwidth]{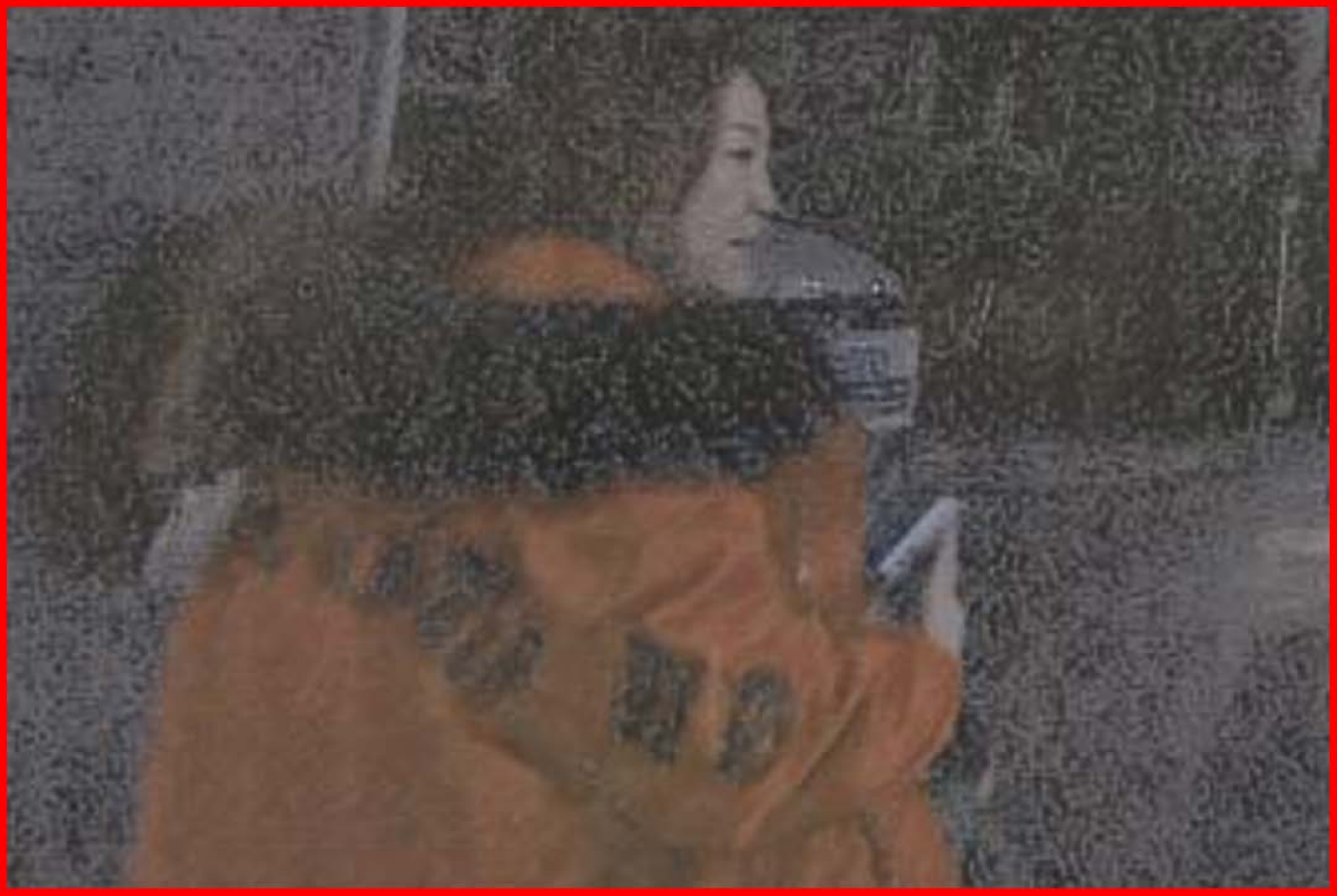}
				\centering \footnotesize  FIDE\\
			\end{minipage}
		}
	\end{minipage}
	\begin{minipage}{0.16\textwidth}
		\subfigure{
			\begin{minipage}{1\textwidth}
				\vspace{0.1cm}
				\includegraphics[width=1\textwidth]{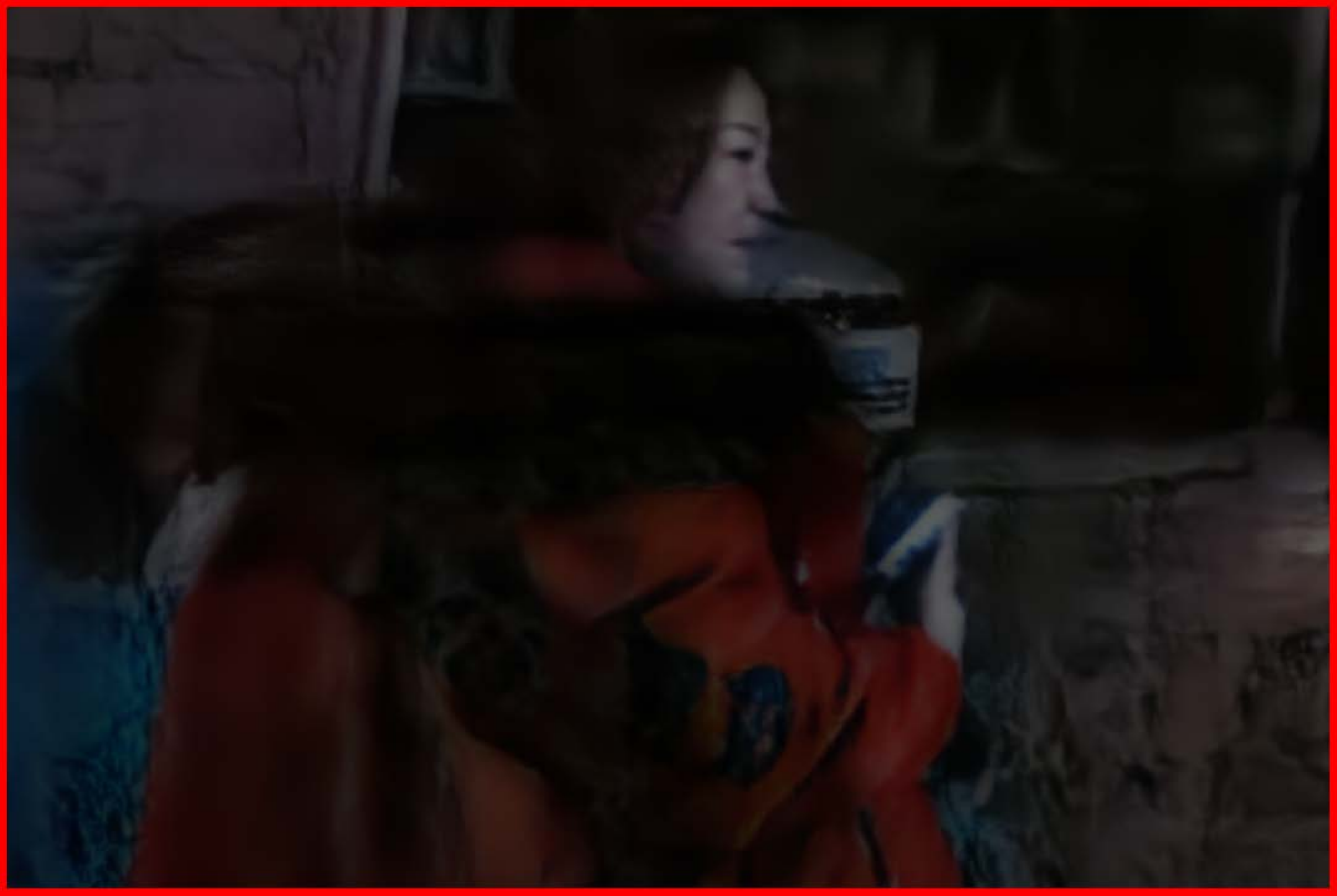}
				\centering \footnotesize KinD \vspace{-0.7em}\\
			\end{minipage}
		}
		\subfigure{
			\begin{minipage}{1\textwidth}
				\vspace{0.1cm}
				\includegraphics[width=1\textwidth]{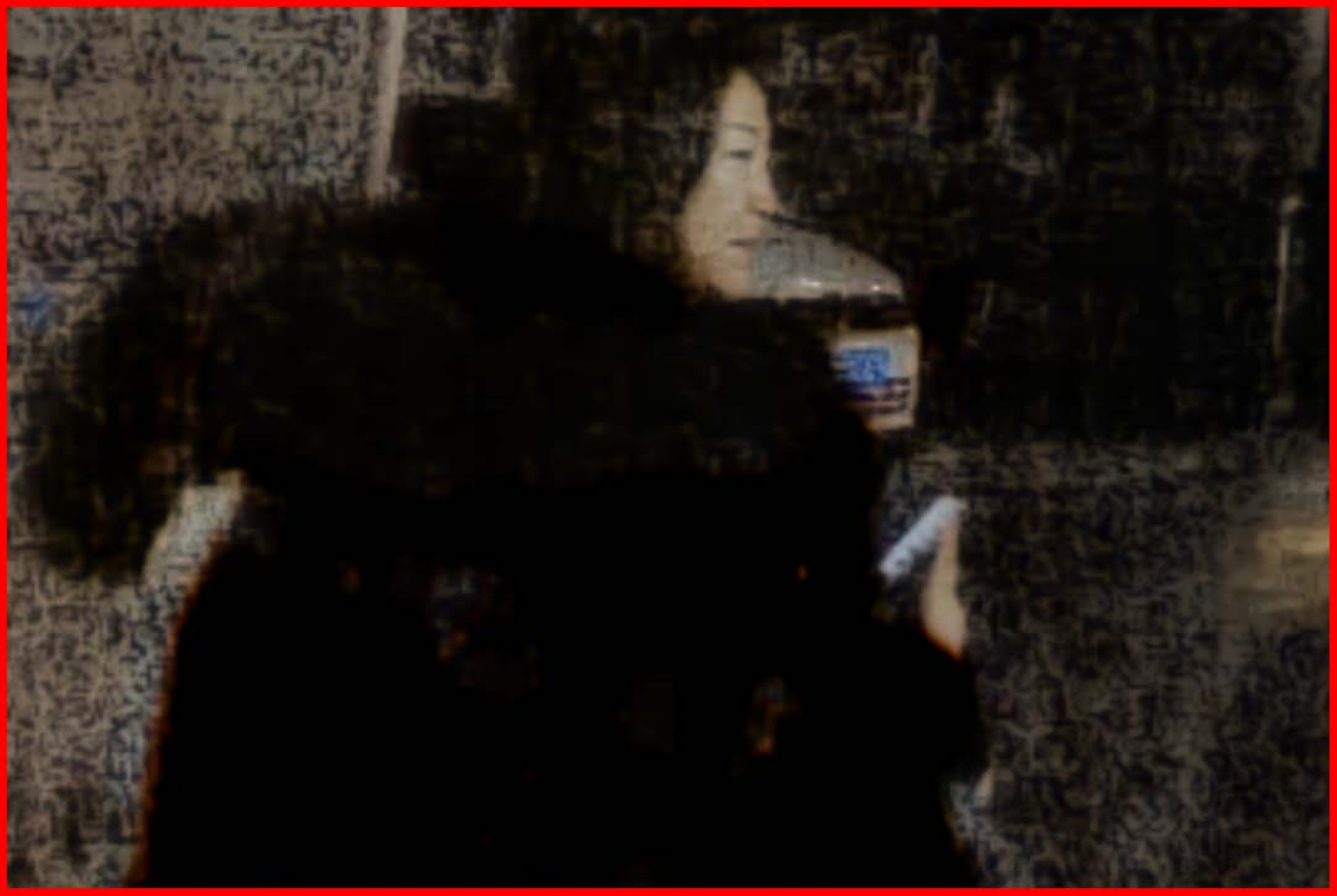}
				\centering \footnotesize DRBN\\
			\end{minipage}
		}
	\end{minipage}
	\begin{minipage}{0.16\textwidth}
		\subfigure{
			\begin{minipage}{1\textwidth}
				\vspace{0.1cm}
				\includegraphics[width=1\textwidth]{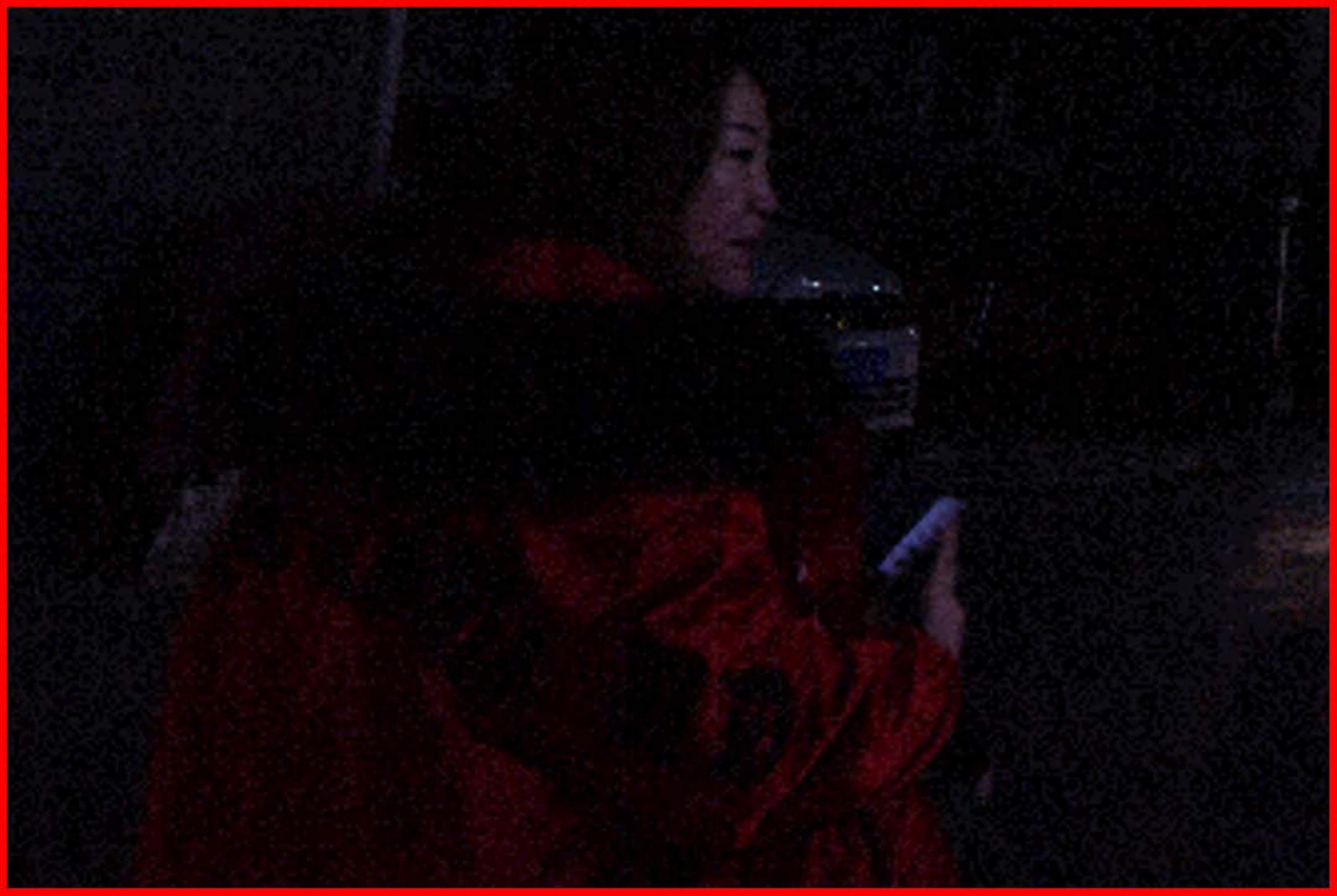}
				\centering \footnotesize DeepUPE \vspace{-0.9em}\\
			\end{minipage}
		}
		\subfigure{
			\begin{minipage}{1\textwidth}
				\vspace{0.1cm}
				\includegraphics[width=1\textwidth]{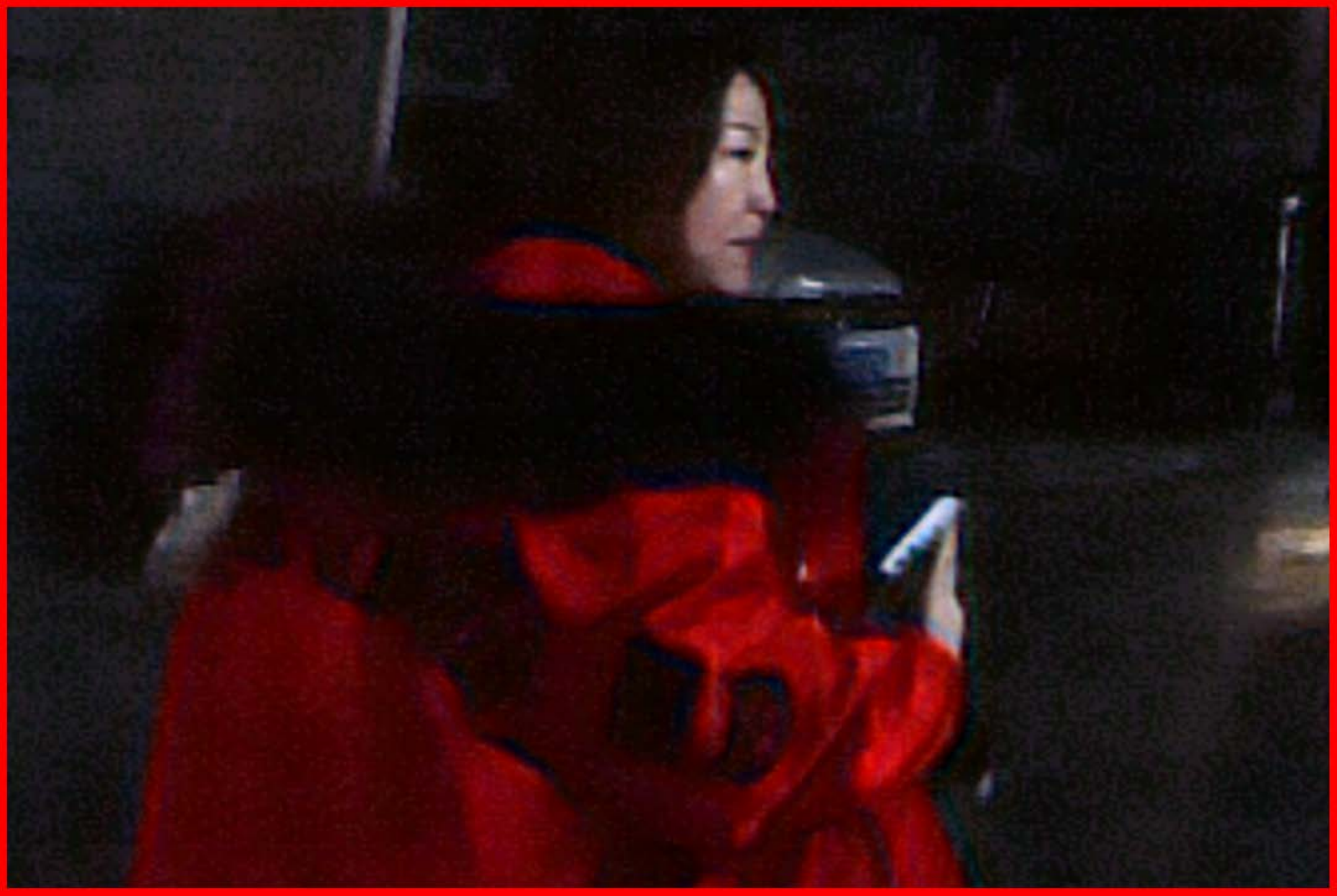}
				\centering \footnotesize Ours\\
			\end{minipage}
		}	
	\end{minipage}
	\begin{minipage}{0.3\textwidth}
		\subfigure{
			\begin{minipage}{1\textwidth}
				\includegraphics[width=1\textwidth]{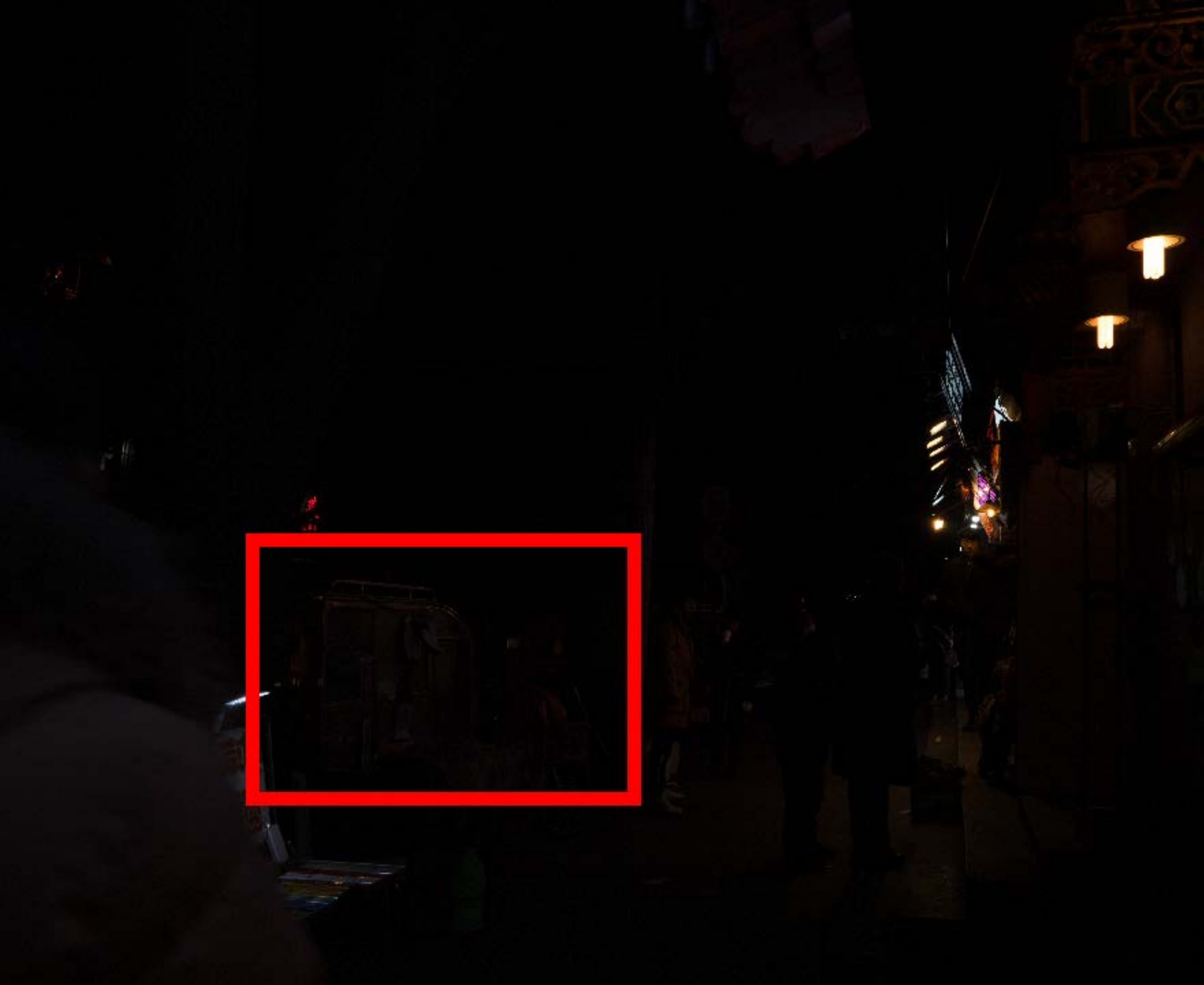}
				\centering  \footnotesize Input\\
			\end{minipage}
		}
	\end{minipage}
	\begin{minipage}{0.16\textwidth}
		\subfigure{
			\begin{minipage}{1\textwidth}
				\vspace{-0.05cm}
				\includegraphics[width=1\textwidth]{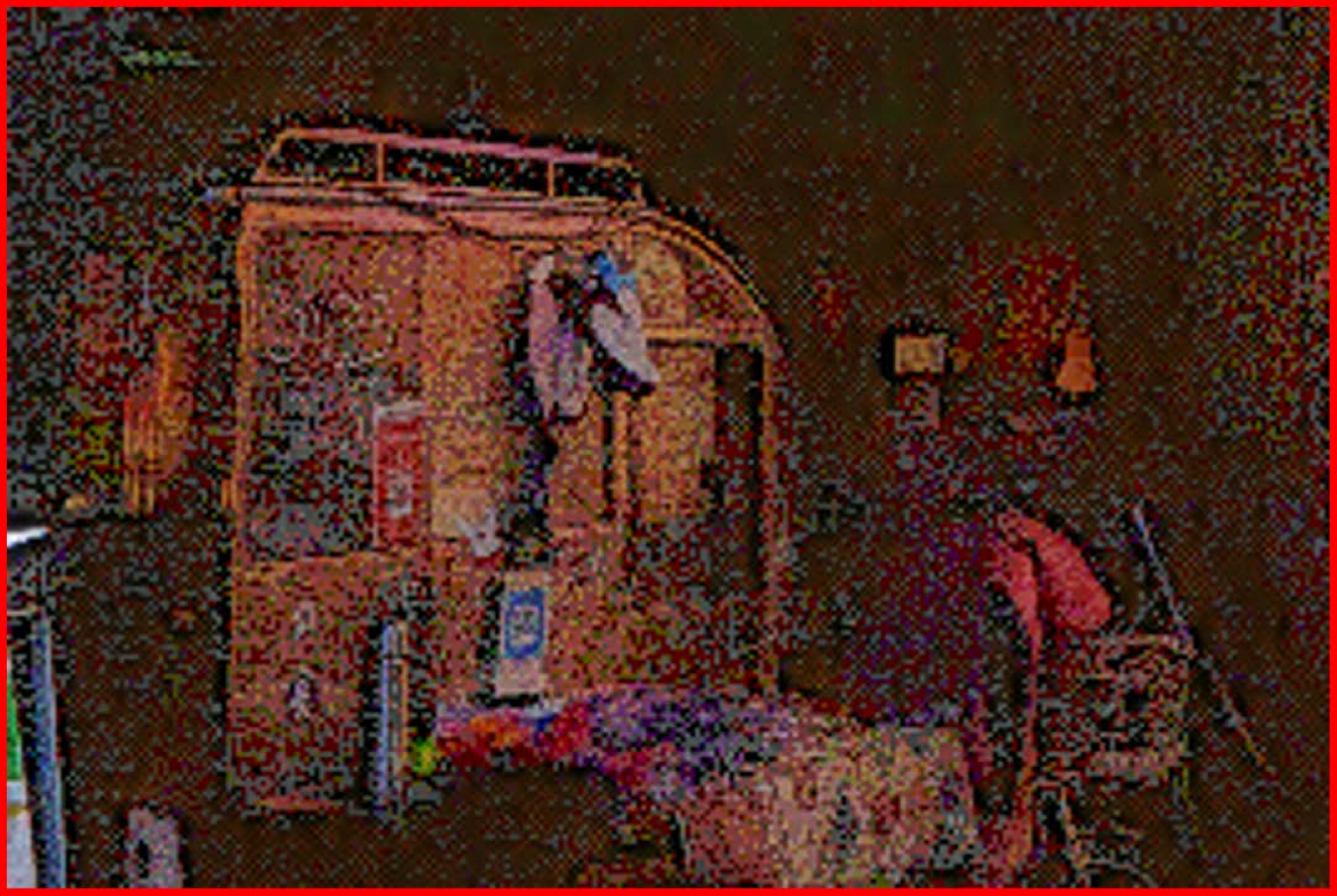}
				\centering \footnotesize RetinexNet \vspace{-0.7em}\\
			\end{minipage}
		}
		\subfigure{
			\begin{minipage}{1\textwidth}
				\vspace{0.1cm}
				\includegraphics[width=1\textwidth]{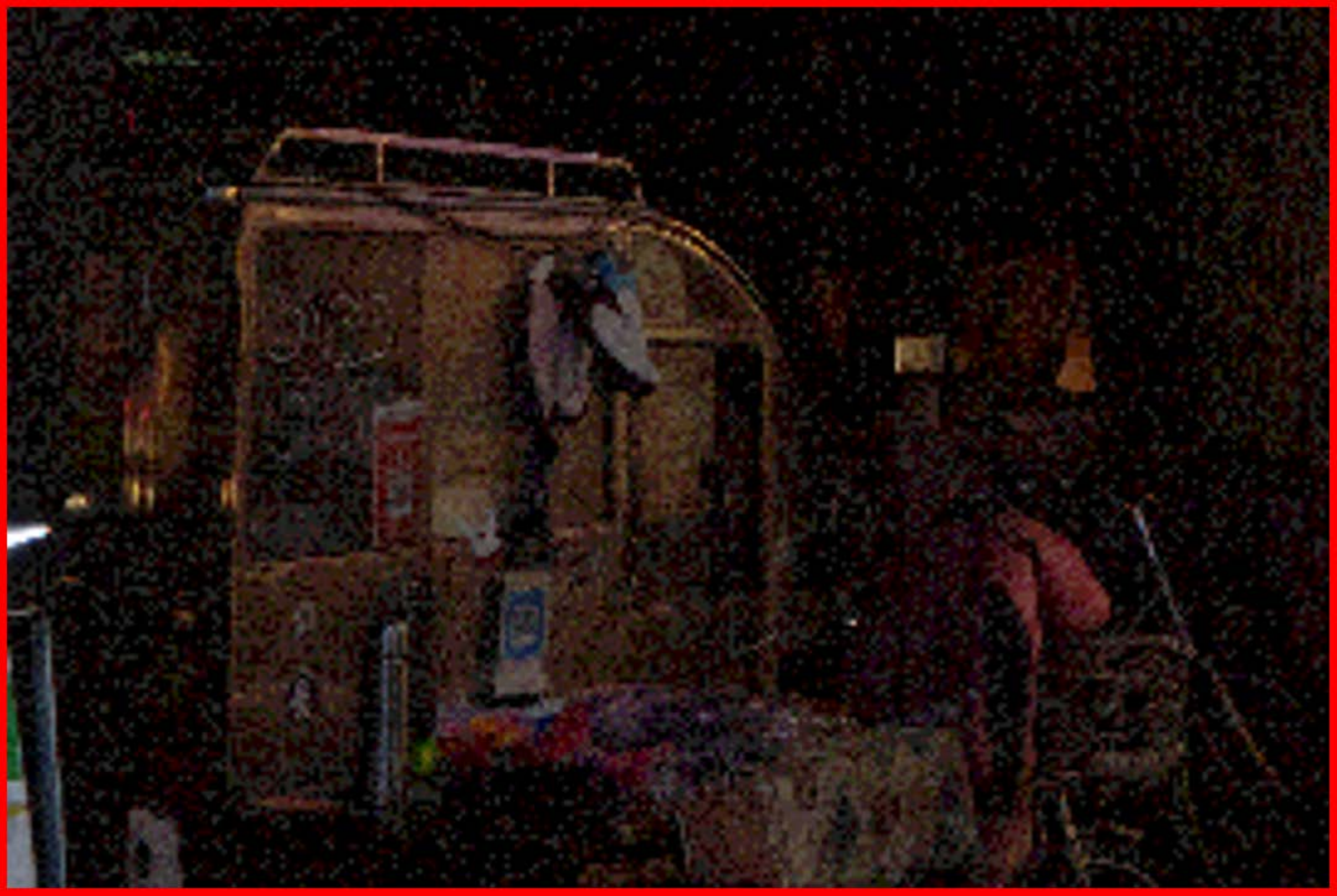}
				\centering \footnotesize ZeroDCE\\
			\end{minipage}
		}
	\end{minipage}
	\begin{minipage}{0.16\textwidth}
		\subfigure{
			\begin{minipage}{1\textwidth}
				\vspace{-0.05cm}
				\includegraphics[width=1\textwidth]{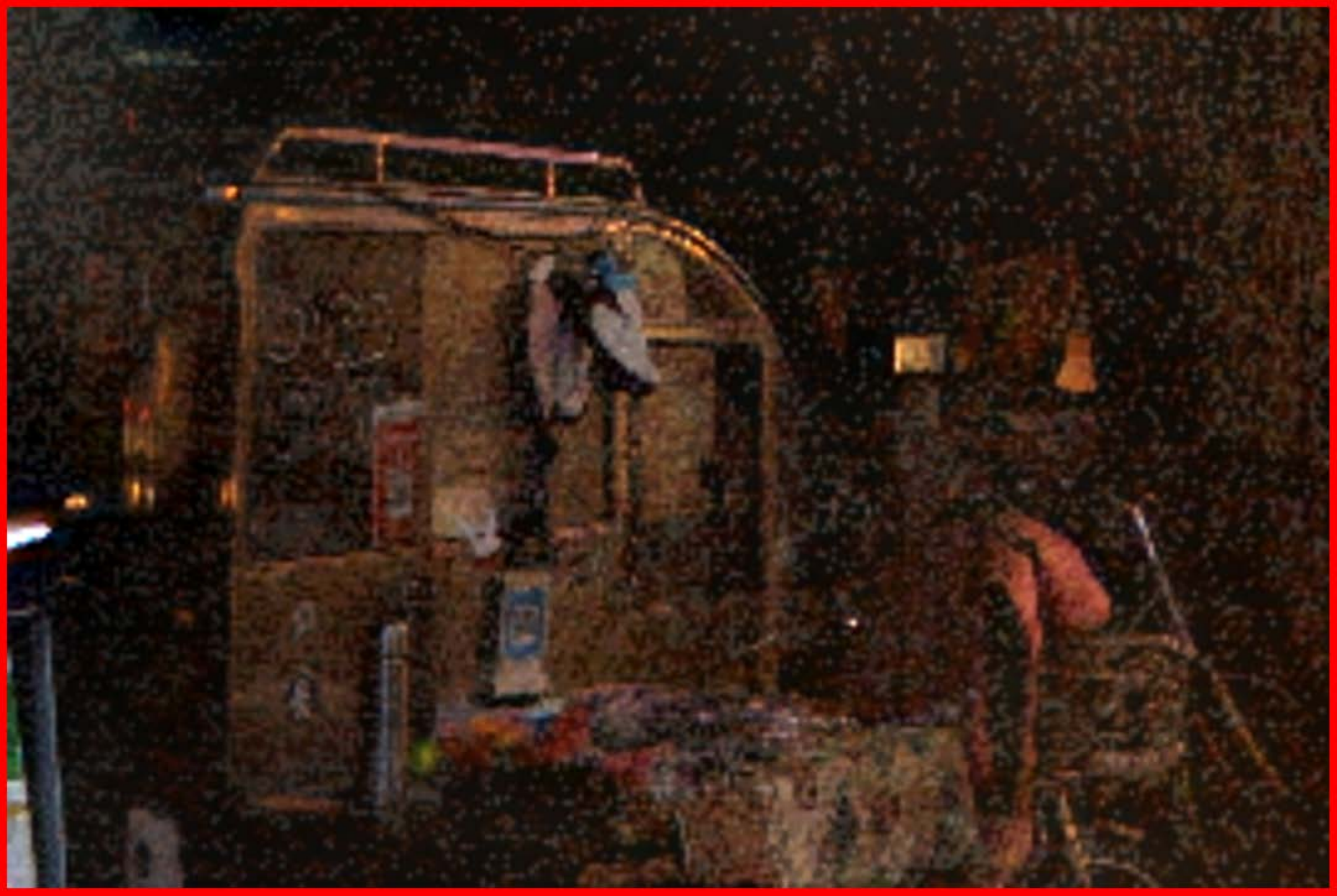}
				\centering \footnotesize  EnGAN \vspace{-0.7em}\\
			\end{minipage}
		}
		\subfigure{
			\begin{minipage}{1\textwidth}
				\vspace{0.1cm}
				\includegraphics[width=1\textwidth]{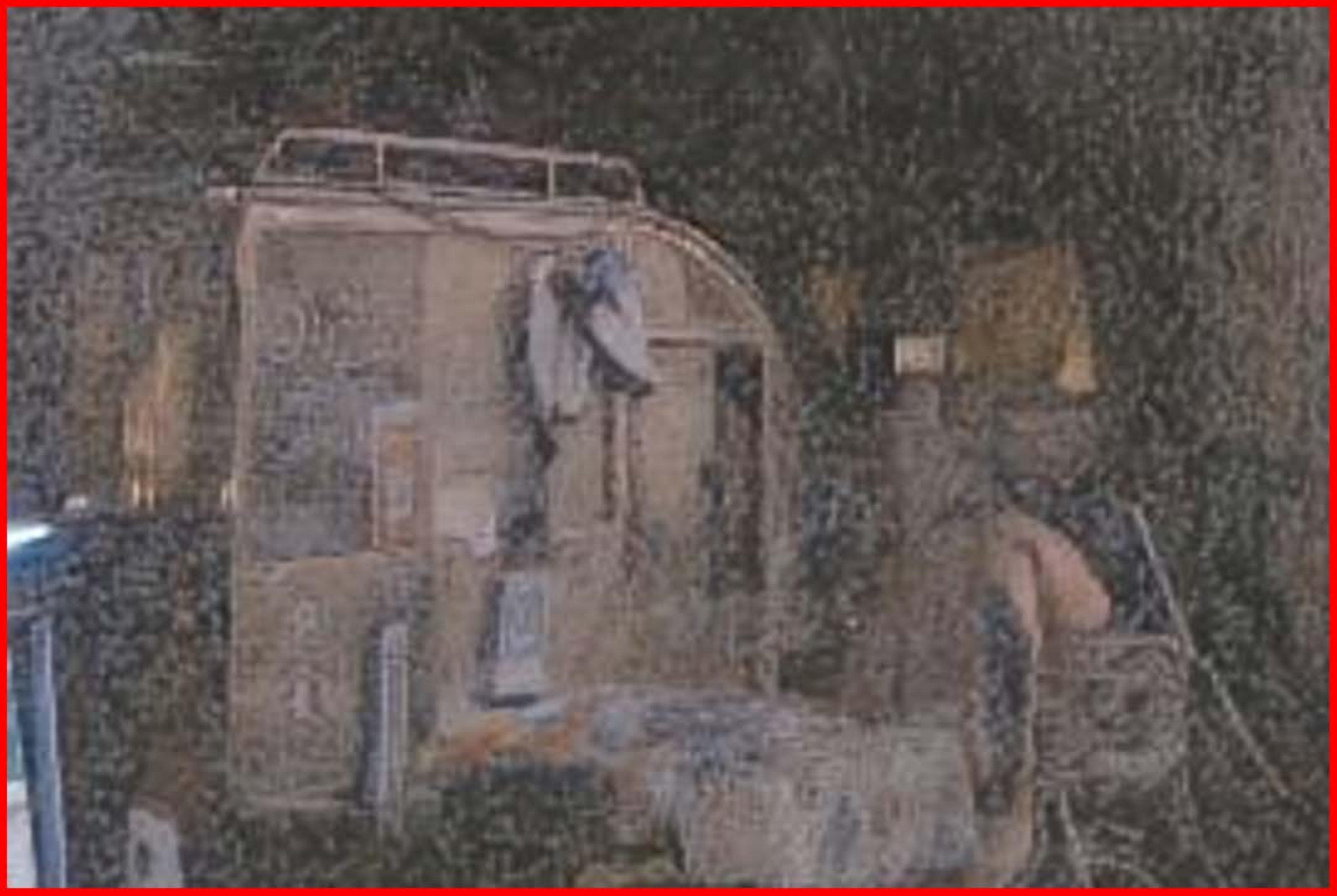}
				\centering \footnotesize  FIDE\\
			\end{minipage}
		}
	\end{minipage}
	\begin{minipage}{0.16\textwidth}
		\subfigure{
			\begin{minipage}{1\textwidth}
				\vspace{-0.05cm}
				\includegraphics[width=1\textwidth]{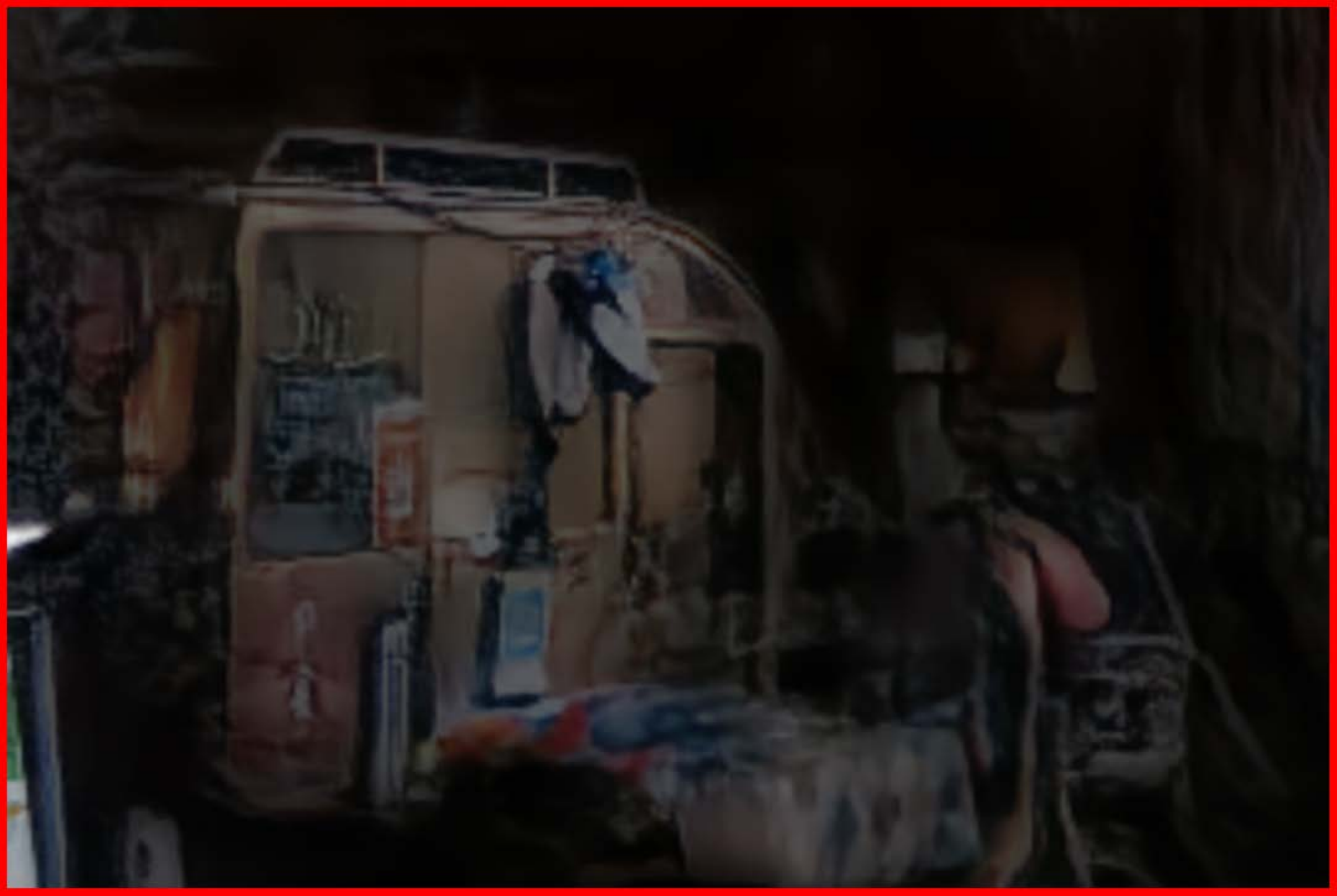}
				\centering \footnotesize KinD \vspace{-0.7em}\\
			\end{minipage}
		}
		\subfigure{
			\begin{minipage}{1\textwidth}
				\vspace{0.1cm}
				\includegraphics[width=1\textwidth]{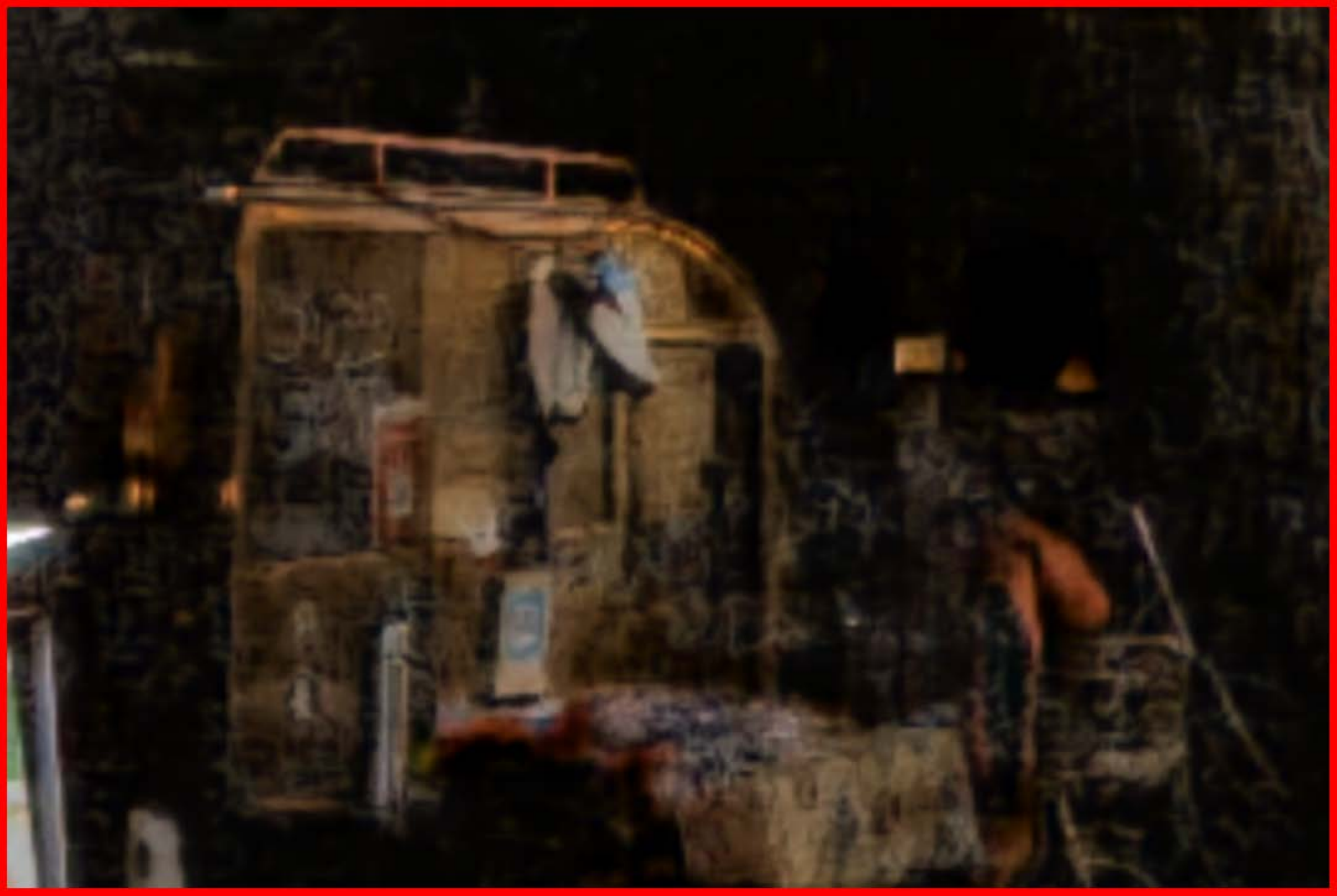}
				\centering \footnotesize DRBN\\
			\end{minipage}
		}
	\end{minipage}
	\begin{minipage}{0.16\textwidth}
		\subfigure{
			\begin{minipage}{1\textwidth}
				\vspace{-0.05cm}
				\includegraphics[width=1\textwidth]{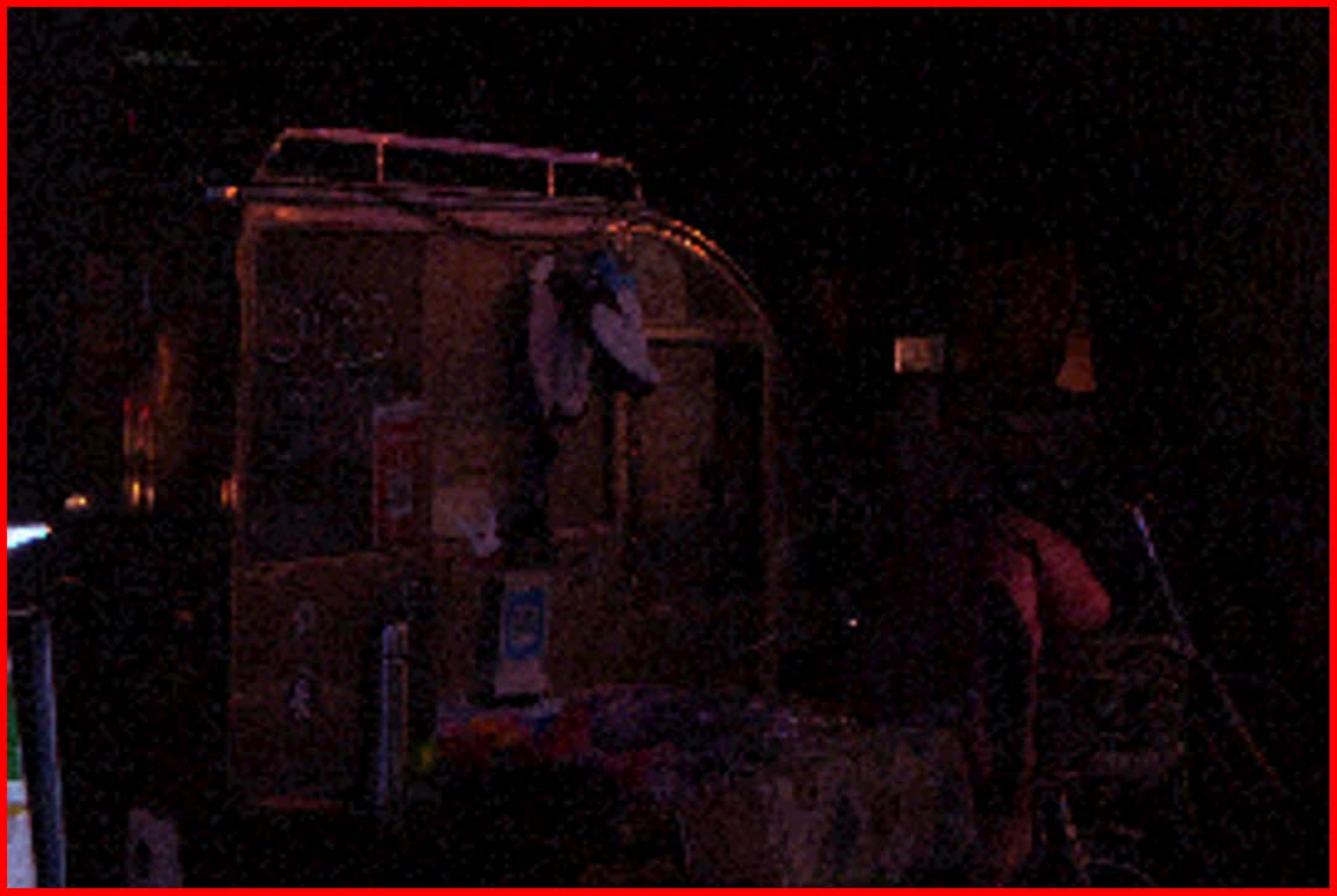}
				\centering \footnotesize DeepUPE \vspace{-0.9em}\\
			\end{minipage}
		}
		\subfigure{
			\begin{minipage}{1\textwidth}
				\vspace{0.1cm}
				\includegraphics[width=1\textwidth]{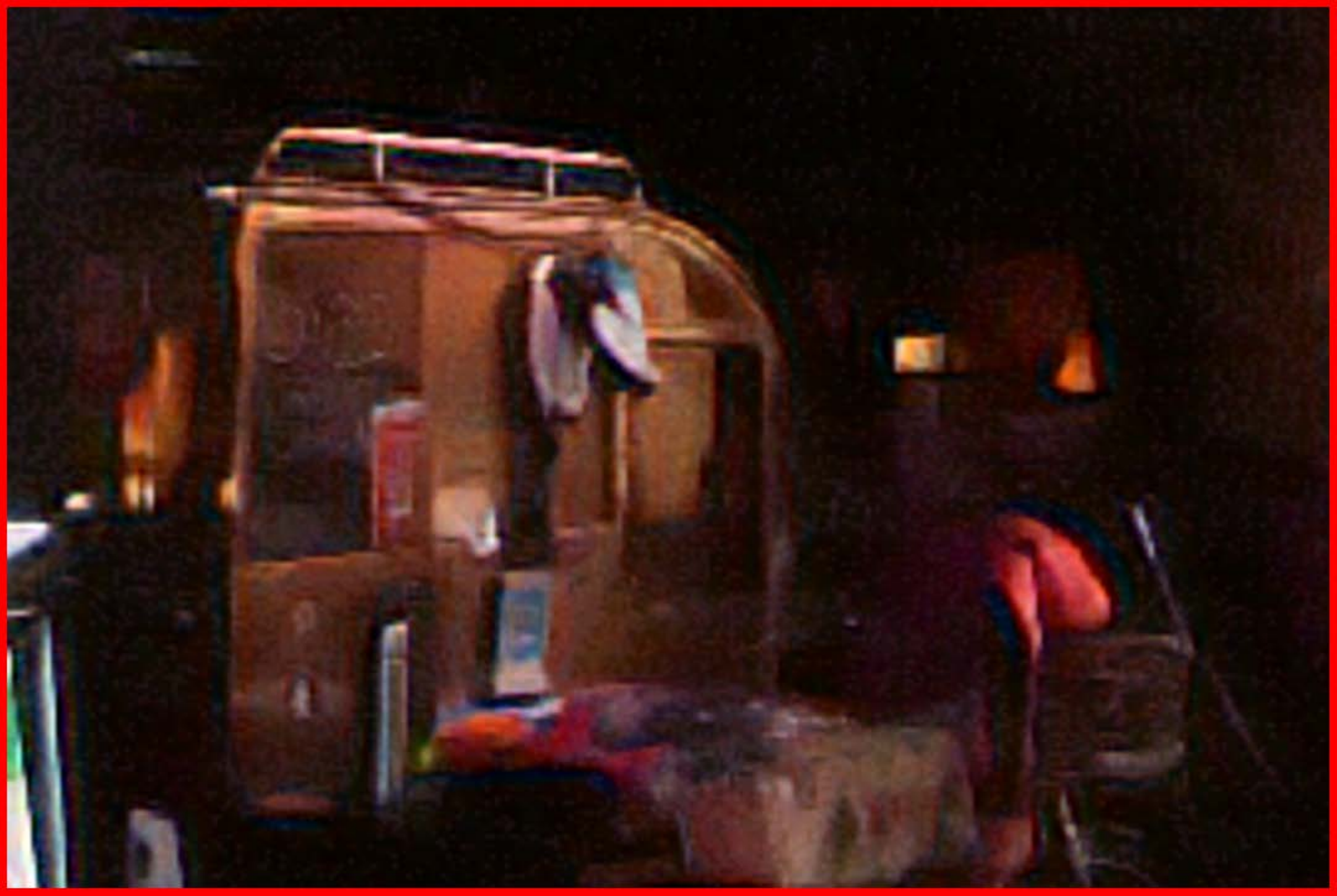}
				\centering \footnotesize Ours\\
			\end{minipage}
		}
	\end{minipage}
	\caption{Visual results of state-of-the-art methods and RUAS on the ExtremelyDarkFace dataset. Red boxes indicate the obvious differences.}
	\label{fig:ExtermelyDarkFace}
\end{figure*}

\subsection{Algorithmic Analyses}
In this part, we made comprehensive analysis in terms of our proposed method from different aspects. 

\textbf{Effects of the warm-start operation. }
Firstly, we evaluated the performance brought by three different warm-start strategies including fix warm-start as $\hat{\mathbf{u}}_\mathtt{i}^0$, update $\hat{\mathbf{u}}_\mathtt{i}^k$ w/o and w/ residual rectification. 
Fig.~\ref{fig: teffect} provided the visual comparison of these warm-start strategies in terms of different components.  
We can observe that the updating strategy of w/o residual rectification indeed supplied the positive over-exposure suppression by comparing it with the naive fix warm-start strategy. Further, by introducing the mechanism of residual rectification, the result performed a more comfortable exposure (see the lampshade) than those by other strategies. In a word, we are able to confirm that our designed warm-start strategy really suppress the over-exposure during the propagations. 

\textbf{Evaluating different versions of RUAS.}
We also perform visual results of RUAS$_\mathtt{S}$, RUAS, and RUAS$_\mathtt{A}$ to demonstrate the difference between them. As shown in the top row of Fig.~\ref{fig:Versions}, RUAS and RUAS$_\mathtt{A}$ significantly improved the visual quality by introducing the denoising mechanism compared with RUAS$_\mathtt{S}$. Further, RUAS$_\mathtt{A}$ avoided denoising when meeting the case without noises (see the bottom row), to provide a higher visual expression. In a word, RUAS$_\mathtt{A}$ is more intelligent which can automatically distinguish whether or not to perform denoising and is applicable to more general low-light scenarios. 

\textbf{Analyzing the iteration number in scene module.}
We analyzed of influence of the total stage number $K$ for low-light image enhancement. As shown in Table~\ref{tab:Stage}, the stage number obviously impacted the quantitative performance. When $K=3$, the algorithm realized the best numerical scores, so we defined $K=3$ as our default setting.

\textbf{Analyzing the search process.}
Firstly, we analyzed the performance using different heuristically-designed architectures on the MIT-Adobe 5K dataset. As shown in Table~\ref{tab:SceneModelingSearch}, even though we adopted the complex supernet that contained massive parameters, the results performed the low PSNR and high time-consuming. As for other heuristically-designed architectures, the performance was also unideal. Briefly, these architectures may not effective enough. It is because that these architectures did not integrate the task cues/principles. By comparison, our searched architecture realized the highest PSNR, with an additional outcome (i.e., less inference time). In a word, this experiment indicated the necessity of searching for architecture and the superiority of our searched architecture.

\textbf{Analyzing the search strategy.}
Actually, the search strategy is a decisive factor for the searched architecture. To this end, we made an evaluation for different search strategies on low-light image enhancement. We considered three search strategies based on how to search the SM and TM. The separate search strategy is to search these two parts one by one. That is, when searching the TM, the searching procedure of the SM has ended. The naive joint search is to view SM and TM as part of an entire architecture, and just need to search for all the architecture once. Table~\ref{tab:SearchStrategy} reported quantitative scores among different strategies. It can be easily observed that our designed cooperative search strategy was significantly, consistently superior to others both in perceptual quality and execution efficiency. 
Additionally, as shown in Fig.~\ref{fig: SearchStrategy}, from the searched architecture, we can see that our searched TM contained more residual-type convolution, it was reasonable that this structure had been proved in some existing denoising works~\cite{zhang2017beyond}. In other words, our cooperative search strategy indeed bridged the scene and task module to realize a valuable collaboration.

\begin{table}[t]
	\renewcommand\arraystretch{1.2} 
	\setlength{\tabcolsep}{3mm}
	\caption{Comparing detection accuracy of different training strategies. }
	\centering
	\begin{tabular}{|c|c|}
		\hline
		\footnotesize Training Strategy&\footnotesize mAP\\
		\hline
		\footnotesize End-to-end training (loss is defined as $\ell_{\mathtt{t}}$)&\footnotesize   0.685\\ 
		\hline
		\footnotesize End-to-end training (loss is defined as $\ell_{\mathtt{t}}+\lambda\ell_{\mathtt{s}}$)&\footnotesize  0.660 \\ 
		\hline
		\footnotesize Hierarchical training&\footnotesize   0.693\\ 
		\hline
	\end{tabular}
	\label{tab:Strategy}
\end{table}

\begin{table}[t]
	\renewcommand\arraystretch{1.2} 
	\setlength{\tabcolsep}{2mm}
	\caption{Comparing detection accuracy of different backbones. }
	\centering
	\begin{tabular}{|c|c|c|c|c|}
		\hline
		\footnotesize Backbone&\footnotesize ResNet-50&\footnotesize ResNet-101&\footnotesize ResNet-152&\footnotesize VGG-16\\
		\hline
		\footnotesize Baseline&\footnotesize 0.618&\footnotesize 0.591&\footnotesize 0.604&\footnotesize 0.611\\
		\hline
		\footnotesize Ours&\footnotesize 0.674&\footnotesize 0.657&\footnotesize 0.661&\footnotesize {0.685}\\
		\hline
	\end{tabular}
	\label{tab: backbone}
\end{table}

\begin{table}[t]
	\renewcommand\arraystretch{1.2} 
	\setlength{\tabcolsep}{3mm}
	\caption{Analyzing the search process of the task module. }
	\centering
	\begin{tabular}{|c|c|c|c|c|}
		\hline
		\footnotesize Method&\footnotesize TM&\footnotesize RUAS$_\mathtt{S}$+TM &\footnotesize Search TM&\footnotesize RUAS\\
		\hline
		\footnotesize mAP&\footnotesize  0.611&\footnotesize  0.639&\footnotesize 0.655 &\footnotesize 0.685 \\ 
		\hline
	\end{tabular}
	\label{tab:DetSearch}
\end{table}

\subsection{Comparisons with State-of-the-Art}

\textbf{Quantitative comparison.}
Firstly, we evaluated these methods in some simple real-world scenarios. We reported the quantitative scores on the MIT-Adobe 5K dataset. As shown in the first three rows in Table~\ref{tab: LLIEquan}, we can easily see that our method achieved the best numerical scores.
Further, we evaluated our method on the LOL dataset qualitatively and quantitatively, where the LOL dataset contained sensible noises to hinder the enhancement. 
The last three rows in Table~\ref{tab: LLIEquan} illustrated the quantitative comparisons. Obviously, our methods obtained a better PSNR, SSIM and LPIPS scores against other state-of-the-art methods. 

Finally, we verified the memory and computation efficiency. The last three rows in Table~\ref{tab: LLIEquan} compared the model size, FLOPs, and running time among different state-of-the-art methods. As can be observed that the FLOPs and running time were calculated on 100 testing images with the size of 600$\times$400 from the LOL dataset. Fortunately, our method just needed very small model size, FLOPs, and time. 
Even though its time consuming was a little higher than EnGAN, it is because that EnGAN ignored introducing the explicit noise removal module. In a word, our proposed RUAS has high-efficiency and fast speed.




\begin{table*}[t]
	\caption{Quantitative results on the DARK FACE dataset among different types of algorithms that contain the pattern of enhancer + detector (with different detectors), existing Low-Light (LL) Detectors and our method. The number in the first column represents the result after finetuning by directly using the corresponding detector on the low-light observations. The best result is in {\textcolor{red}{\textbf{red}}} whereas the second best one is in {\textcolor{blue}{\textbf{blue}}}.}
	\renewcommand\arraystretch{1.2} 
	\setlength{\tabcolsep}{0.8mm}
	\centering
	\begin{tabular}{|c|c|cccccccccc||c|c||c|}
		\hline
		\multicolumn{12}{|c||}{\footnotesize \footnotesize Enhancer + Detector}&\multicolumn{2}{c||}{\footnotesize LL Detector}&\footnotesize \multirow{2}{*}{Ours}\\		
		\cline{1-14}
		\footnotesize {Detector}&\footnotesize {Model}&\footnotesize MBLLEN&\footnotesize GLADNet&\footnotesize RetinexNet&\footnotesize EnGAN&\footnotesize SSIENet&\footnotesize KinD&\footnotesize DeepUPE&\footnotesize ZeroDCE&\footnotesize FIDE&\footnotesize DRBN&\footnotesize HLA&\footnotesize REG&\footnotesize ~\\
		\hline
		{\footnotesize DSFD}&\footnotesize Pre-train& \footnotesize 0.437&\footnotesize 0.520&\footnotesize 0.502&\footnotesize 0.476&\footnotesize 0.477&\footnotesize 0.421&\footnotesize 0.484&\footnotesize 0.526&\footnotesize 0.149&\footnotesize 0.354&\multirow{6}{*}{\footnotesize 0.607}& \multirow{6}{*}{\footnotesize 0.514}&\footnotesize \multirow{6}{*}{\footnotesize \textcolor{red}{\textbf{0.693}}} \\
		\cline{2-12}
		\footnotesize (0.596) &\footnotesize Finetune & \footnotesize 0.619&\footnotesize 0.643&\footnotesize 0.637&\footnotesize 0.621&\footnotesize 0.547&\footnotesize 0.605&\footnotesize 0.632&\footnotesize 0.657&\footnotesize 0.530&\footnotesize 0.607&\footnotesize ~&\footnotesize ~&\footnotesize \\
		\cline{1-12}
		{\footnotesize S3FD}&\footnotesize Pre-train&\footnotesize 0.361&\footnotesize 0.519&\footnotesize 0.458&\footnotesize 0.428&\footnotesize 0.416&\footnotesize 0.362&\footnotesize 0.444&\footnotesize 0.492&\footnotesize 0.128&\footnotesize 0.291&\footnotesize &\footnotesize &\footnotesize \\
		\cline{2-12}
		\footnotesize (0.619)&\footnotesize Finetune& \footnotesize 0.589&\footnotesize 0.638&\footnotesize 0.617&\footnotesize 0.634&\footnotesize 0.587&\footnotesize 0.566&\footnotesize 0.647&\footnotesize 0.645&\footnotesize 0.393&\footnotesize 0.559&\footnotesize &\footnotesize &\footnotesize \\
		\cline{1-12}
		{\footnotesize PyramidBox}&\footnotesize Pre-train & \footnotesize 0.440&\footnotesize 0.519&\footnotesize 0.500&\footnotesize 0.487&\footnotesize 0.484&\footnotesize 0.432&\footnotesize 0.484&\footnotesize 0.534&\footnotesize 0.127&\footnotesize 0.363& & &\footnotesize \\
		\cline{2-12}
		\footnotesize (0.592)&\footnotesize Finetune&\footnotesize  0.613& \footnotesize 0.648&\footnotesize 0.622&\footnotesize 0.639&\footnotesize 0.579&\footnotesize 0.594&\footnotesize 0.639&\footnotesize \textcolor{blue}{\textbf{0.658}}&\footnotesize 0.401&\footnotesize 0.562&\footnotesize ~&\footnotesize ~&\footnotesize ~\\
		\hline
	\end{tabular}
	\label{tab:detection}
\end{table*}

%

%

\begin{figure*}[t]
	\centering
	\begin{tabular}{c@{\extracolsep{0.6em}}c}
		\includegraphics[width=0.485\linewidth]{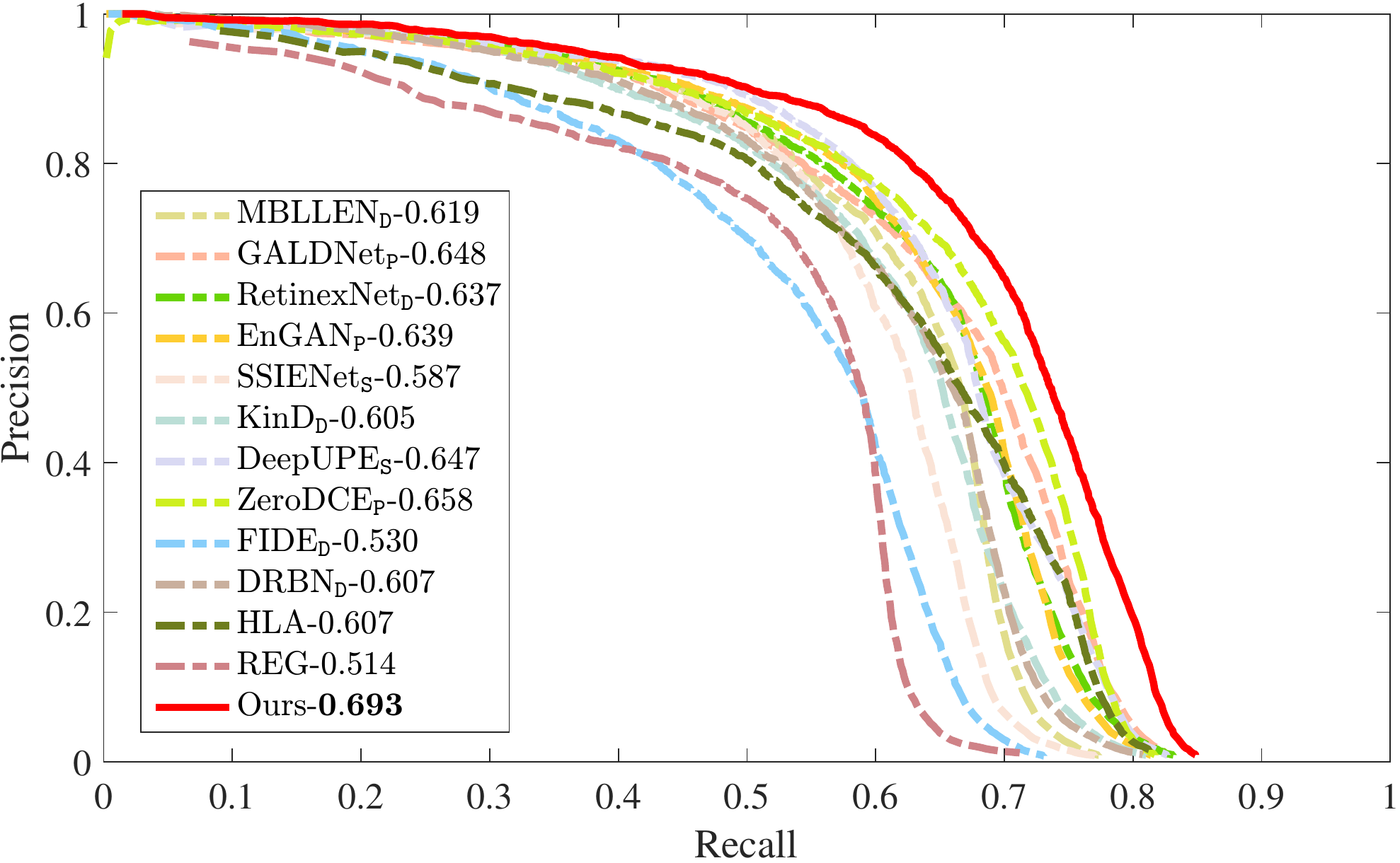}&
		\includegraphics[width=0.485\linewidth]{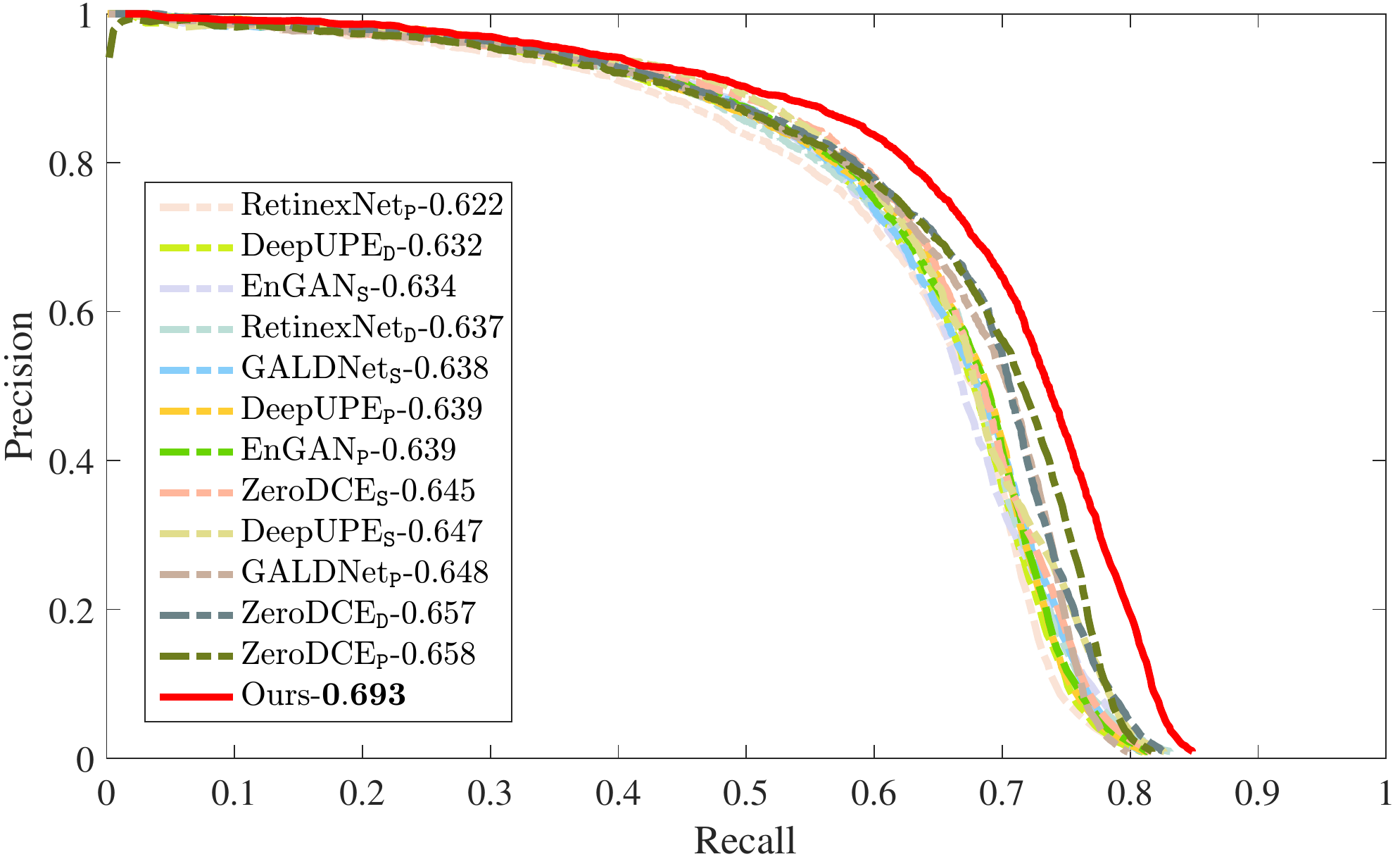}\\
		\footnotesize (a) &\footnotesize (b)\
	\end{tabular}	
	\caption{Precision-Recall (PR) curves for (a) all the compared methods that are with the best score among all cases in Table~\ref{tab:detection} and (b) the top thirteen methods appeared in  Table~\ref{tab:detection}. Note that the mark in the right-hand corner represents the detector, that is, ``$\mathtt{D}$'', ``$\mathtt{S}$'', ``$\mathtt{P}$'' denote the DSFD, S3FD, PyramidBox, respectively. } 
	\label{fig:Det1}
\end{figure*}

\begin{table*}[t]
	\centering
	\caption{Quantitative results of object detection on the ExDark dataset about the finetuned detector (SSD) on the enhanced results generated by all the compared methods. The best result is in {\textcolor{red}{\textbf{red}}} whereas the second best one is in {\textcolor{blue}{\textbf{blue}}}.}
	\renewcommand\arraystretch{1.2} 
	\setlength{\tabcolsep}{2.4mm}
	\begin{tabular}{|c|cccccccccccc|c|}
		\hline 
		\footnotesize Method&\footnotesize Bus&\footnotesize Car&\footnotesize Cat&\footnotesize Bicycle 
		&\footnotesize Dog &\footnotesize Boat&\footnotesize Cup&\footnotesize People&\footnotesize Chair&\footnotesize Motorbike&\footnotesize Bottle&\footnotesize Table&\footnotesize mAP\\
		\hline 
		\footnotesize MBLLEN&\footnotesize \textcolor{red}{\textbf{0.775}} &\footnotesize 0.613 &\footnotesize 0.576 &\footnotesize 0.592 &\footnotesize 0.522 &\footnotesize 0.556 &\footnotesize 0.423 &\footnotesize 0.460 &\footnotesize 0.431 &\footnotesize 0.376 &\footnotesize 0.400 &\footnotesize 0.195 &\footnotesize 0.493\\
		\hline 
		\footnotesize GLADNet&\footnotesize 0.665 &\footnotesize 0.564 &\footnotesize 0.548 &\footnotesize 0.553 &\footnotesize 0.468 &\footnotesize 0.545 &\footnotesize 0.301 &\footnotesize 0.417 &\footnotesize 0.378 &\footnotesize 0.258 &\footnotesize 0.365 &\footnotesize 0.164 &\footnotesize 0.436 \\
		\hline 
		
		\footnotesize RetinexNet&\footnotesize 0.574&\footnotesize 0.584&\footnotesize 0.545&\footnotesize 0.645&\footnotesize 0.567 &\footnotesize 0.547&\footnotesize 0.433&\footnotesize 0.449&\footnotesize 0.433&\footnotesize 0.321&\footnotesize 0.395&\footnotesize 0.194&\footnotesize 0.474\\
		\hline 
		
		\footnotesize DeepUPE&\footnotesize 0.693&\footnotesize 0.613&\footnotesize 0.623&\footnotesize  \textcolor{blue}{\textbf{0.650}}&\footnotesize  0.593&\footnotesize 0.596&\footnotesize 0.429&\footnotesize \textcolor{red}{\textbf{0.504}}&\footnotesize \textcolor{blue}{\textbf{0.487}}&\footnotesize 0.397&\footnotesize 0.429&\footnotesize 0.185&\footnotesize \textcolor{blue}{\textbf{0.516}}\\
		\hline 
		
		\footnotesize SSIENet&\footnotesize 0.724 &\footnotesize 0.615 &\footnotesize 0.614 &\footnotesize 0.528 &\footnotesize 0.489 &\footnotesize 0.493 &\footnotesize 0.385 &\footnotesize 0.430 &\footnotesize 0.389 &\footnotesize 0.372 &\footnotesize 0.388 &\footnotesize 0.184 &\footnotesize 0.467 \\
		\hline 
		
		\footnotesize EnGAN&\footnotesize 0.614&\footnotesize 0.559&\footnotesize 0.622&\footnotesize  \textcolor{red}{\textbf{0.664}}&\footnotesize  0.567&\footnotesize 0.520&\footnotesize 0.425&\footnotesize \textcolor{blue}{\textbf{0.496}}&\footnotesize \textcolor{red}{\textbf{0.492}}&\footnotesize 0.283&\footnotesize \textcolor{red}{\textbf{0.457}}&\footnotesize \textcolor{blue}{\textbf{0.206}}&\footnotesize 0.492\\
		\hline 
		
		\footnotesize ZeroDCE&\footnotesize 0.685&\footnotesize \textcolor{blue}{\textbf{0.650}}&\footnotesize 0.594&\footnotesize  0.610&\footnotesize  0.531&\footnotesize 0.553&\footnotesize 0.460&\footnotesize 0.487&\footnotesize 0.421&\footnotesize \textcolor{blue}{\textbf{0.423}}&\footnotesize \textcolor{blue}{\textbf{0.445}}&\footnotesize 0.137&\footnotesize 0.500\\
		\hline 
		\footnotesize FIDE&\footnotesize 0.634&\footnotesize 0.586&\footnotesize 0.607&\footnotesize  0.593&\footnotesize  0.557&\footnotesize 0.575&\footnotesize 0.417&\footnotesize 0.430&\footnotesize 0.367&\footnotesize 0.357&\footnotesize 0.407&\footnotesize 0.161&\footnotesize 0.474\\
		\hline 
		\footnotesize DRBN&\footnotesize 0.637&\footnotesize 0.562&\footnotesize 0.527&\footnotesize 0.541&\footnotesize  0.593&\footnotesize 0.491&\footnotesize 0.411&\footnotesize 0.454&\footnotesize 0.481&\footnotesize 0.353&\footnotesize 0.386&\footnotesize 0.193&\footnotesize 0.469\\
		\hline 
		
		\footnotesize KinD&\footnotesize \textcolor{blue}{\textbf{0.727}}&\footnotesize \textcolor{red}{\textbf{0.654}}&\footnotesize \textcolor{blue}{\textbf{0.625}}&\footnotesize  0.578 &\footnotesize 0.597&\footnotesize \textcolor{red}{\textbf{0.622}}&\footnotesize \textcolor{blue}{\textbf{0.462}}&\footnotesize 0.464&\footnotesize 0.451&\footnotesize 0.391&\footnotesize 0.394 &\footnotesize 0.194&\footnotesize 0.513\\
		\hline
		\footnotesize Ours&\footnotesize 0.716&\footnotesize 0.638&\footnotesize \textcolor{red}{\textbf{0.724}}&\footnotesize 0.569&\footnotesize  \textcolor{red}{\textbf{0.629}}&\footnotesize \textcolor{blue}{\textbf{0.597}}&\footnotesize \textcolor{red}{\textbf{0.463}}&\footnotesize 0.476 &\footnotesize 0.478&\footnotesize \textcolor{red}{\textbf{0.472}} &\footnotesize 0.420&\footnotesize \textcolor{red}{\textbf{0.256}}&\footnotesize \color{red}{\textbf{0.536}}\\
		\hline 
	\end{tabular}
	\label{tab: ExDark}
\end{table*}

\begin{figure*}[t]
	\centering
	\begin{tabular}{c@{\extracolsep{0.3em}}c@{\extracolsep{0.3em}}c@{\extracolsep{0.3em}}c@{\extracolsep{0.3em}}c@{\extracolsep{0.3em}}c@{\extracolsep{0.3em}}c@{\extracolsep{0.3em}}c@{\extracolsep{0.3em}}c@{\extracolsep{0.3em}}c}
		\includegraphics[width=0.093\linewidth]{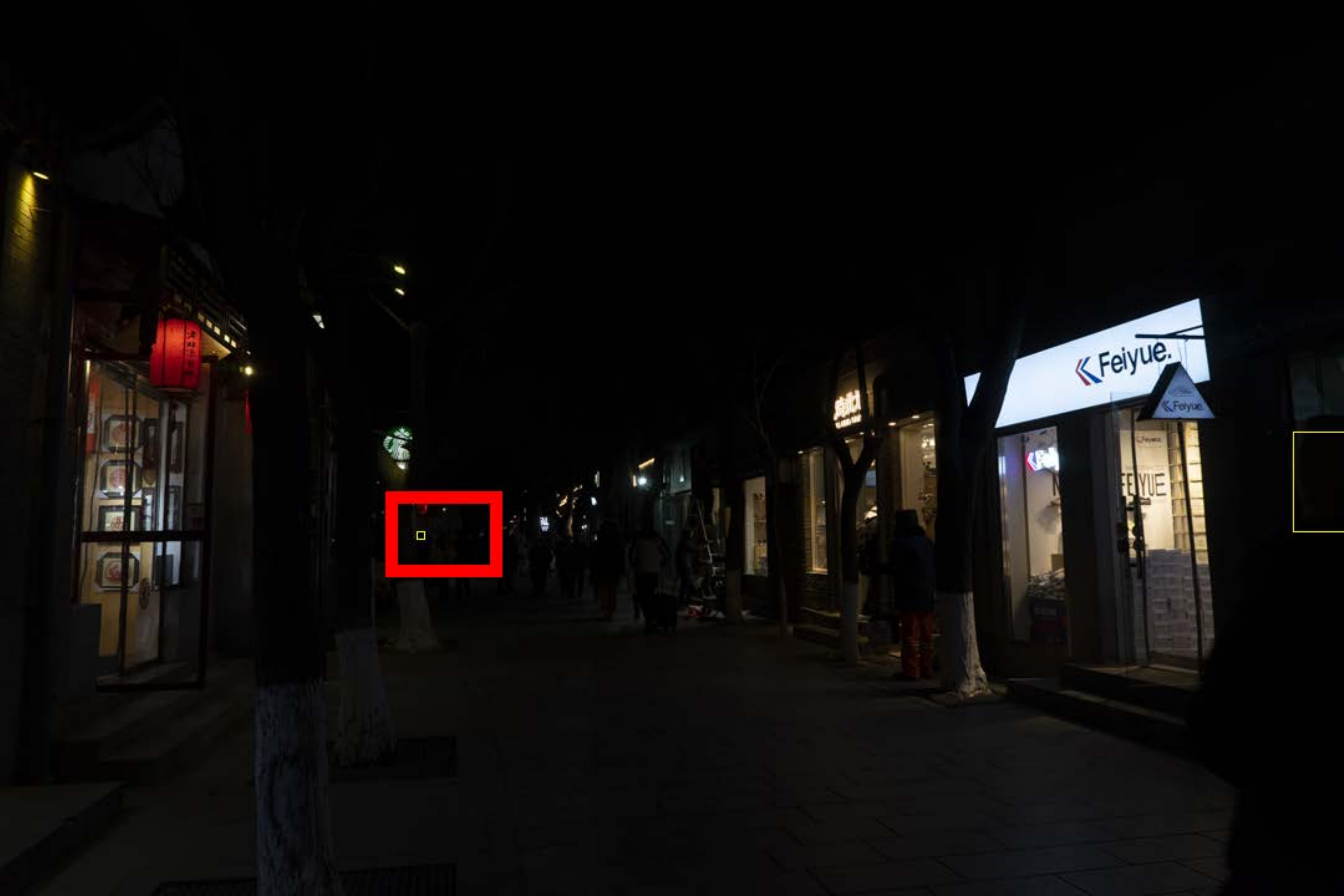}&
		\includegraphics[width=0.093\linewidth]{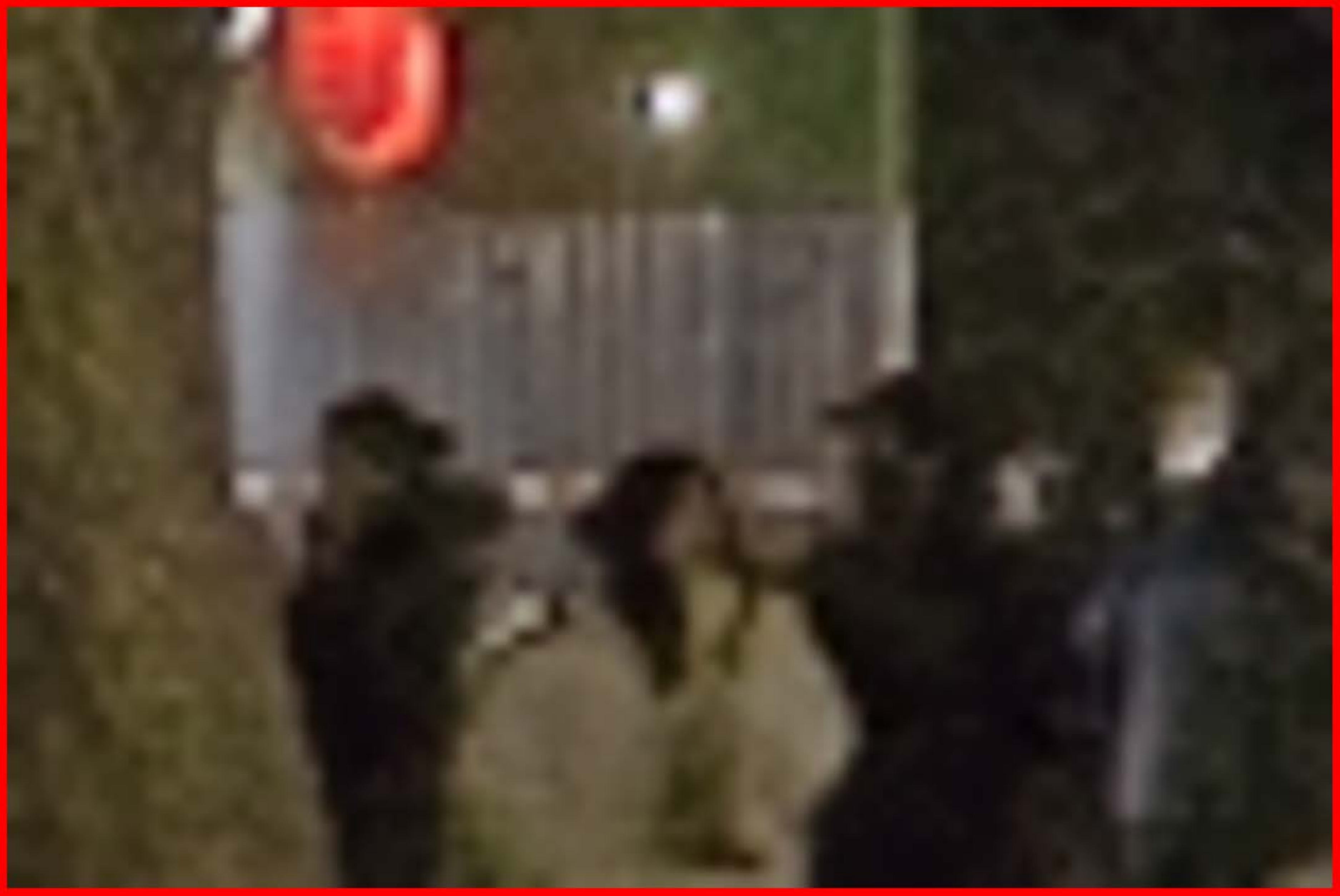}&
		\includegraphics[width=0.093\linewidth]{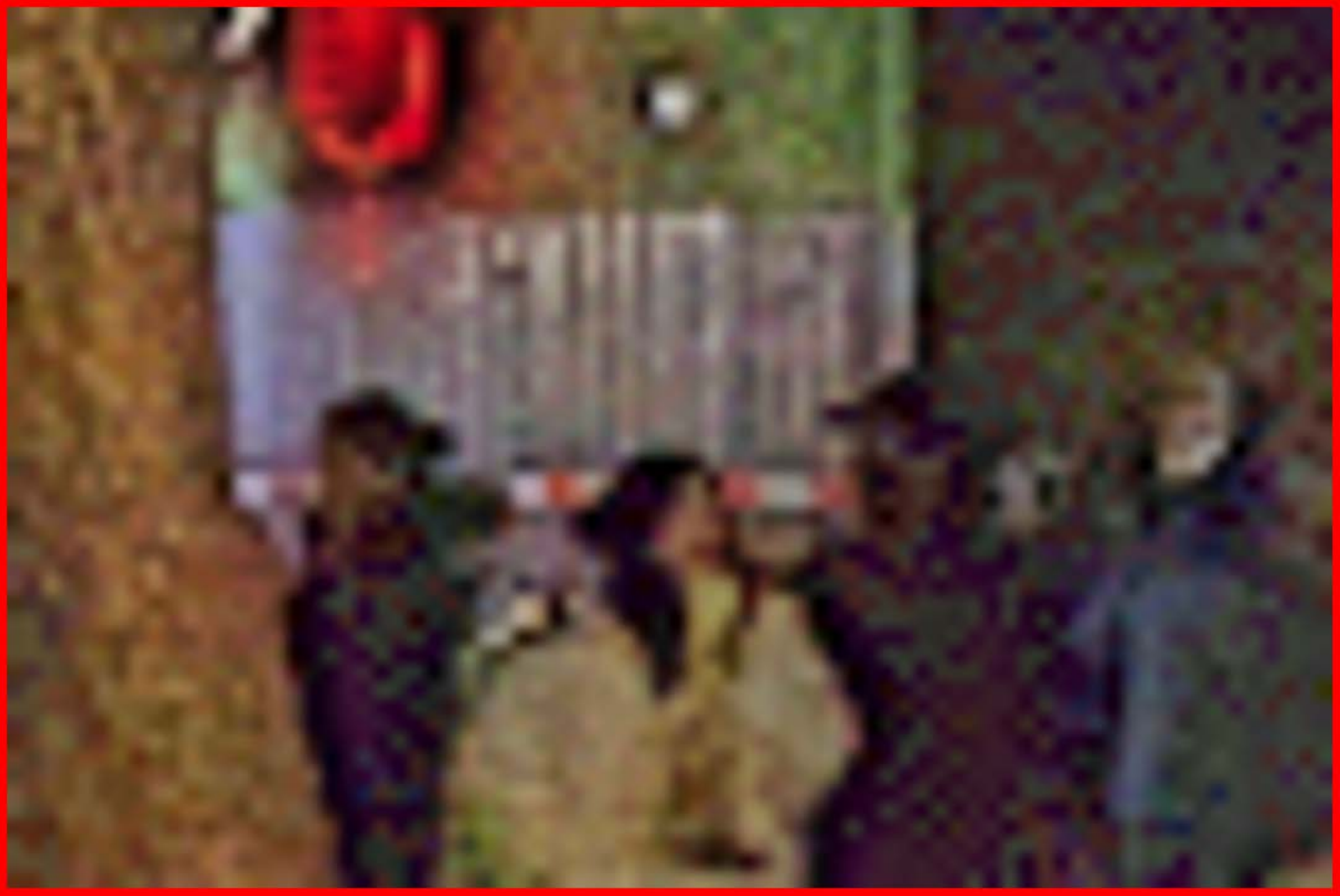}&
		\includegraphics[width=0.093\linewidth]{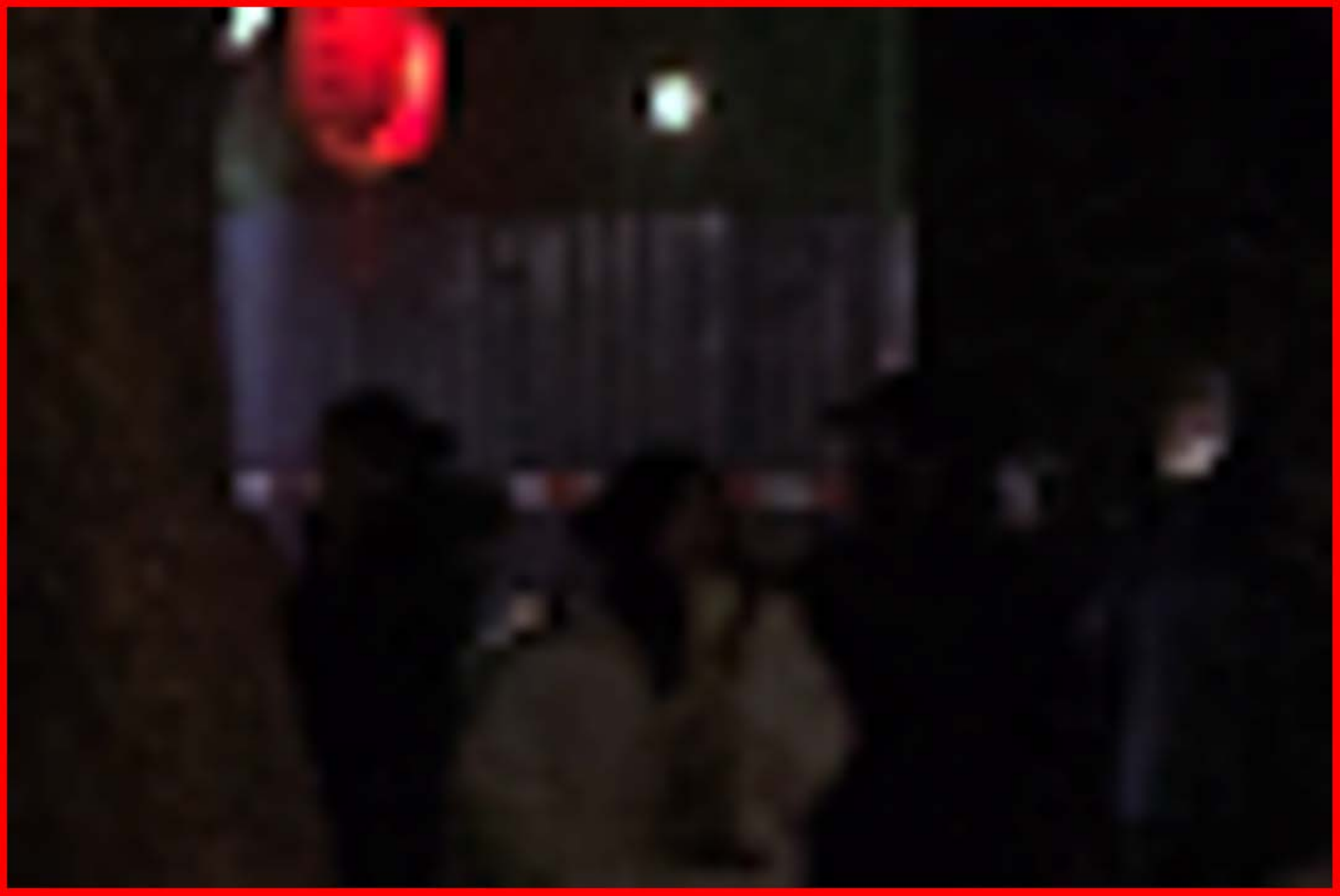}&
		\includegraphics[width=0.093\linewidth]{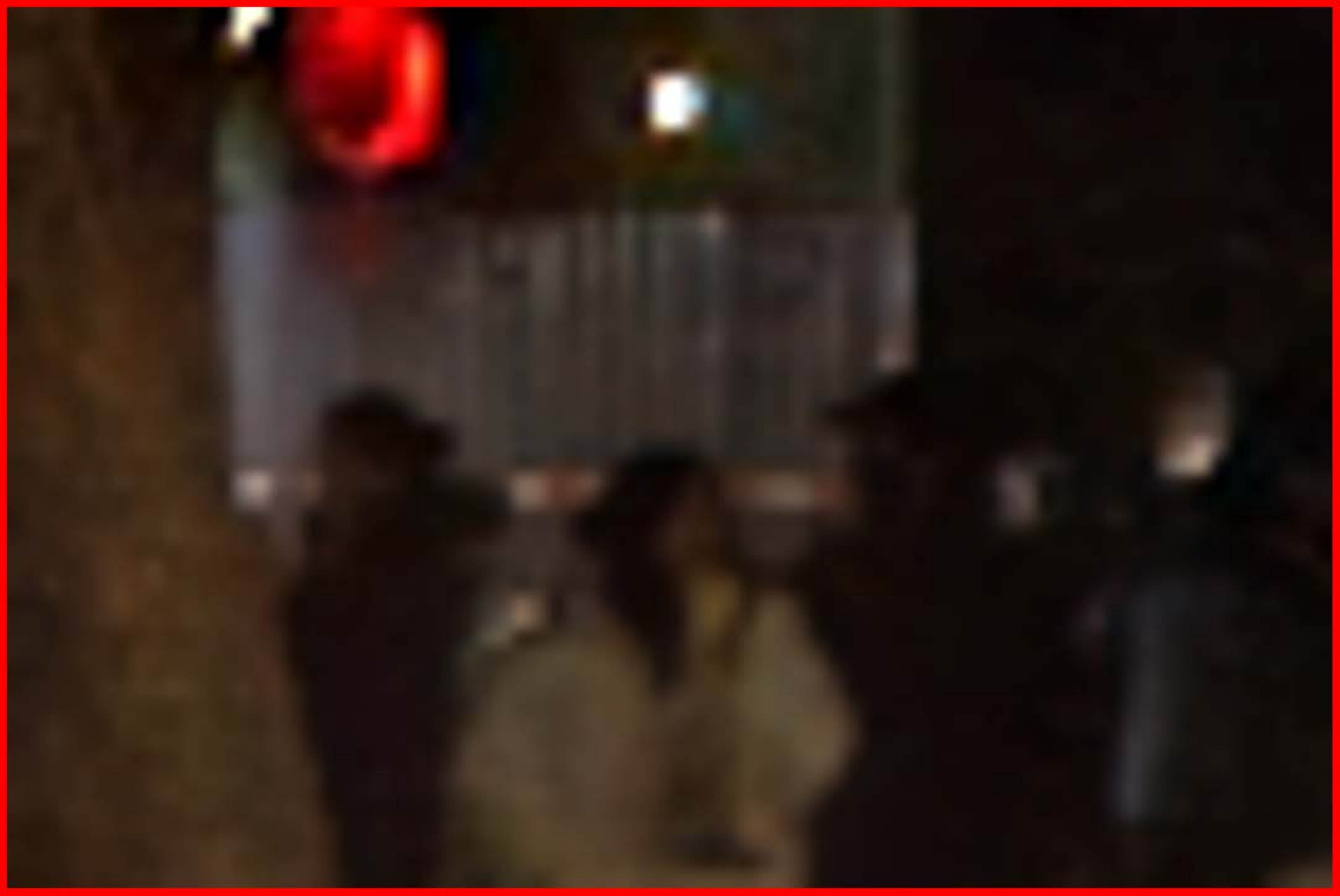}&
		\includegraphics[width=0.093\linewidth]{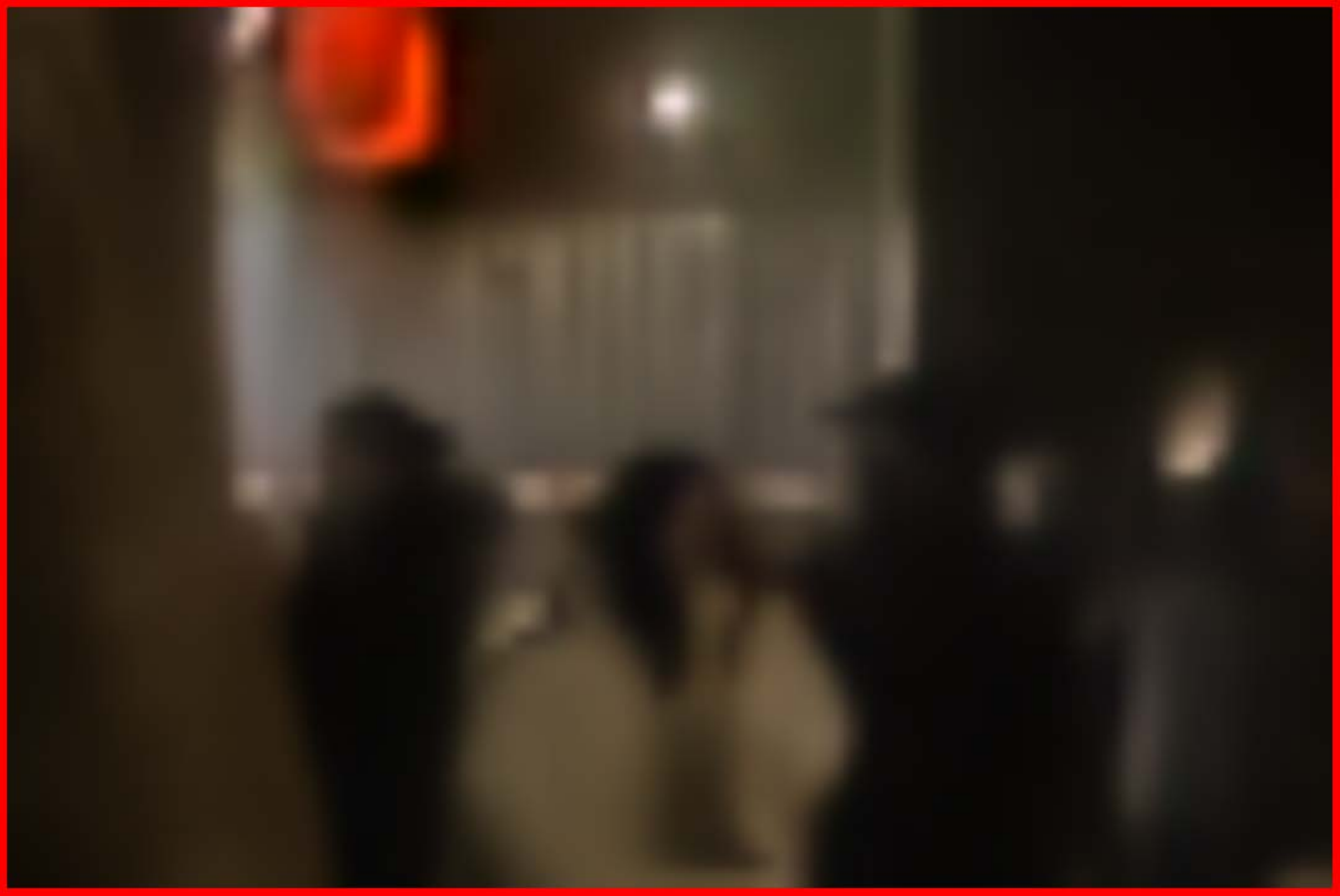}&
		\includegraphics[width=0.093\linewidth]{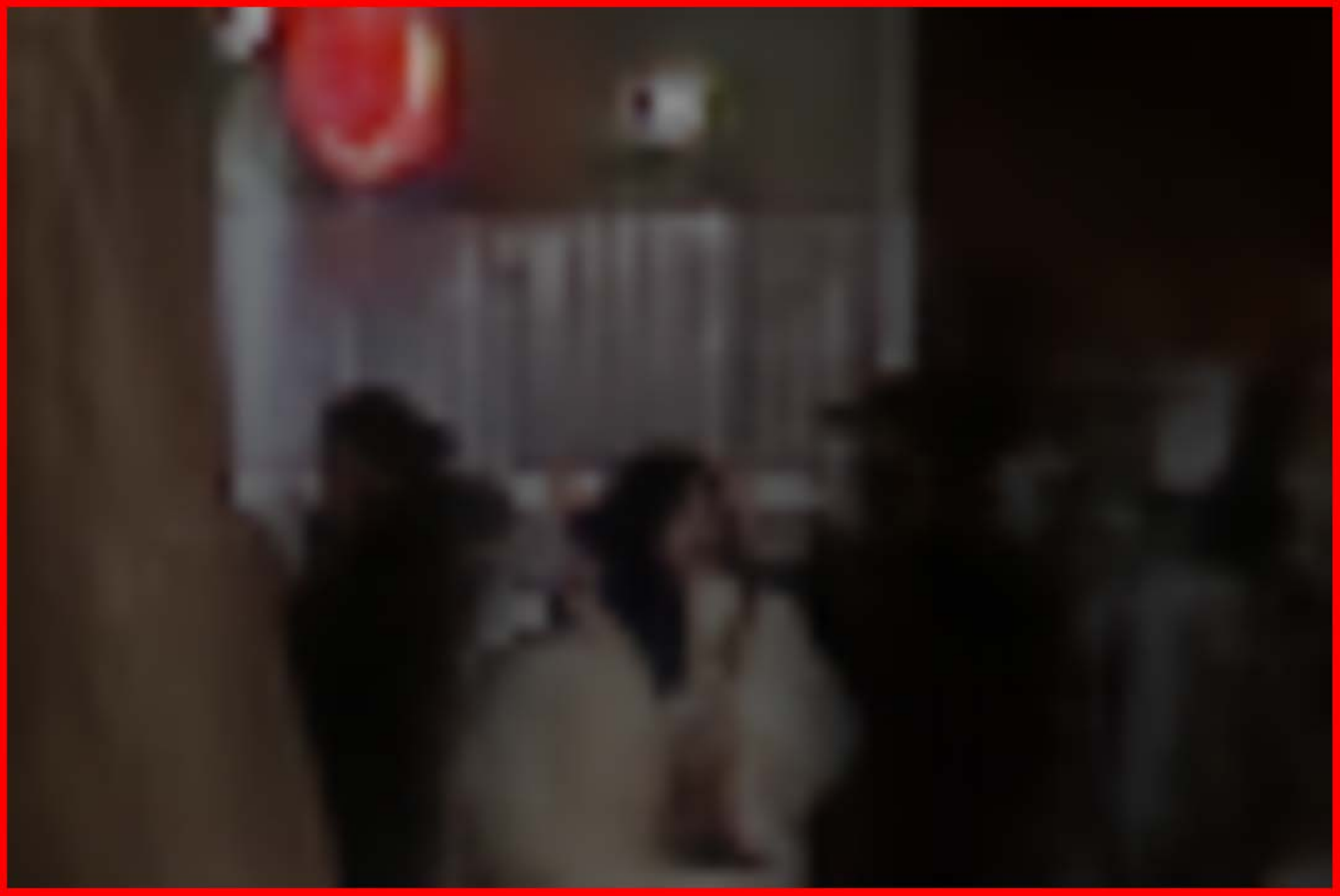}&
		\includegraphics[width=0.093\linewidth]{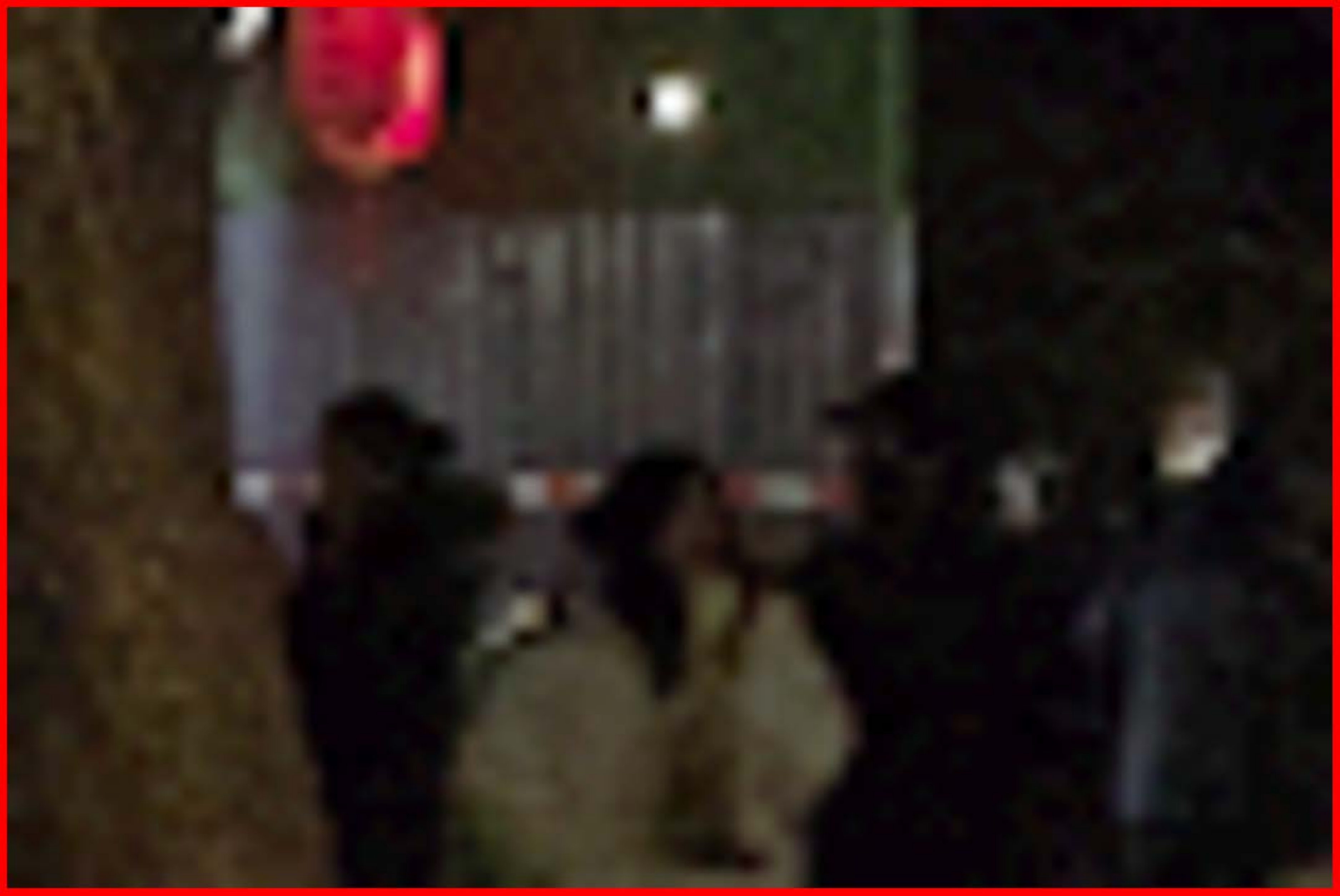}&
		\includegraphics[width=0.093\linewidth]{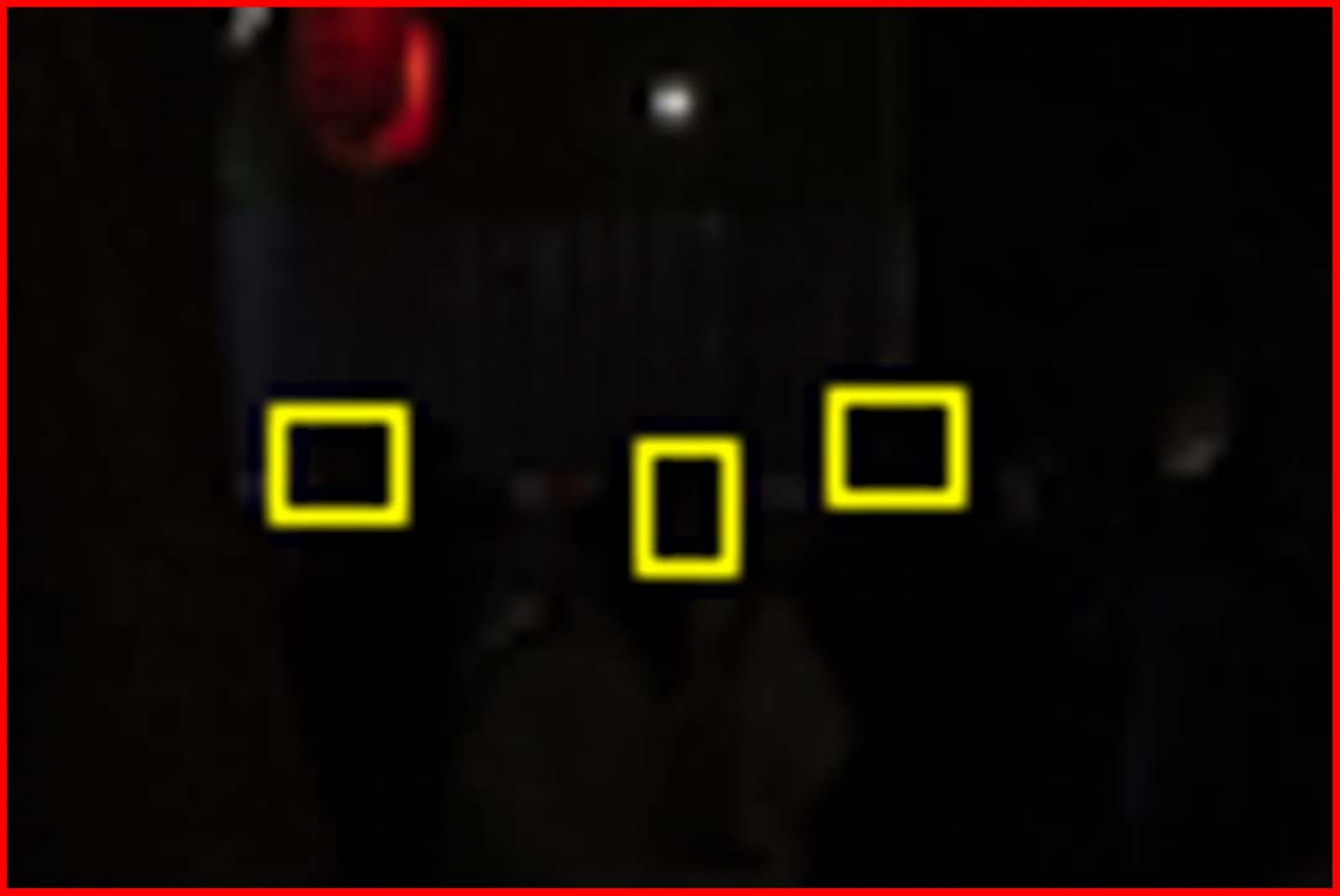}&
		\includegraphics[width=0.093\linewidth]{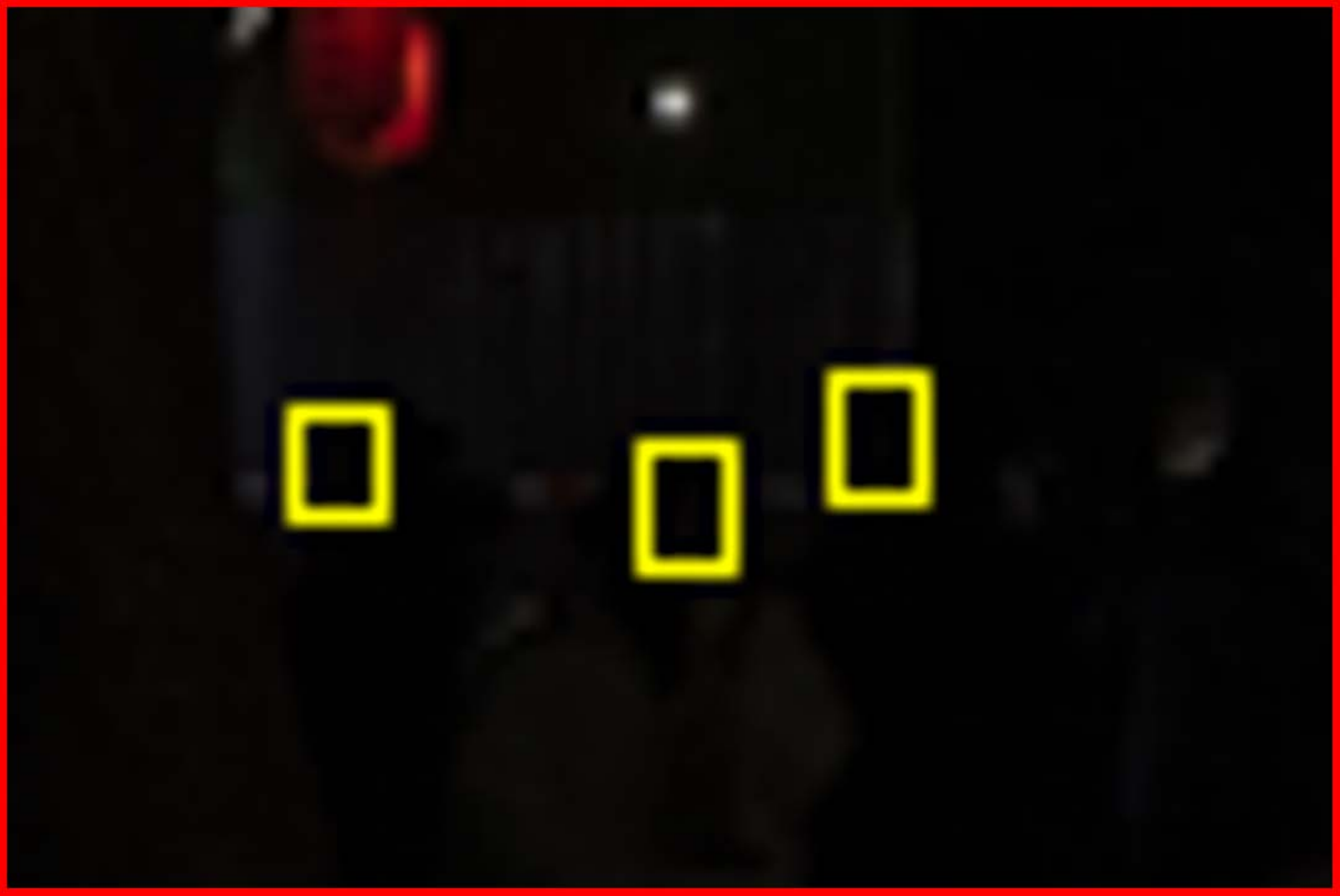}\\
		\includegraphics[width=0.093\linewidth]{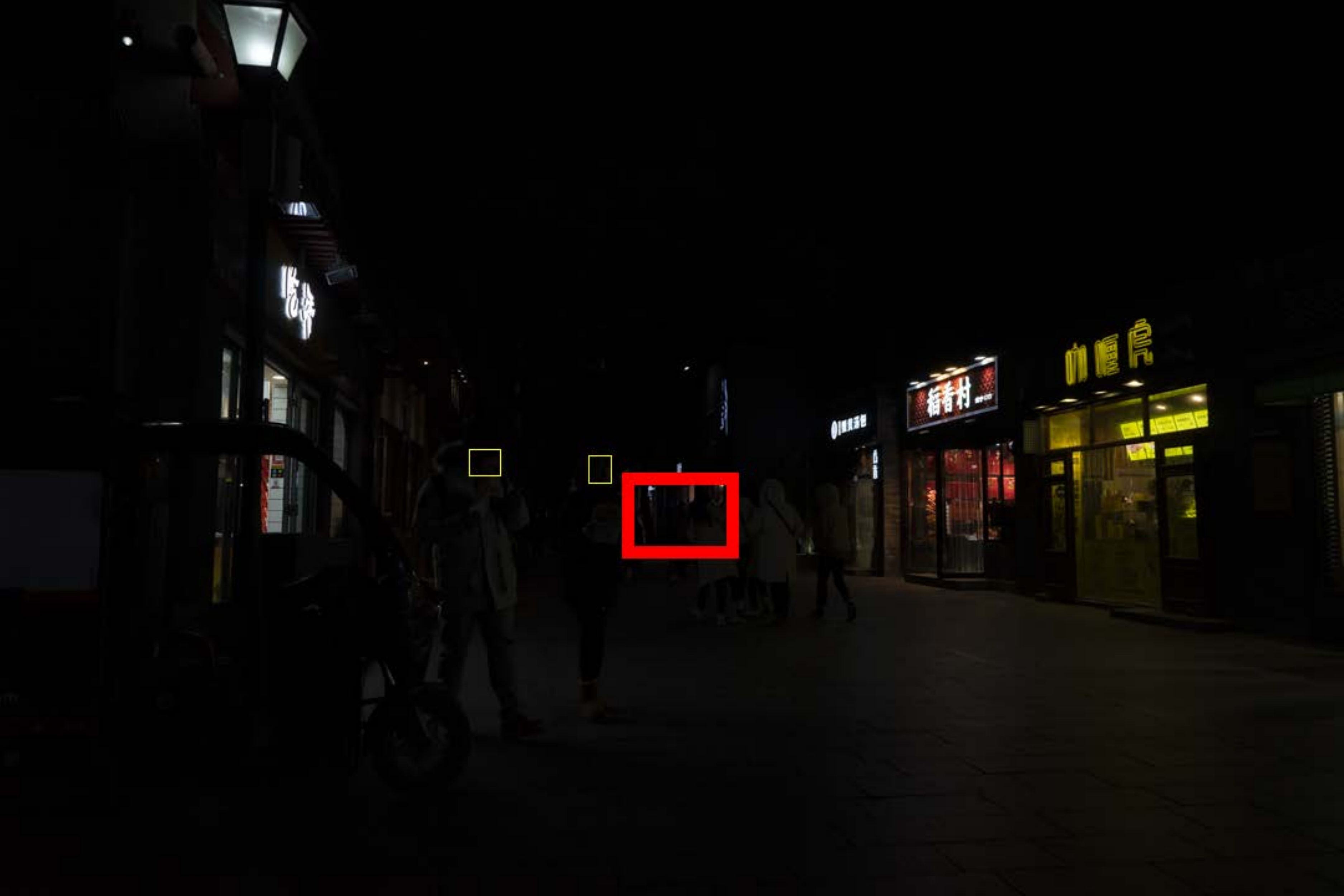}&
		\includegraphics[width=0.093\linewidth]{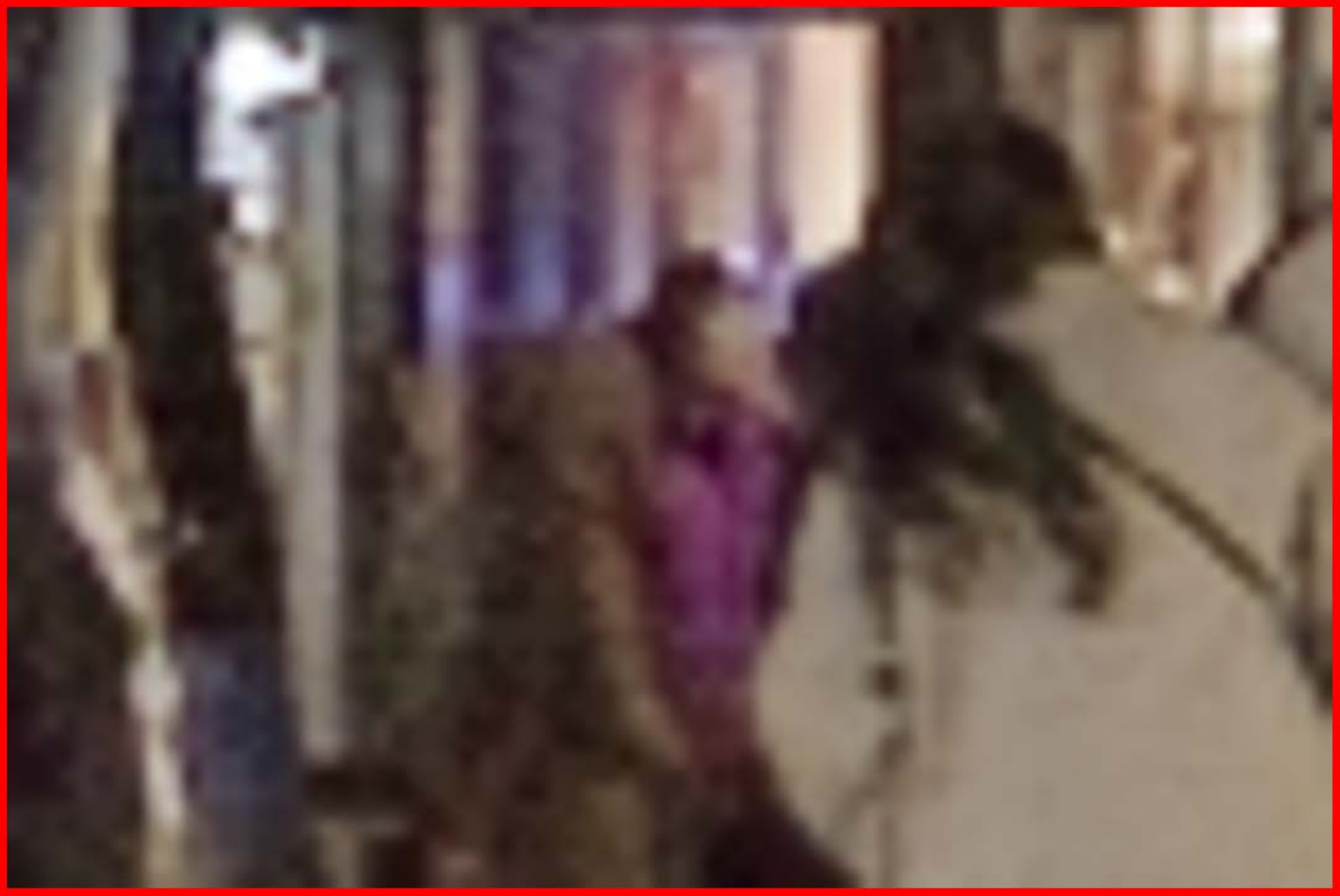}&
		\includegraphics[width=0.093\linewidth]{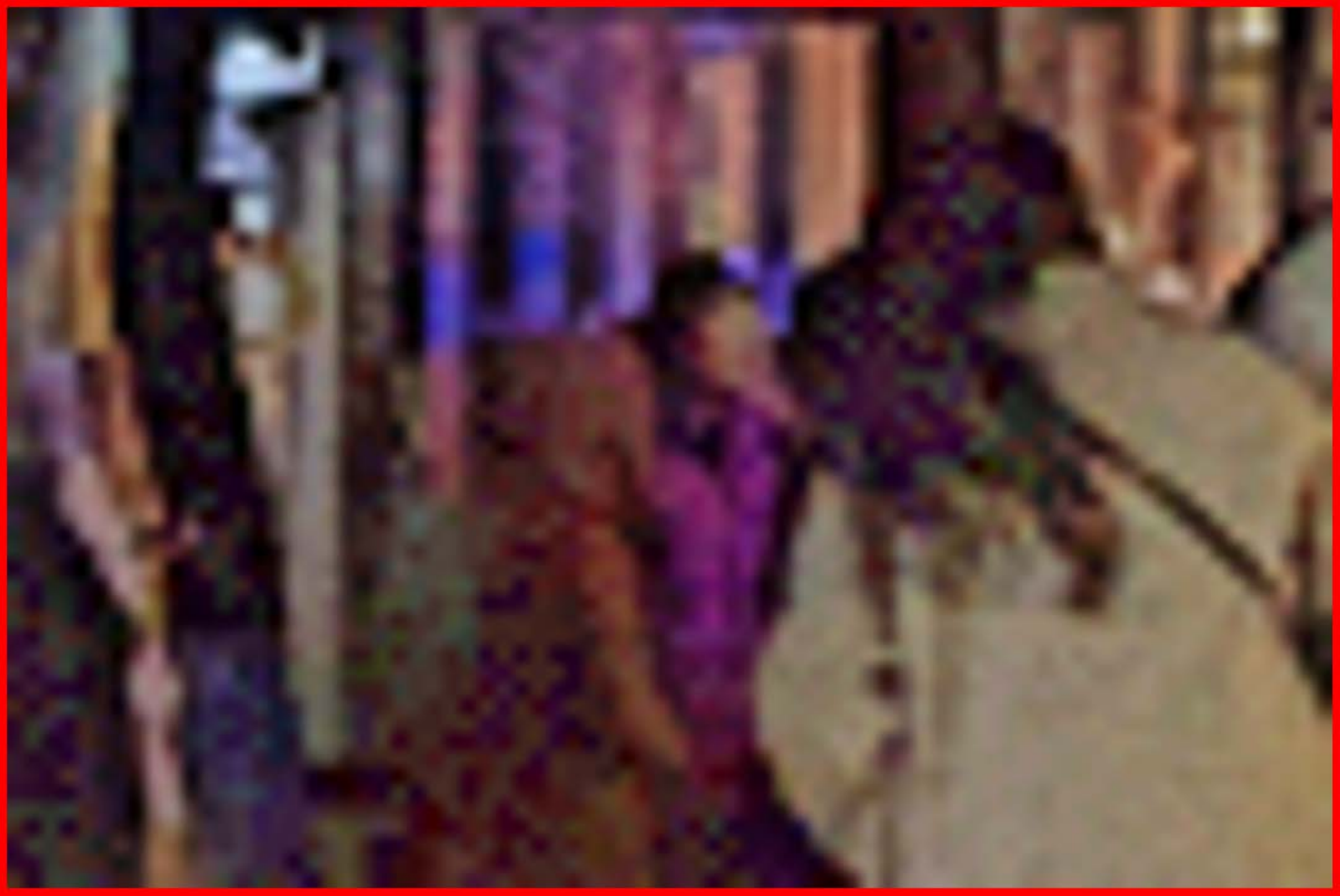}&
		\includegraphics[width=0.093\linewidth]{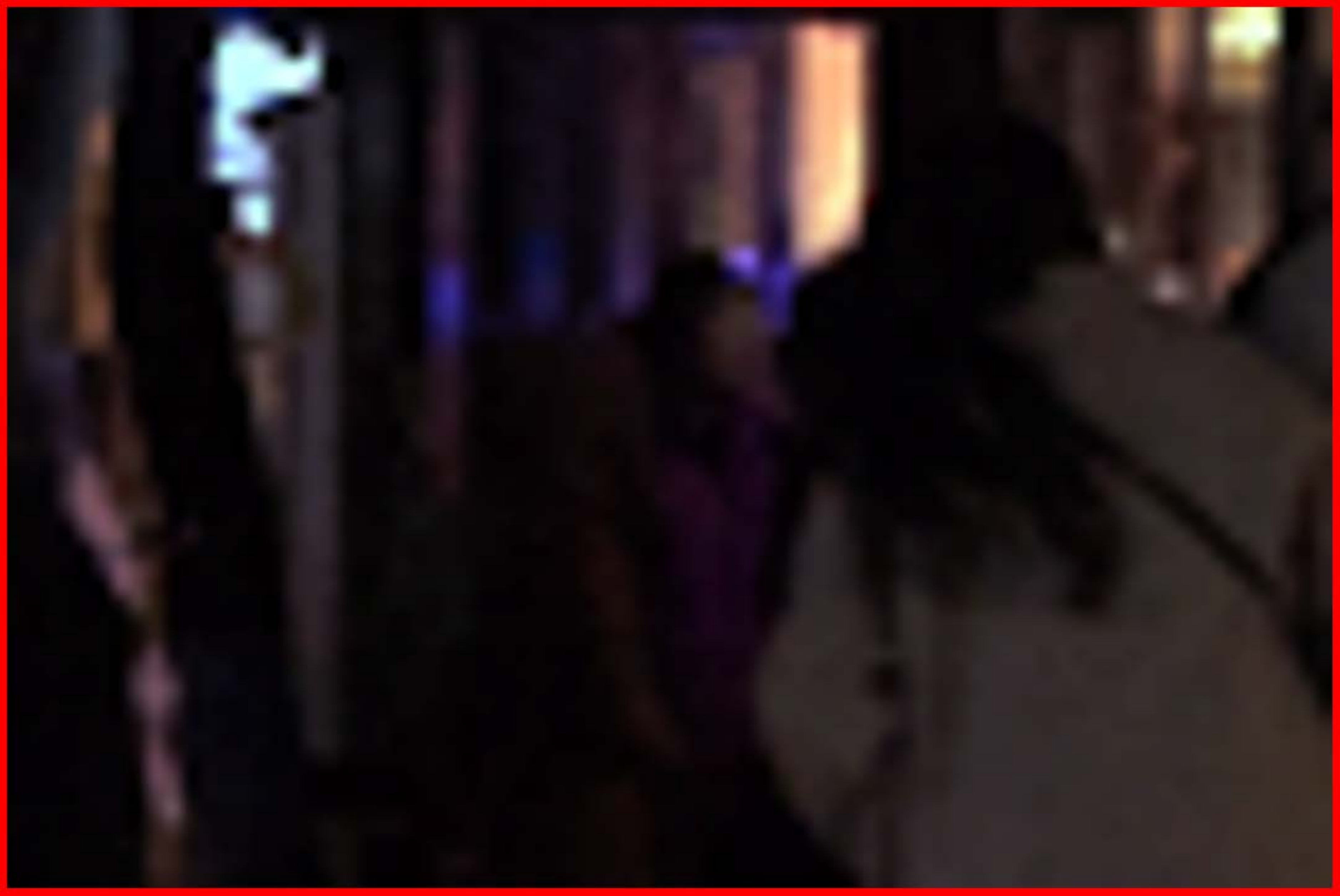}&
		\includegraphics[width=0.093\linewidth]{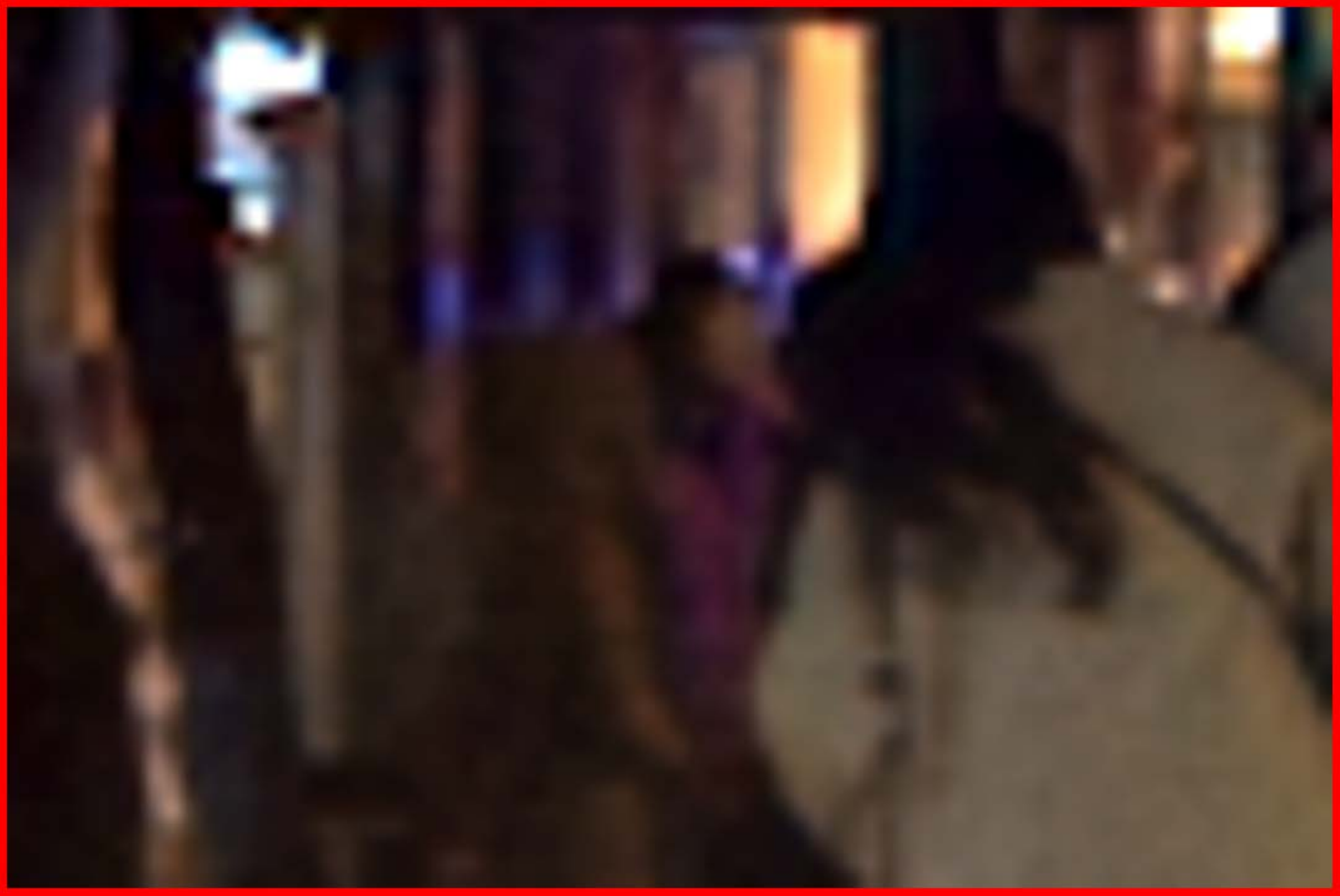}&
		\includegraphics[width=0.093\linewidth]{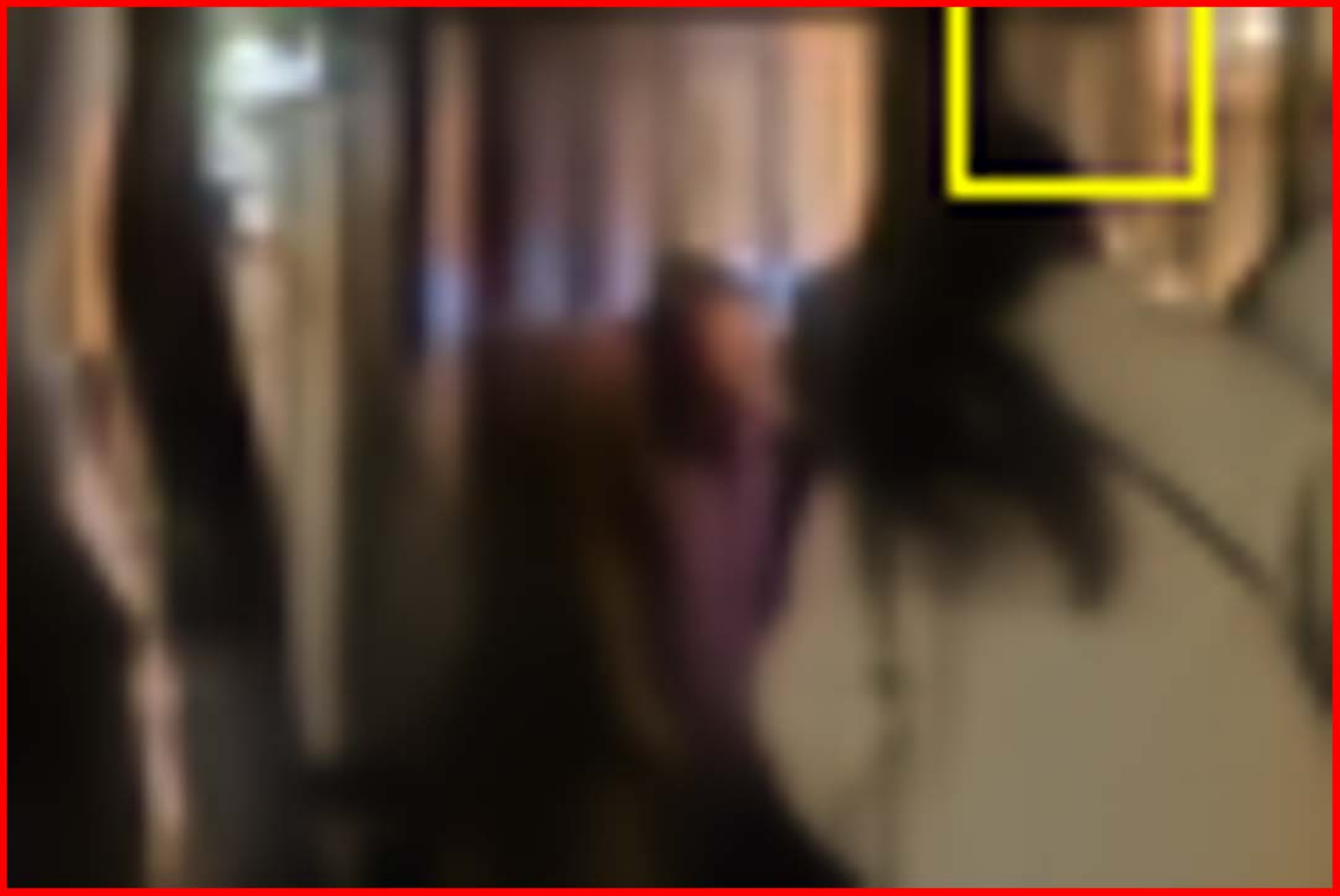}&
		\includegraphics[width=0.093\linewidth]{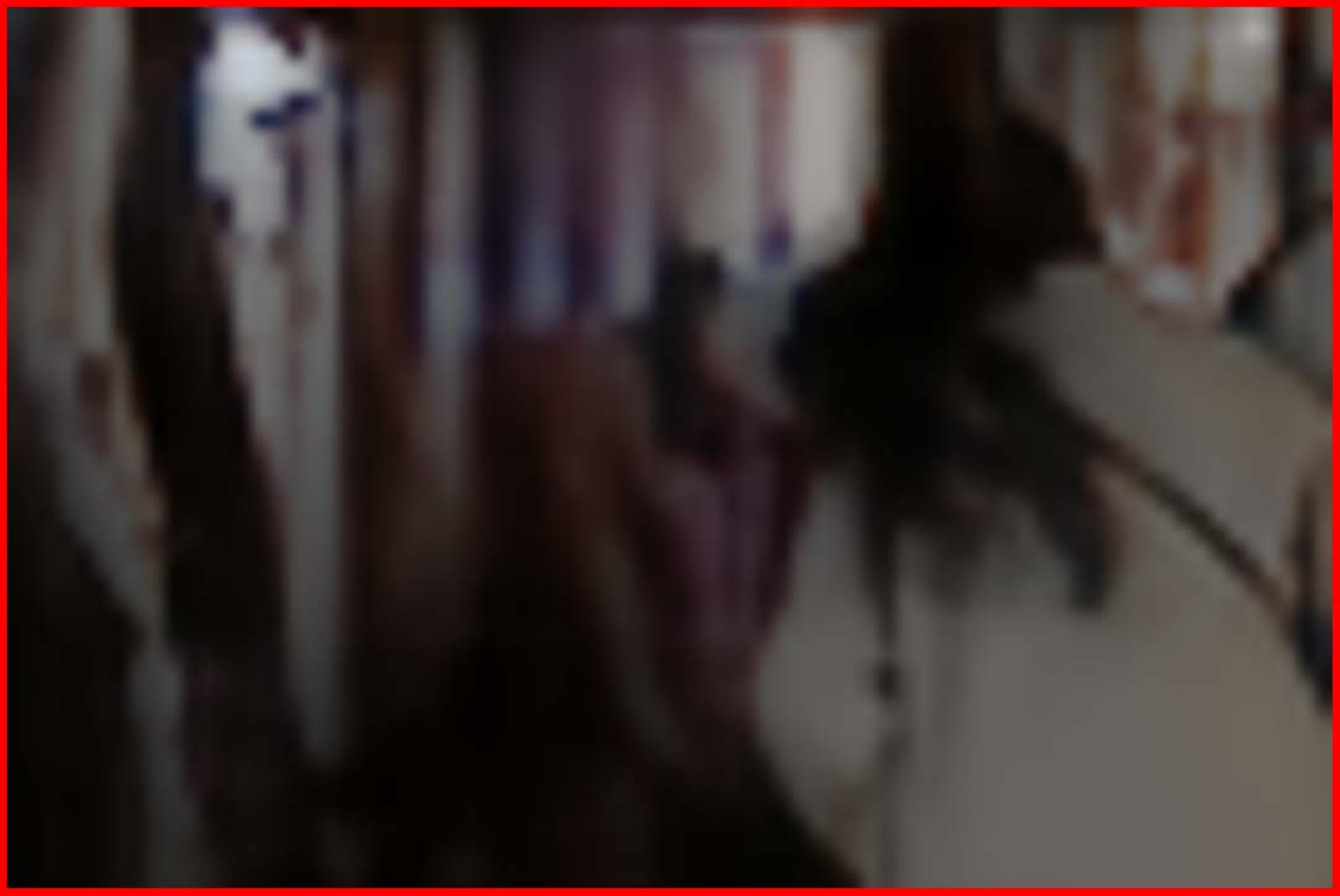}&
		\includegraphics[width=0.093\linewidth]{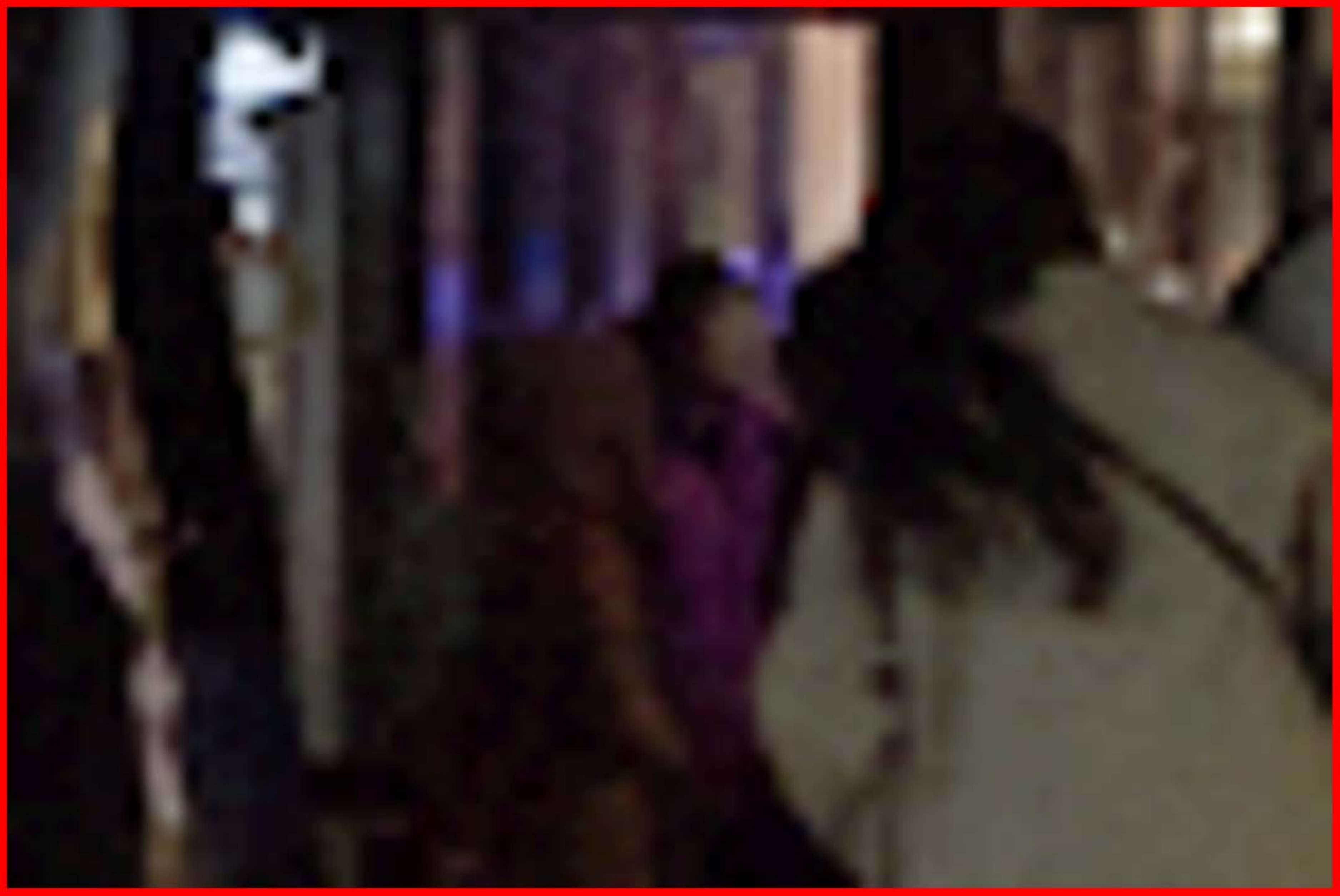}&
		\includegraphics[width=0.093\linewidth]{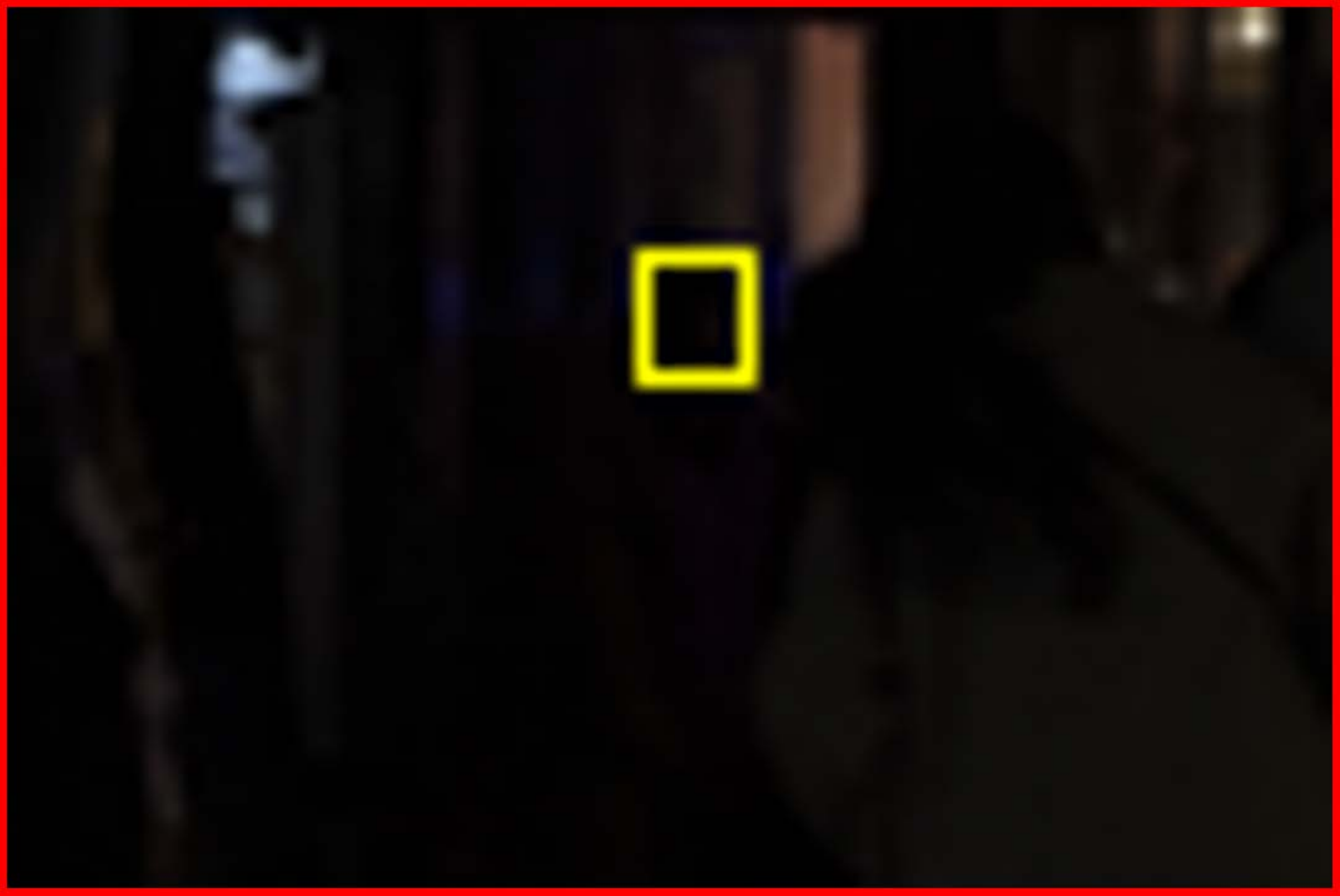}&
		\includegraphics[width=0.093\linewidth]{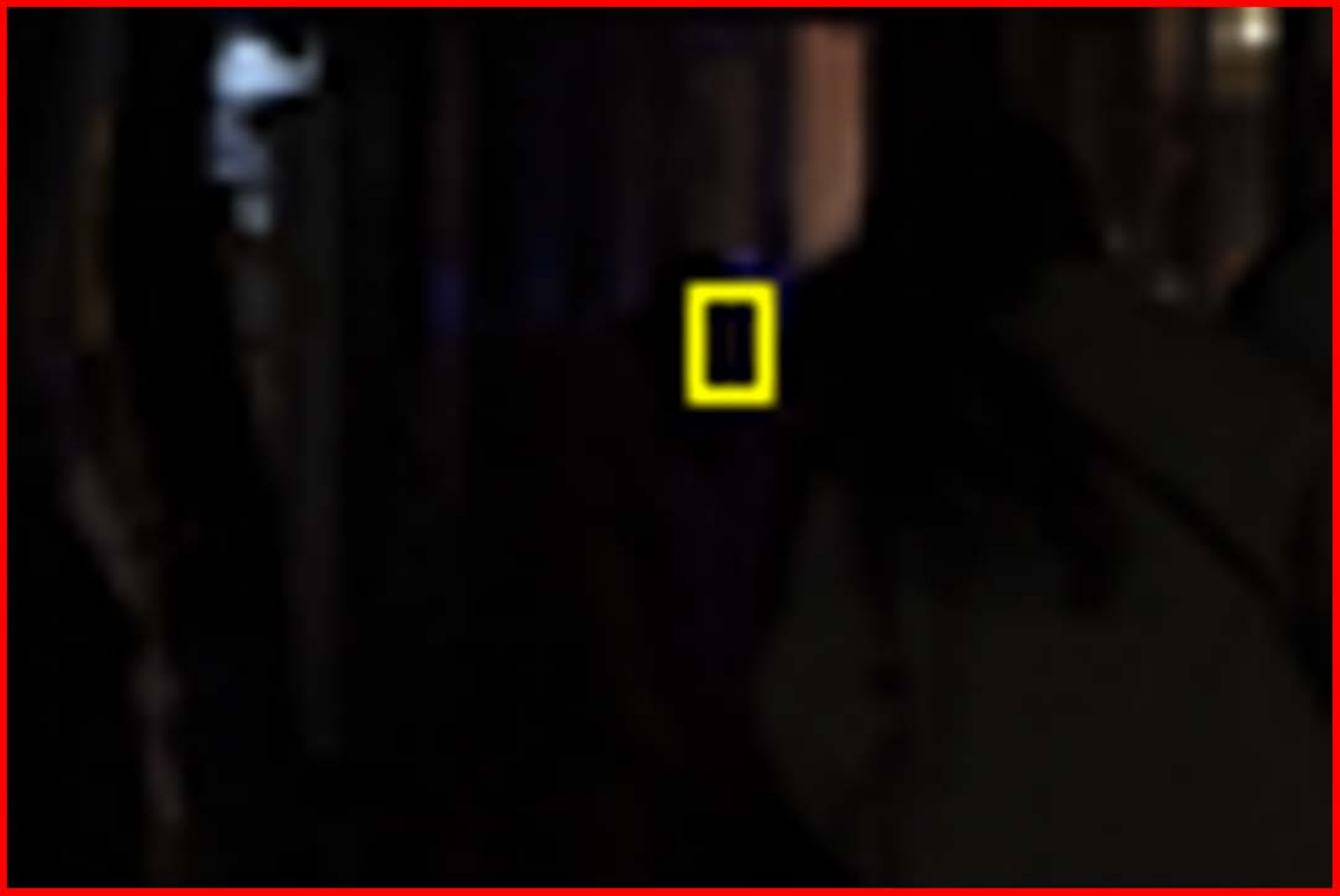}\\
		\includegraphics[width=0.093\linewidth]{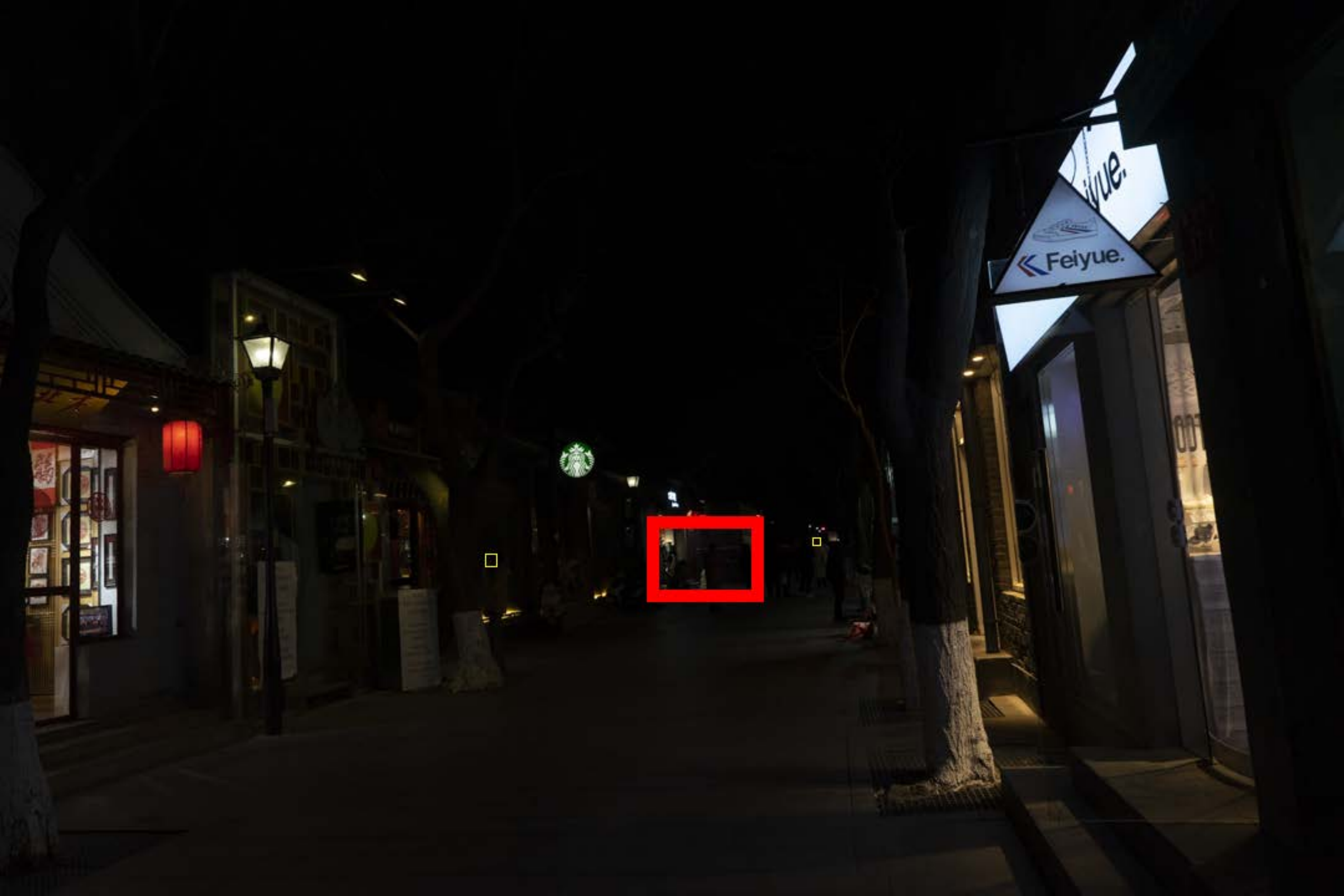}&
		\includegraphics[width=0.093\linewidth]{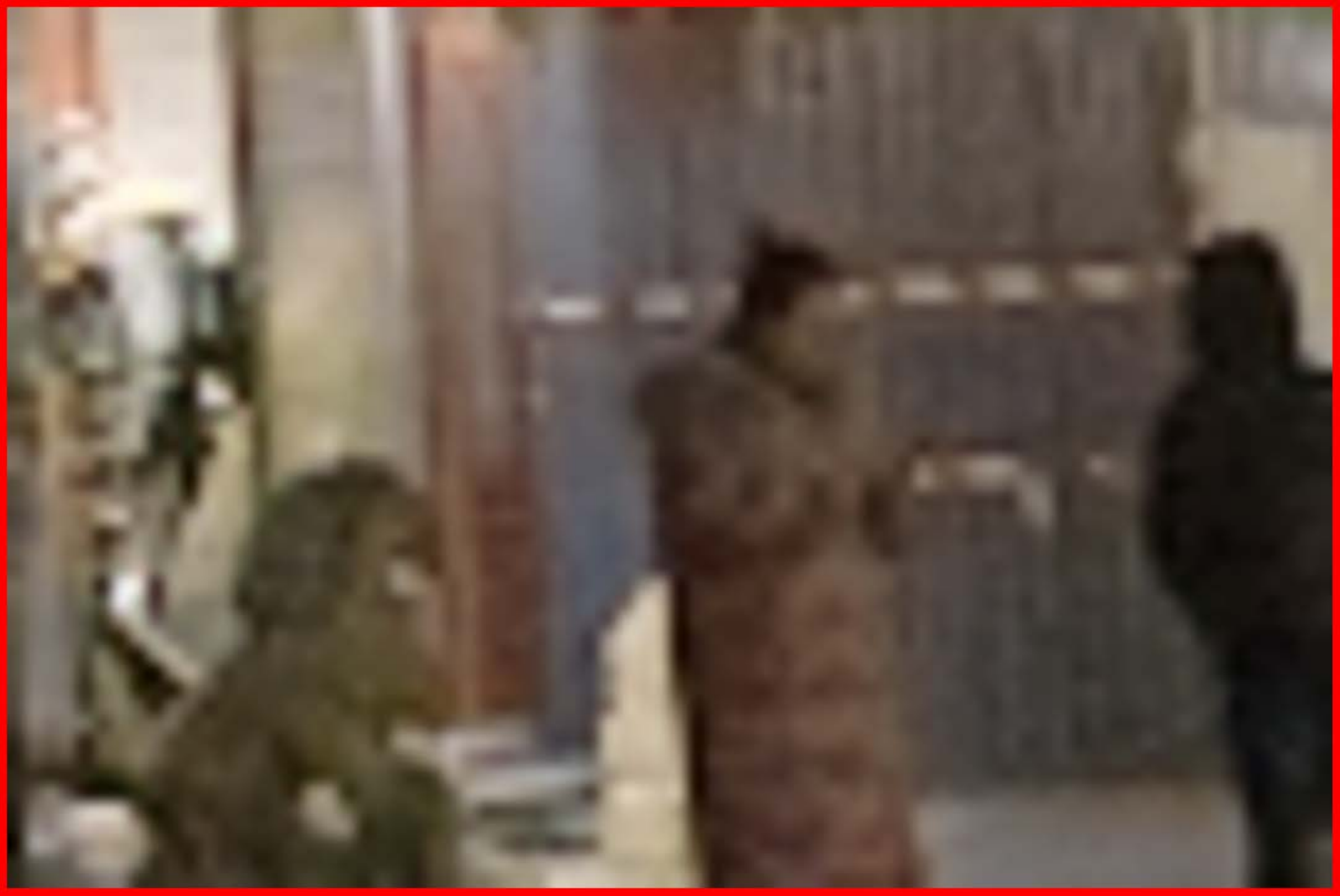}&
		\includegraphics[width=0.093\linewidth]{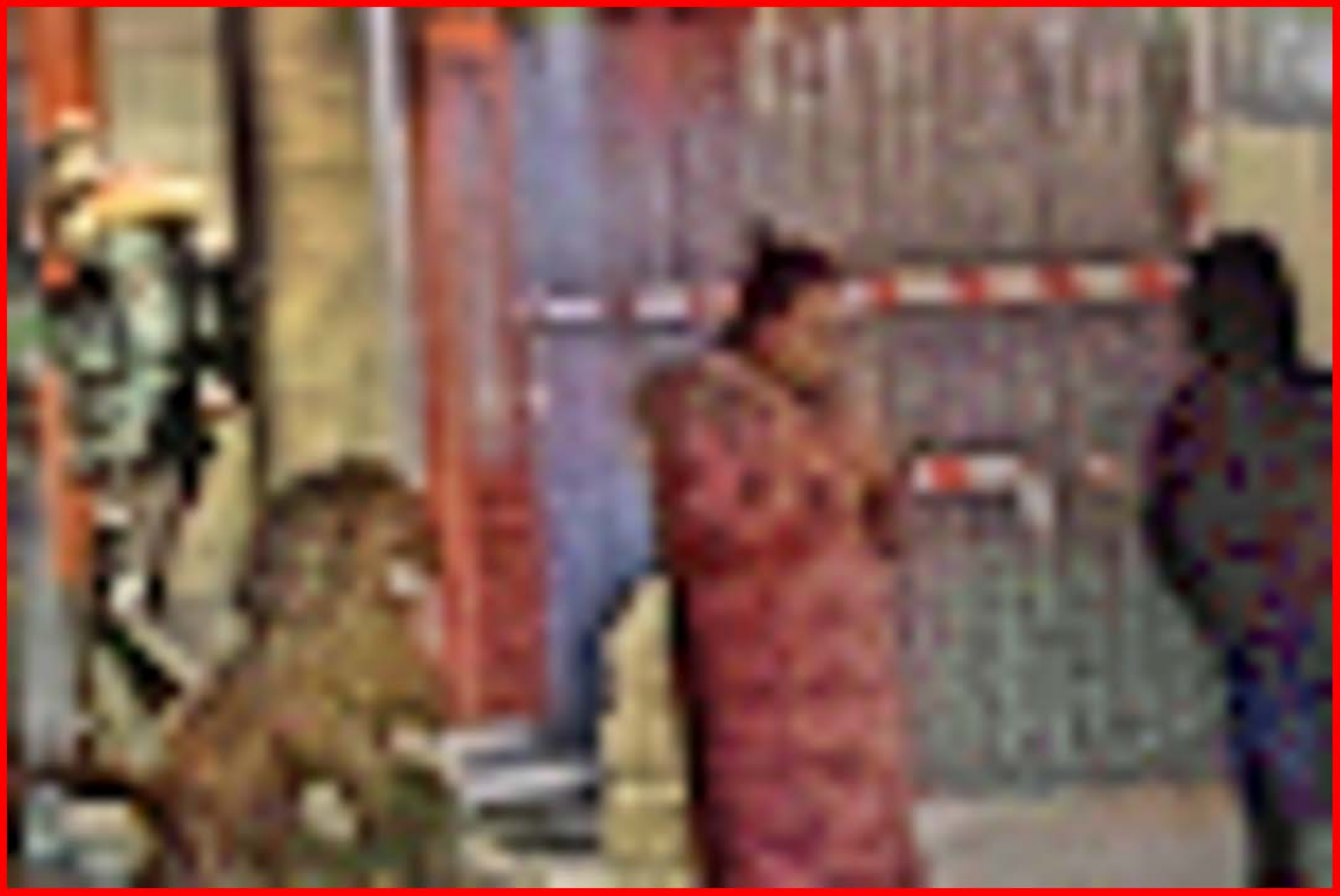}&
		\includegraphics[width=0.093\linewidth]{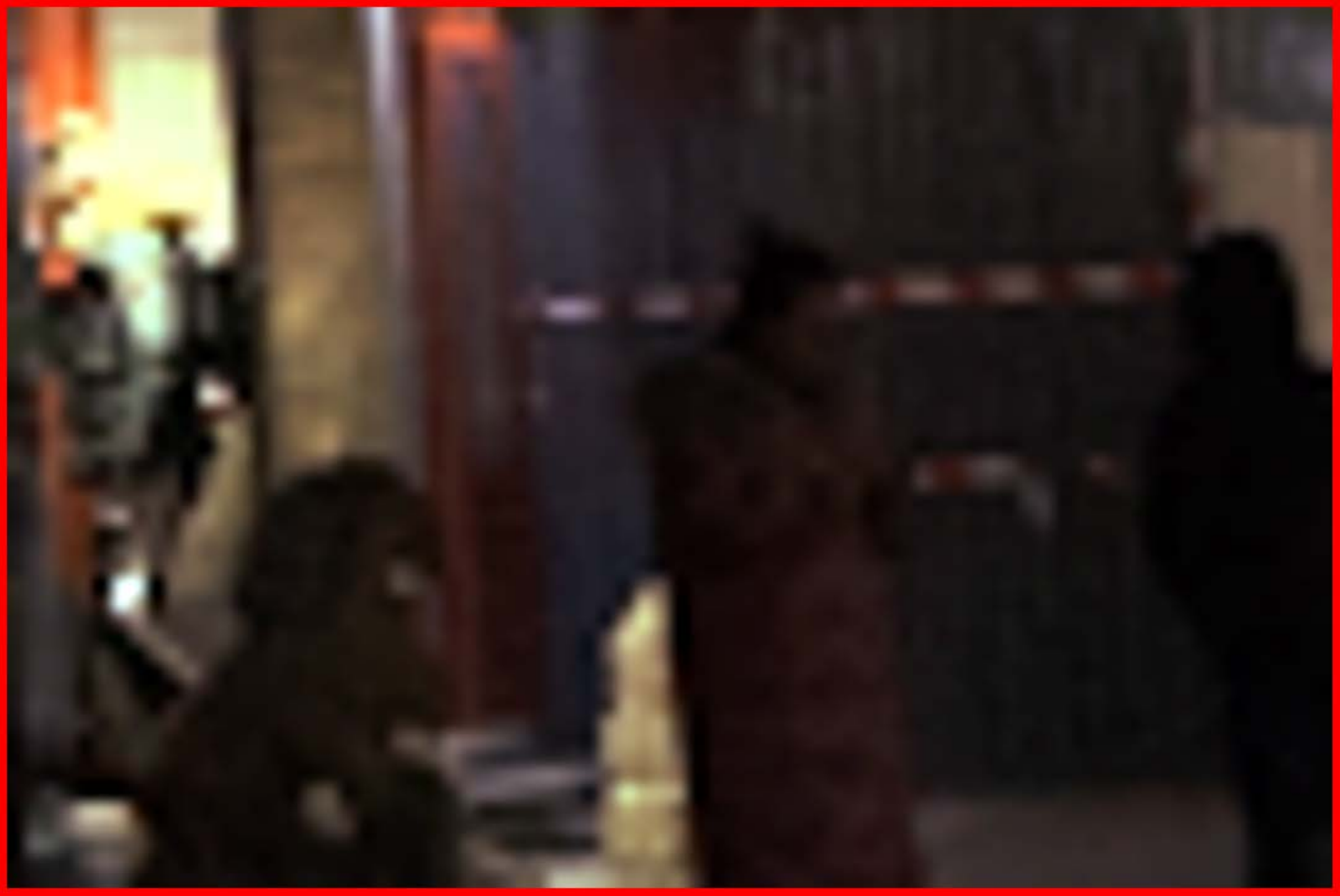}&
		\includegraphics[width=0.093\linewidth]{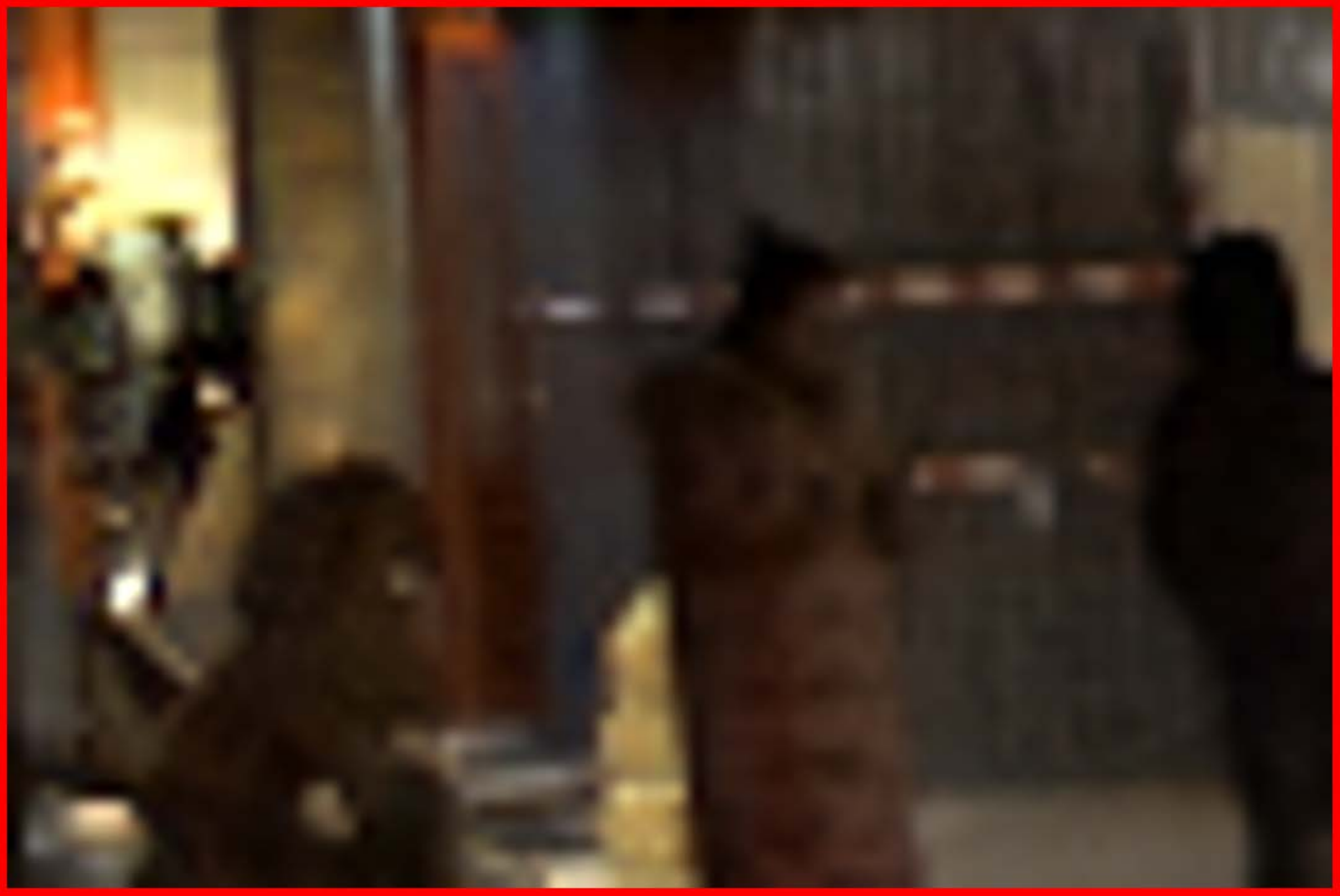}&
		\includegraphics[width=0.093\linewidth]{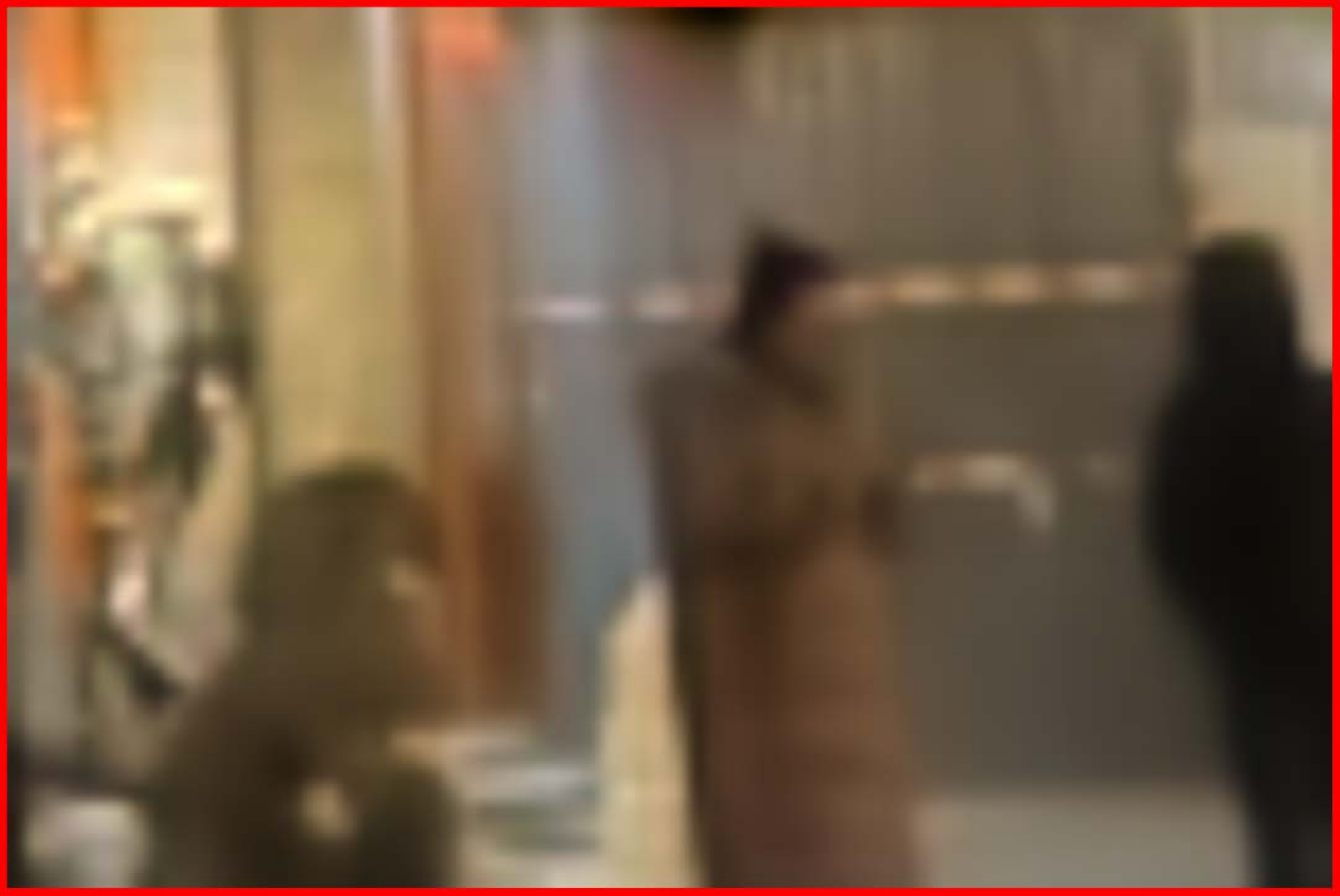}&
		\includegraphics[width=0.093\linewidth]{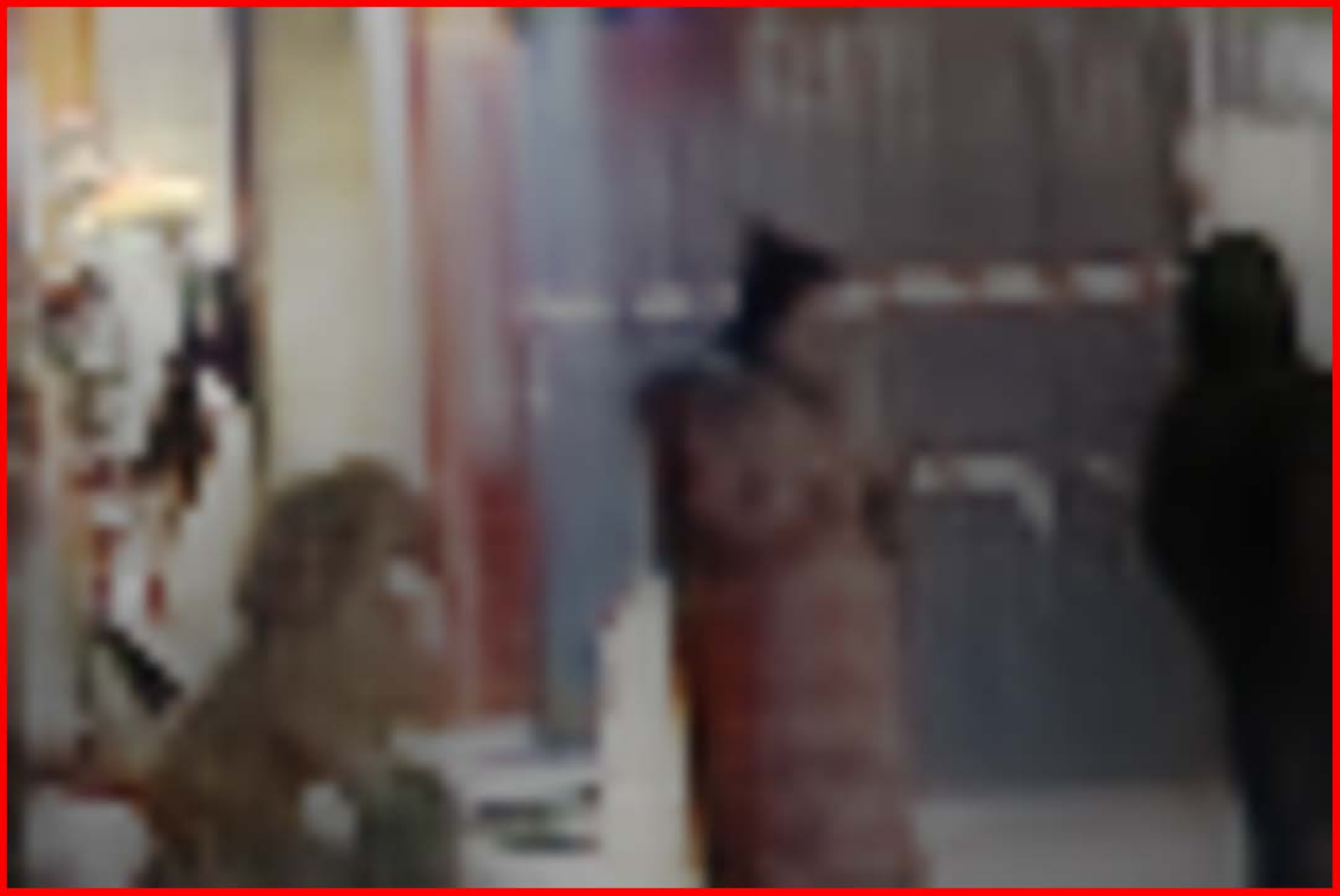}&
		\includegraphics[width=0.093\linewidth]{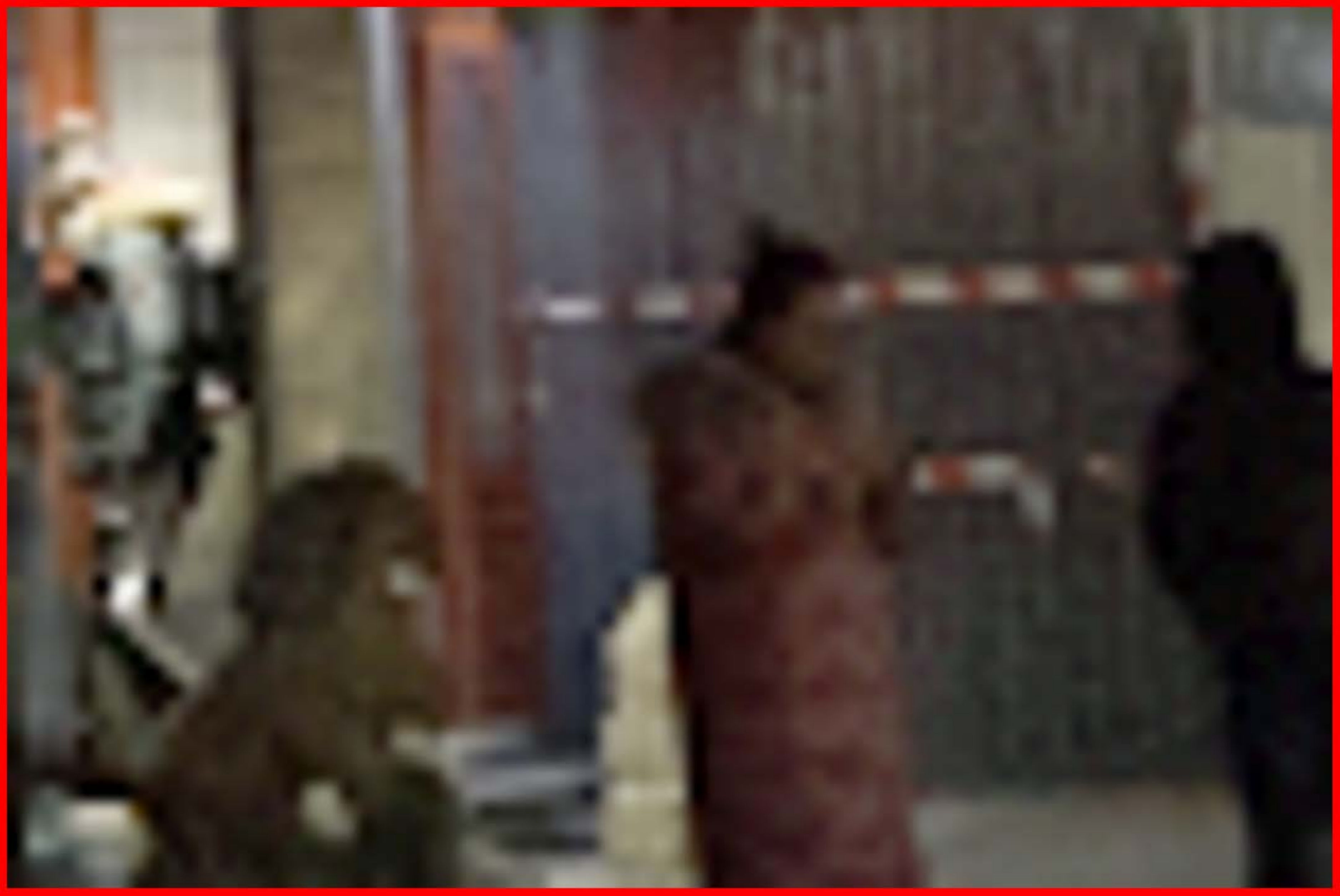}&
		\includegraphics[width=0.093\linewidth]{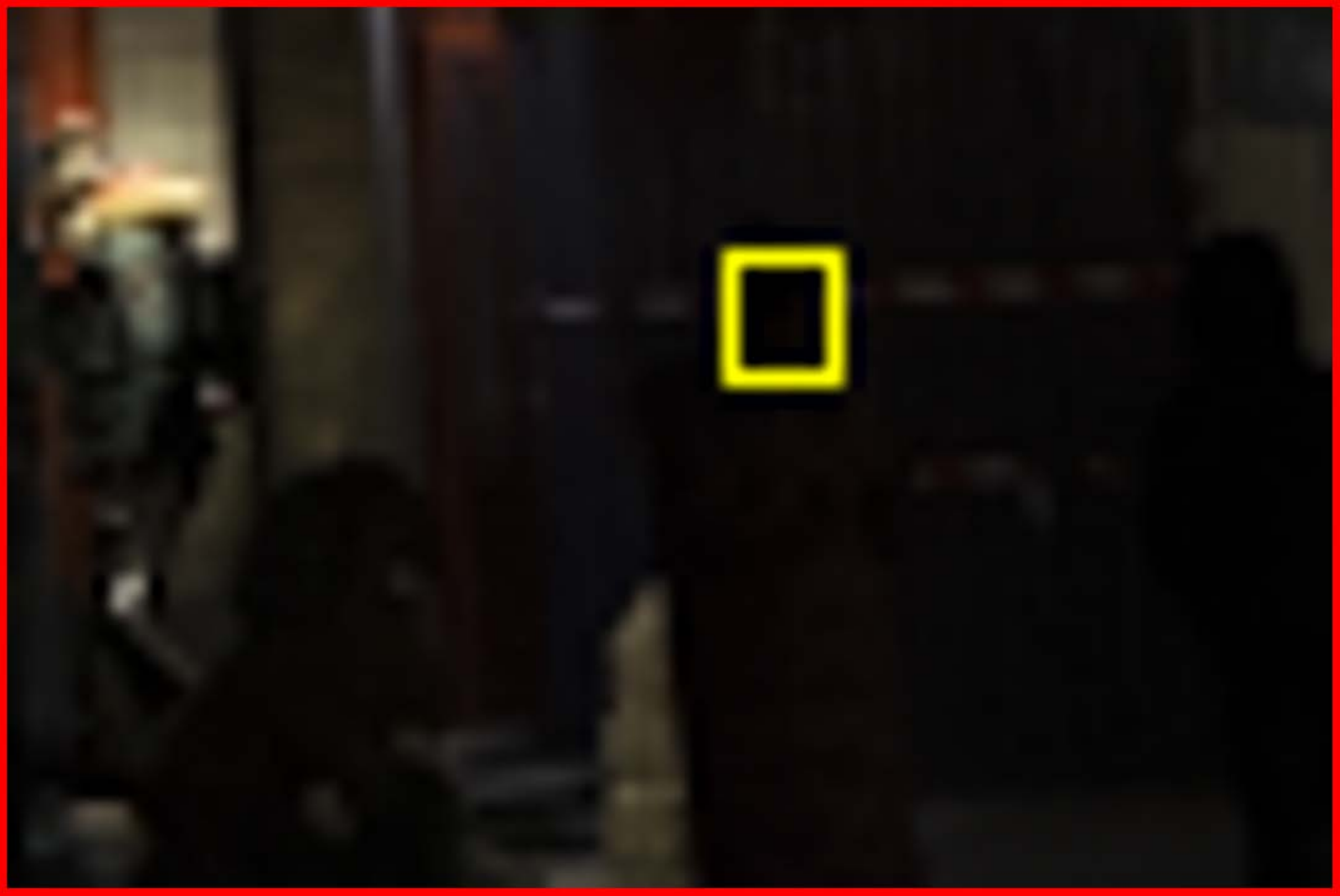}&
		\includegraphics[width=0.093\linewidth]{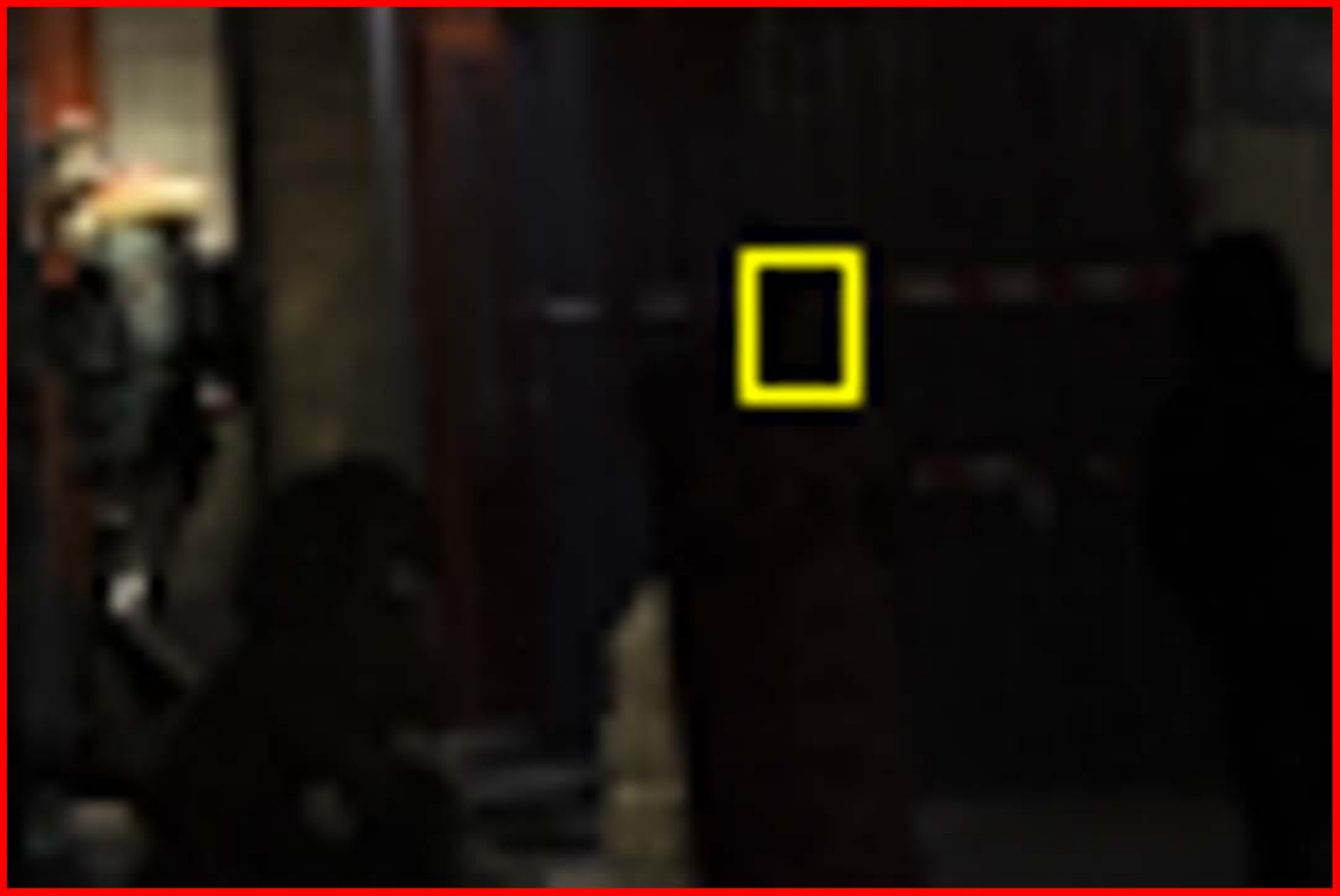}\\
		\includegraphics[width=0.093\linewidth]{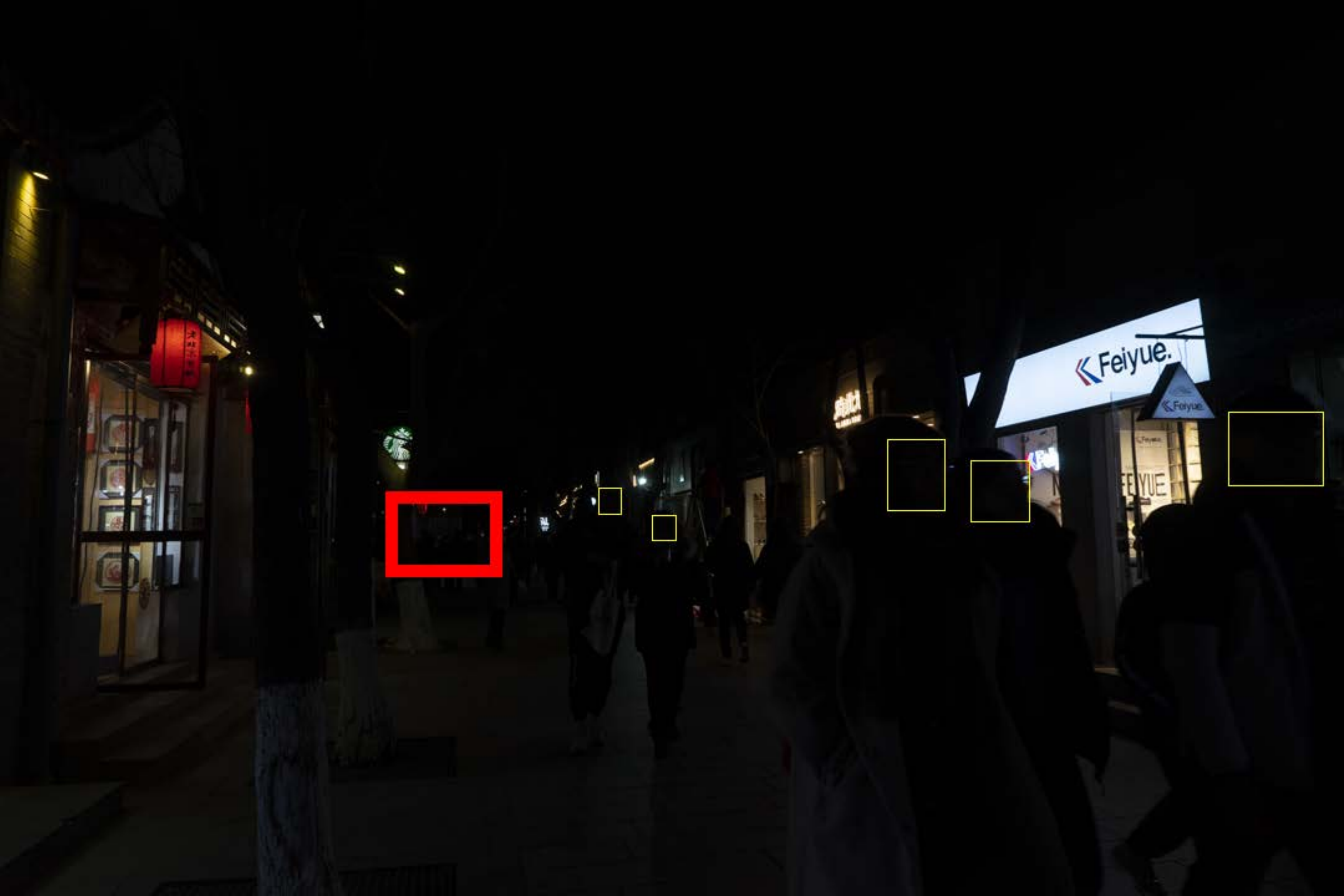}&
		\includegraphics[width=0.093\linewidth]{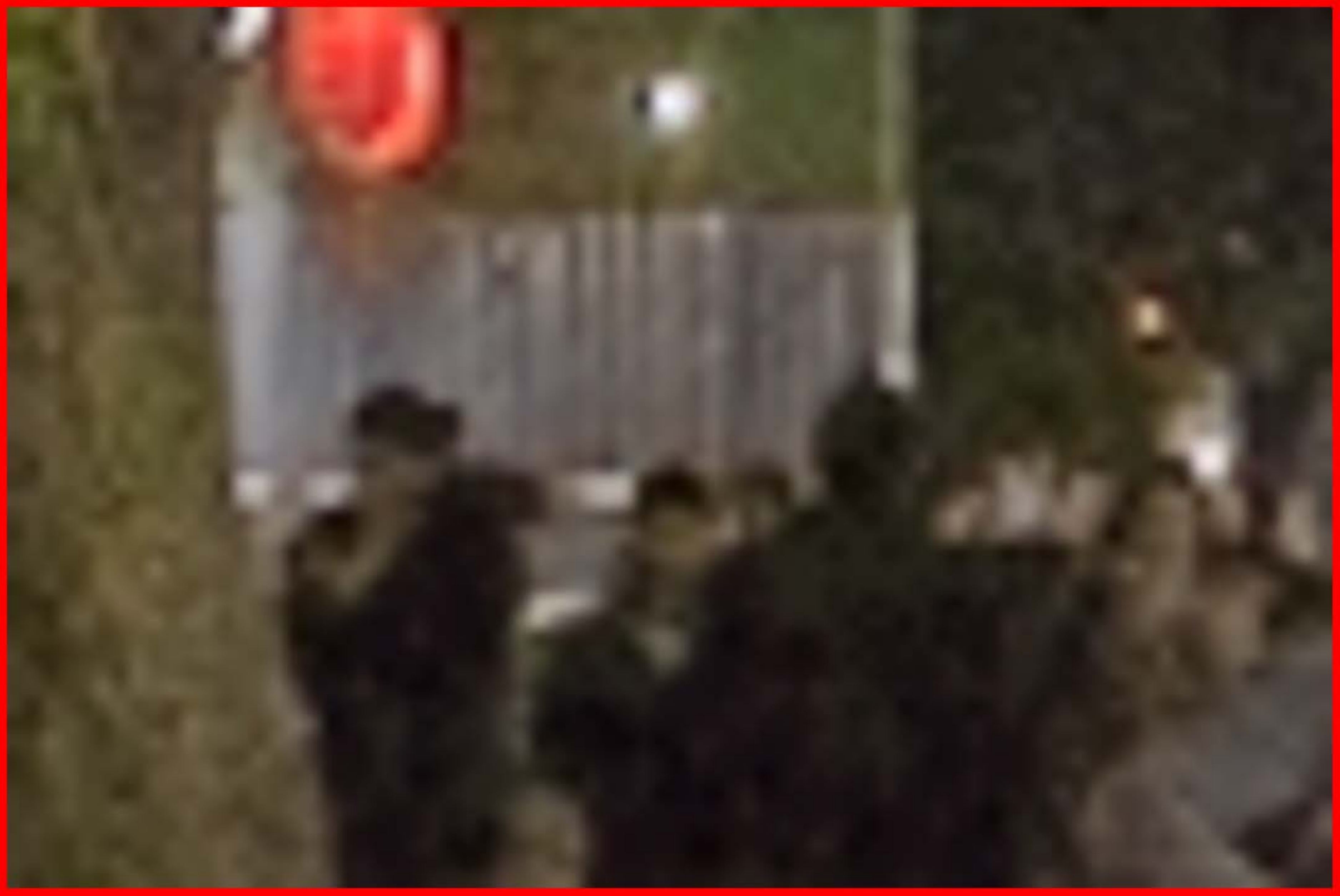}&
		\includegraphics[width=0.093\linewidth]{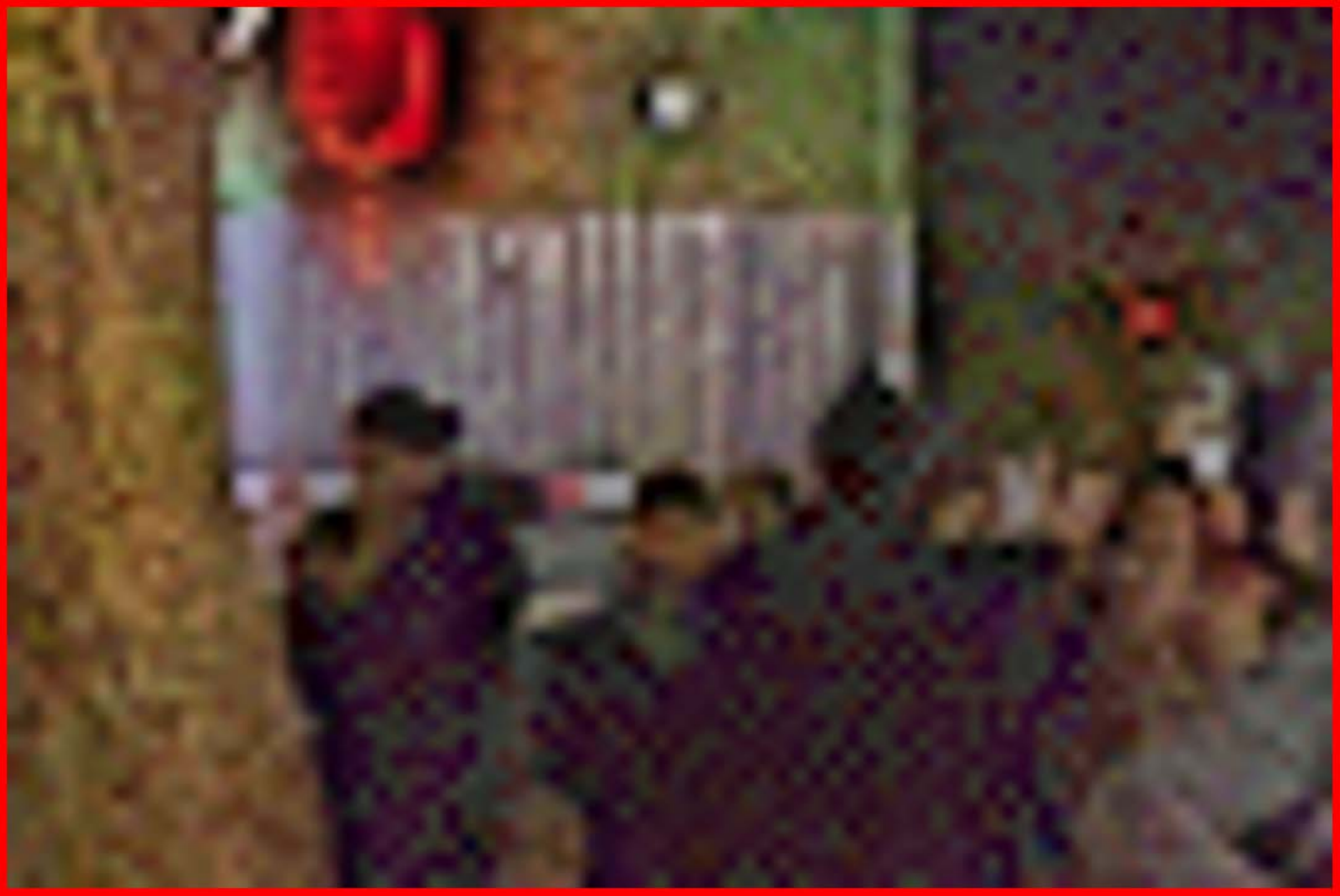}&
		\includegraphics[width=0.093\linewidth]{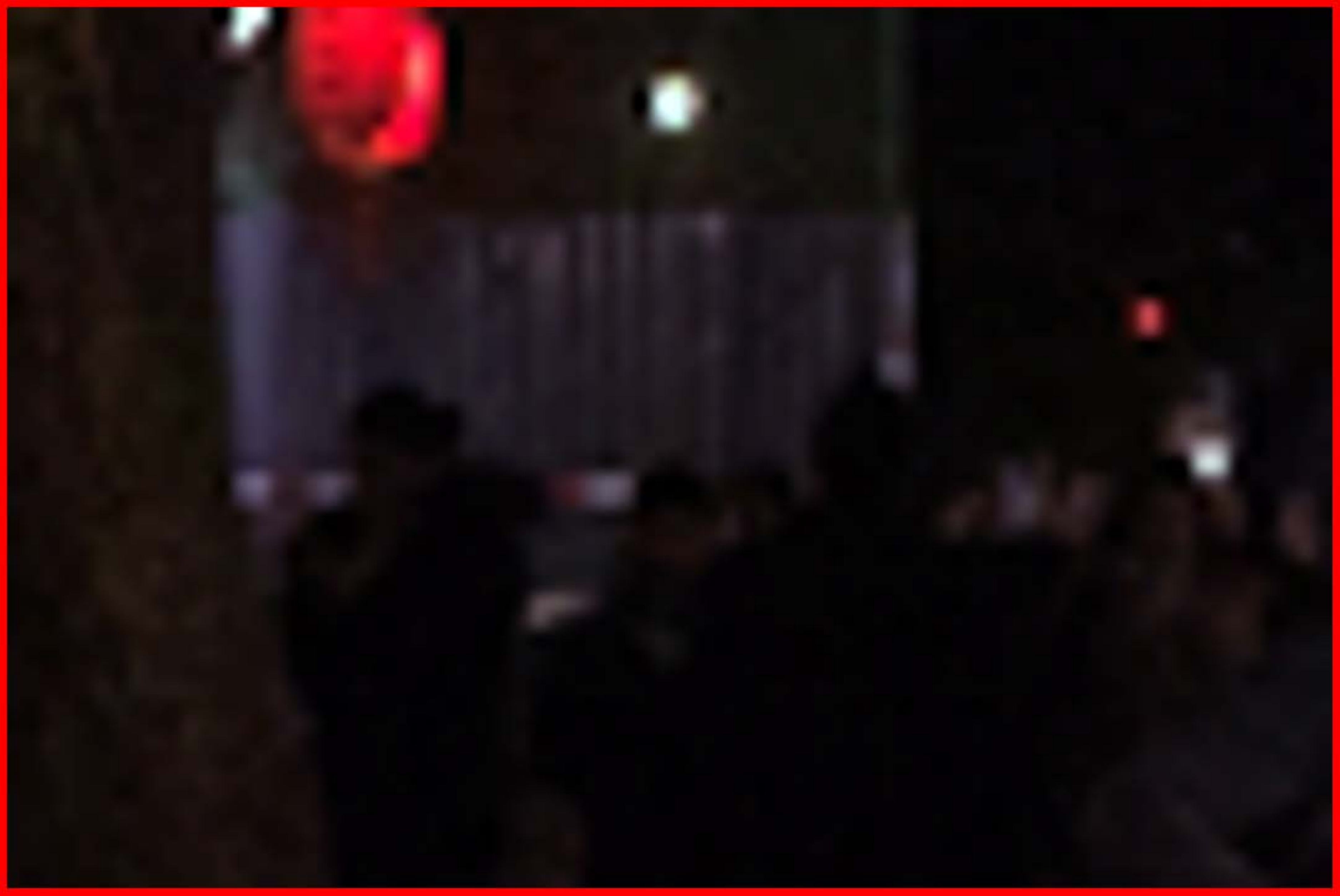}&
		\includegraphics[width=0.093\linewidth]{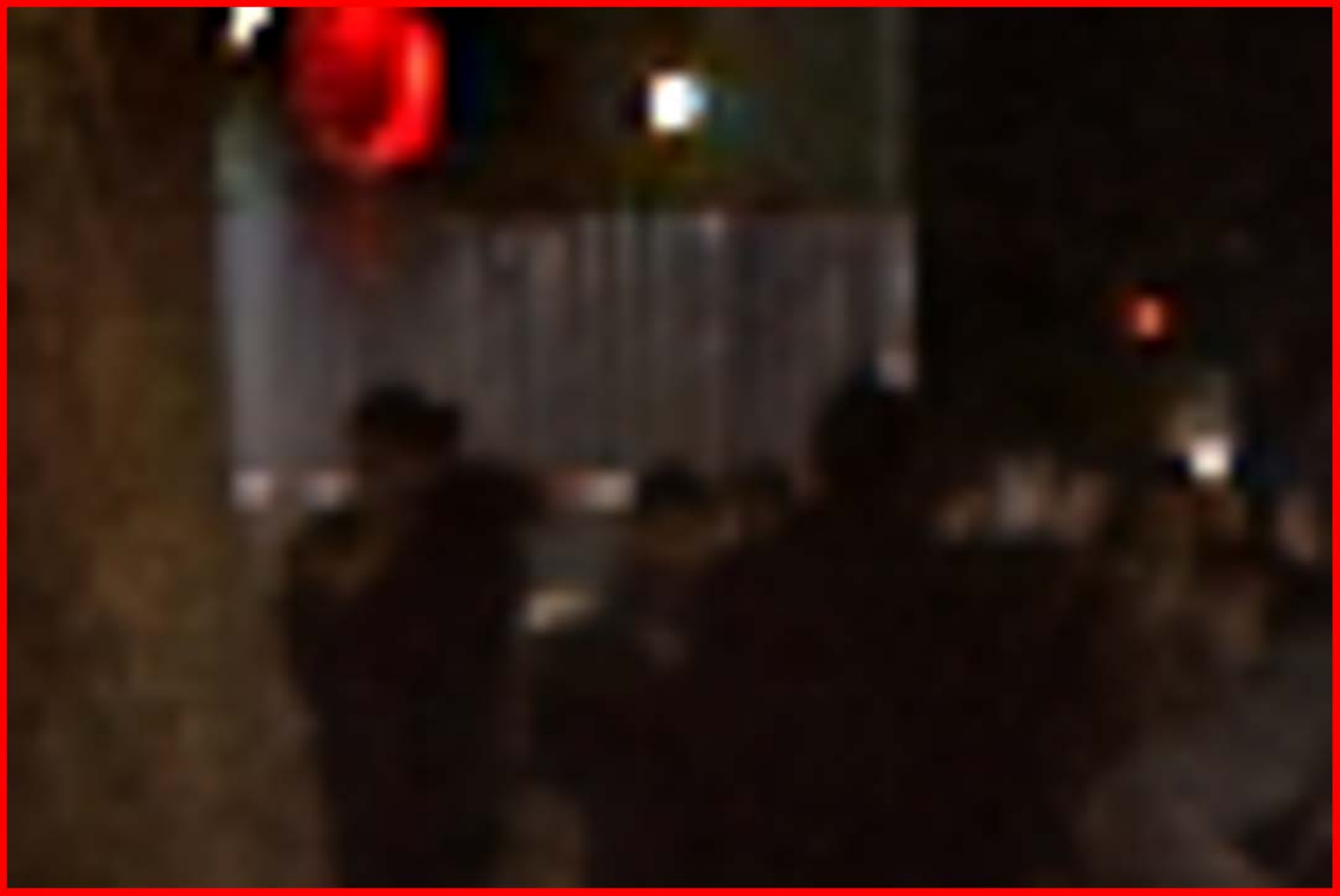}&
		\includegraphics[width=0.093\linewidth]{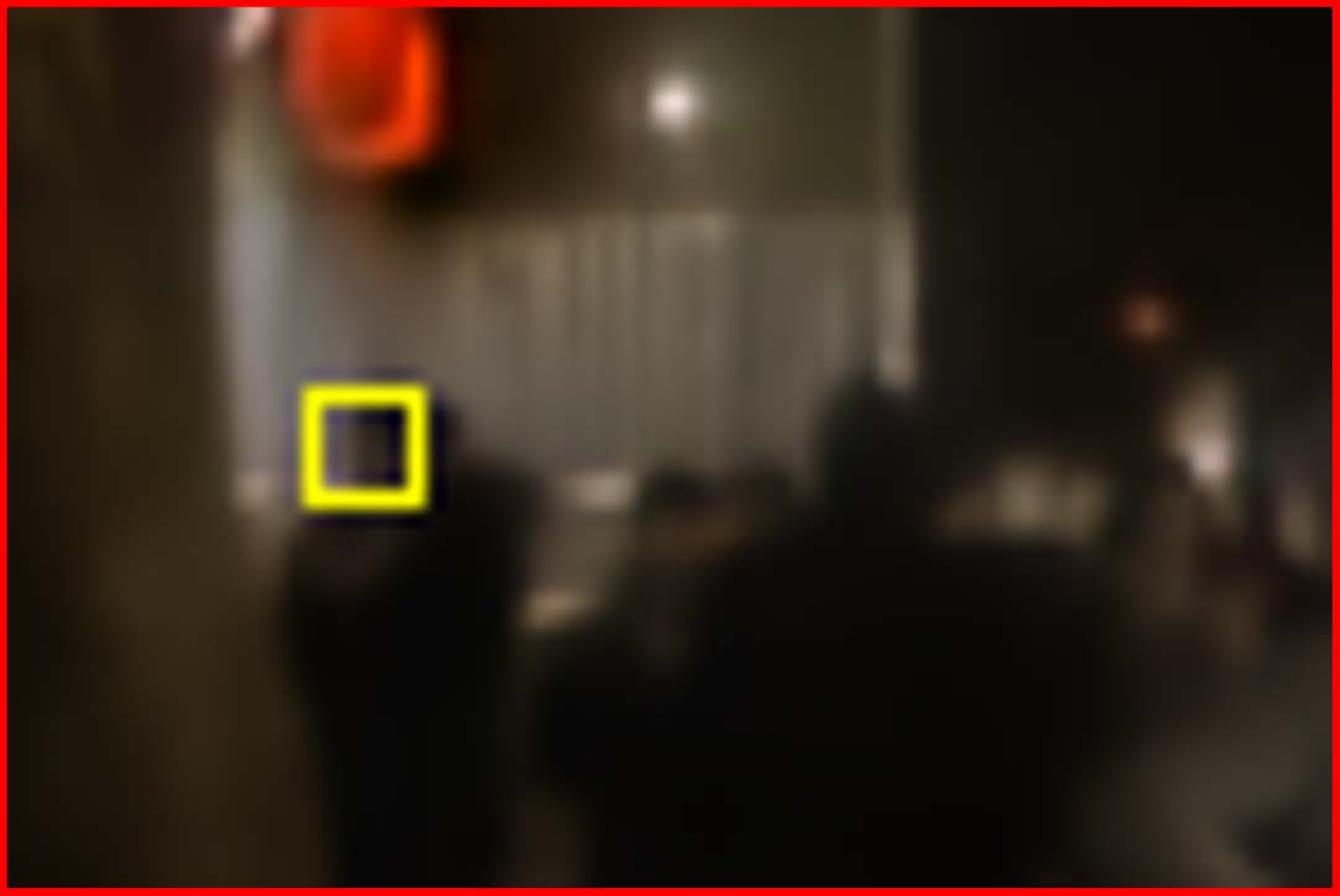}&
		\includegraphics[width=0.093\linewidth]{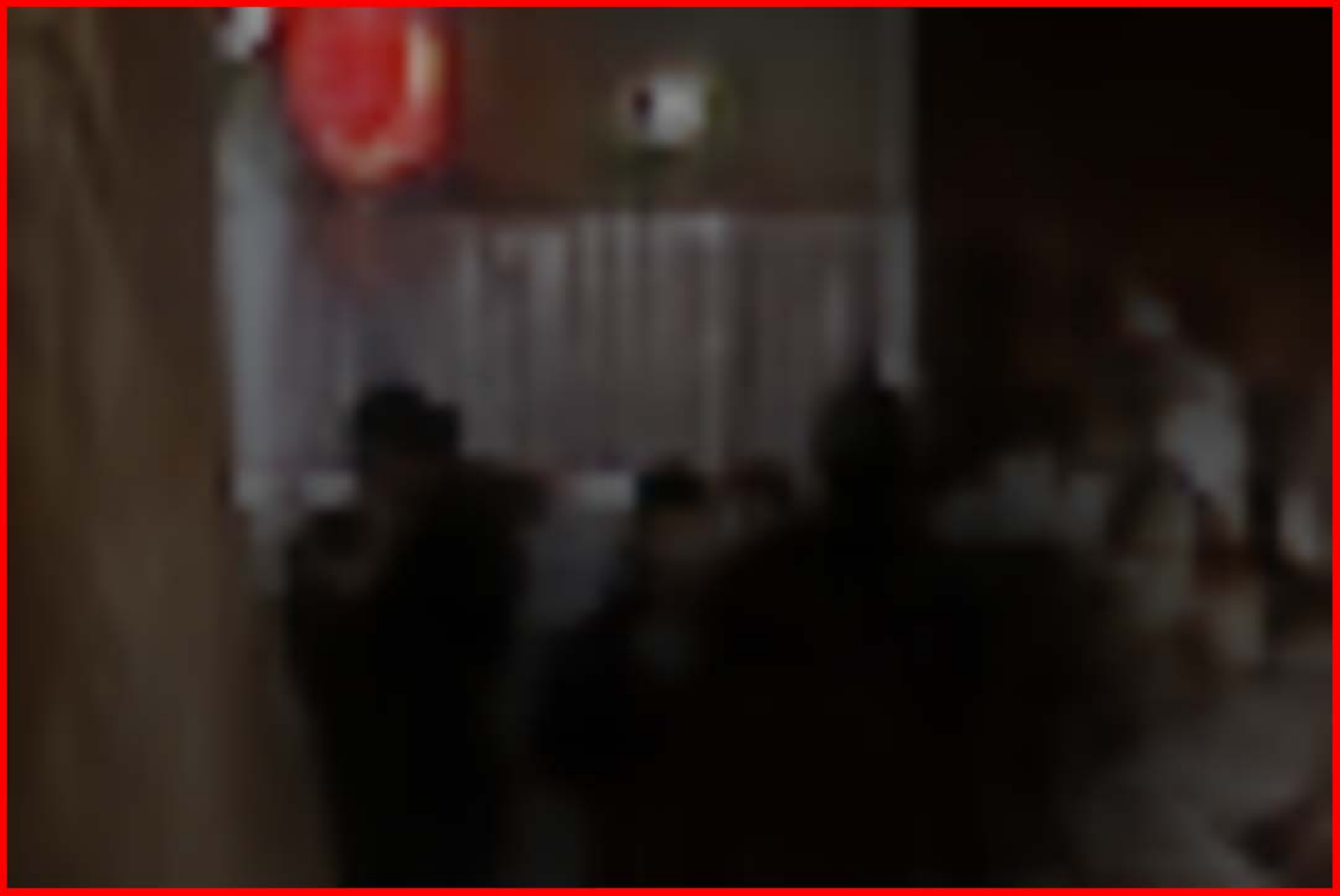}&
		\includegraphics[width=0.093\linewidth]{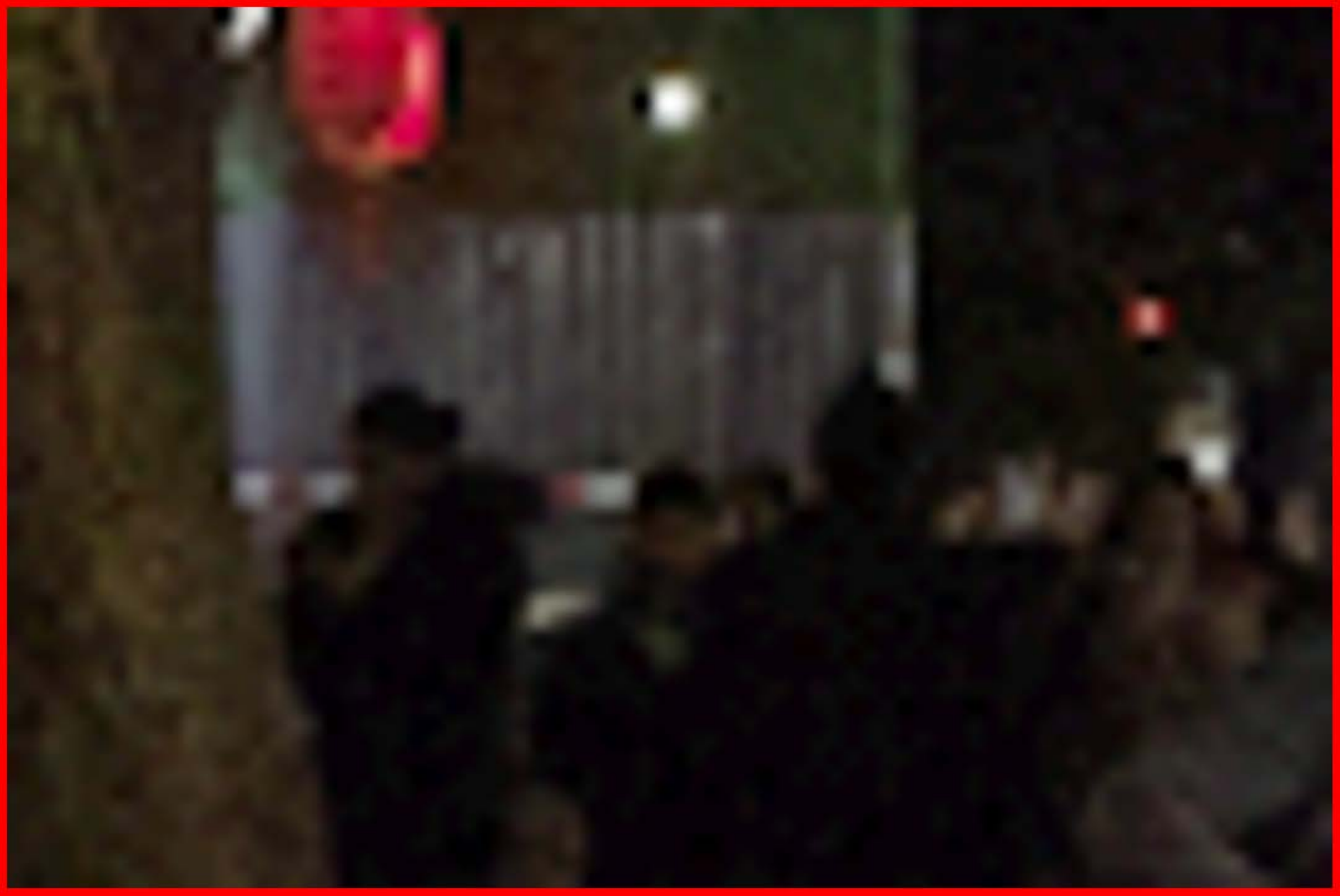}&
		\includegraphics[width=0.093\linewidth]{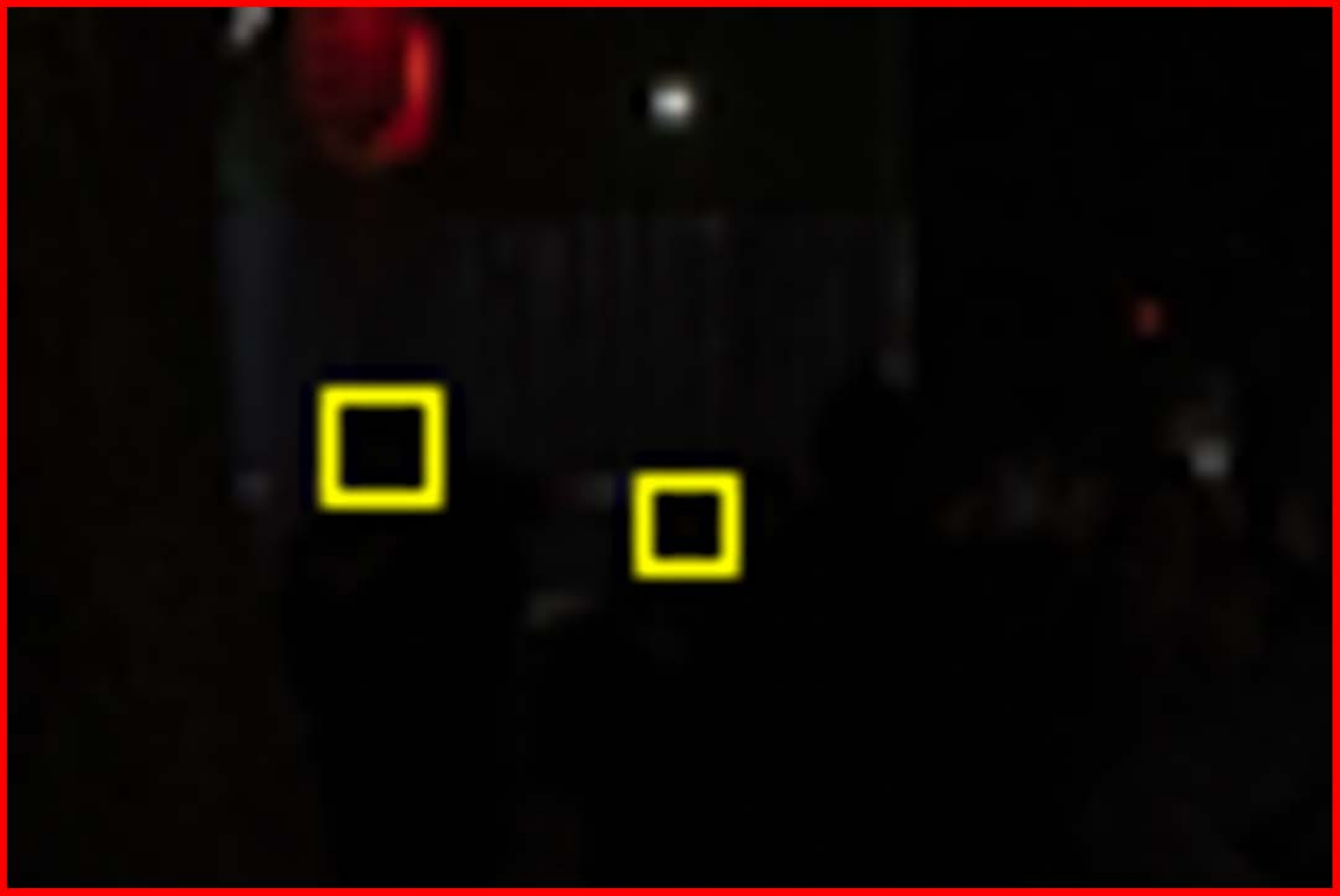}&
		\includegraphics[width=0.093\linewidth]{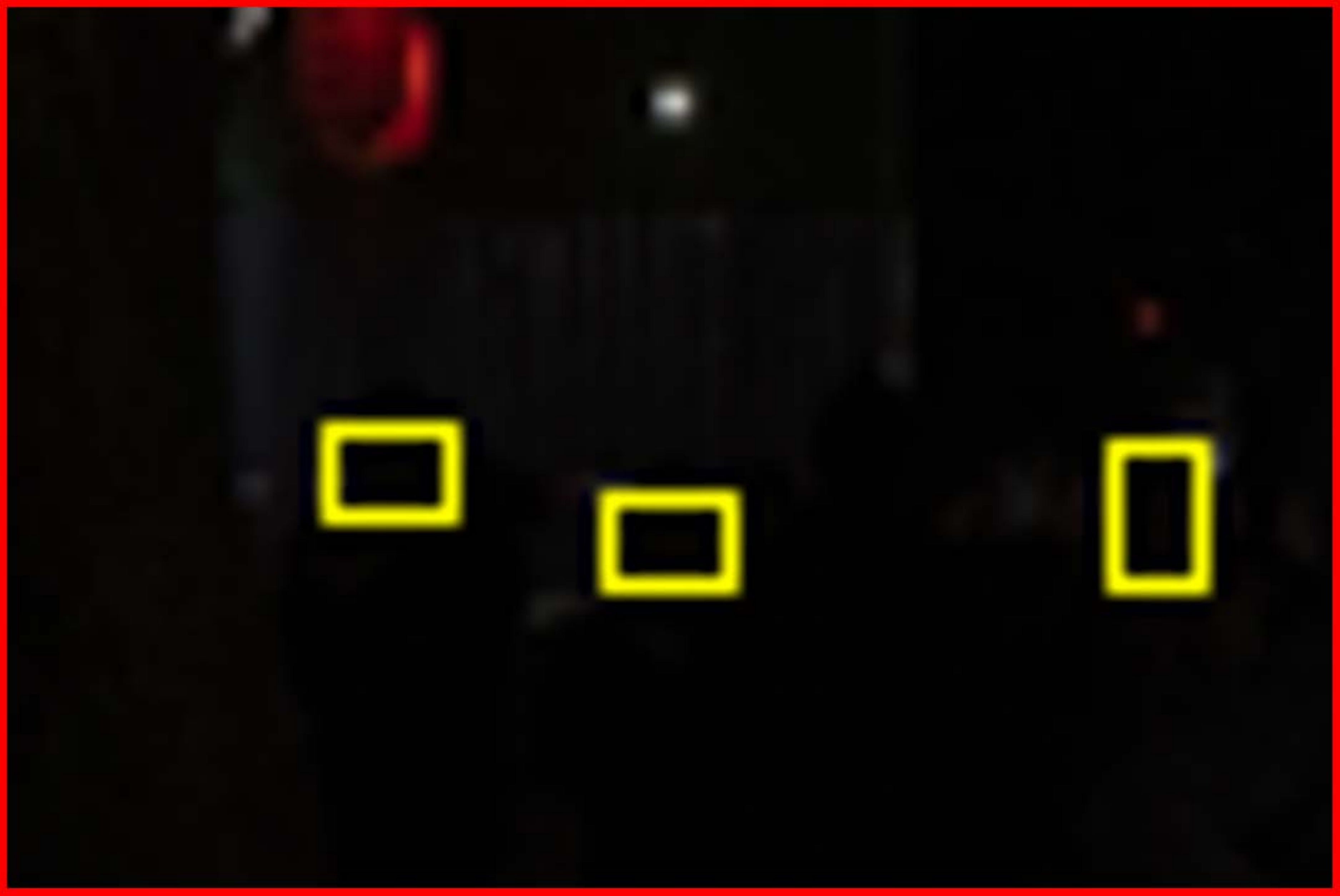}\\
		\footnotesize Input&\footnotesize GLADNet$_\mathtt{P}$&\footnotesize RetinexNet$_\mathtt{D}$&\footnotesize DeepUPE$_\mathtt{S}$&\footnotesize EnGAN$_\mathtt{P}$&\footnotesize DRBN$_\mathtt{D}$&\footnotesize KinD$_\mathtt{D}$&\footnotesize ZeroDCE$_\mathtt{P}$&\footnotesize Ours&\footnotesize Label\\
	\end{tabular}
	\caption{Visual comparison of face detection on DARK FACE dataset. Red boxes indicate the obvious differences.}
	\label{fig:DarkFaceDetection}
\end{figure*}

\begin{figure*}[!htb]
	\centering
	\begin{tabular}{c@{\extracolsep{0.3em}}c@{\extracolsep{0.3em}}c@{\extracolsep{0.3em}}c}
		\includegraphics[width=0.242\linewidth]{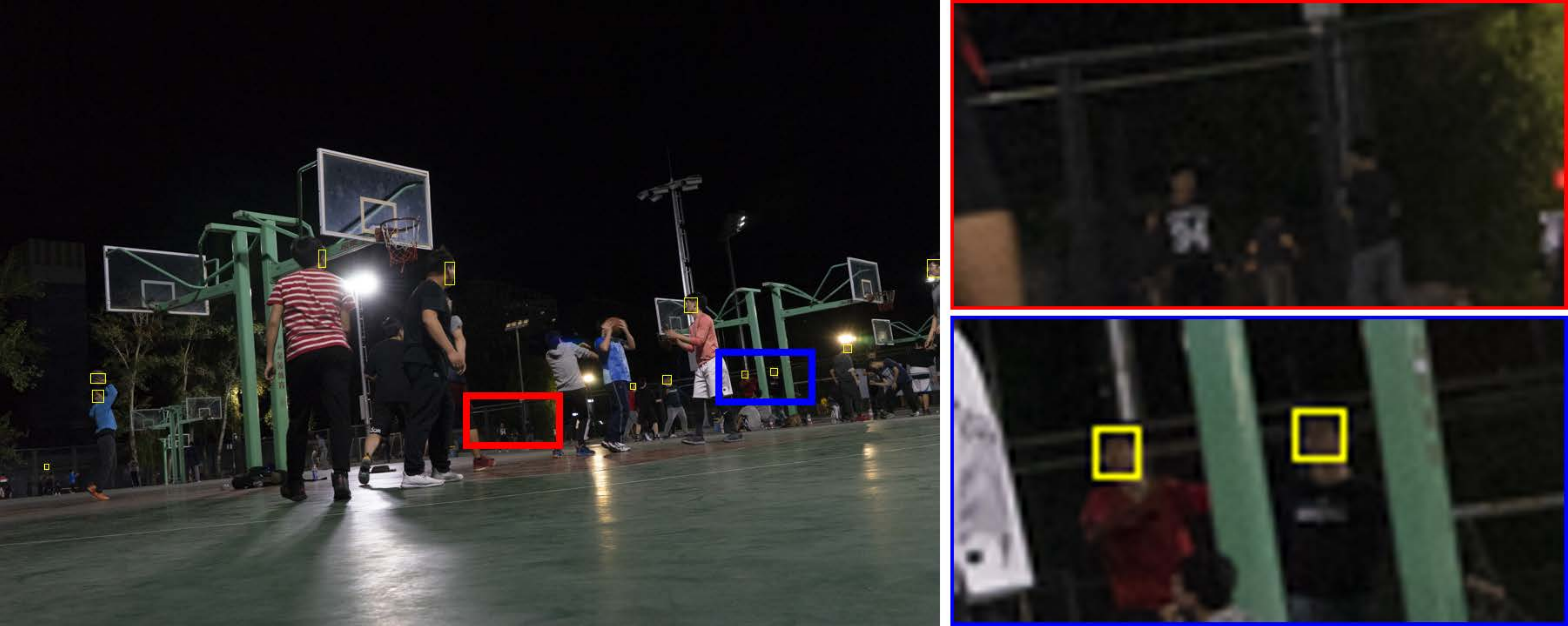}&
		\includegraphics[width=0.242\linewidth]{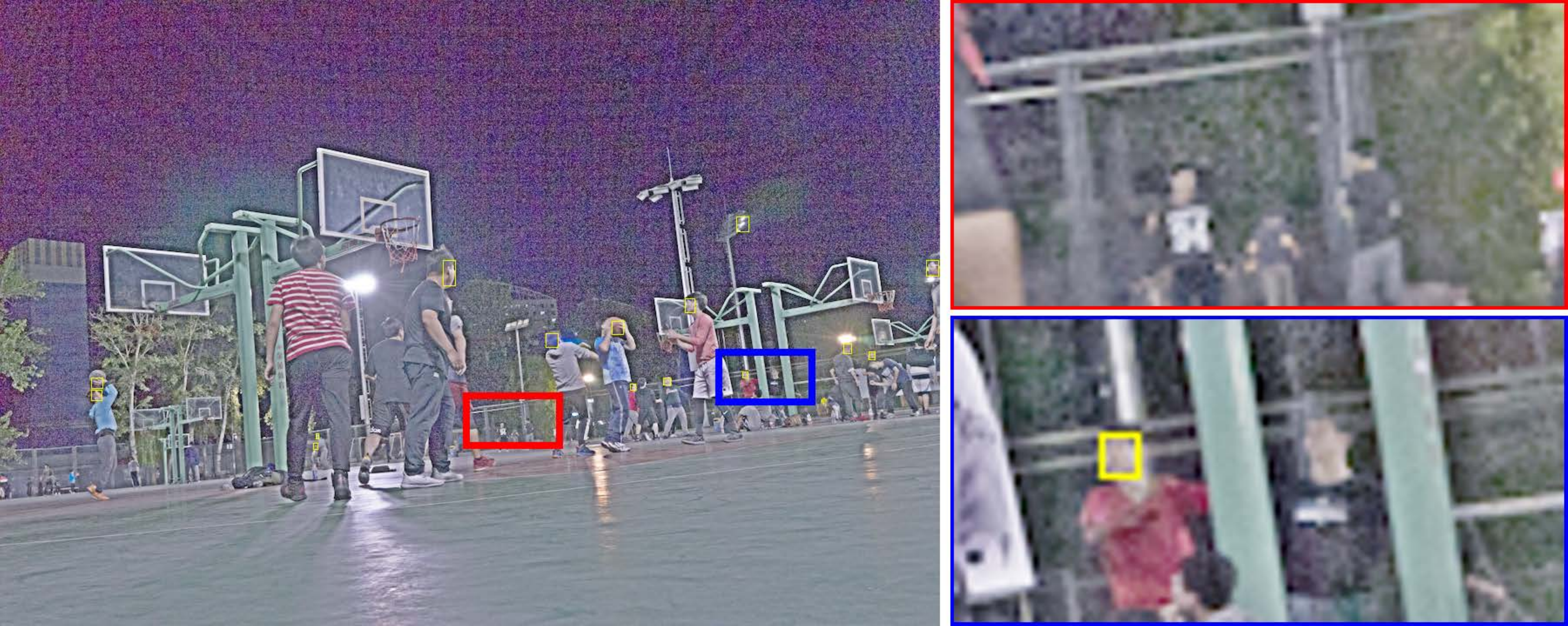}&
		\includegraphics[width=0.242\linewidth]{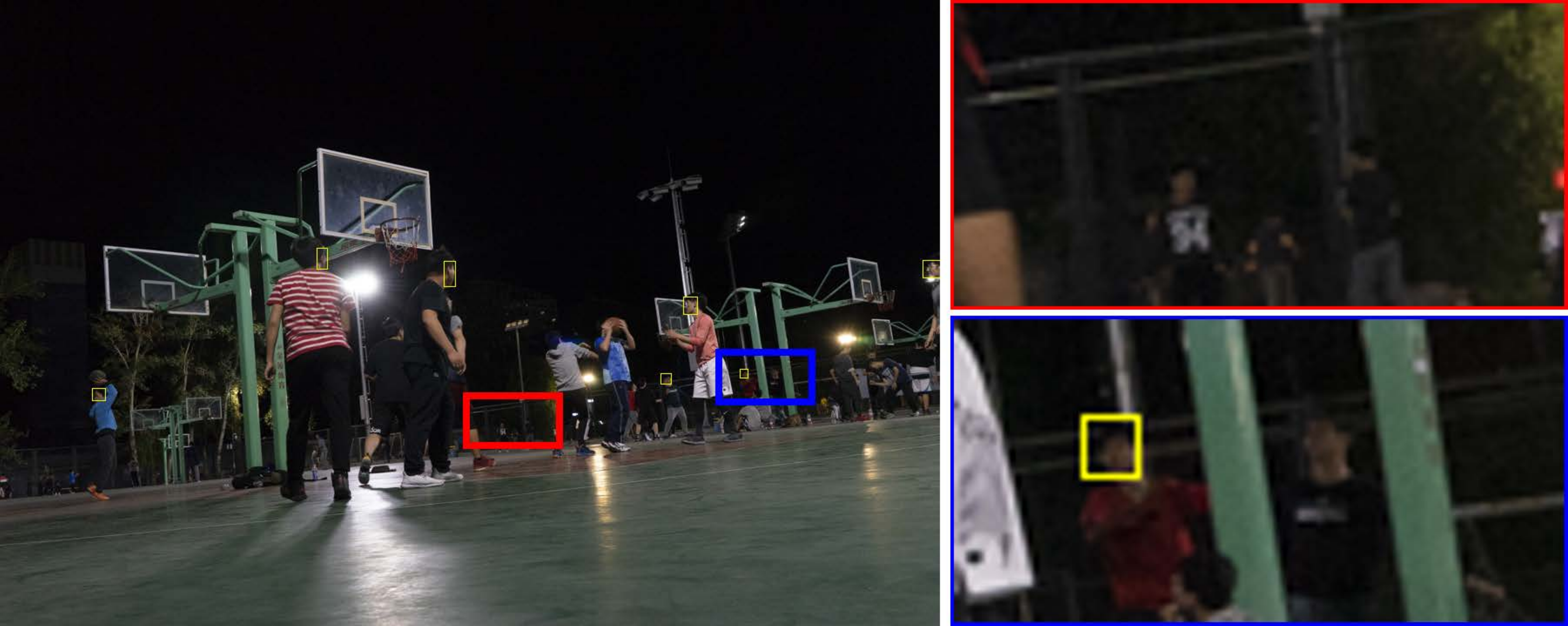}&
		\includegraphics[width=0.242\linewidth]{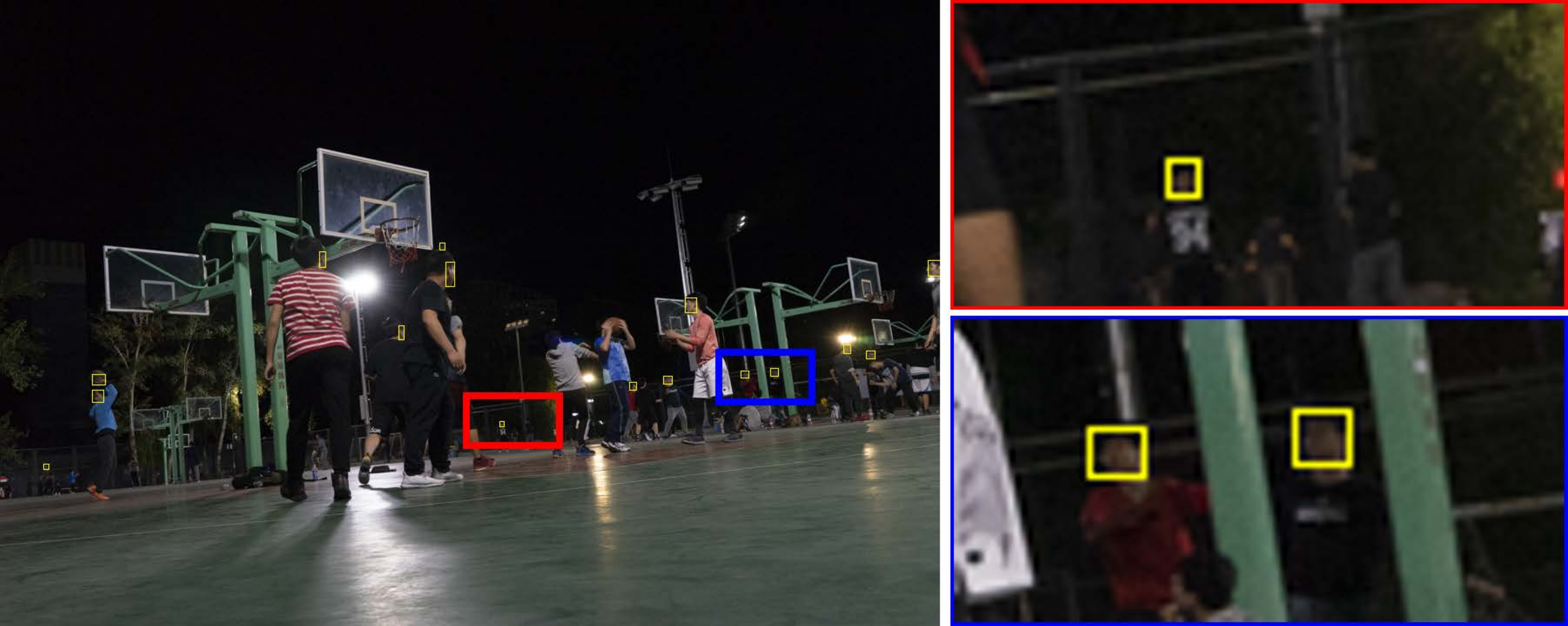}\\
		\footnotesize Input& \footnotesize HLA&\footnotesize REG& \footnotesize Ours\\
	\end{tabular}
	\caption{Visual comparison of face detection on DARK FACE dataset among two recently-proposed detection algorithms (HLA~\cite{wang2021hla}, REG~\cite{liang2021recurrent}). Red and blue boxes indicate the obvious differences.}
	\label{fig:DarkFaceDetection2}
\end{figure*}

\begin{figure*}[t]
	\centering
	\begin{tabular}{c@{\extracolsep{0.3em}}c@{\extracolsep{0.3em}}c@{\extracolsep{0.3em}}c@{\extracolsep{0.3em}}c@{\extracolsep{0.3em}}c@{\extracolsep{0.3em}}c@{\extracolsep{0.3em}}c@{\extracolsep{0.3em}}c}
		\includegraphics[width=0.104\linewidth]{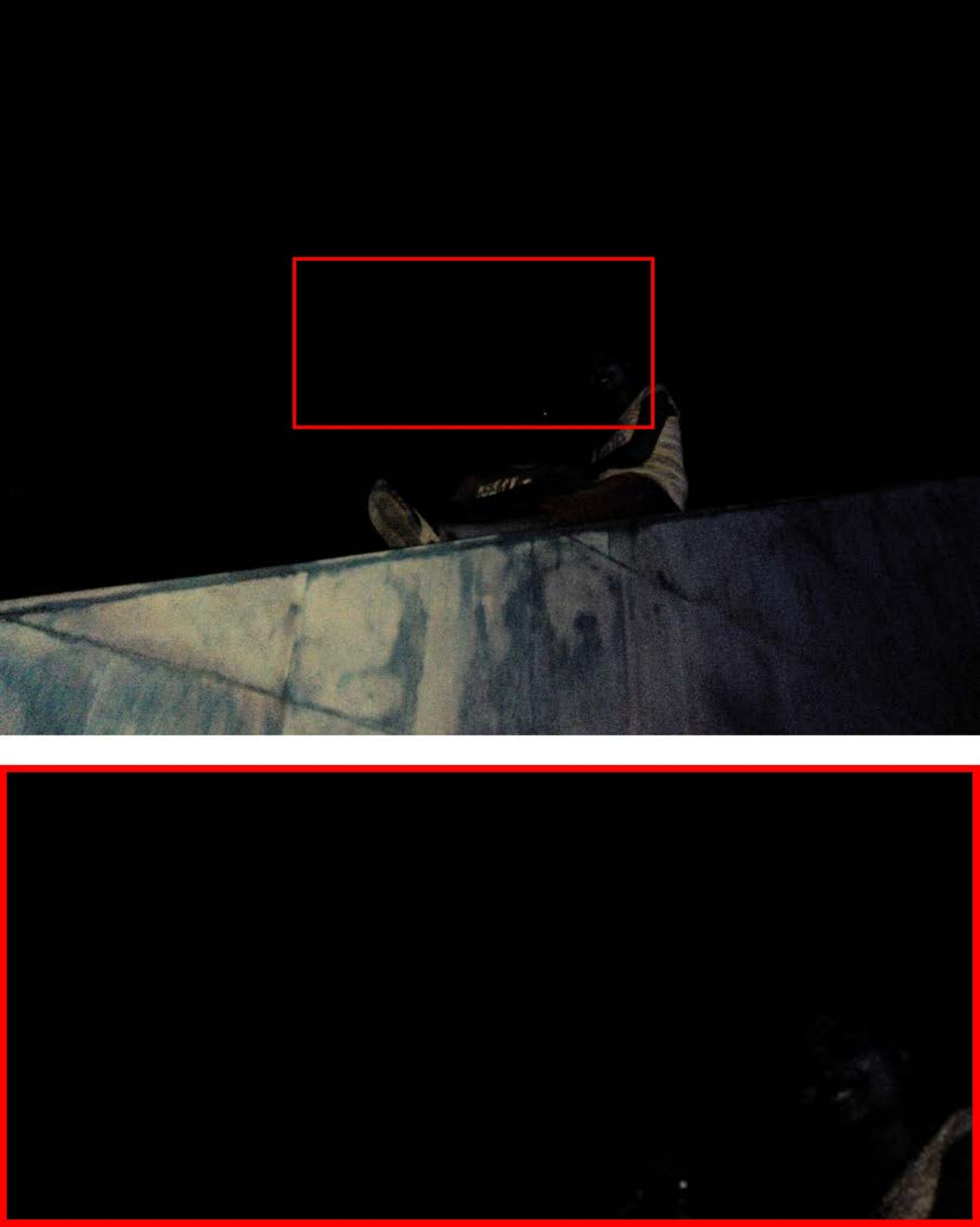}&
		\includegraphics[width=0.104\linewidth]{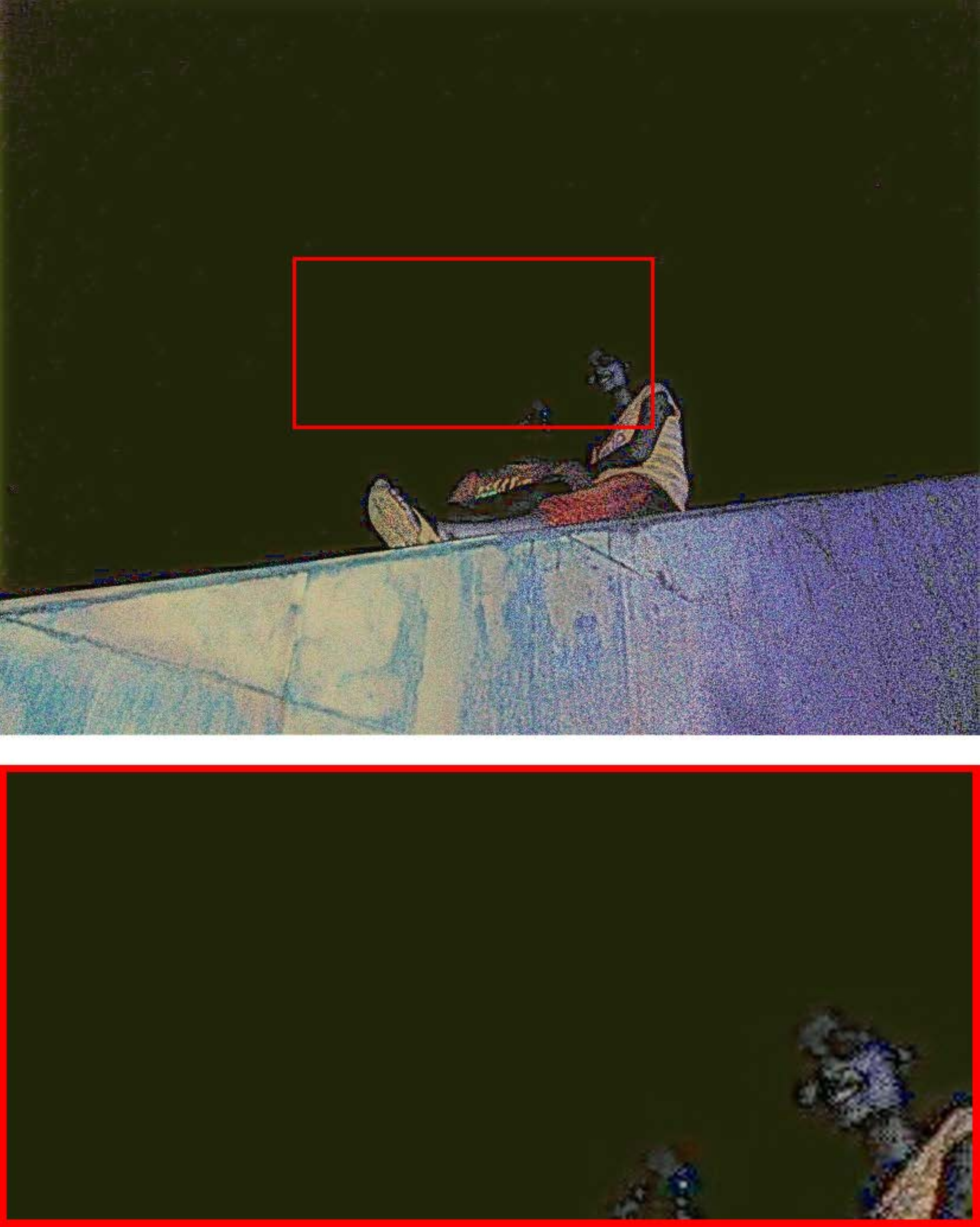}&
		\includegraphics[width=0.104\linewidth]{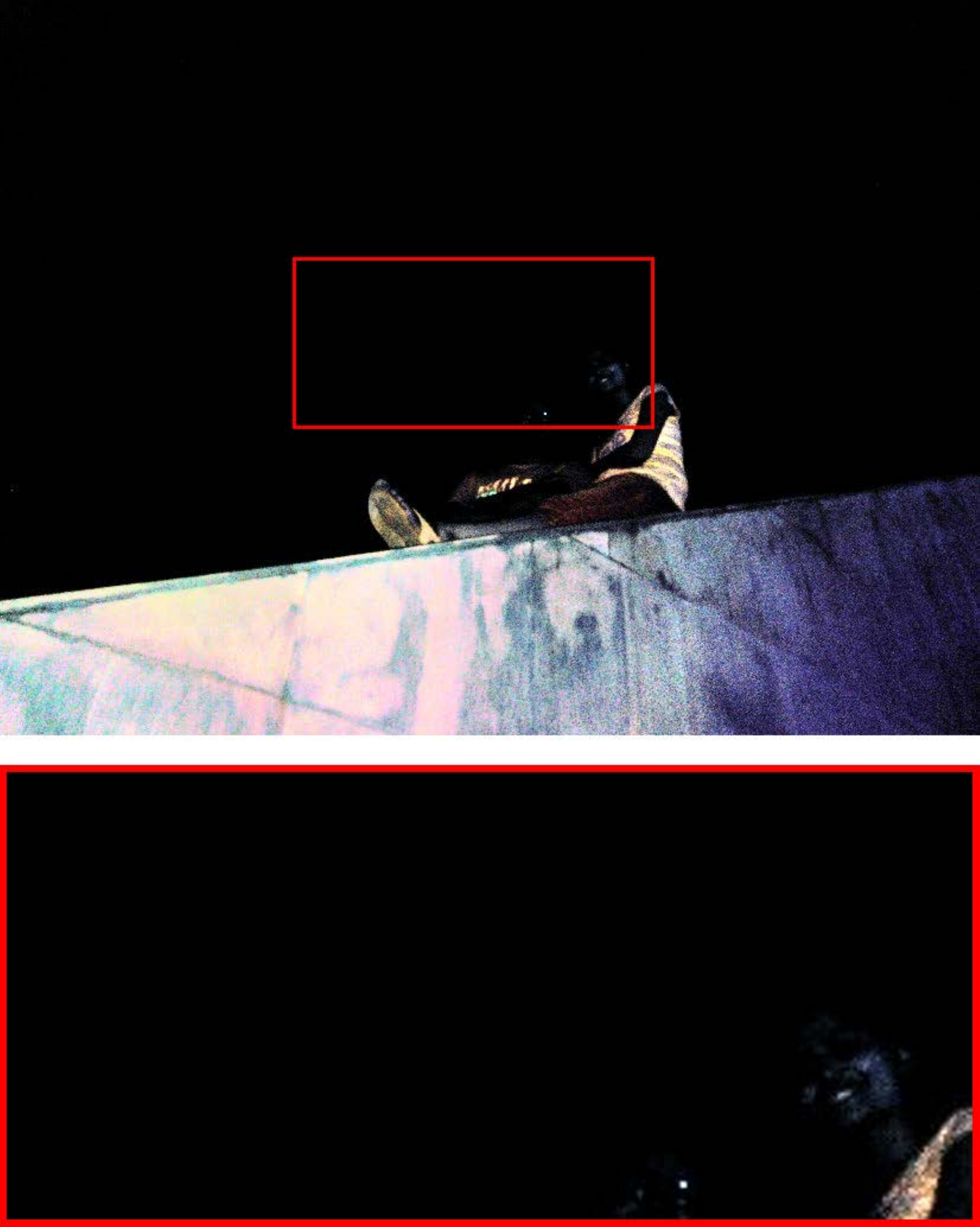}&
		\includegraphics[width=0.104\linewidth]{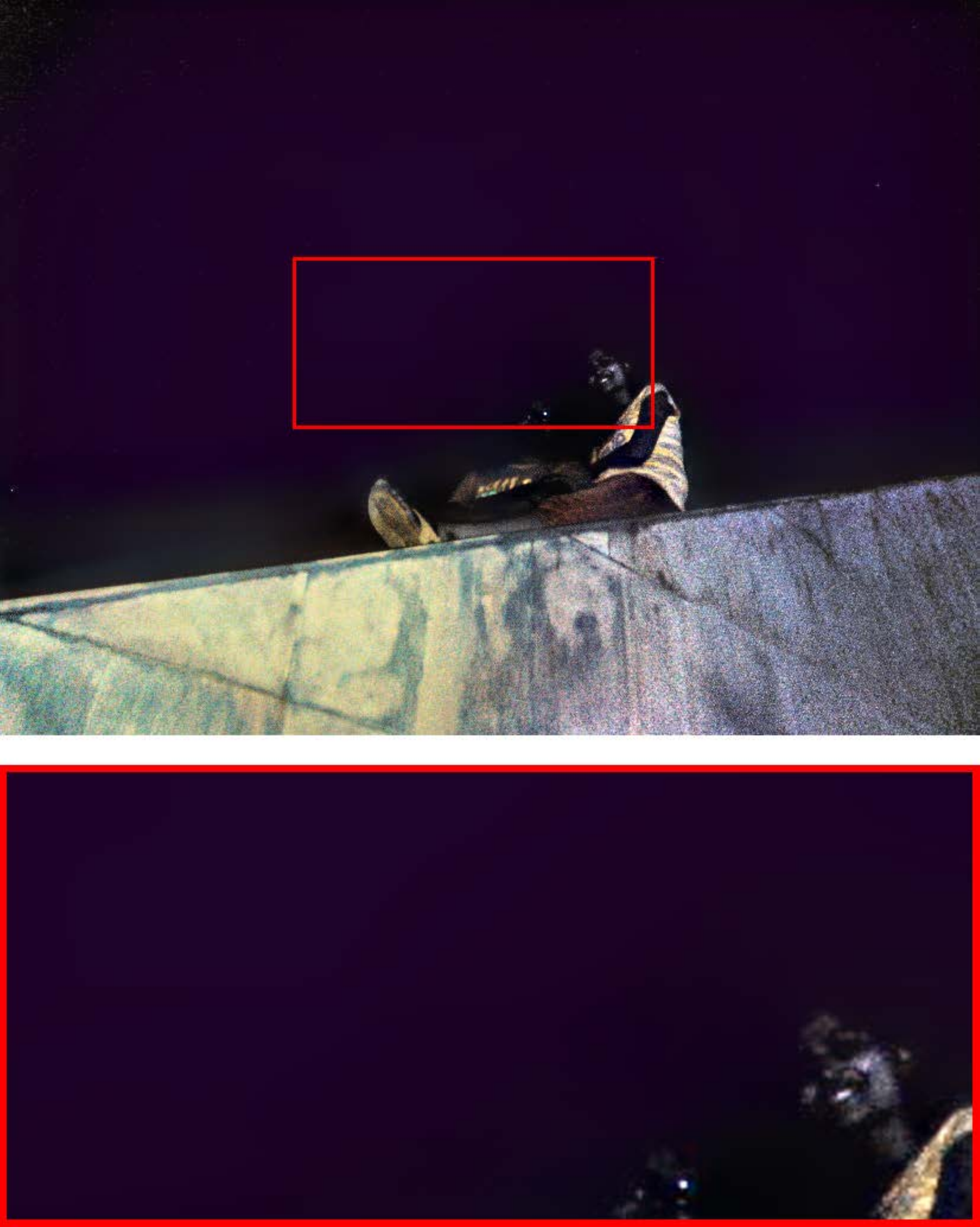}&
		\includegraphics[width=0.104\linewidth]{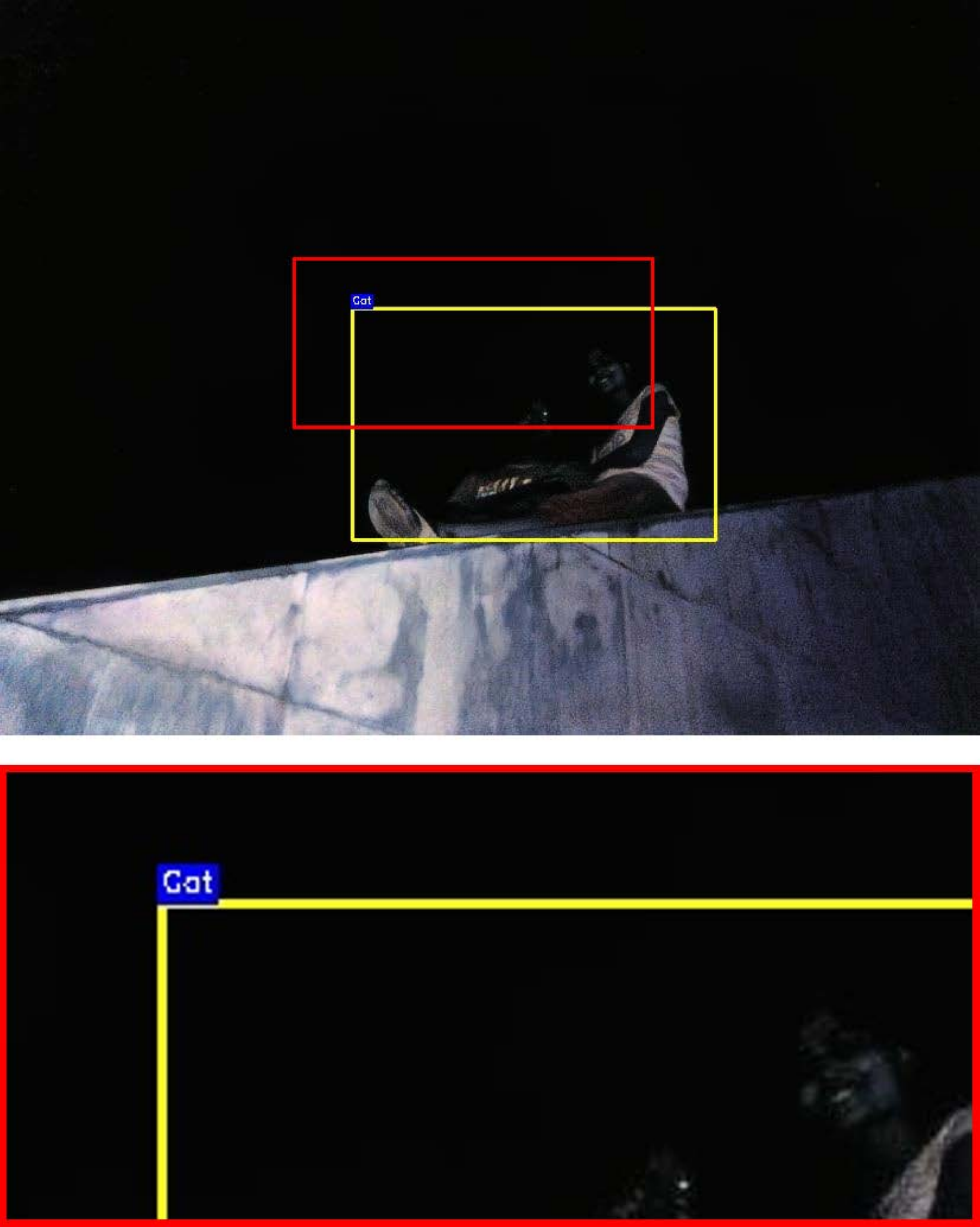}&
		\includegraphics[width=0.104\linewidth]{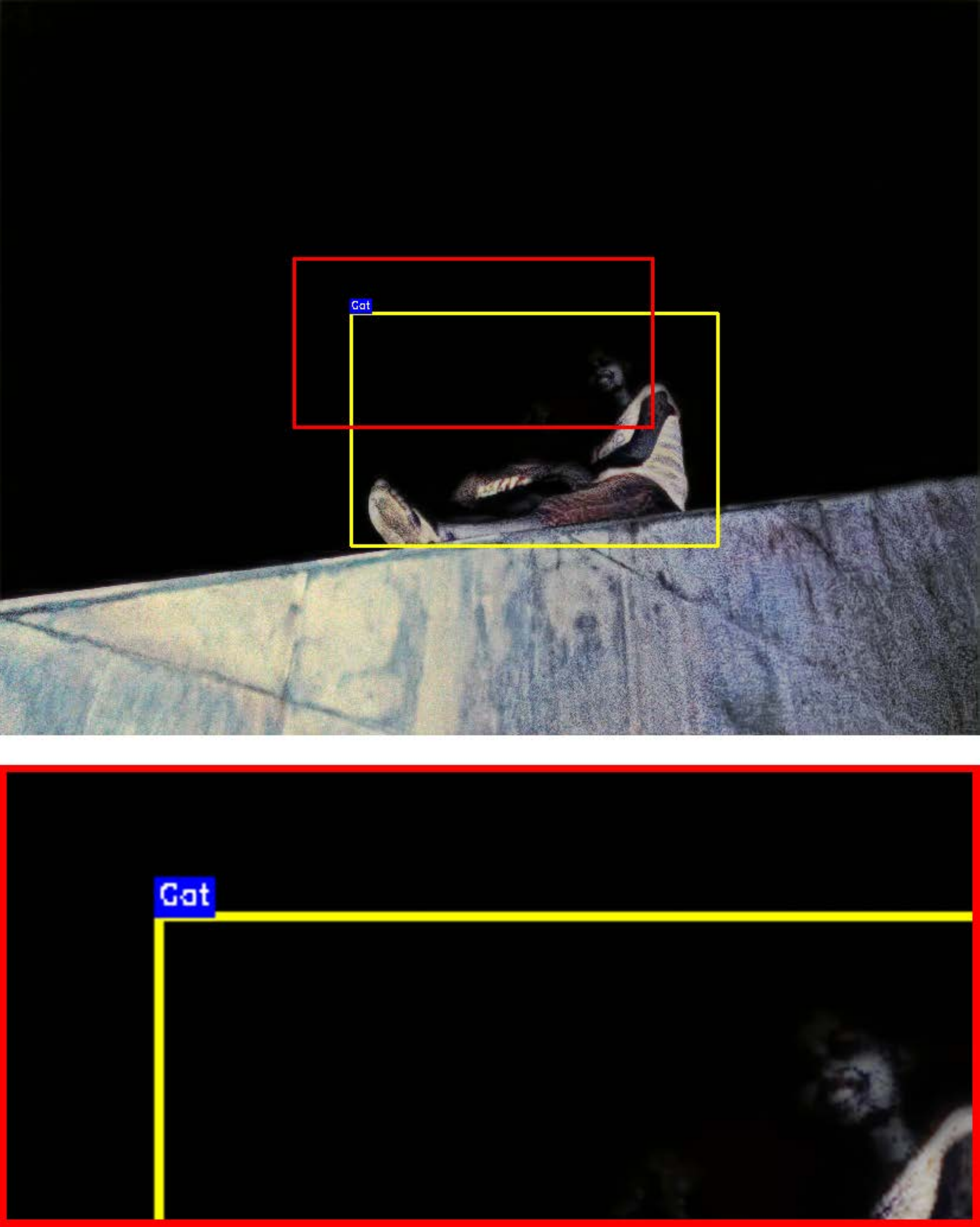}&
		\includegraphics[width=0.104\linewidth]{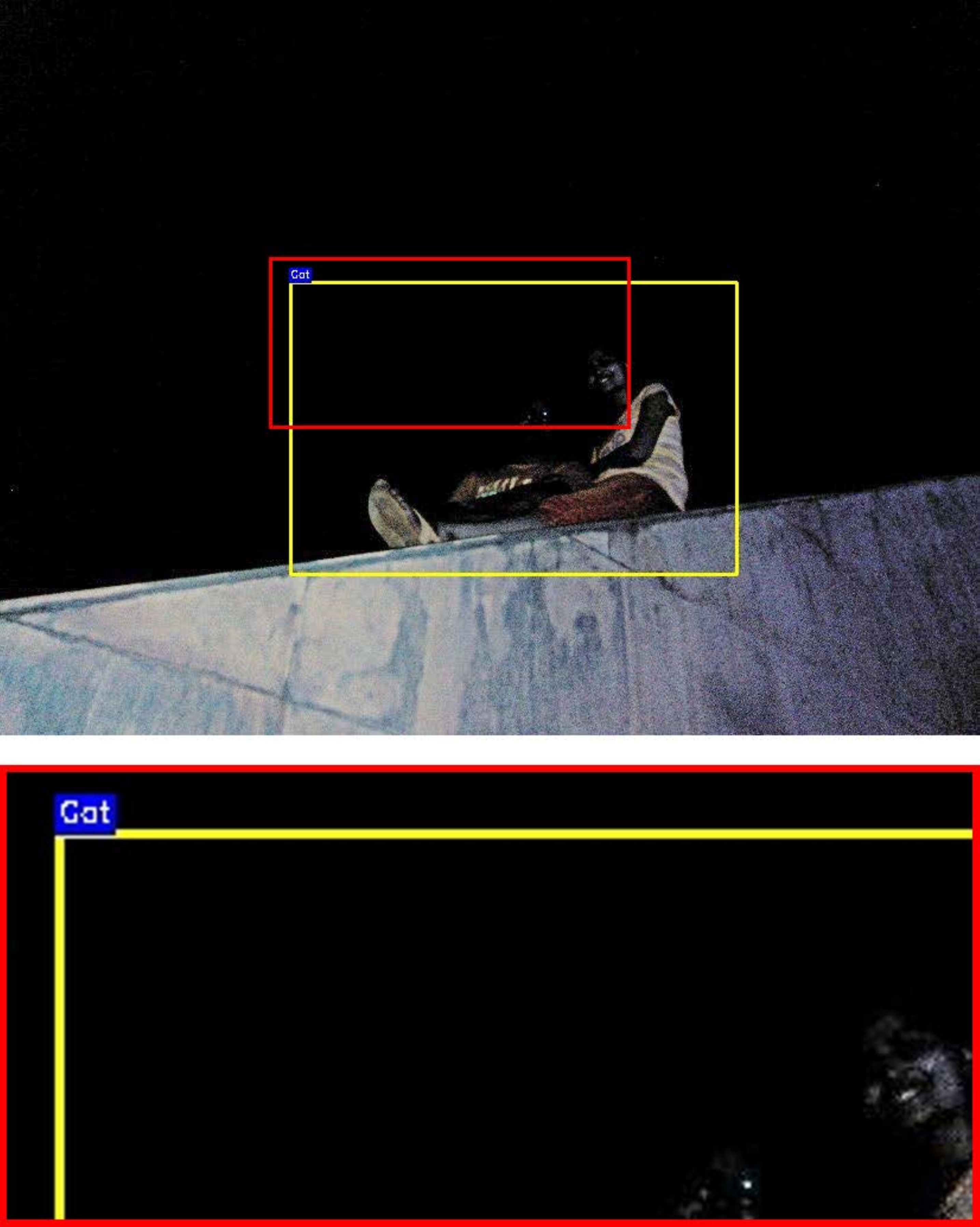}&
		\includegraphics[width=0.104\linewidth]{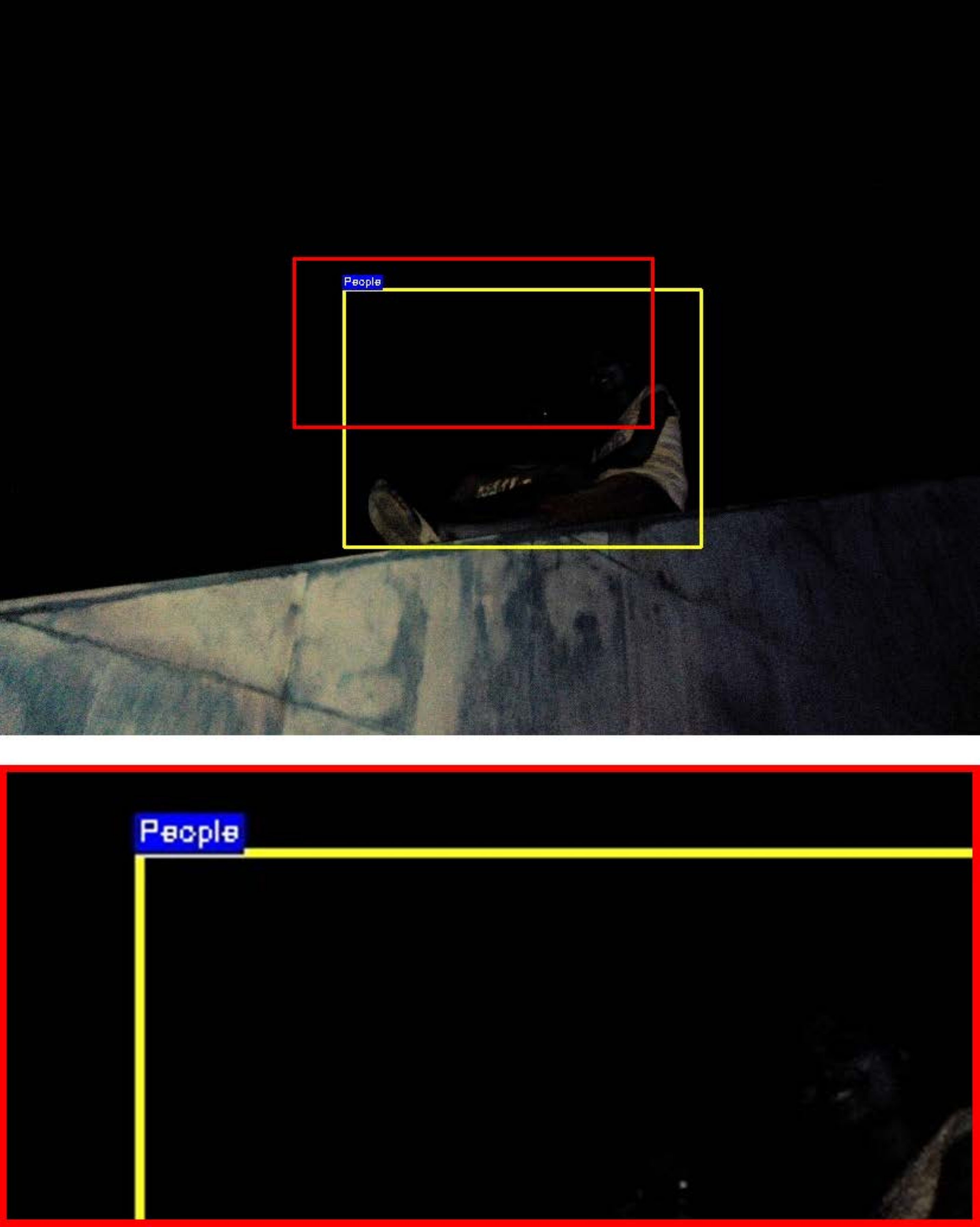}&
		\includegraphics[width=0.104\linewidth]{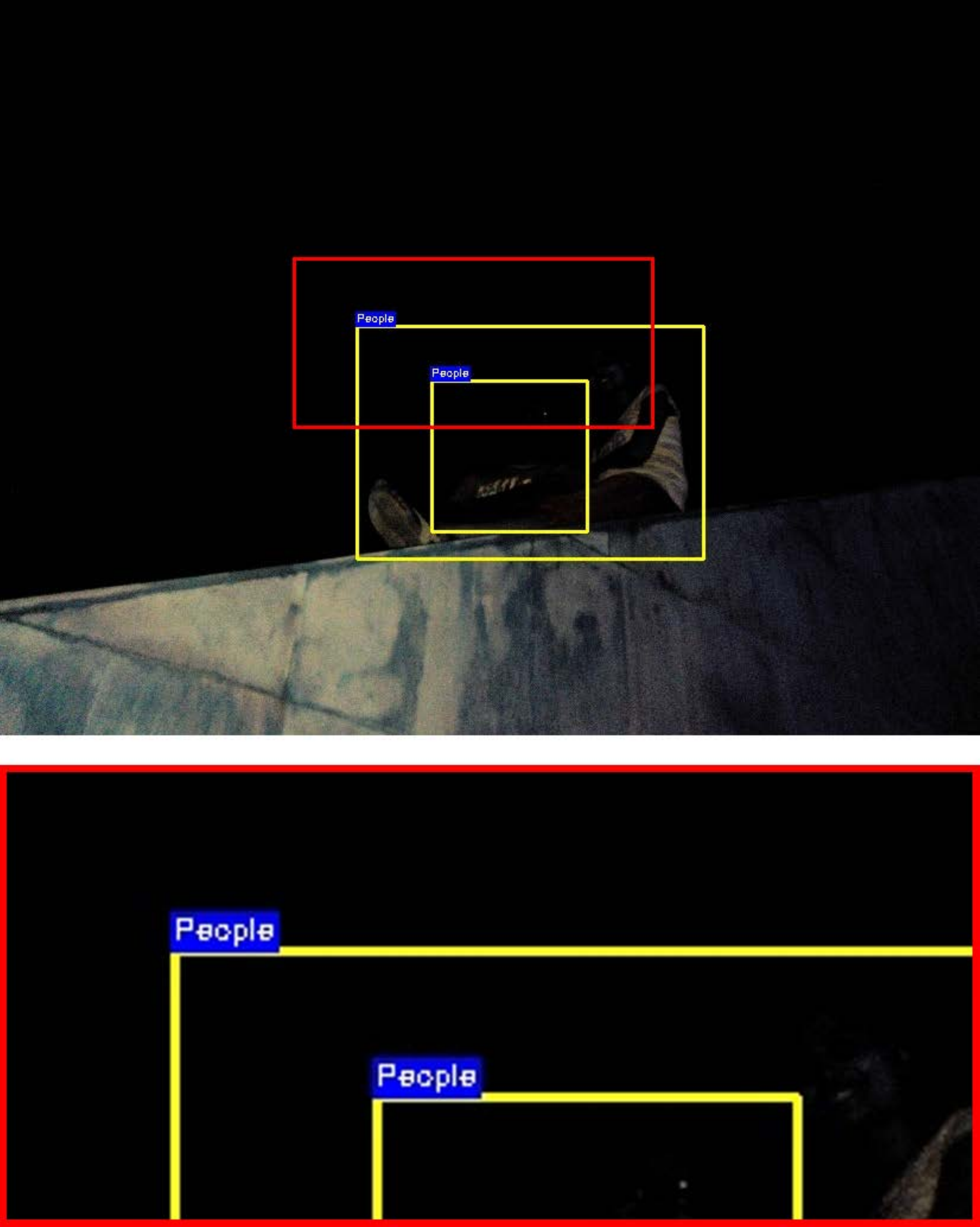}\\
		\includegraphics[width=0.104\linewidth]{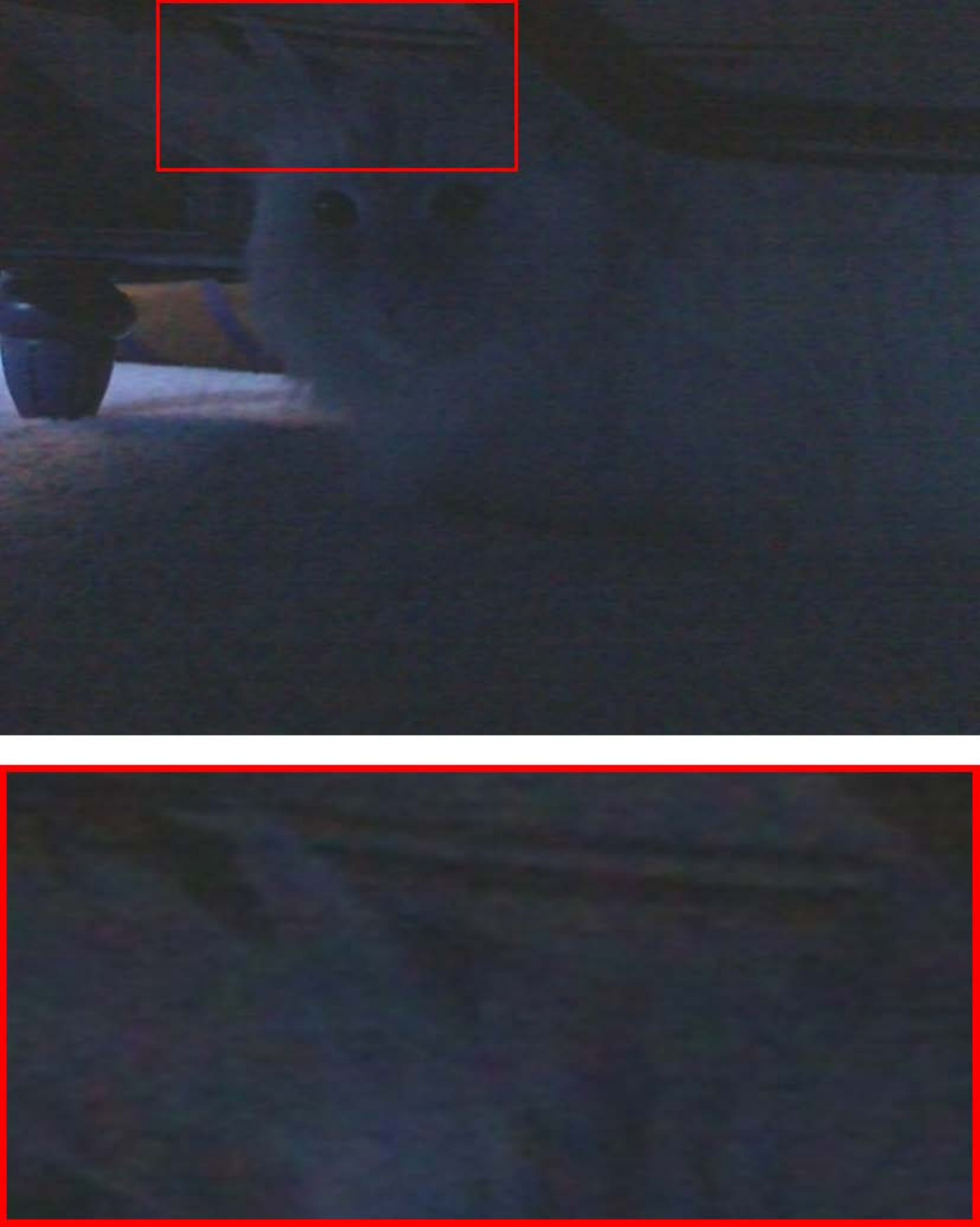}&
		\includegraphics[width=0.104\linewidth]{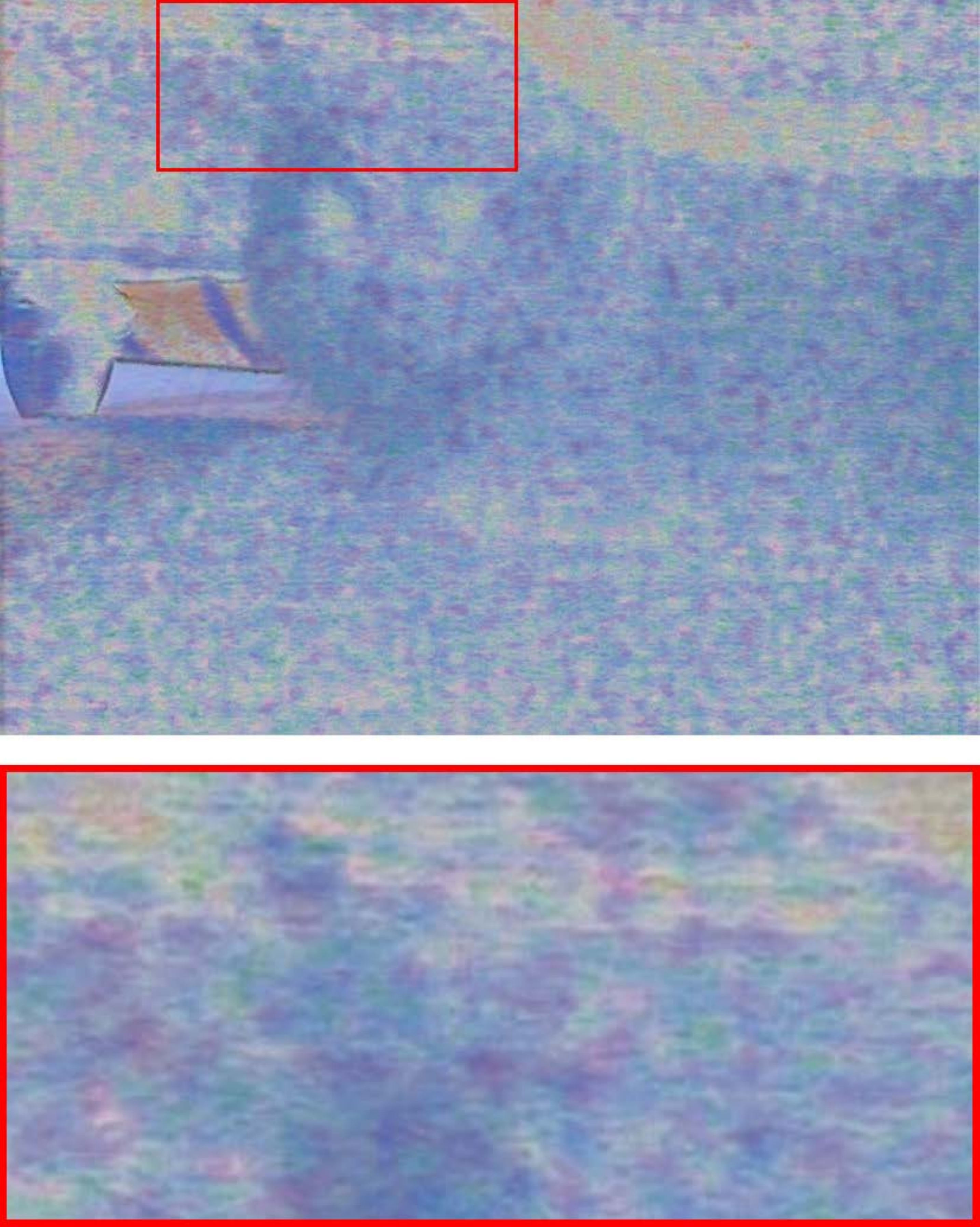}&
		\includegraphics[width=0.104\linewidth]{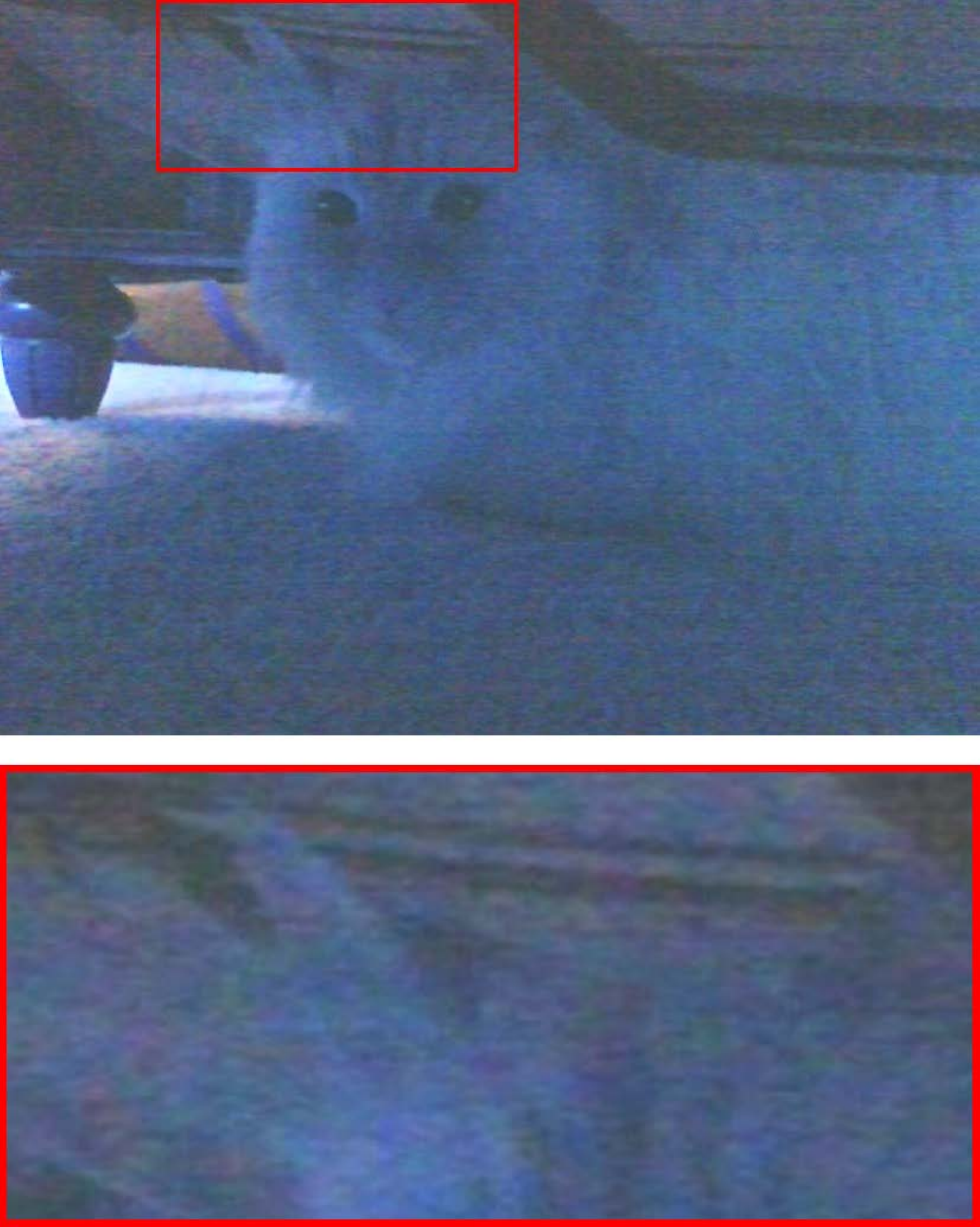}&
		\includegraphics[width=0.104\linewidth]{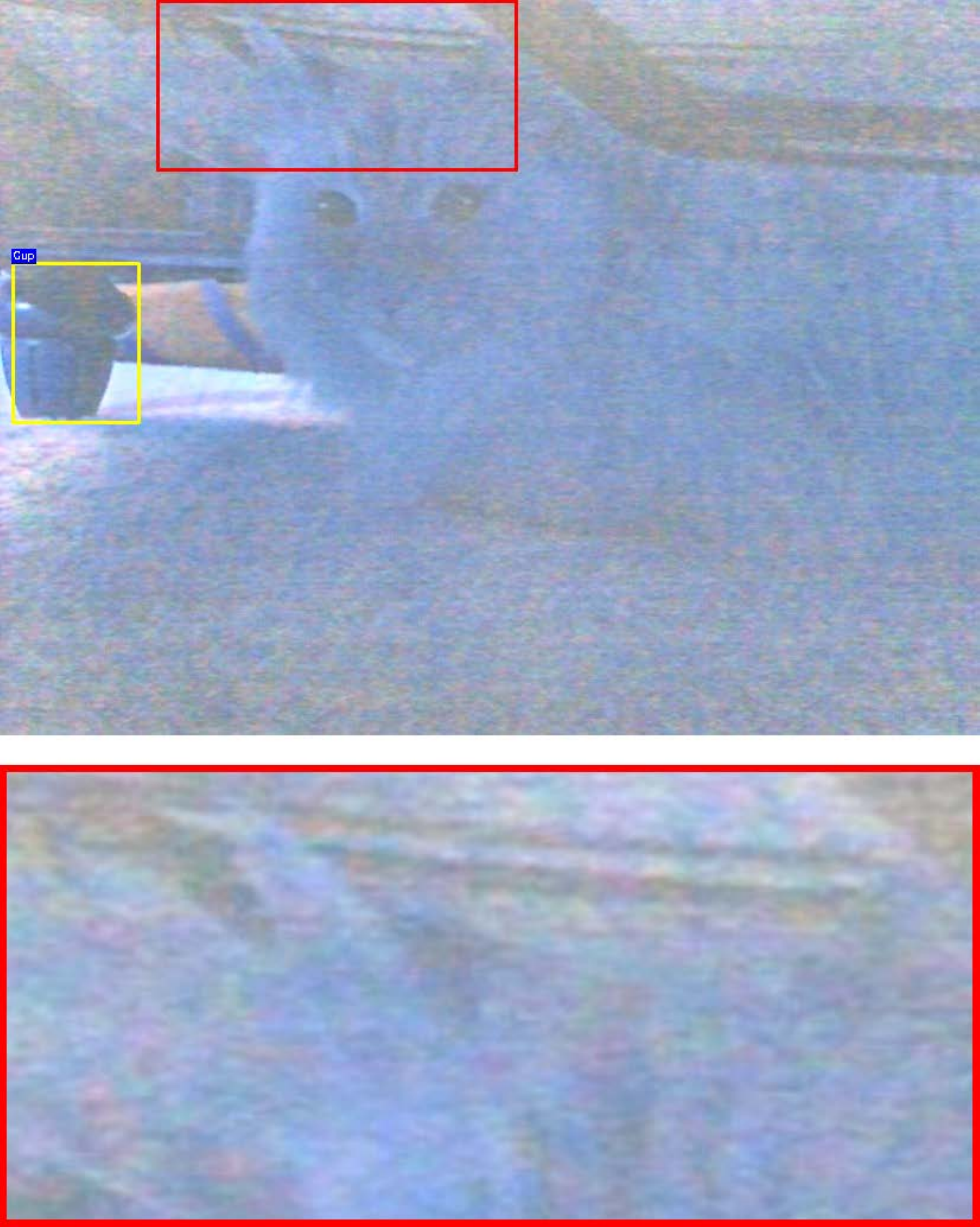}&
		\includegraphics[width=0.104\linewidth]{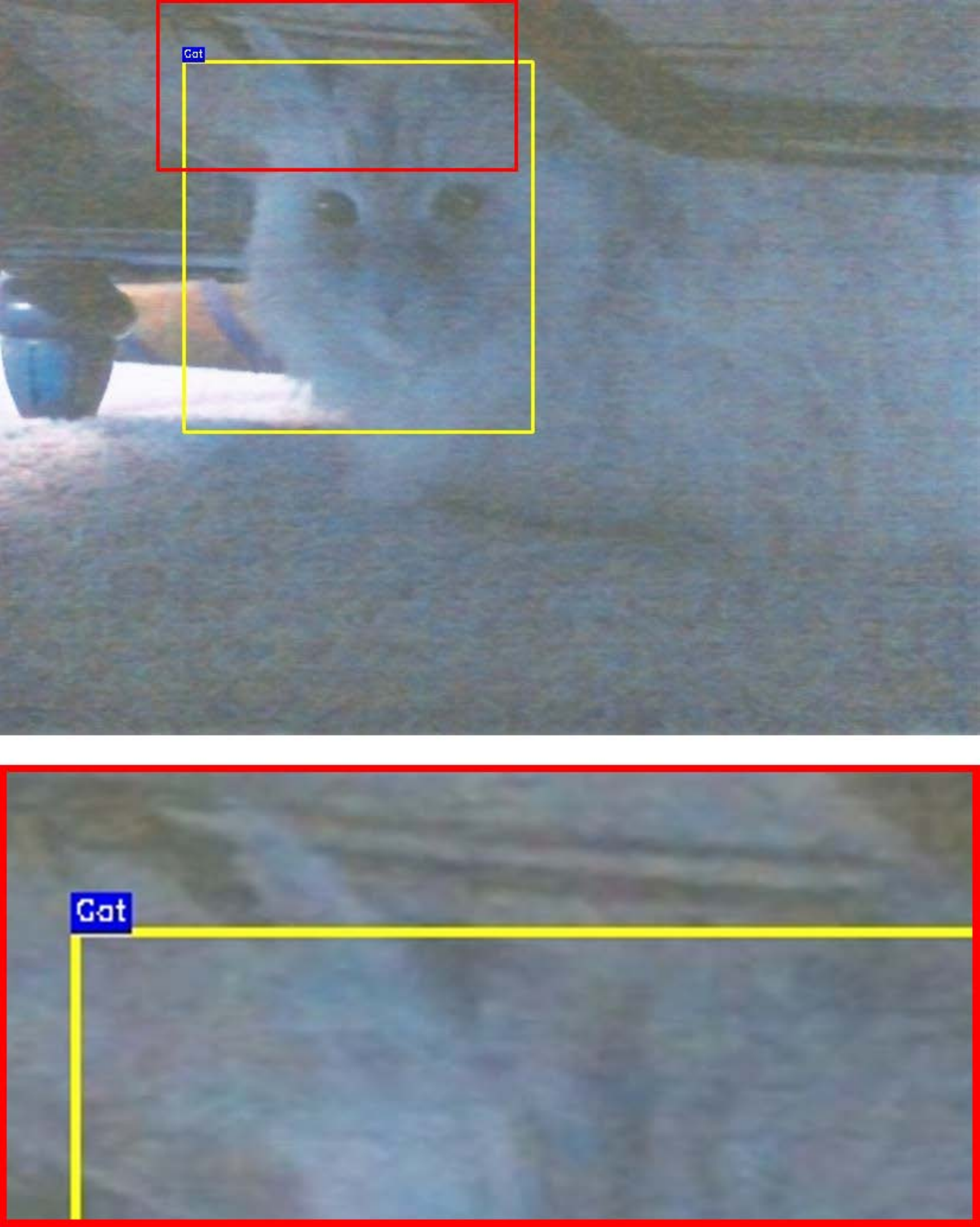}&
		\includegraphics[width=0.104\linewidth]{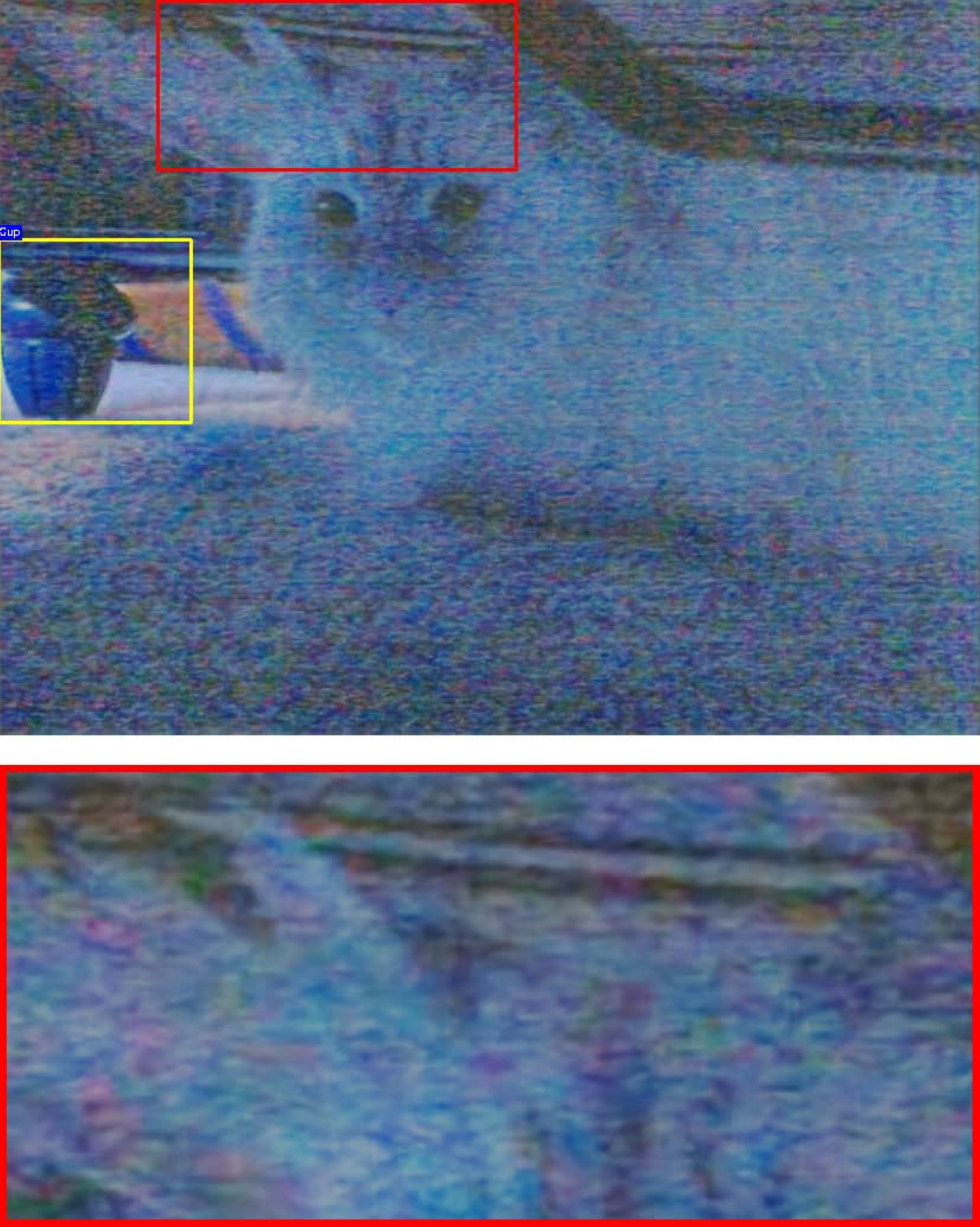}&
		\includegraphics[width=0.104\linewidth]{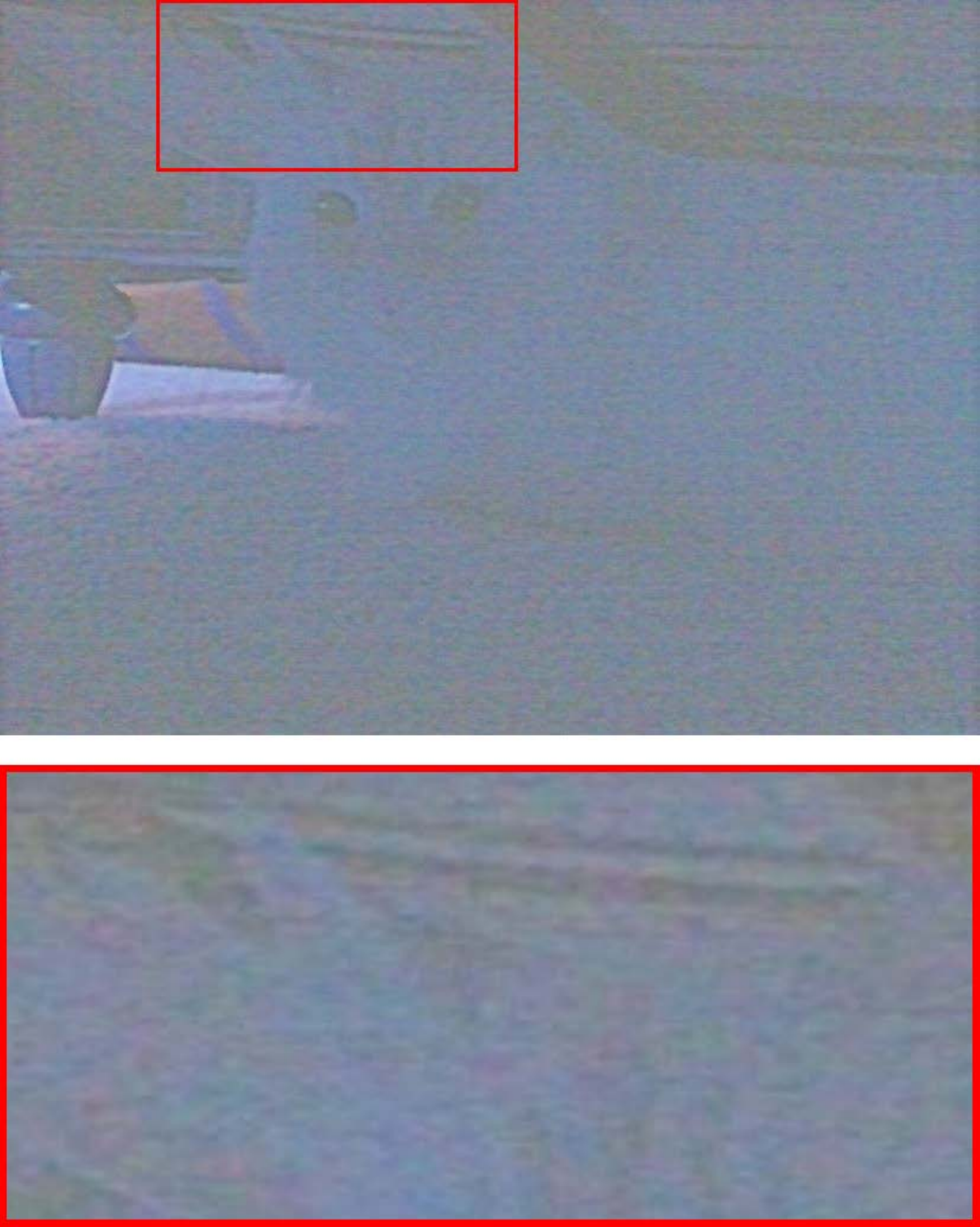}&
		\includegraphics[width=0.104\linewidth]{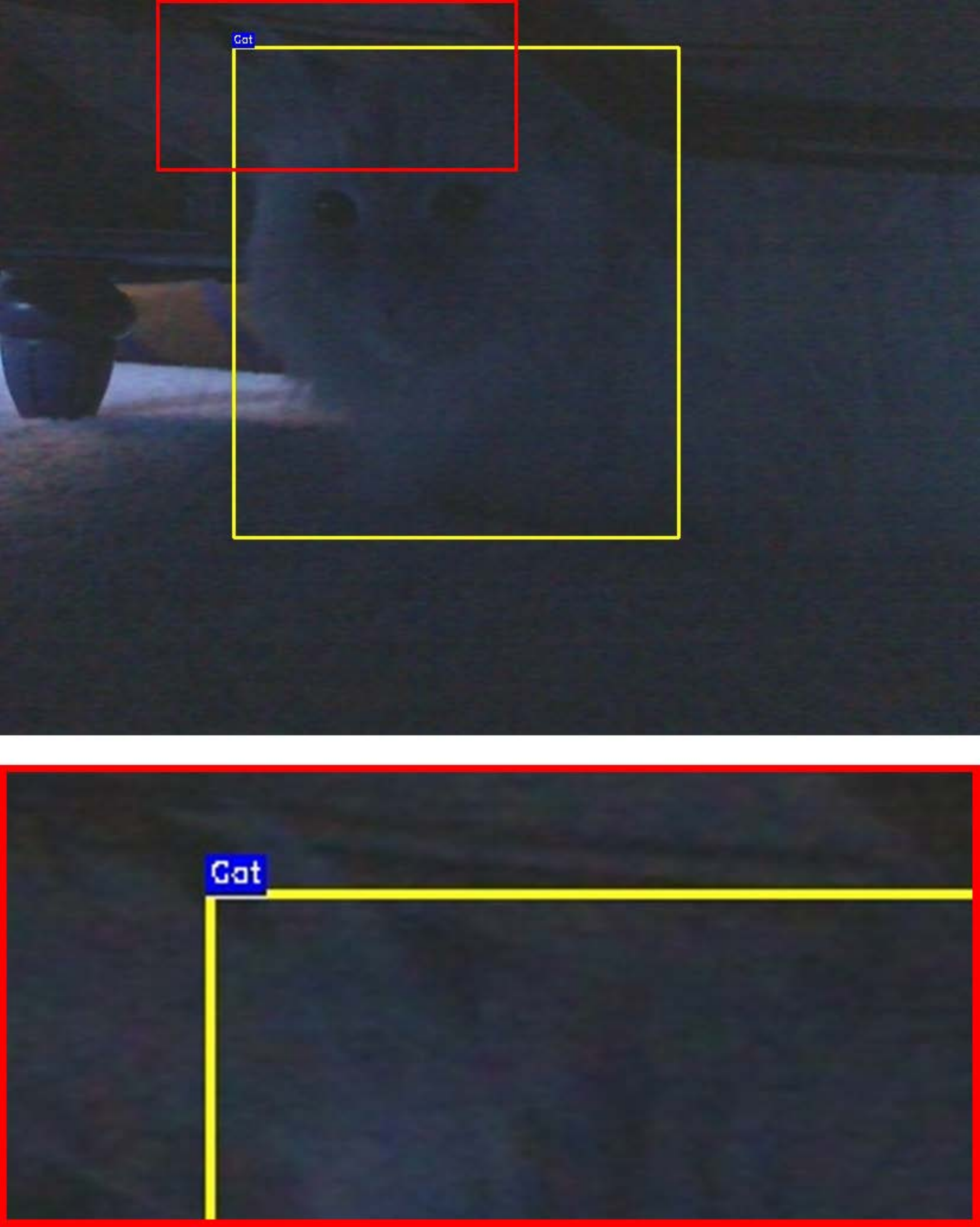}&
		\includegraphics[width=0.104\linewidth]{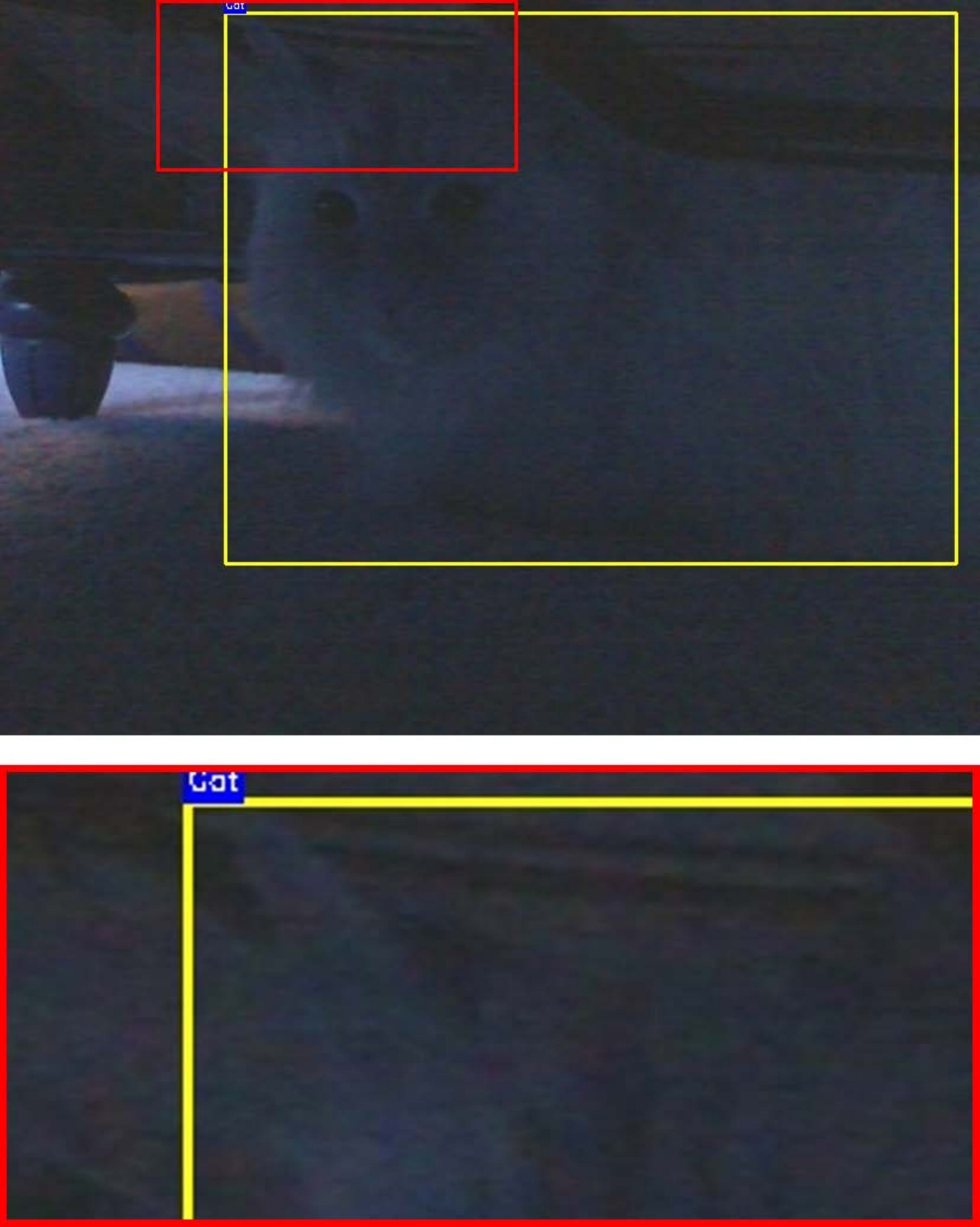}\\
		\includegraphics[width=0.104\linewidth]{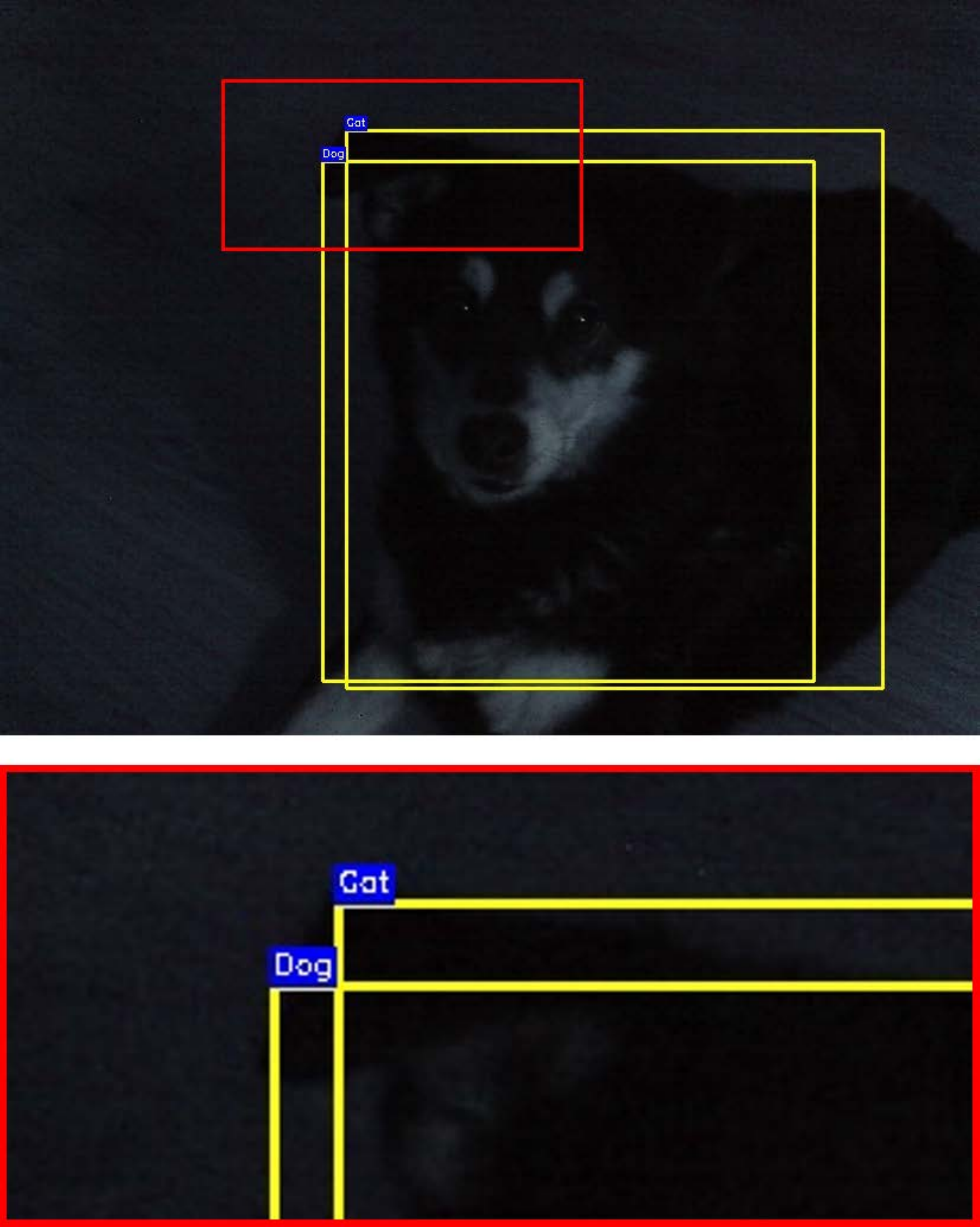}&
		\includegraphics[width=0.104\linewidth]{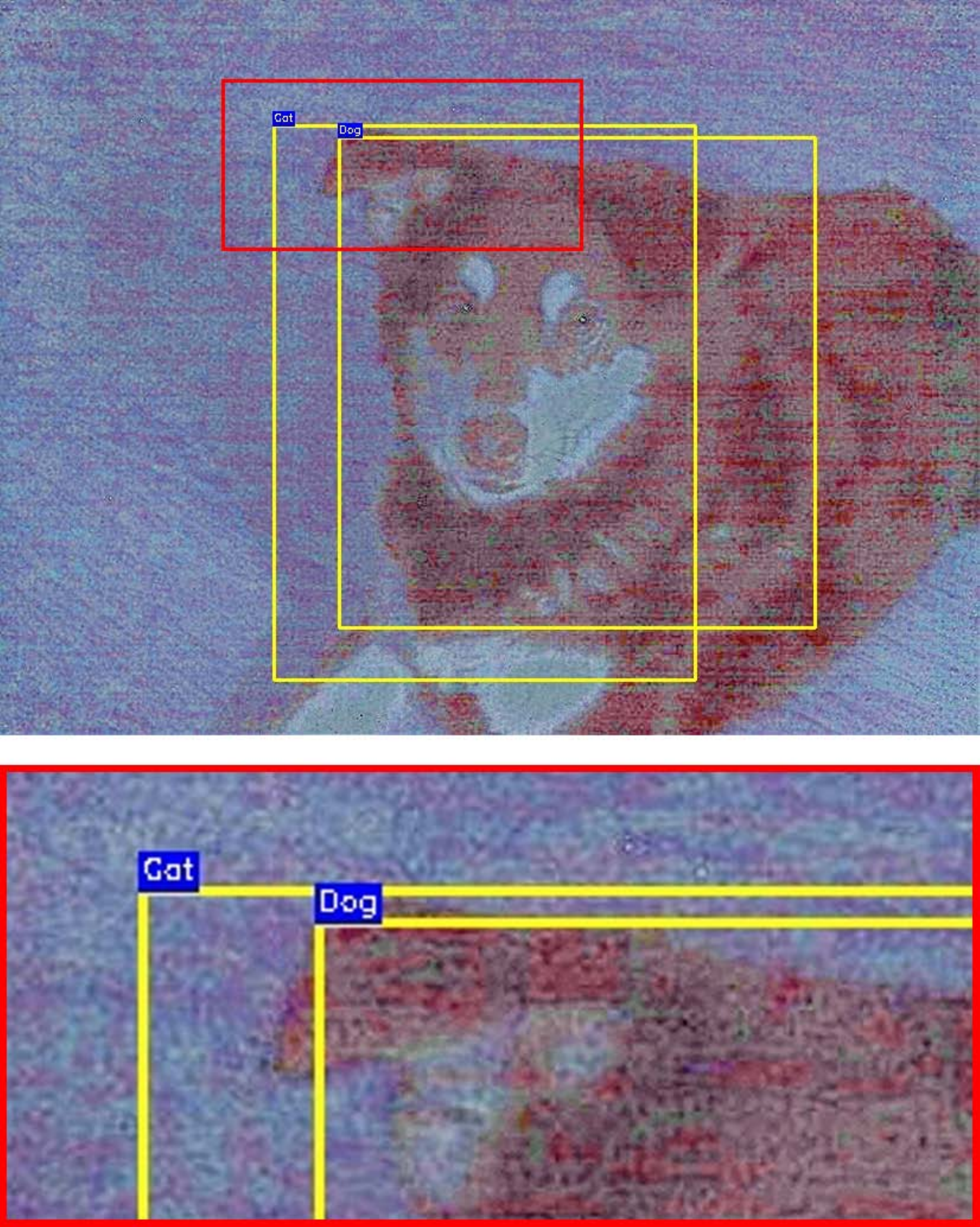}&
		\includegraphics[width=0.104\linewidth]{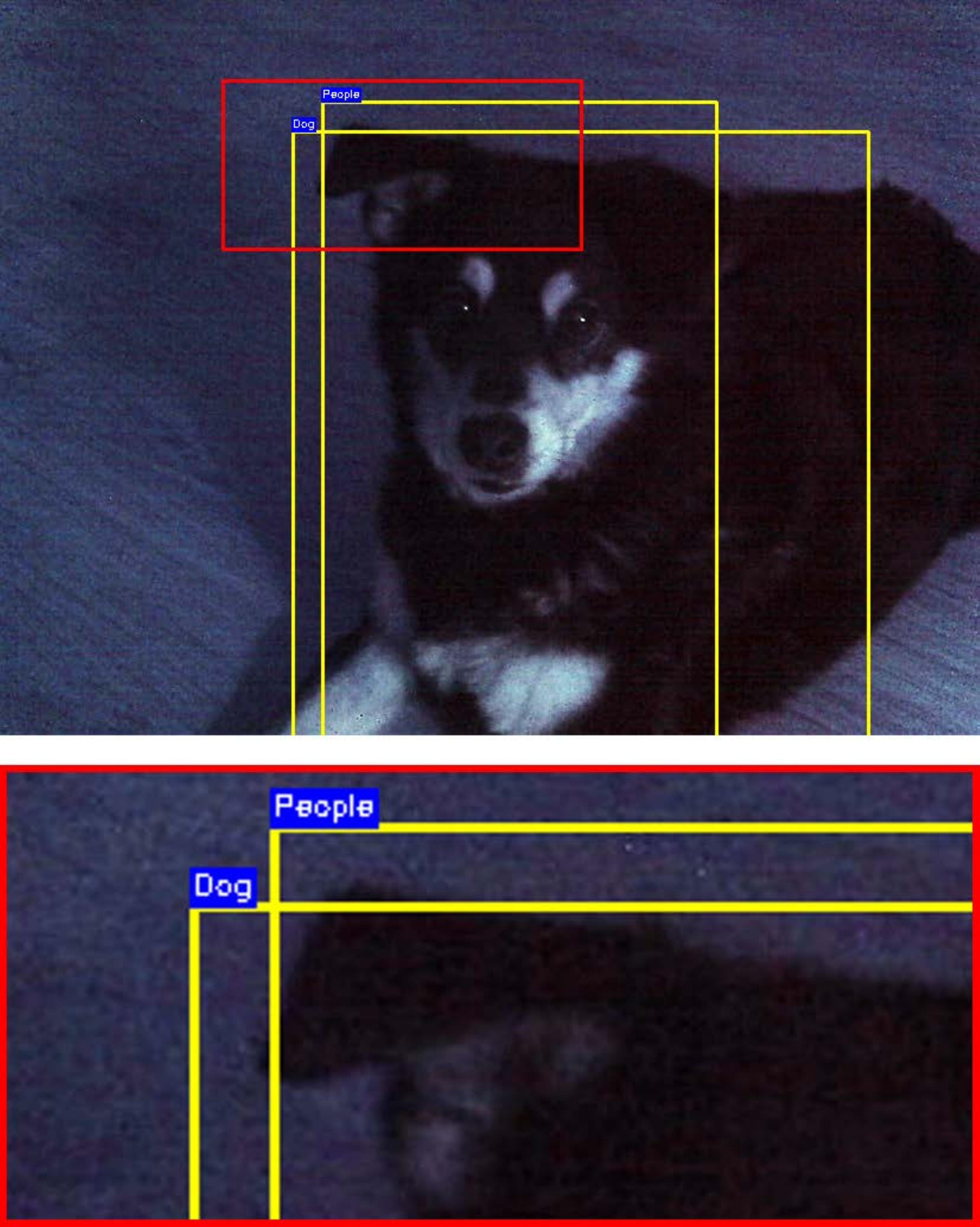}&
		\includegraphics[width=0.104\linewidth]{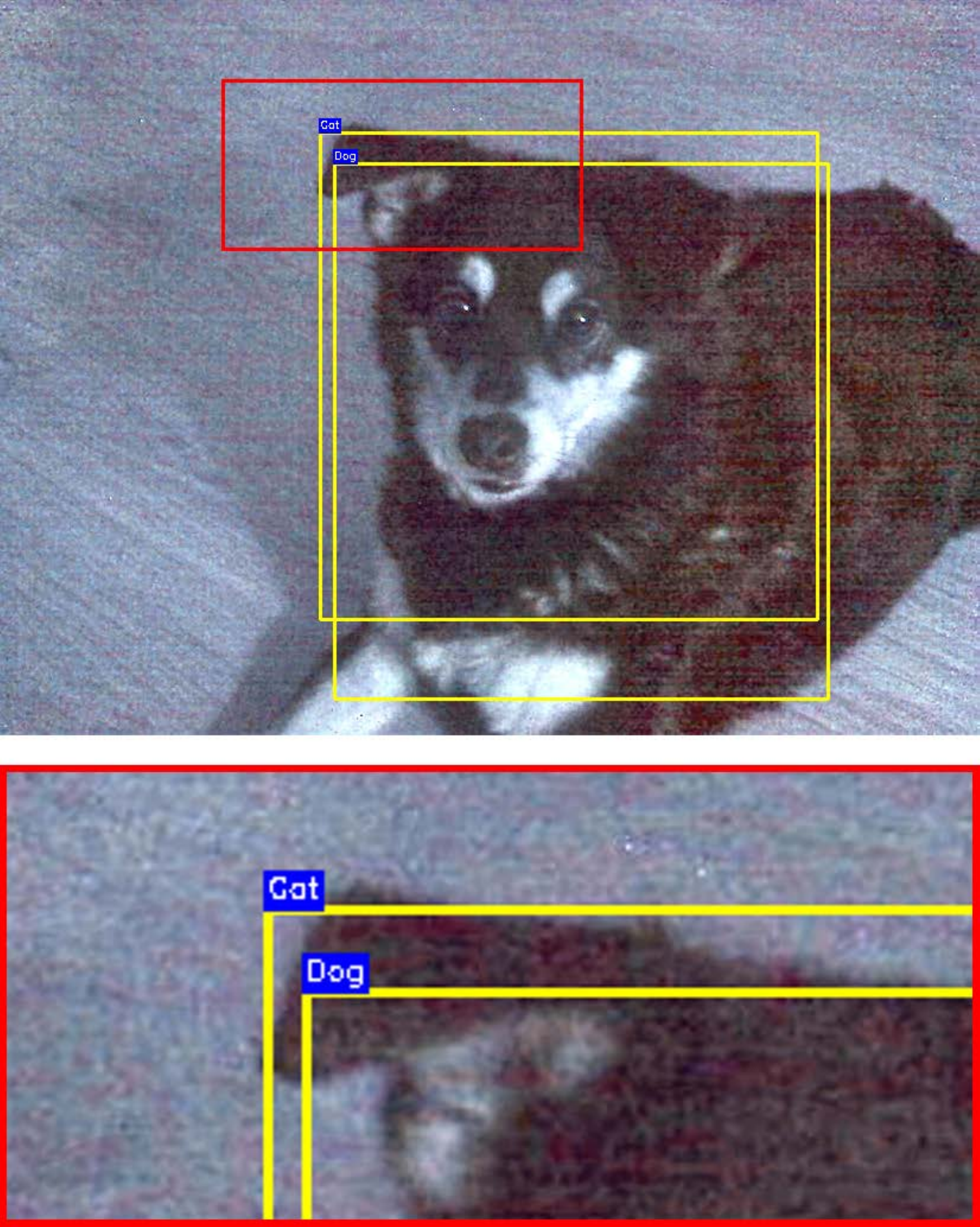}&
		\includegraphics[width=0.104\linewidth]{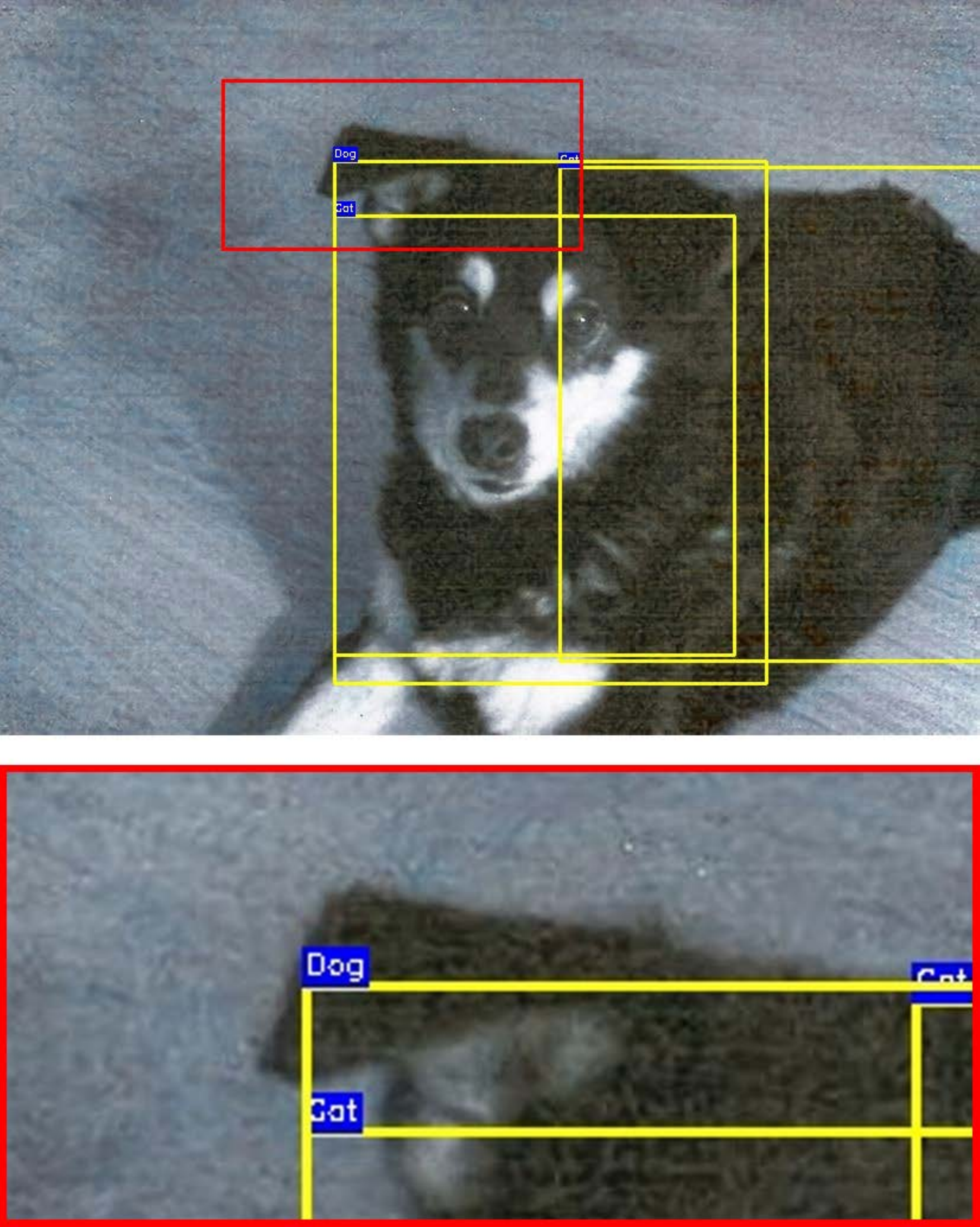}&
		\includegraphics[width=0.104\linewidth]{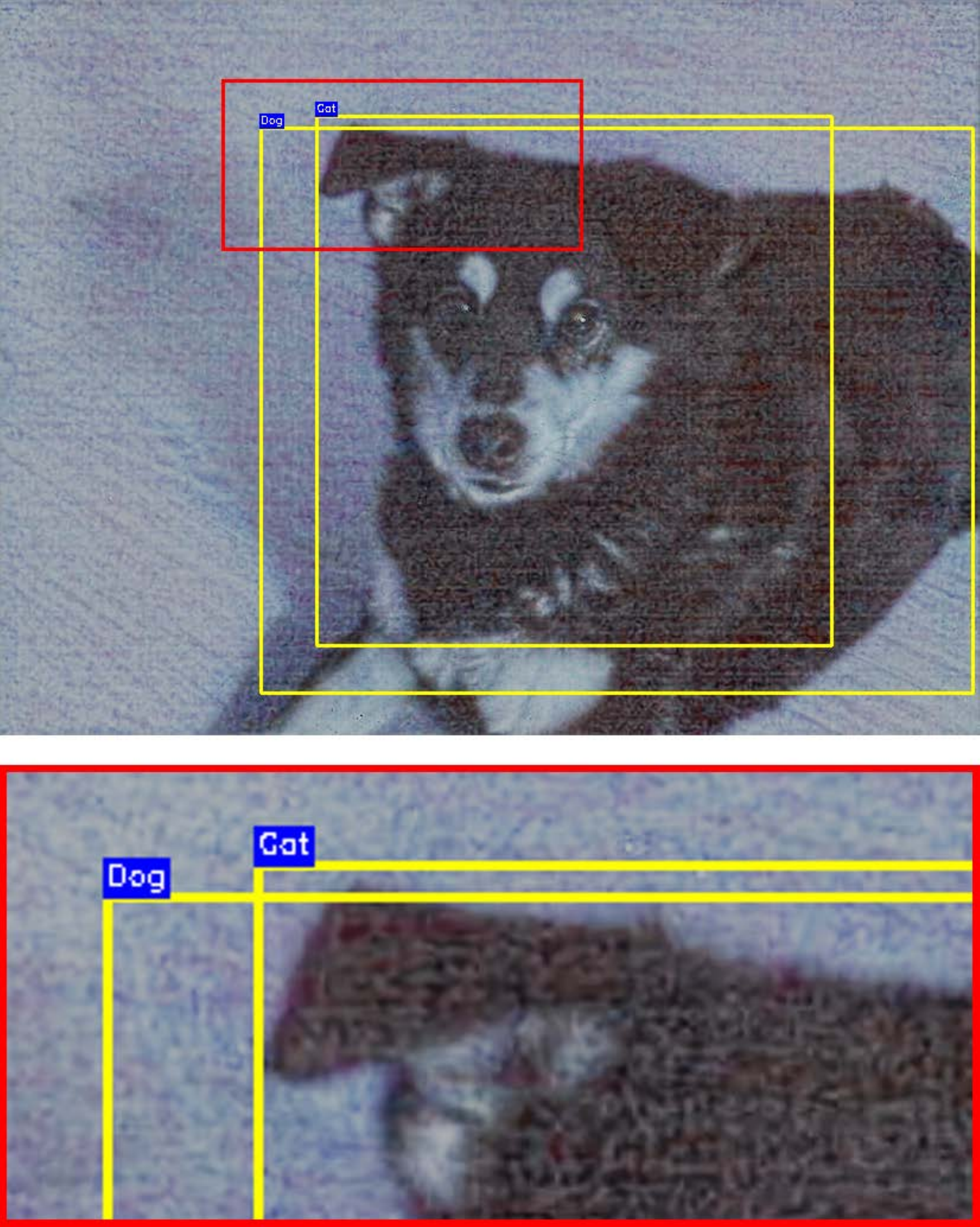}&
		\includegraphics[width=0.104\linewidth]{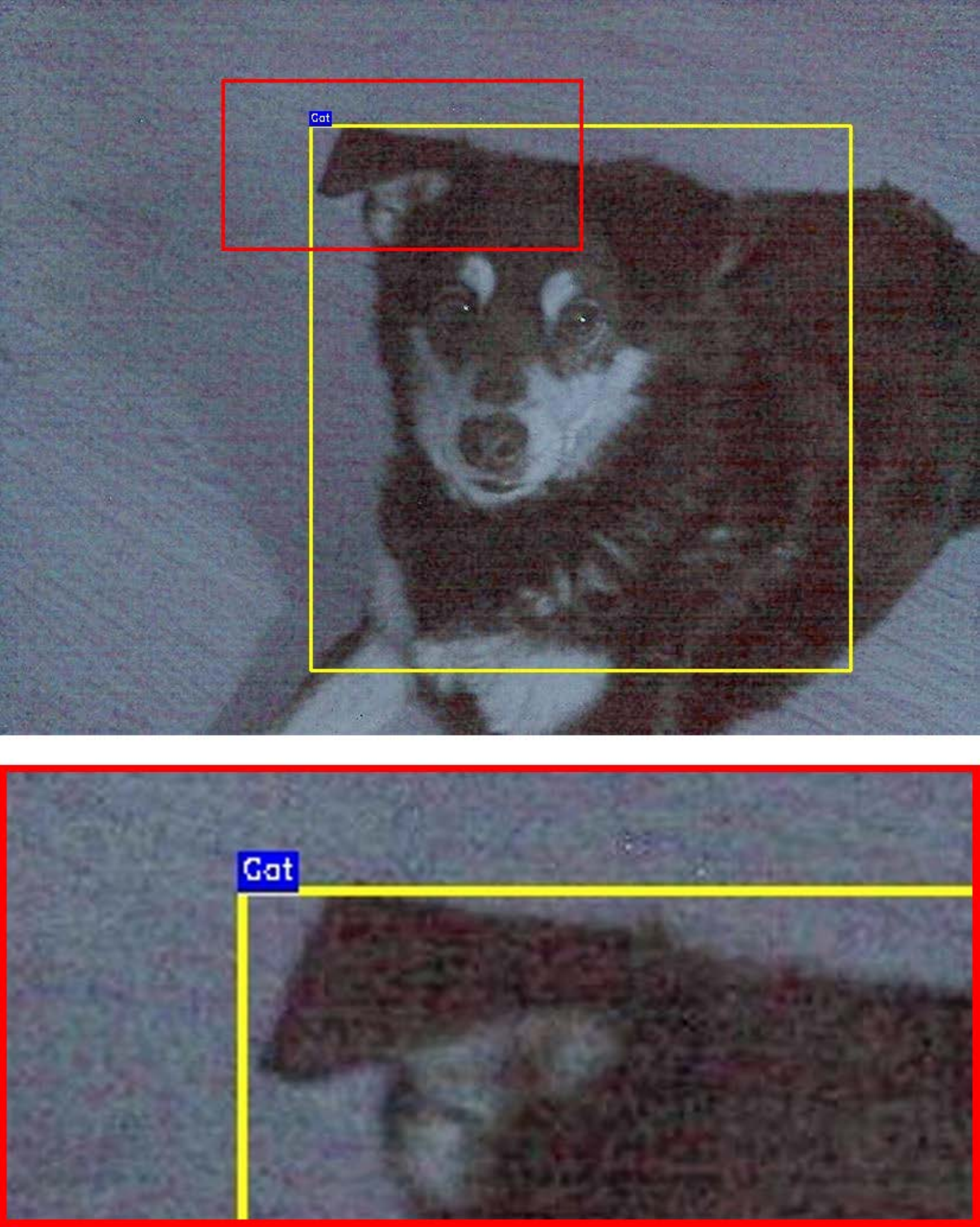}&
		\includegraphics[width=0.104\linewidth]{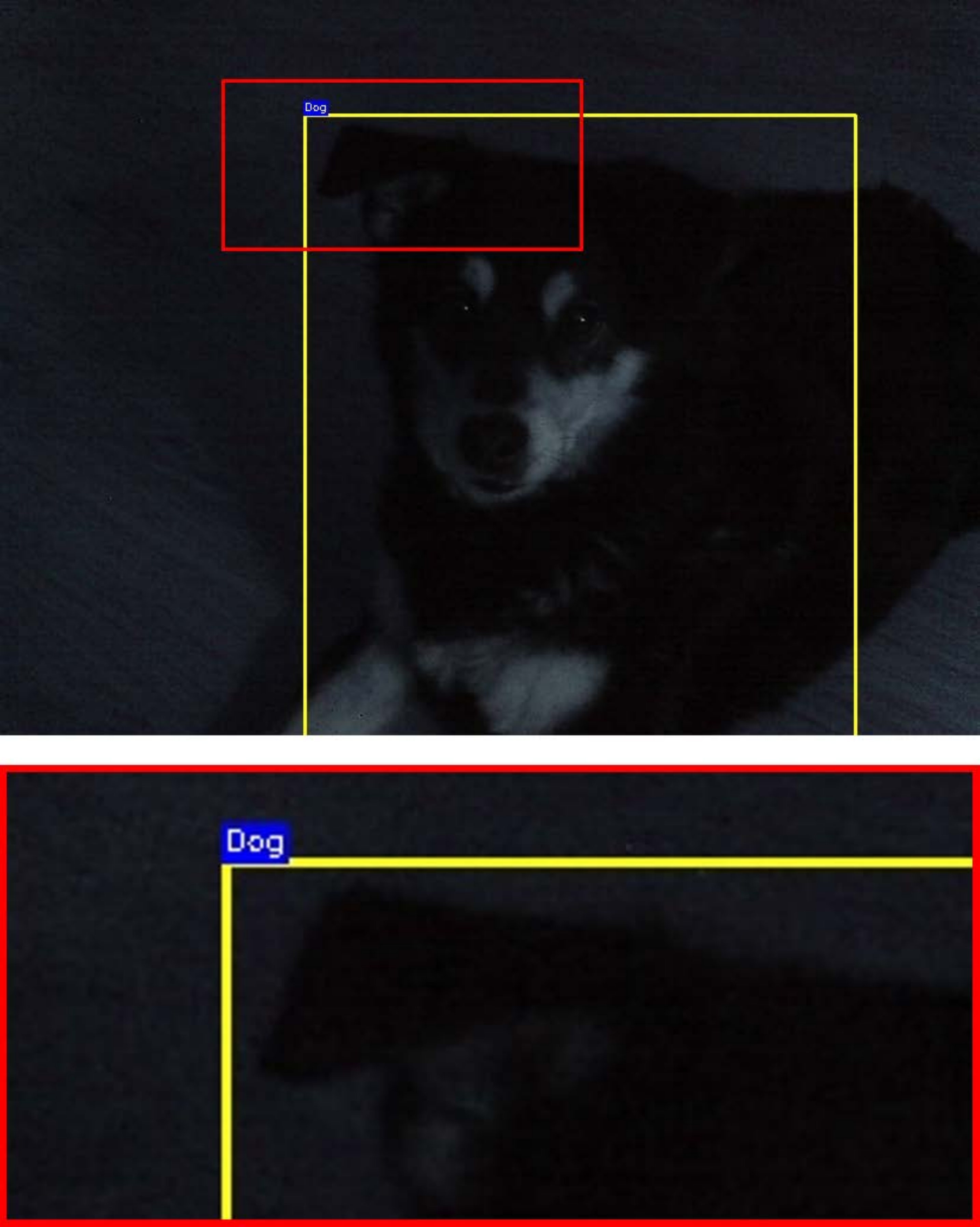}&
		\includegraphics[width=0.104\linewidth]{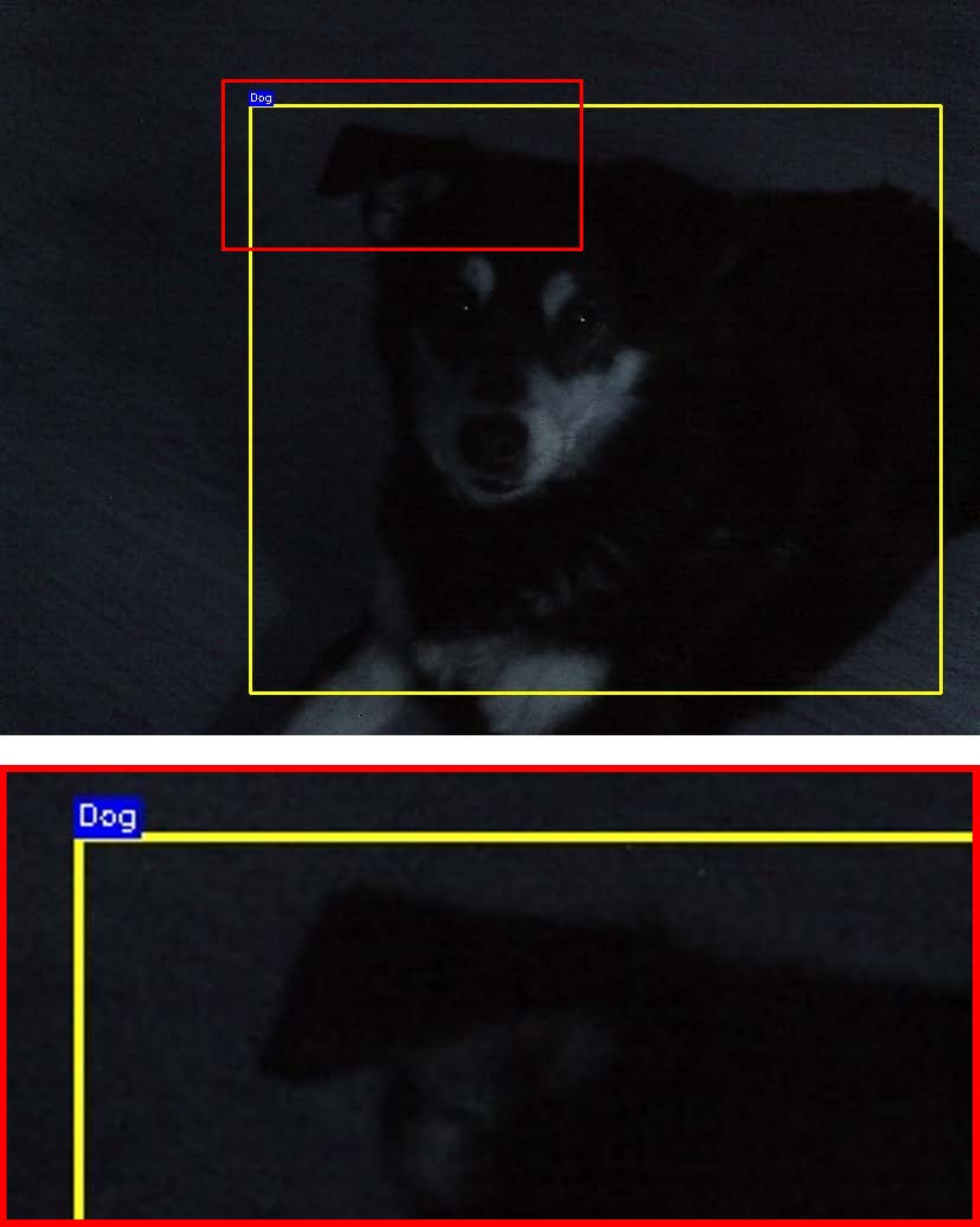}\\
		\footnotesize Input&\footnotesize RetinexNet&\footnotesize DeepUPE&\footnotesize EnGAN&\footnotesize FIDE&\footnotesize KinD&\footnotesize ZeroDCE&\footnotesize Ours &\footnotesize Label\\
	\end{tabular}
	\caption{Visual comparison of object detection on ExDark dataset. Red boxes indicate the obvious differences. }
	\label{fig:ExDark}
\end{figure*}

\textbf{Qualitative comparison.}
Here we evaluated the visual performance in some challenging real-world scenarios. Fig.~\ref{fig:DarkFace} demonstrated three groups of visual comparisons on DARK FACE dataset~\cite{yang2020advancing}. 
Although some methods were able to enhance the brightness successfully, they failed to restore the clear image textures.
On the contrary, our RUAS can restore the brightness and the details perfectly at the same time. 
In addition, we performed the visual comparison on LOL dataset that contained visible noises. As shown in Fig.~\ref{fig:LOL}, all compared methods failed to take on vivid and true colors. KinD and DRBN indeed removed noises but introduced some unknown artifacts, while our results presented vivid colors and removed undesired noises.
Then some extreme examples from the ExtremelyDarkFace dataset were shown in Fig.~\ref{fig:ExtermelyDarkFace}. These advanced deep networks being compared indeed improved the lightness, but lots of adverse artifacts became visible significantly. The recently-proposed methods, FIDE and DRBN even destroyed the color system, e.g., the overcoat should be red. 
By comparison, our RUAS improved brightness well and had great advantages in both detail restoration and noise removal.

\begin{table*}[t]
	\caption{Quantitative results of nighttime semantic segmentation on the ACDC dataset. The symbol set \{RO, SI, BU, WA, FE, PO, TL, TS, VE, TE, SK, PE, RI, CA, TR, MO, BI\} represents \{road, sidewalk, building, wall, fence, pole, traffic light, traffic sign, vegetation, terrain, sky, person, rider, car, train, motorcycle, bicycle\}. Notice that we retrained the segmentation model on the enhanced results that were generated by all the compared methods. The best result is in {\textcolor{red}{\textbf{red}}} whereas the second best one is in {\textcolor{blue}{\textbf{blue}}}.}
	\renewcommand\arraystretch{1.2} 
	\setlength{\tabcolsep}{2mm}
	\centering
	\begin{tabular}{|c|ccccccccccccccccc|c|}
		\hline 
		\footnotesize Method&\footnotesize RO&\footnotesize SI&\footnotesize BU&\footnotesize WA &\footnotesize FE &\footnotesize PO&\footnotesize TL&\footnotesize TS&\footnotesize VE&\footnotesize TE&\footnotesize SK&\footnotesize PE&\footnotesize RI&\footnotesize CA&\footnotesize TR&\footnotesize MO&\footnotesize BI&\footnotesize mIoU\\
		\hline 
		\footnotesize DeepLab-v3+&\footnotesize 90.0 &\footnotesize 61.4 &\footnotesize  74.2&\footnotesize 32.8 &\footnotesize 34.4 &\footnotesize \textcolor{red}{\textbf{45.7}} &\footnotesize \textcolor{blue}{\textbf{49.8}} &\footnotesize \footnotesize \textcolor{red}{\textbf{31.2}}  &\footnotesize 68.8&\footnotesize \textcolor{blue}{\textbf{14.6}} &\footnotesize  \textcolor{blue}{\textbf{80.4}} &\footnotesize  27.1 &\footnotesize 12.6 &\footnotesize \textcolor{red}{\textbf{62.1}}  &\footnotesize \textcolor{blue}{\textbf{76.3}} &\footnotesize 7.3 &\footnotesize  14.4 &\footnotesize \textcolor{blue}{\textbf{46.1}} \\
		\hline 
		\footnotesize GLADNet&\footnotesize {{89.3}} &\footnotesize  58.9 &\footnotesize 74.0 &\footnotesize 31.9 &\footnotesize 29.1 &\footnotesize 44.7 &\footnotesize 47.9 &\footnotesize 26.0 &\footnotesize 67.7 &\footnotesize 11.9 &\footnotesize 78.7 &\footnotesize 26.6 &\footnotesize \textcolor{blue}{\textbf{15.9}} &\footnotesize 56.8&\footnotesize 67.9 &\footnotesize 4.5 &\footnotesize 10.6 &\footnotesize 43.7\\
		\hline 
		\footnotesize DeepUPE&\footnotesize 90.1 &\footnotesize 62.0 &\footnotesize 70.0 &\footnotesize 30.3 &\footnotesize 33.6 &\footnotesize 44.3 &\footnotesize \textcolor{red}{\textbf{50.1}} &\footnotesize \footnotesize \textcolor{blue}{\textbf{28.4}}   &\footnotesize 68.8 &\footnotesize  13.1 &\footnotesize 80.1 &\footnotesize 26.3 &\footnotesize 6.3 &\footnotesize 57.4&\footnotesize 73.2 &\footnotesize 6.9 &\footnotesize 9.4 &\footnotesize 44.3 \\
		\hline 
		\footnotesize EnGAN&\footnotesize 89.7 &\footnotesize 58.9 &\footnotesize 73.7 &\footnotesize 32.8 &\footnotesize 31.8 &\footnotesize  44.7 &\footnotesize 49.2 &\footnotesize 26.2 &\footnotesize 67.3 &\footnotesize 14.2 &\footnotesize 77.8 &\footnotesize 25.0 &\footnotesize 10.6 &\footnotesize {{59.0}}  &\footnotesize 71.2 &\footnotesize 6.9 &\footnotesize 7.8 &\footnotesize 43.9 \\
		\hline
		\footnotesize FIDE&\footnotesize 90.0 &\footnotesize 60.7 &\footnotesize  {{72.8}} &\footnotesize 32.4 &\footnotesize 34.1 &\footnotesize 43.3 &\footnotesize  47.9 &\footnotesize  26.1&\footnotesize 67.0 &\footnotesize 13.7&\footnotesize 78.0&\footnotesize 26.5 &\footnotesize 5.8 &\footnotesize  57.1&\footnotesize 71.0 &\footnotesize  4.8 &\footnotesize 12.4 &\footnotesize  43.7\\
		\hline 
		\footnotesize ZeroDCE&\footnotesize \textcolor{blue}{\textbf{90.6}} &\footnotesize 59.9 &\footnotesize 73.9 &\footnotesize 32.6 &\footnotesize 31.7 &\footnotesize 44.3 &\footnotesize 46.2 &\footnotesize 25.8 &\footnotesize 67.2 &\footnotesize \textcolor{blue}{\textbf{14.6}}&\footnotesize 79.1 &\footnotesize 24.7 &\footnotesize 7.7 &\footnotesize \textcolor{blue}{\textbf{59.4}} &\footnotesize 66.8 &\footnotesize 6.2 &\footnotesize 13.9 &\footnotesize 43.8\\
		\hline
		\footnotesize DANNet &\footnotesize 89.9 &\footnotesize 61.7 &\footnotesize 71.1 &\footnotesize 32.5 &\footnotesize 36.0 &\footnotesize 20.9 &\footnotesize 37.7 &\footnotesize 22.2 &\footnotesize \textcolor{blue}{\textbf{69.5}}  &\footnotesize \textcolor{red}{\textbf{21.6}} &\footnotesize \textcolor{blue}{\textbf{80.4}} &\footnotesize \textcolor{red}{\textbf{32.5}}  &\footnotesize \textcolor{red}{\textbf{16.6}} &\footnotesize 42.5 &\footnotesize 61.6 &\footnotesize \textcolor{blue}{\textbf{8.6}} &\footnotesize \textcolor{red}{\textbf{46.9}} &\footnotesize 44.3 \\
		\hline
		\footnotesize Ours&\footnotesize \textcolor{red}{\textbf{91.7}} &\footnotesize \textcolor{red}{\textbf{65.3}} &\footnotesize  \textcolor{red}{\textbf{75.7}} &\footnotesize \textcolor{red}{\textbf{38.5}}  &\footnotesize \textcolor{red}{\textbf{38.5}}  &\footnotesize \textcolor{blue}{\textbf{45.3}}&\footnotesize {{45.7}} &\footnotesize {{27.0}} &\footnotesize \textcolor{red}{\textbf{70.0}} &\footnotesize 10.4&\footnotesize \textcolor{red}{\textbf{82.1}} &\footnotesize \textcolor{blue}{\textbf{29.2}}  &\footnotesize {{11.6}} &\footnotesize {{59.0}}   &\footnotesize \textcolor{red}{\textbf{79.8}} &\footnotesize \textcolor{red}{\textbf{10.4}} &\footnotesize \textcolor{blue}{\textbf{43.9}} &\footnotesize \textcolor{red}{\textbf{48.5}} \\
		\hline 
	\end{tabular}
	\label{tab: Segmentation}
\end{table*}

\begin{figure*}[t]
	\centering
	\begin{tabular}{c@{\extracolsep{0.3em}}c@{\extracolsep{0.3em}}c@{\extracolsep{0.3em}}c@{\extracolsep{0.3em}}c@{\extracolsep{0.3em}}c@{\extracolsep{0.3em}}c@{\extracolsep{0.3em}}c@{\extracolsep{0.3em}}c}
		\includegraphics[width=0.104\linewidth]{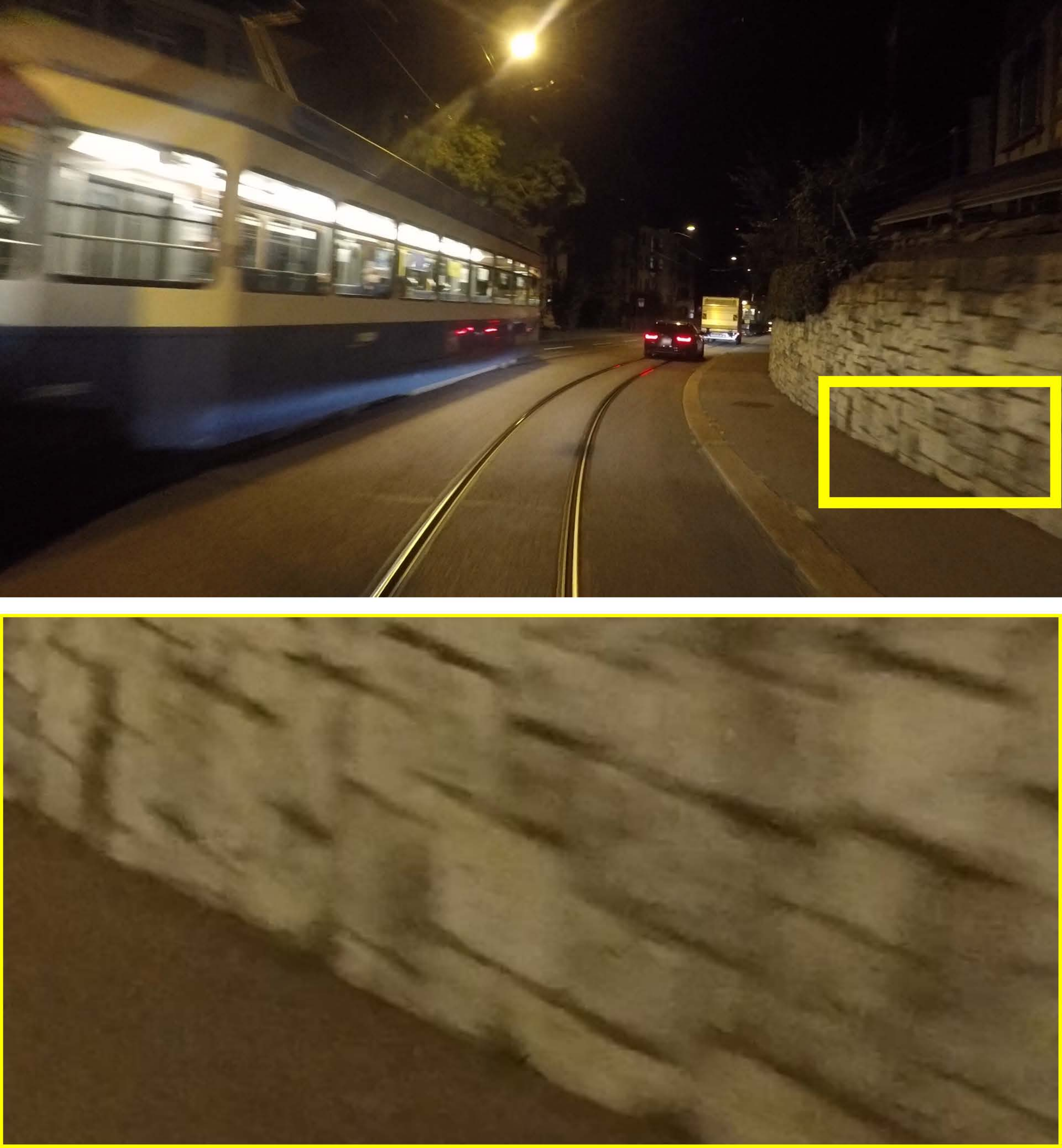}&
		\includegraphics[width=0.104\linewidth]{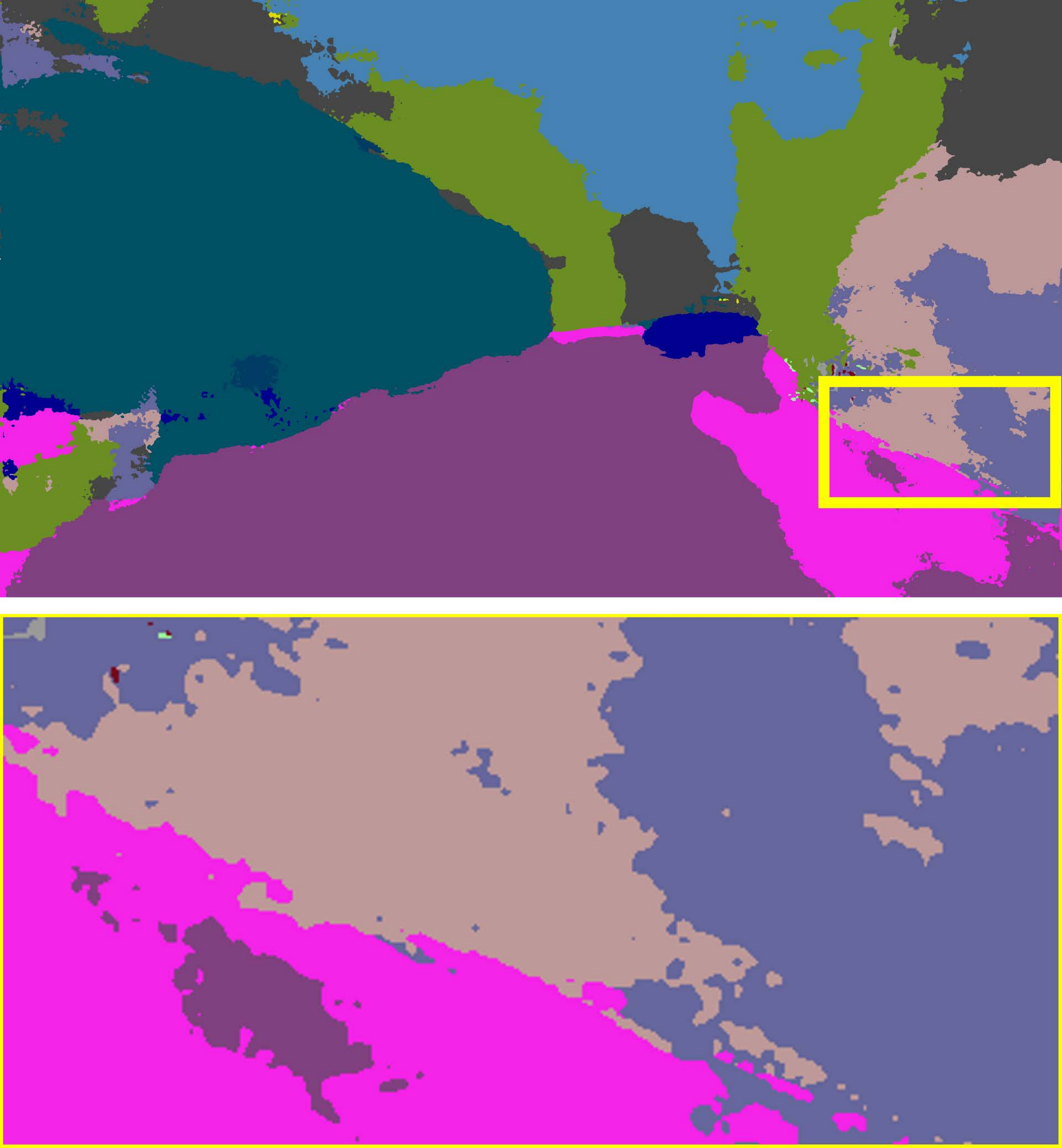}&
		\includegraphics[width=0.104\linewidth]{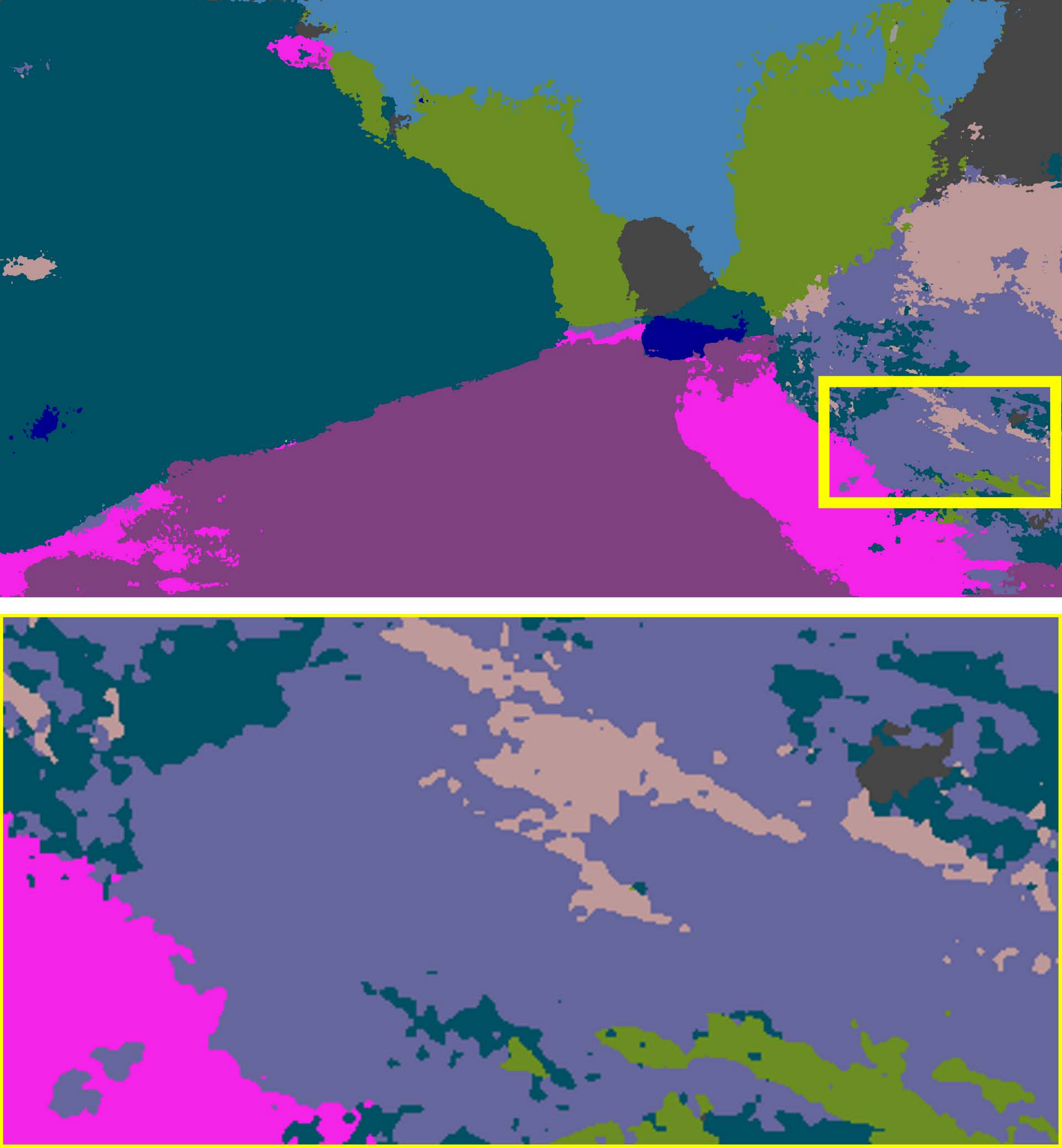}&
		\includegraphics[width=0.104\linewidth]{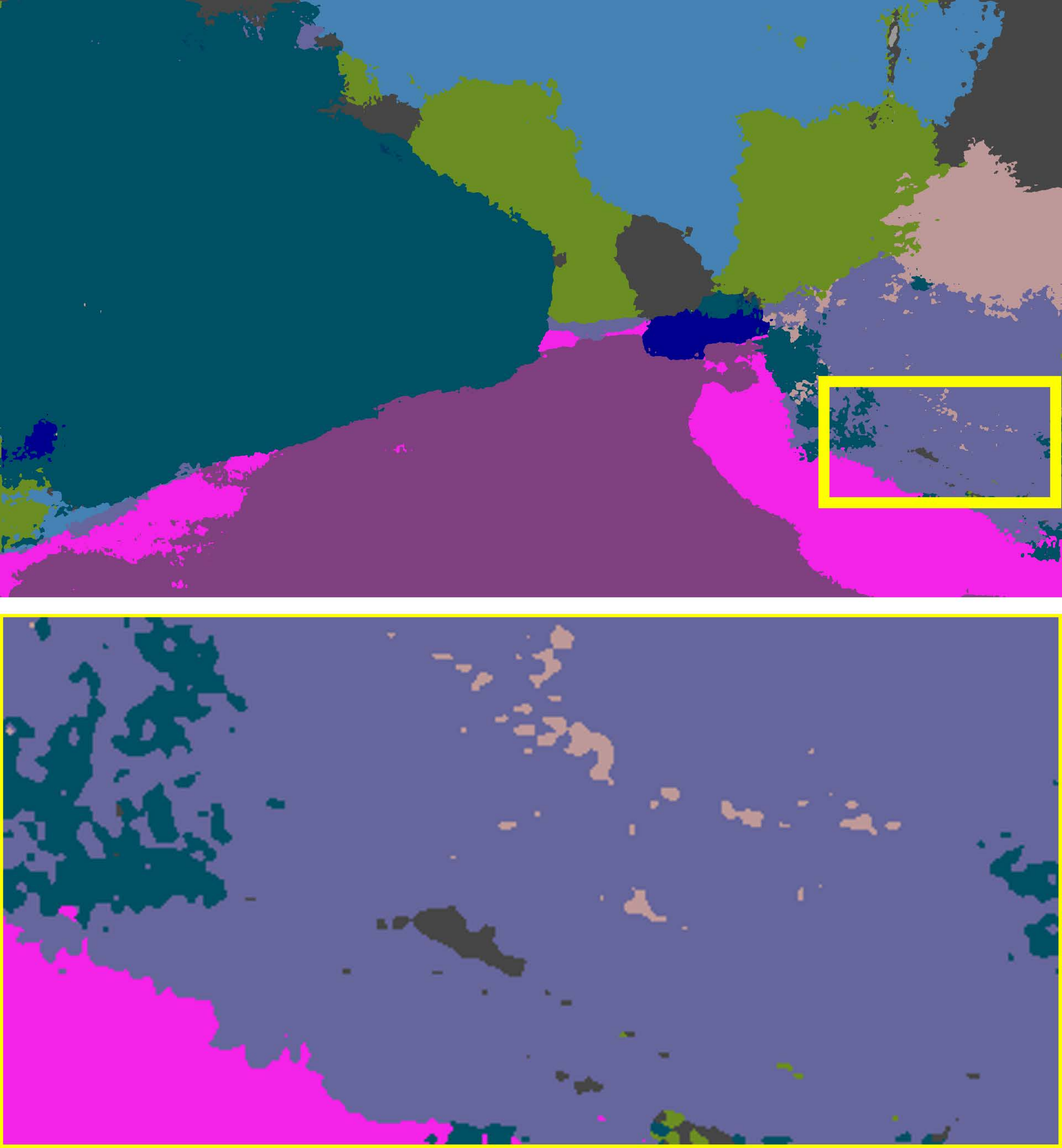}&
		\includegraphics[width=0.104\linewidth]{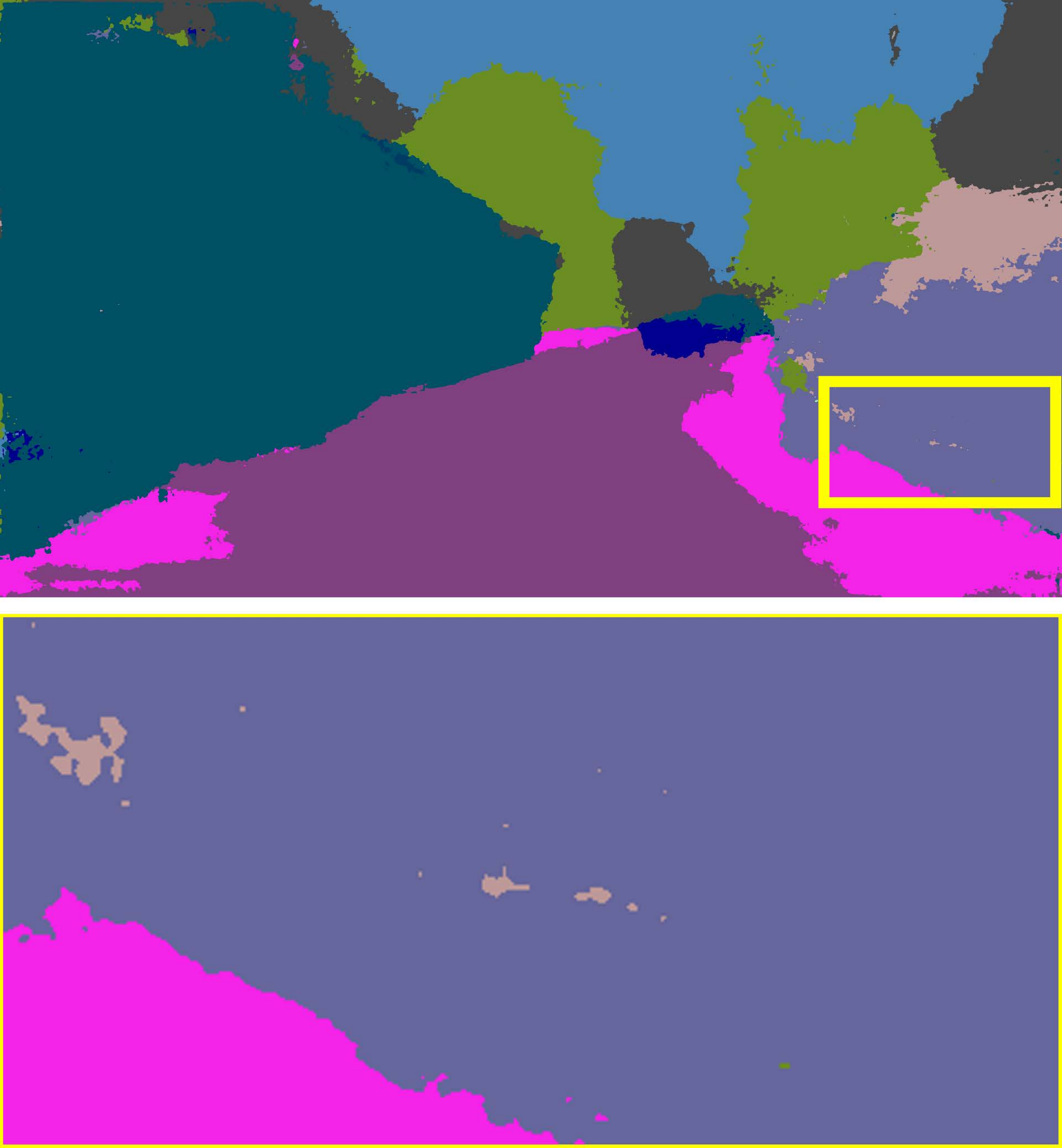}&
		\includegraphics[width=0.104\linewidth]{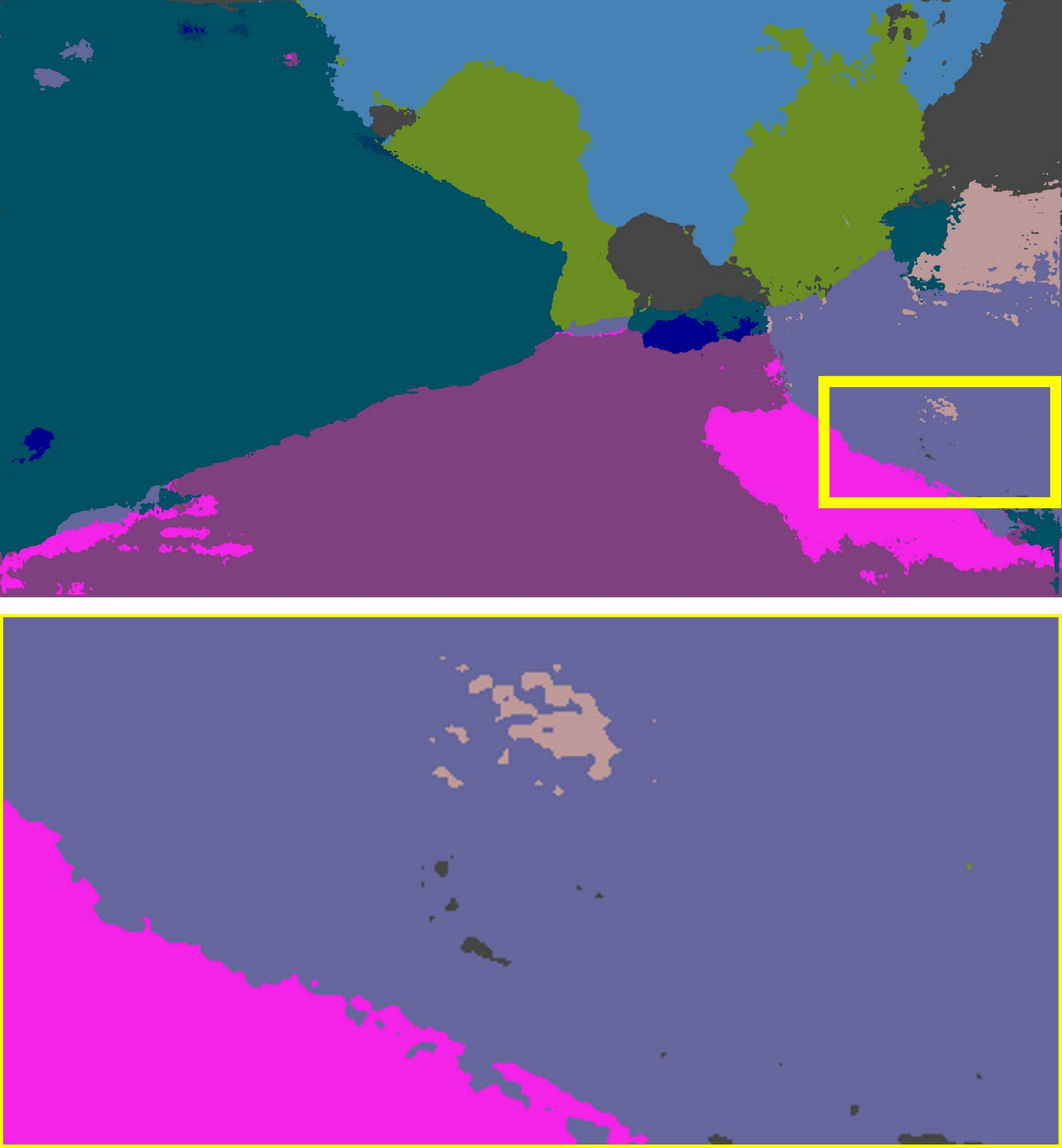}&
		\includegraphics[width=0.104\linewidth]{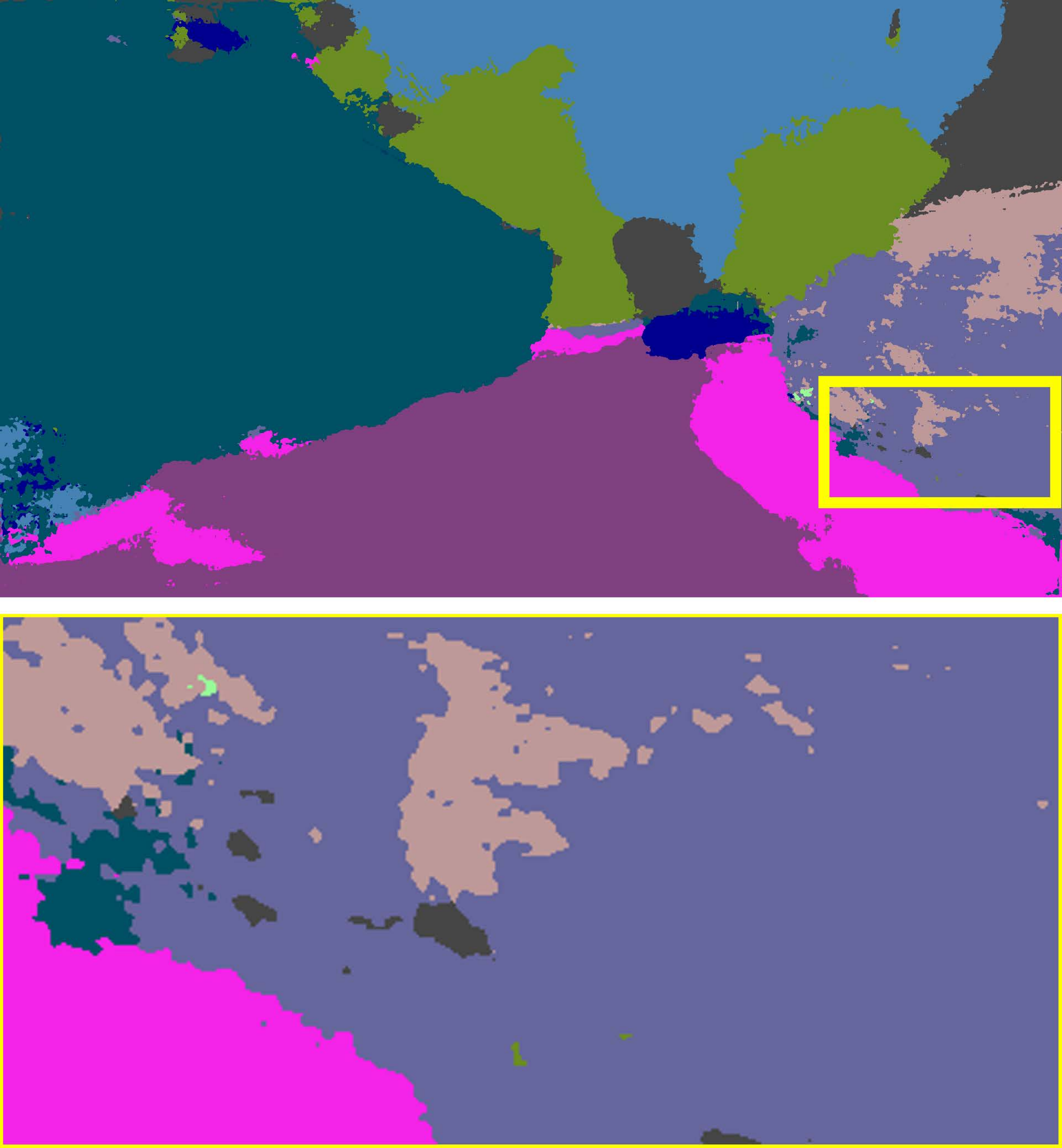}&
		\includegraphics[width=0.104\linewidth]{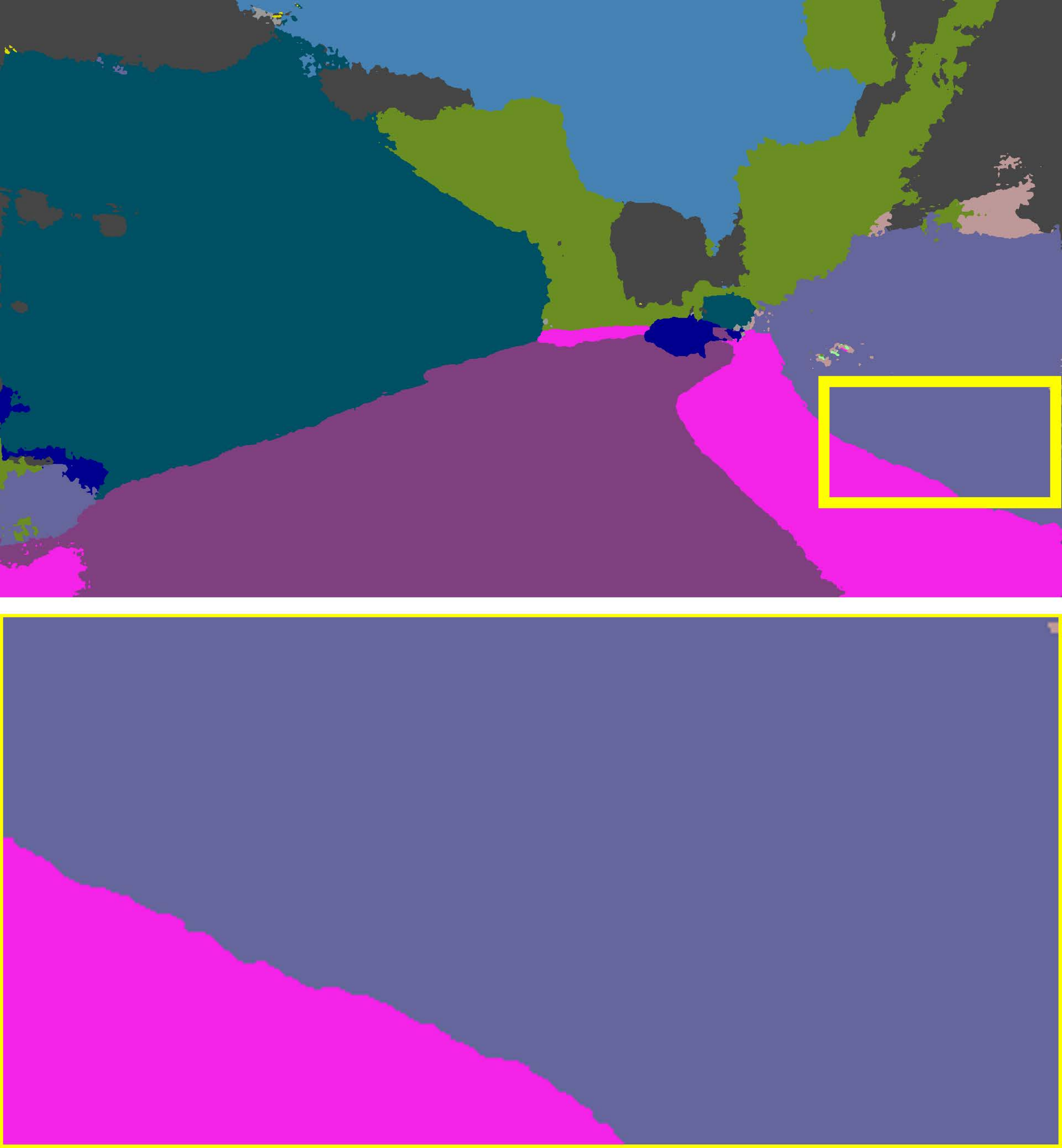}&
		\includegraphics[width=0.104\linewidth]{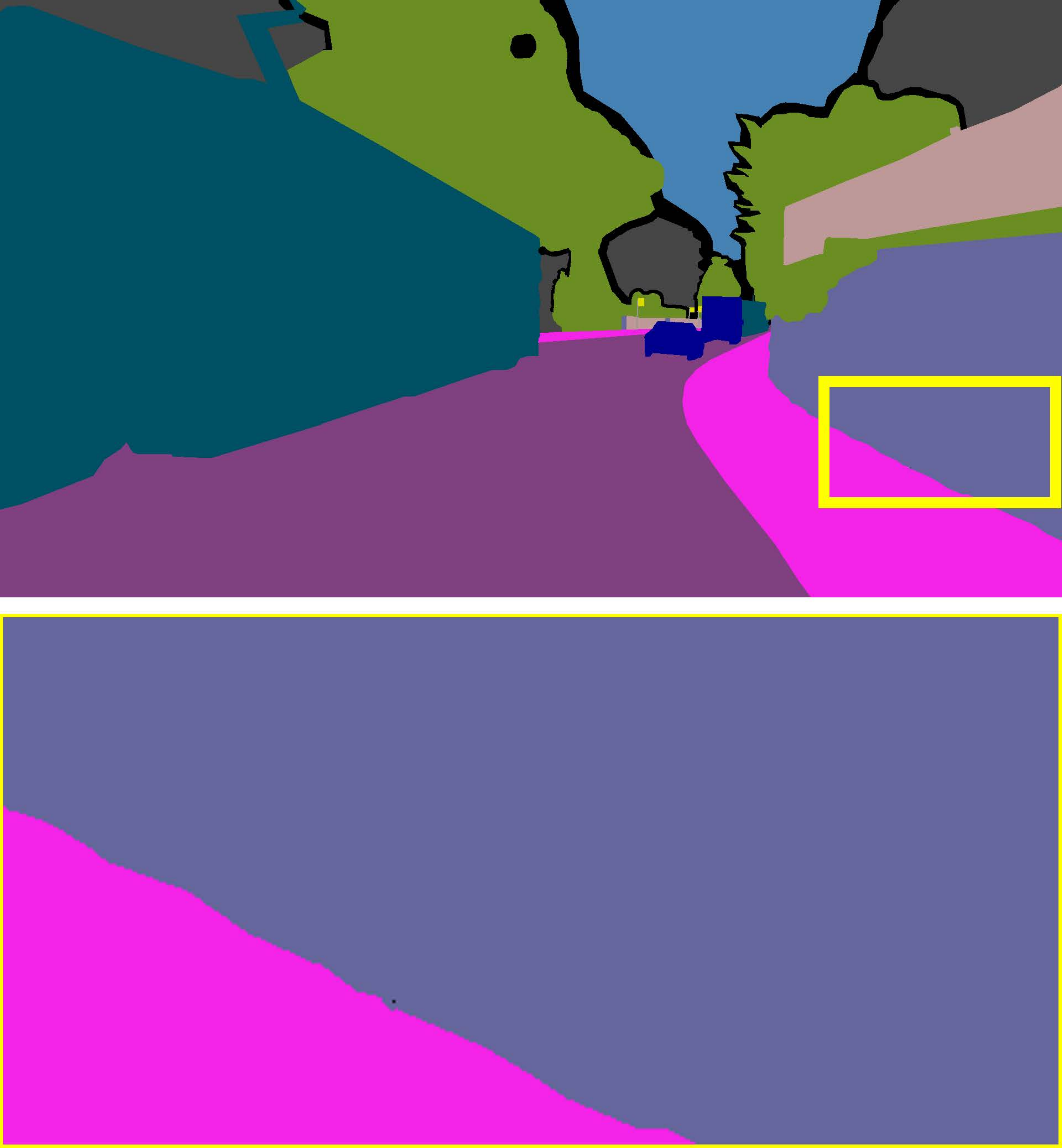}\\
		\includegraphics[width=0.104\linewidth]{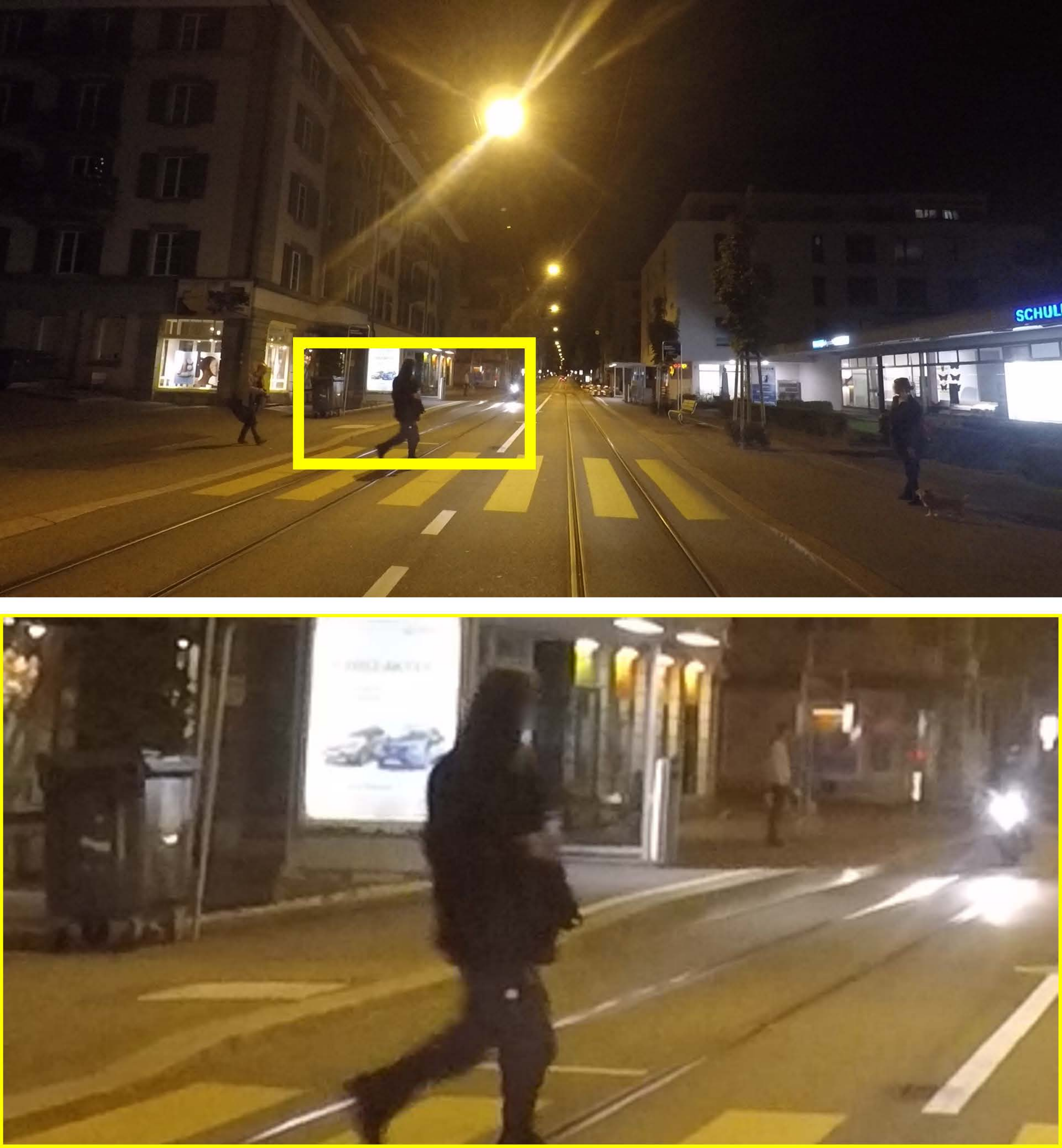}&
		\includegraphics[width=0.104\linewidth]{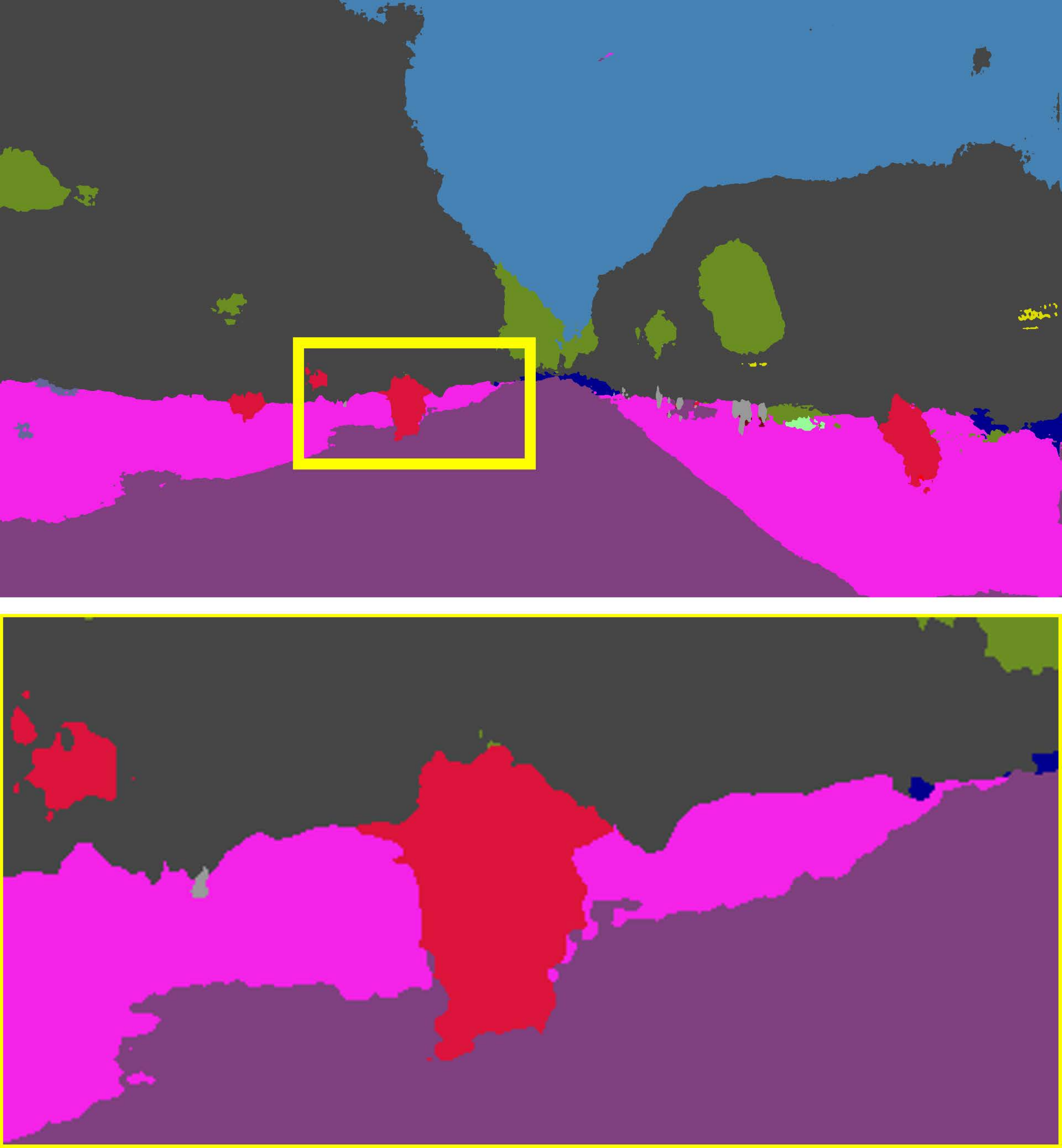}&
		\includegraphics[width=0.104\linewidth]{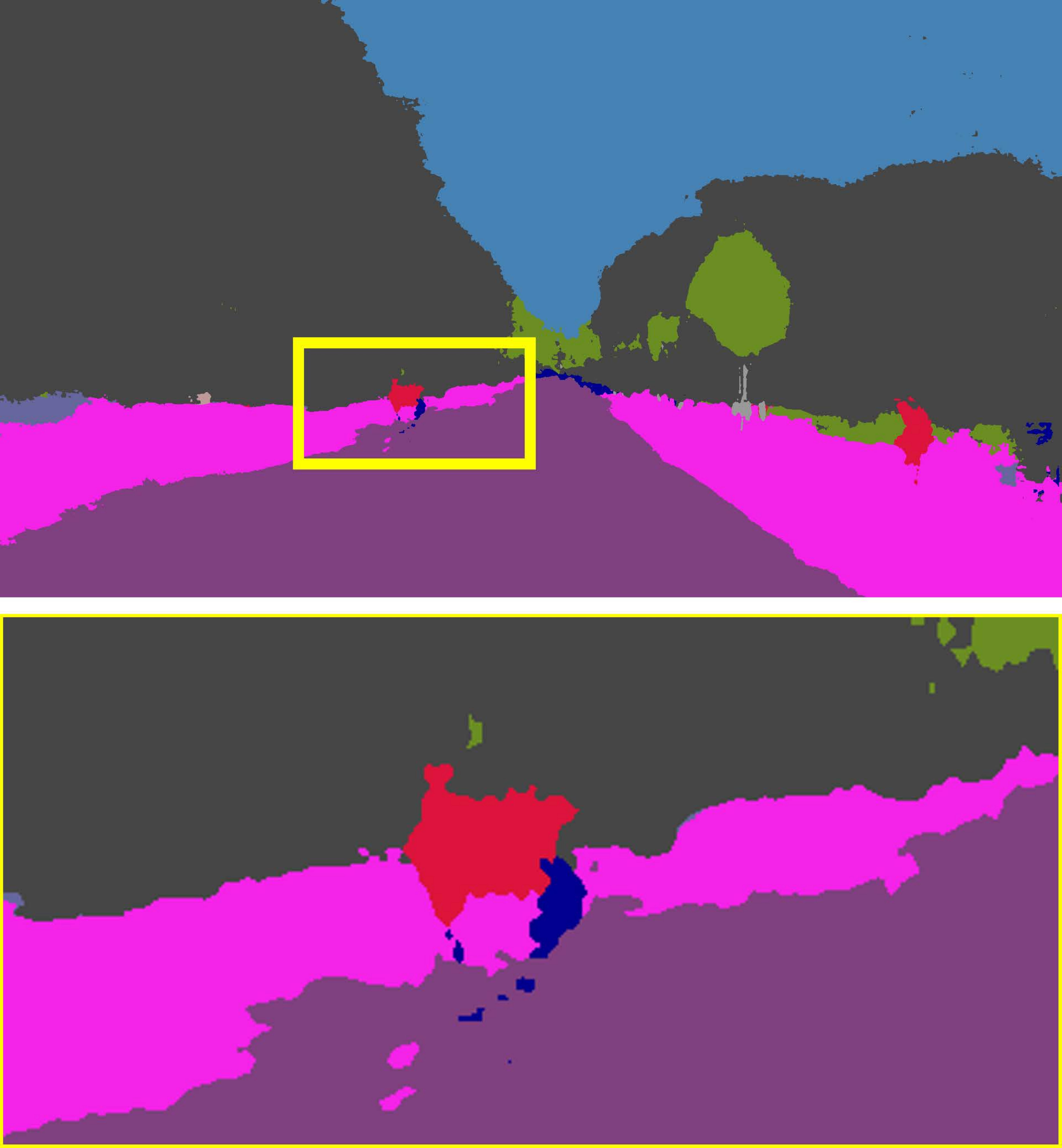}&
		\includegraphics[width=0.104\linewidth]{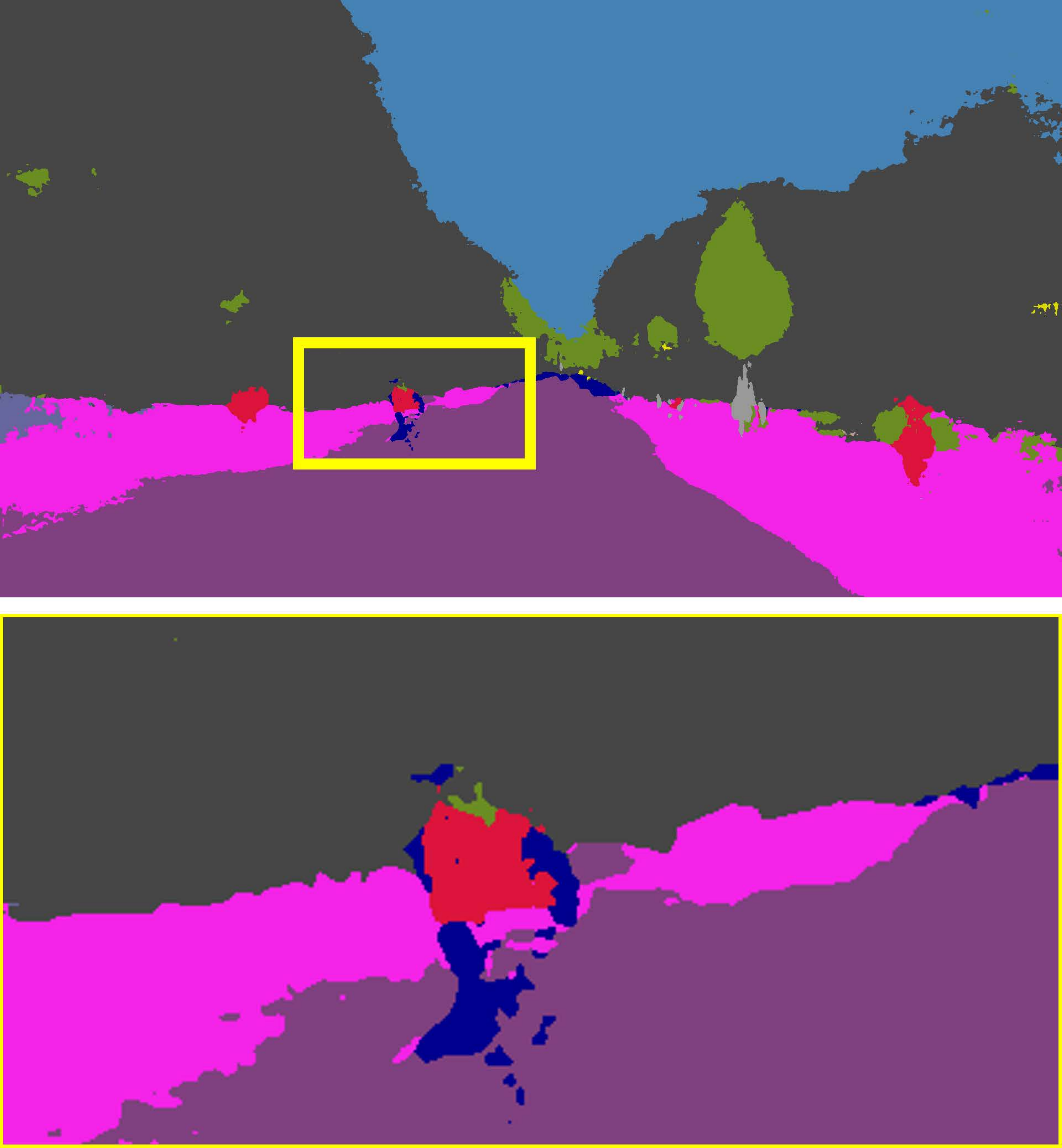}&
		\includegraphics[width=0.104\linewidth]{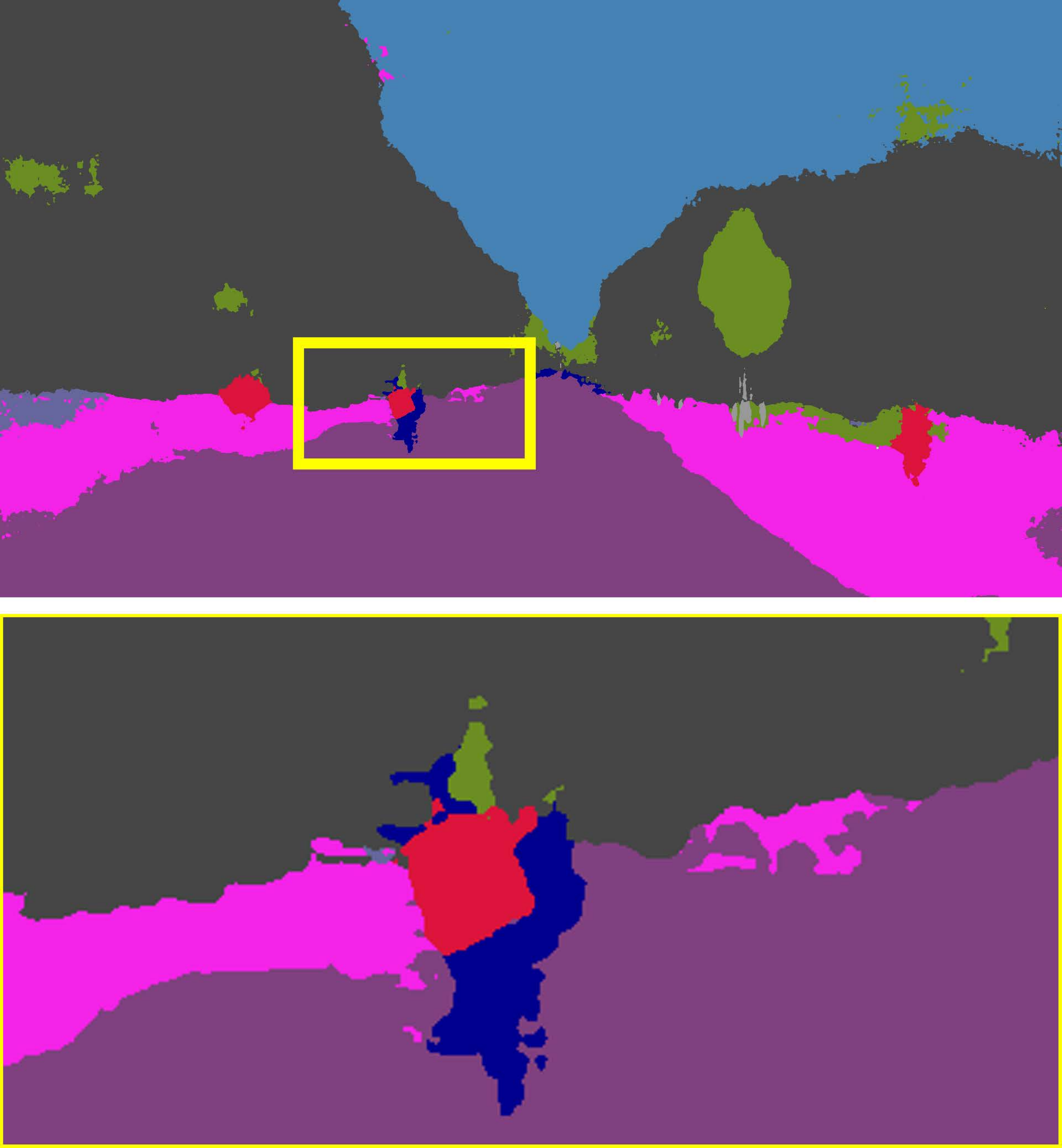}&
		\includegraphics[width=0.104\linewidth]{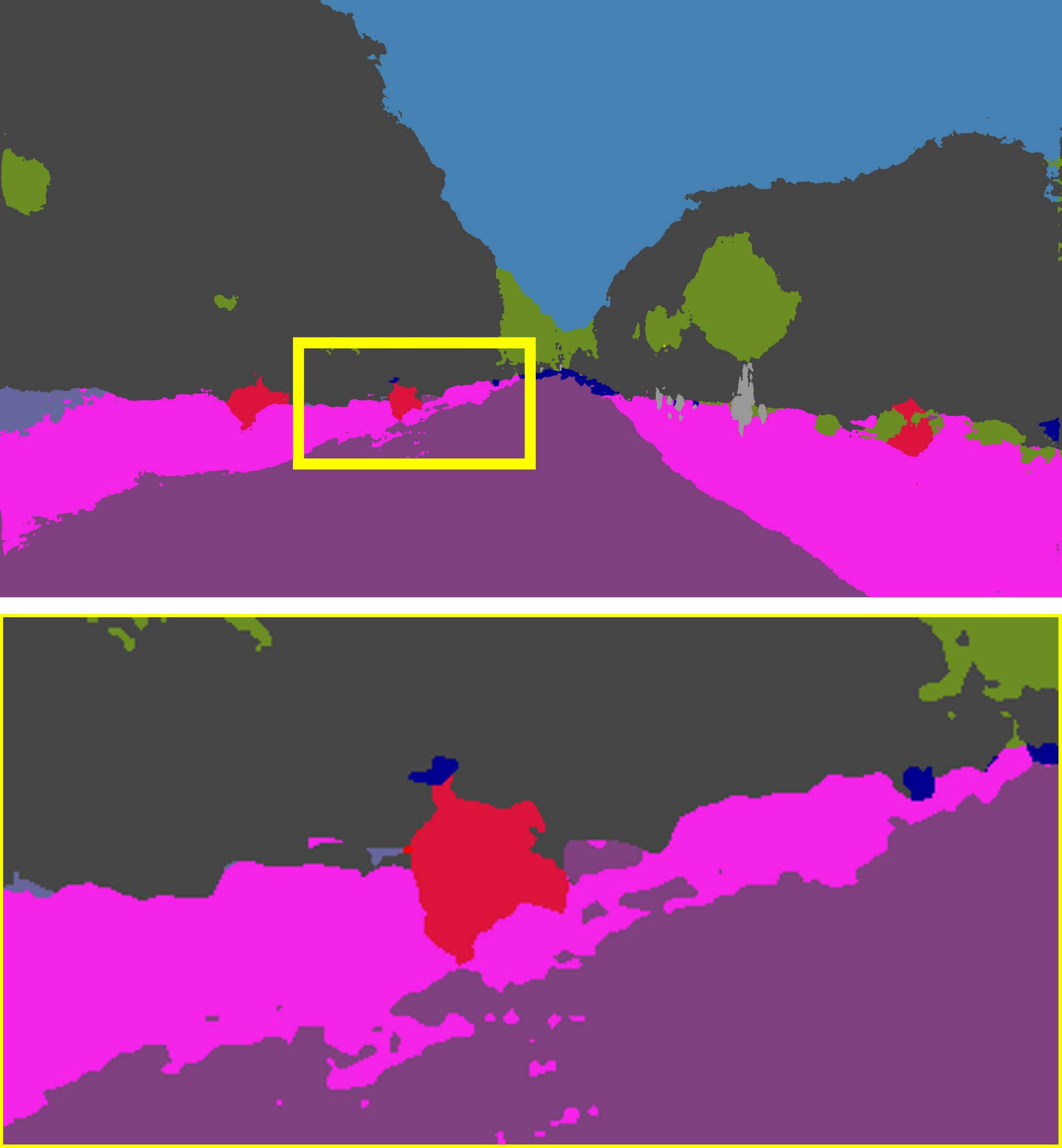}&
		\includegraphics[width=0.104\linewidth]{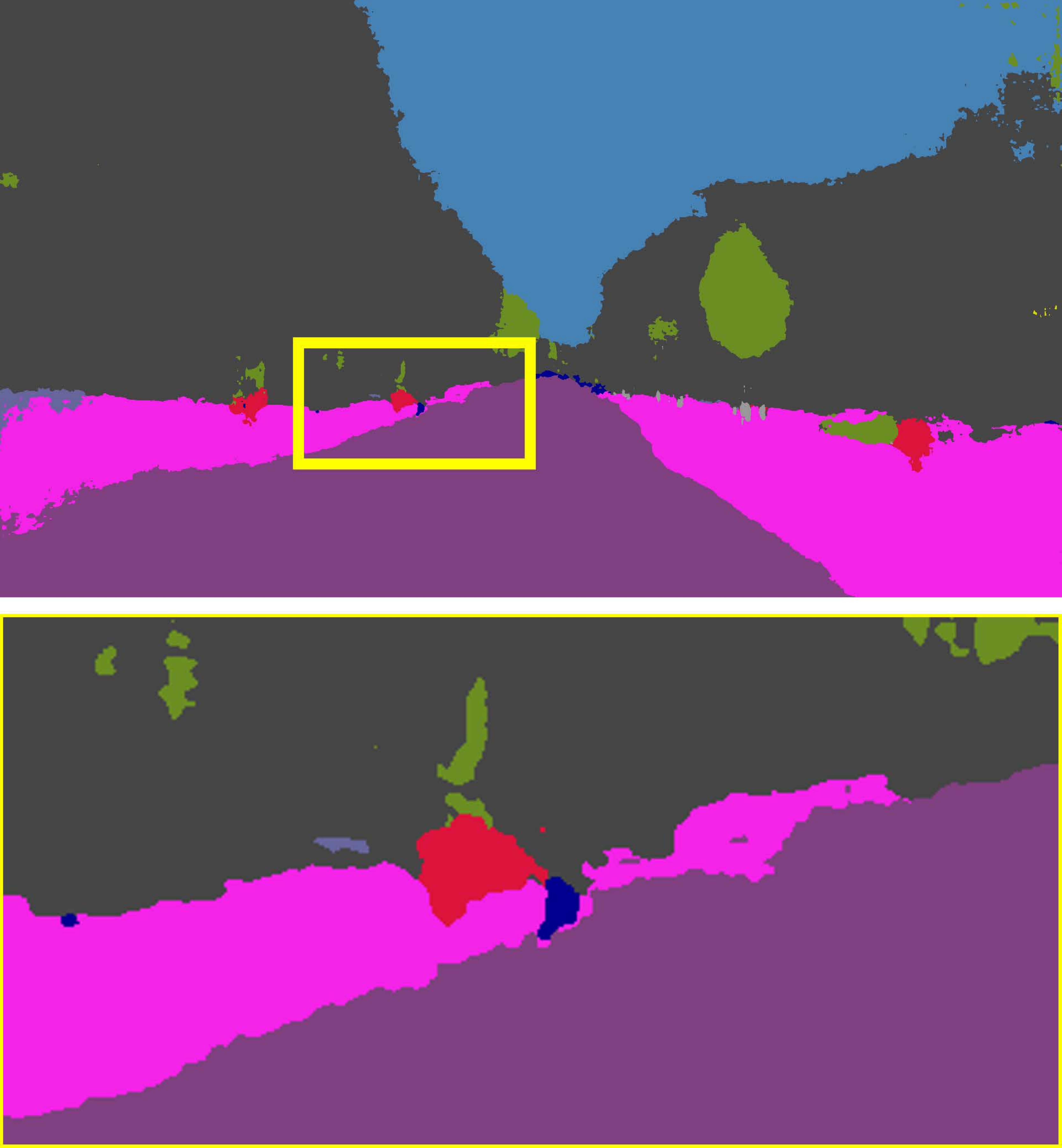}&
		\includegraphics[width=0.104\linewidth]{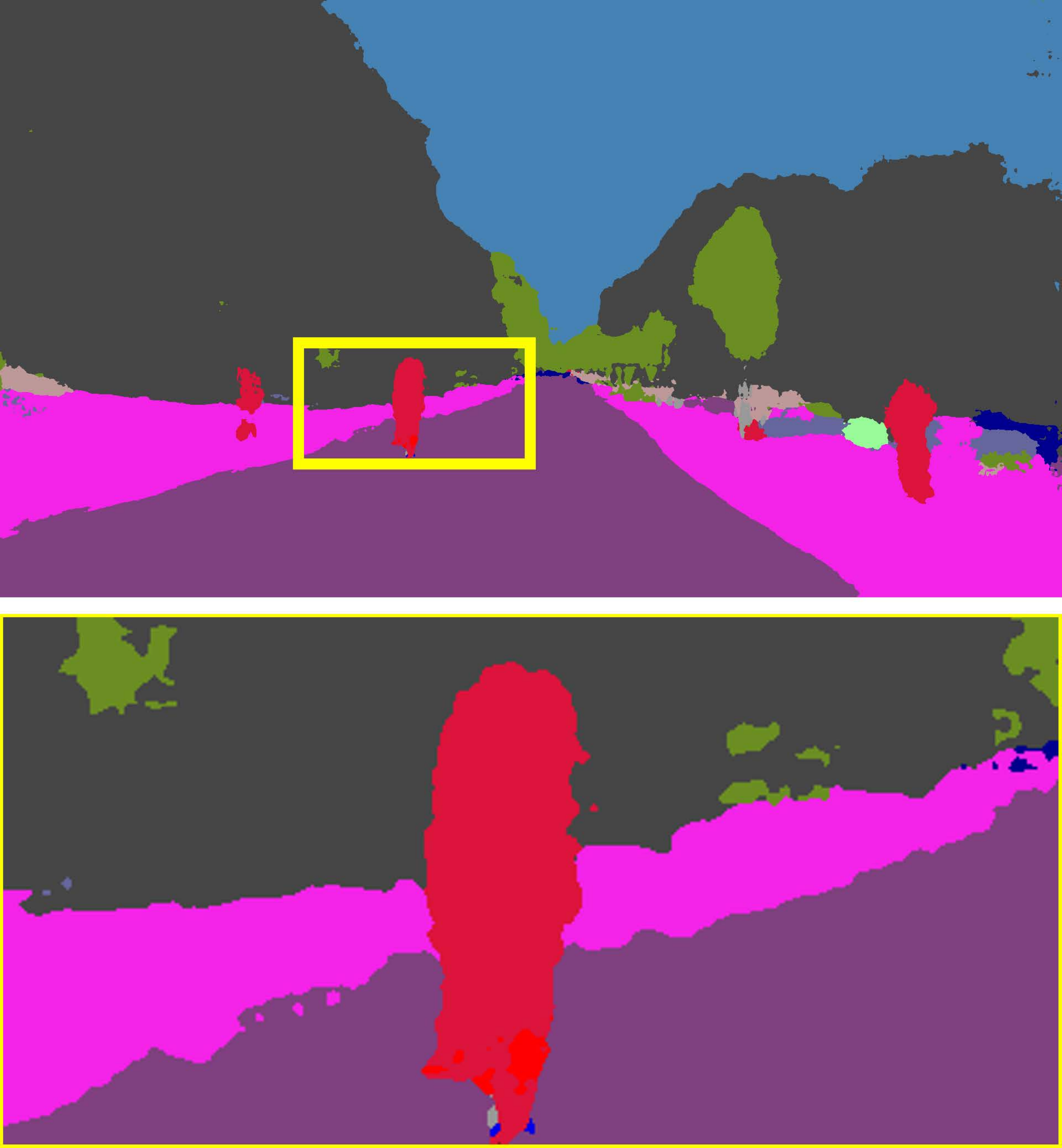}&
		\includegraphics[width=0.104\linewidth]{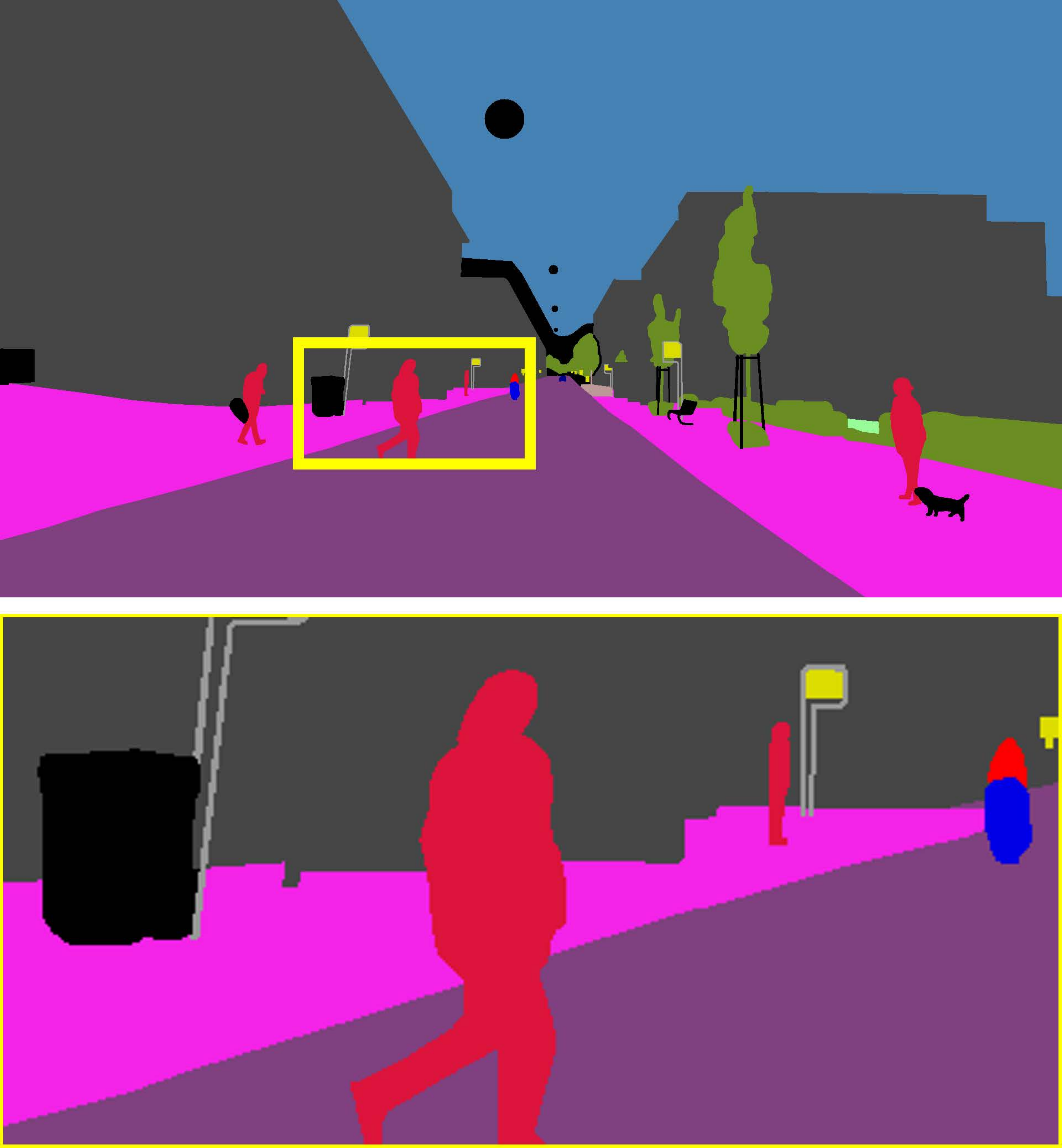}\\
		\footnotesize Input&\footnotesize DeepLab-v3+&\footnotesize DeepUPE&\footnotesize EnGAN &\footnotesize FIDE&\footnotesize GLADNet&\footnotesize ZeroDCE&\footnotesize Ours &\footnotesize Label\\
	\end{tabular}
	\caption{Visual results of semantic segmentation on the ACDC dataset~\cite{ACDC}. Yellow boxes indicate the obvious differences.}
	\label{fig:Segmentation}
\end{figure*}

\section{Applications for High-level LLVs}

\subsection{Low-Light Object Detection}
To fully evaluate the performance of detection, we considered two challenging datasets, including the DARK FACE dataset~\cite{yang2020advancing} (face detection) and the ExDark dataset (multiple objects, e.g., car, dog, table, and so on). In which, we utilized the detection model S3FD~\cite{zhang2017s3fd}, PyramidBox~\cite{tang2018pyramidbox}, DSFD~\cite{li2019dsfd} as the baseline in all methods for DARK FACE and SSD~\cite{liu2016ssd} for ExDark dataset, respectively. Note that our method searched the neck and head (the basic architecture was built based on SSD~\cite{liu2016ssd}) by using our proposed cooperative search strategy to acquire the architectures toward low-light scenarios.  
As for the DARK FACE dataset, we randomly sampled 5000 images for searching and training, 1000 images for testing.  As for the ExDark dataset, we randomly sampled 5200 images for searching and training, 1200 images for testing. Except for the methods of ``enhancer + detector'', we also considered two recently-proposed low-light face detection methods HLA~\cite{wang2021hla} and REG~\cite{liang2021recurrent}. 

In the architecture search phase for detection, we set the maximum epoch as 30, the batch size as 1, and chose the initial learning rate as $3\times10^{-4}$. The momentum parameter was randomly sampled from (0.5, 0.999) and the weight decay was set as $5\times10^{-4}$. As for the training phase (with searched architecture), we used the same training losses as that in the search phase for detection.

\textbf{Analyzing different training strategy.}
As shown in Table~\ref{tab:Strategy}, we compared the numerical accuracy of different training strategies. It can be easily observed that the ``pretrain+finetune'' strategy acquired the highest score. It also verifies our intention in the beginning. In addition, the accuracy of ``end-to-end training $\Psi$ on $\ell_{\mathtt{t}}+\lambda\ell_{\mathtt{s}}$'' was poorer than ``end-to-end training $\Psi$ on $\ell_{\mathtt{t}}$''. It revealed that the enhancement-friendly loss cannot assist in improving the accuracy of detection.

\textbf{Evaluations on different backbones.}
In our designed framework, the backbone was fixed in the search process. Here, we explored the influence of different backbones in detection. Table~\ref{tab: backbone} reported the numerical scores among different backbones. We can easily see that the VGG-16 obtained the best scores. 

\textbf{Analyzing the search process.}
Table~\ref{tab:DetSearch} reported mAP results among different cases for task module in terms of object detection. We can easily observe that after introducing our designed scene module, the detection accuracy acquired an evident improvement which attributed to our designed scene module. Further, if only searching SSD by using our designed search process, the performance improvement was more extrusive, which manifested the necessity of search. Fortunately, our constructed cooperative search in our designed scene-task module framework can achieve the best score.

\textbf{Quantitative comparison.}
Table~\ref{tab:detection} demonstrated PR-Curve which reported the quantitative result (Precision and Recall) on DARK FACE dataset. In which, we considered two cases to make a comprehensive evaluation, including executing detection on the enhanced images using the pre-trained model and retraining the detection model on the enhanced images. We could easily see that all compared methods can realize the highest performance after retraining. It indicates that the necessity of retraining detection model on the enhanced data. Fortunately, our method attained the highest numerical scores. 

Further, we performed detection on the ExDark dataset which contained multiple objects and poorer image quality than DARK FACE dataset. Table~\ref{tab: ExDark} reported the detection results on the specific class. Here we only considered the finetune result. It can be easily observed that our designed framework was significantly superior to other methods. 


\textbf{Qualitative comparison.}
Fig.~\ref{fig:DarkFaceDetection} demonstrated the visual comparison among different methods in the pattern ``enhancement+detector''. It can be easily observed that our method detected more objects than others obviously. Further, we also provided the visual comparison between our algorithm and two recent low-light face detection methods in Fig.~\ref{fig:DarkFaceDetection2}. These two methods are unstable and fail to recognize the correct objects that were recognized in the baseline. By contrast, our method is significantly superior to these different types of methods.

Fig.~\ref{fig:ExDark} provided the visual comparison among different state-of-the-art methods on the challenging ExDark dataset. As shown in the first row in Fig.~\ref{fig:ExDark}, most of the compared methods can detect the cat but perform a lower accuracy than ours. As for the second and third rows presented in Fig.~\ref{fig:ExDark}, these methods can detect the correct goal, but containing the wrong identification. As for the rest rows in Fig.~\ref{fig:ExDark}, all compared methods mostly cannot recognize the object because of too poor image quality. Compared with them, our method successfully recognize all objects due to our designed framework.

\subsection{Low-Light Semantic Segmentation}
Here we utilized the pre-trained model acquired in~\cite{chen2018encoder} as our default baseline for segmentation. We evaluated the performance on the newly-published ACDC dataset~\cite{ACDC} which contained 506 nighttime observations and the corresponding segmentation map. We randomly sampled 400 paired images for searching and training, 106 images for testing.  We adopted the mean Intersection-over-Union (mIoU) as the evaluation metric, and the higher the better. Additionally, except for the methods of ``enhancement + segmentation'', we also compared with a recently-proposed low-light semantic segmentation method DANNet~\cite{wu2021dannet}. 

In the architecture search phase for segmentation, we set the maximum epoch as 30, the batch size as 1, and chose the initial learning rate as $3\times10^{-4}$. The momentum parameter was randomly sampled from (0.5, 0.999) and the weight decay was set as $5\times10^{-4}$. As for the training phase (with searched architecture), we use the same training losses as that in the search phase.

Table~\ref{tab: Segmentation} reported numerical results among different approaches on different classes of the ACDC testing dataset. Obviously, our method obtained the best scores on quite a few categories, such as road, pole, motorcycle, bicycle, and so on. Our method also realized the best result in terms of the averaged performance. Further, we provided visual results in Fig.~\ref{fig:Segmentation}. It can be easily observed that our method significantly outperformed other methods, especially the characterization for the overall structure of the specific target. For example, as shown in the zoomed-in regions of the first example, all compared methods produced some unknown artifacts to damage the quality of the generated segmentation map. 

\section{Conclusions}
This paper developed RUAS, a general and principled learning framework to fully explore low-light scene information for handling a series of Low-Light Vision (LLV) problems. We provided a scene energy constrained learning formulation for LLVs, and established two kinds of training strategies by introducing a Retinex-inspired unrolling scheme and an unsupervised scene-oriented loss. Moreover, we defined a cooperative differentiable search strategy to automatically discover the architectures of scene and task modules. Finally, we also demonstrated how to apply RUAS on real-world applications, such as low-light image enhancement, object detection, and semantic segmentation.

\section*{Acknowledgments}
This work is partially supported by the National Key R\&D Program of China (2020YFB1313503), the National Natural Science Foundation of China  (No. 61922019), and the Fundamental Research Funds for the Central Universities.

%

%
%
%
%
%
%

\ifCLASSOPTIONcaptionsoff
\newpage
\fi



%

\bibliographystyle{IEEEtran}
\bibliography{egbib}

%
%
%
%
%




\end{document}